\RequirePackage{fix-cm}

\documentclass[twocolumn]{svjour3}          %

\smartqed  %

\usepackage{graphicx}

\journalname{International Journal of Computer Vision}

\usepackage{mathptmx}      %
\usepackage{booktabs}
\usepackage{makecell}
\usepackage[final]{microtype}
\usepackage[USenglish]{babel}
\usepackage{etoolbox}
\usepackage{hyperref}
\usepackage[T1]{fontenc}
\usepackage{amsmath}
\usepackage{amssymb}
\usepackage{natbib}

\usepackage{afterpage}
\usepackage{placeins}

\usepackage{bm}
\usepackage[margin=0pt]{subfig}
\usepackage{graphbox}
\usepackage{tabularx}

\newcommand*\vect[1]{\mathbf{#1}}

\newcommand{\vx}{\vect{x}}

\DeclareMathOperator*{\concat}{||}

\newcommand{\FC}{\mathrm{FC}}

\usepackage{xspace}

\makeatletter
\DeclareRobustCommand\onedot{\futurelet\@let@token\@onedot}
\def\@onedot{\ifx\@let@token.\else.\null\fi\xspace}

\def\eg{\emph{e.g}\onedot} 
\def\ie{\emph{i.e}\onedot}

\makeatother

\begin{document}

\title{Shape My Face%
}
\subtitle{Registering 3D Face Scans by Surface-to-Surface Translation}

\author{Mehdi Bahri \and
        Eimear O' Sullivan \and
        Shunwang Gong \and
        Feng Liu \and
        Xiaoming Liu \and
        Michael M. Bronstein \and
        Stefanos Zafeiriou
}

\institute{M. Bahri, E. O' Sullivan, S. Gong, M. M. Bronstein, and S. Zafeiriou are with \at
              Department of Computing\\
              Imperial College London, London, UK\\
              \email{\{m.bahri, e.o-sullivan16, shunwang.gong16, m.bronstein, s.zafeiriou\}@imperial.ac.uk}           %
            \and
            M. M. Bronstein is also with Twitter, UK\at
            \and
          F. Liu and X. Liu are with \at
              Department of Computer Science and Engineering\\
              Michigan State University, East Lansing, MI, USA\\
              \email{isliuf1990@gmail.com, liuxm@cse.msu.edu}
            \and
            Corresponding author: M. Bahri (ORCID 0000-0002-2409-0261)
}

\date{Received: date / Accepted: date}

\maketitle

\begin{abstract}
Standard registration algorithms need to be independently applied to each surface to register, following careful pre-processing and hand-tuning. Recently, learning-based approaches have emerged that reduce the registration of new scans to running inference with a previously-trained model. The potential benefits are multifold: inference is typically orders of magnitude faster than solving a new instance of a difficult optimization problem, deep learning models can be made robust to noise and corruption, and the trained model may be re-used for other tasks, \eg through transfer learning. In this paper, we cast the registration task as a surface-to-surface translation problem, and design a model to reliably capture the latent geometric information  directly from raw 3D face scans. We introduce Shape-My-Face (SMF), a powerful encoder-decoder architecture based on an improved point cloud encoder, a novel visual attention mechanism, graph convolutional decoders with skip connections, and a specialized mouth model that we smoothly integrate with the mesh convolutions. Compared to the previous state-of-the-art learning algorithms for non-rigid registration of face scans, SMF only requires the raw data to be rigidly aligned (with scaling) with a pre-defined face template. Additionally, our model provides topologically-sound meshes with minimal supervision, offers faster training time, has orders of magnitude fewer trainable parameters, is more robust to noise, and can generalize to previously unseen datasets. We extensively evaluate the quality of our registrations on diverse data. We demonstrate the robustness and generalizability of our model with in-the-wild face scans across different modalities, sensor types, and resolutions. Finally, we show that, by learning to register scans, SMF produces a hybrid linear and non-linear morphable model. Manipulation of the latent space of SMF allows for shape generation, and morphing applications such as expression transfer in-the-wild. We train SMF on a dataset of human faces comprising 9 large-scale databases on commodity hardware.
\end{abstract}

\section*{Declarations}

\paragraph*{Funding:}~M. Bahri was supported by a Department of Computing scholarship and a Qualcomm Innovation Fellowship. E. O' Sullivan was supported by a Department of Computing scholarship. We thank Amazon for AWS Cloud Credits for Research.

\paragraph*{Conflicts of interest:}~The authors declare no conflicts of interest.

\paragraph*{Availability of data and material:}~All databases of Table 1 are available upon request to their respective authors and sufficient for reproducing the main results of the paper. 4DFAB and MeIn3D (used to train SMF+) and 3DMD (used for testing) are not publicly available currently.

\paragraph*{Code availability:}~A pre-trained model will be released publicly along with code.

\section{Introduction}

\begin{figure*}[t]
    \centering
    \includegraphics[width=174mm]{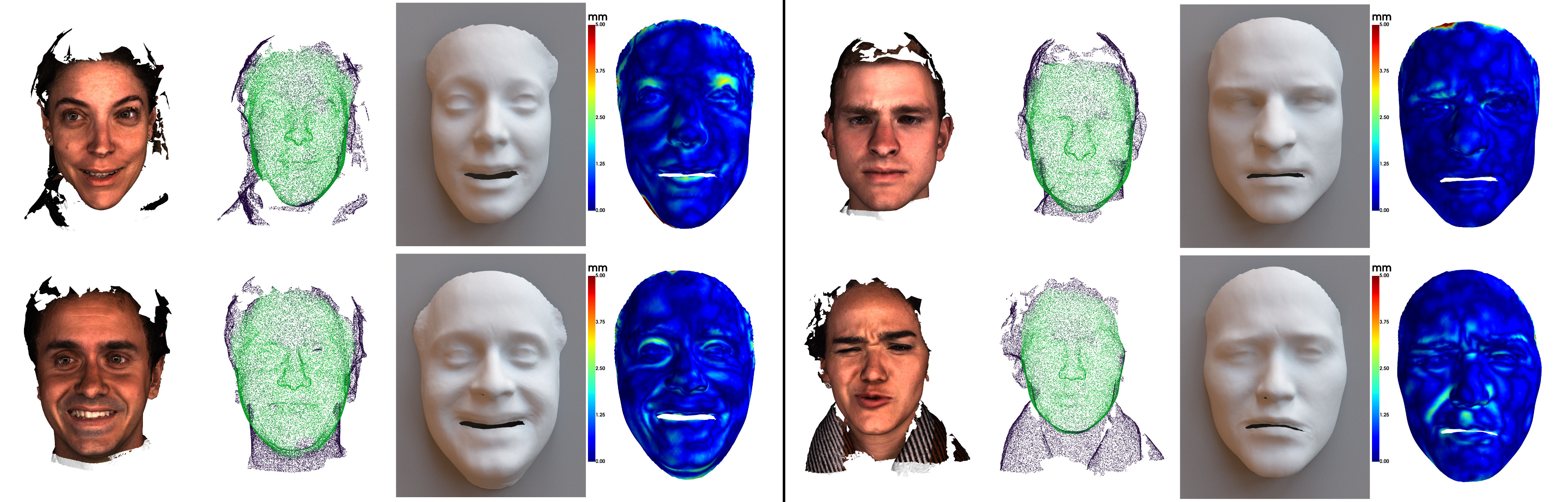}
    \caption{\textbf{Sample test scans and their registration.} Left to right: textured mesh, input point cloud sampled uniformly from the mesh (black) and the attention mask predicted by the model (green), registration, and heatmap of the surface error.}
    \label{fig:teaser}
\end{figure*}

3D shapes come in a variety of representations, including range images, voxel grids, point clouds, implicit surfaces, and meshes. Human face scans, in particular, are often given as either range images, or meshes, but typically do not share a common parameterization (\ie, the output of the 3D scanner does not typically have a fixed connectivity, sampling rate etc.). 
Fundamentally, this diversity of representations is only a by-product of the inability of computers to represent continuous surfaces, but the latent geometric information to be represented is the same. In practice, this poses a challenge: two surfaces represented with two different parameterizations are not easily compared, which makes exploiting the geometric information difficult. Finding a shared representation while preserving the geometry is the task of dense surface registration, a cornerstone in both 3D computer vision and graphics \citep{Amberg2007,Salazar2014}.

The design and construction of a shared shape representation is often implemented by means of a common template, which has a predefined number of vertices and vertex connectivity. After choosing the common template, a fitting method is implemented to bring the raw facial scans in dense correspondence with the chosen template. The use of a common template is a crucial step towards learning a statistical model of the face shape, also know as 3D Morphable Models (3DMMs) \citep{Blanz1999,Booth2016}, which is a very important tool for shape representation and has been used for a wide range of applications spanning from 3D face reconstruction from images \citep{Blanz2003,booth20183d} to diagnosis and treatment of face disorders \citep{knoops2019machine,mueller2011missing}. 

Arguably, the current methods of choice for establishing dense correspondences are variants of Non-rigid Iterative Closest Point (NICP) \citep{Amberg2007}, and non-rigid registration approaches whose regularization properties are defined by statistical \citep{cheng2017statistical} and non-statistical \citep{luthi_gaussian_2016} models. The application of deep learning techniques to the problem of establishing dense correspondences was only recently possible after the design of proper layered structures that directly consumes point clouds and respect the permutation invariance of points in the input data (\eg, PointNet \citep{Qi2017a}).

To the best of our knowledge the only technique that tries to solve the problem of establishing dense correspondences on unstructured point-cloud data and learning a face model on a common template has been presented in \cite{Liu_2019_ICCV}. The method uses a PointNet to summarise (\ie, encode) the information of an unstructured facial point cloud. Then, fully-connected layers (similar to the ones used in dense statistical models \citep{Blanz1999,Booth2016}) are used to reconstruct (\ie, decode) the geometric information in the topology of the common template. In this paper, we work on a similar line of research and we make a series of important contributions in three different areas. In particular,
\begin{itemize}
    \item \textbf{Network architecture}. We propose architectural modifications of the point cloud CNN framework that improve on restrictions of \cite{Qi2017a}. That is, in order to avoid having to adopt heuristic noise reduction and cropping strategies we incorporate a learned attention mechanism in the network structure. We demonstrate that the proposed architecture is better suited for in-the-wild captured data. Furthermore, we propose a variant of PointNet better suited for small batches, hence able to consume higher resolution raw-scans. Our morphable model part of the network (\ie, the decoder) comprises of a series of mesh-convolutional layers \citep{bouritsas2019neural,Gong2019} with novel (in the mesh processing literature) skip connections that can capture better details and local structures. Finally, our network structure is also considerably smaller than the state-of-the-art.
    \item \textbf{Engineering/Implementation}. One of the major challenges when establishing dense correspondences in raw facial scans is the large deformations of the mouth area, especially in extreme expressions. We propose a very carefully engineered approach that smoothly  incorporates a statistical mouth model. We demonstrate our method captures the mouth area very robustly.
    \item \textbf{Application}. Our emphasis in this work is on robustness to noise in the scans (\eg sensor noise, background contamination, and points from the inside of the mouth), compactness of the model, and generalization. The model we develop should be readily usable on, \eg, embedded 3D scanners to produce both a registered scan and a set of latent representations that can be leveraged in downstream tasks. We present extensive experiments to demonstrate the power of our algorithm, such as expression transfer and interpolation between in the wild scans across modalities and resolution. One of the major outcomes of our paper is a novel morphable model trained on 9 diverse large scale datasets, which will be made public.
\end{itemize}
Figure \ref{fig:teaser} shows some test textured scans and their corresponding registrations and attention masks.

\subsection{Structure of the Paper}

We provide an extensive summary of prior published work in Section \ref{sec:previous_art}, covering relevant areas of the morphable models, registration, and 3D deep learning literature. Section \ref{sec:comparison_3dfc} is dedicated to reviewing the current state of the art model, which we use as a baseline in our experiments, and to highlight the limitations and challenges we tackle. We introduce our model, Shape My Face (SMF) in Section \ref{sec:method}, and provide detailed descriptions of its different components, how they provide solutions to the challenges identified in Section \ref{sec:comparison_3dfc}, and how they allow us to frame the registration task as a surface-to-surface translation problem. We also introduce our model trained on a very large dataset comprising 9 large human face scans databases. For the sake of clarity, we split our experimental evaluation into two parts. Section \ref{sec:registration_experiments} studies the performance of SMF for registration, and presents a statistical analysis of the model's stability, as well as an ablation study. Section \ref{sec:experiments_morphable} evaluates SMF on morphable model applications and studies properties of the latent representations; in particular, in Section \ref{sec:in_the_wild} we evaluate SMF on surface-to-surface translation applications entirely in the wild.

\paragraph{Notations} Throughout the paper, matrices and vectors are denoted by upper and lowercase bold letters (\eg, $\mathbf{X}$ and  ($\mathbf{x}$), respectively.  
$\mathbf{I}$ denotes the identity matrix of compatible dimensions. The $i^{th}$ column of $\mathbf{X}$ is denoted as $\mathbf{x}_{i}$. The sets of real numbers is denoted by $\mathbb{R}$. A {\em graph} $\mathcal{G} = (\mathcal{V}, \mathcal{E})$ consists of 
{\em vertices} $\mathcal{V}=\{1,\hdots, n\}$ and {\em edges} $\mathcal{E} \subseteq \mathcal{V}\times \mathcal{V}$. The graph structure can be encoded in the {\em adjacency matrix} $\mathbf{A}$, where $a_{ij} = 1$ if $(i,j) \in \mathcal{E}$ (in which case $i$ and $j$ are said to be {\em adjacent}) and zero otherwise.  The {\em degree matrix} $\mathbf{D}$ is a diagonal matrix with elements $d_{ii} = \sum_{j=1}^n a_{ij}$. 
The {\em neighborhood} of vertex $i$, denoted by $\mathcal{N}(i) = \{j : (i,j) \in \mathcal{E} \}$, is the set of vertices adjacent to $i$. 

\section{Related Work}
\label{sec:previous_art}

Although primarily a fast registration method with a focus on generalizability to unseen data, our approach also makes important progress towards learning an accurate part-based non-linear 3D morphable model of the human face, as well as a generative model with applications to surface-to-surface translation. We first review the relevant literature across the related fields. Then, we devote Section \ref{sec:comparison_3dfc} to exposing the limitations of the current state of the art algorithm that motivate the choices made in this work.

\subsection{Surface Registration and Statistical Morphable Models}

Surface registration is the task of finding a common parameterization for heterogeneous surfaces. It is a necessary pre-processing step for a range of downstream tasks that assume a consistent representation of the data, such as statistical analysis and building 3D morphable models. As such, it is a fundamental problem in 3D computer vision and graphics.

\subsubsection{Surface registration}
\label{sec:lit_rev_registration}

Two main classes of methods coexist for surface registration. Image-based registration methods first require finding a mapping between the surface to align and a two-dimensional parameter space; most commonly, a UV parameterization is computed for a textured mesh, typically using a cylindrical projection. Image registration methods are then applied to align the unwrapped surface with a template, for instance using optical flow analysis \citep{Horn1981,Lefebure2001}, or thin plate spline warps \citep{Bookstein1989}. UV-space registration is computationally efficient and relies on mature image processing techniques, but the flattening step unavoidably leads to a loss of information, and sampling of the UV space is required to reconstruct a surface. For this reason, the second main class of surface registration methods operates directly in 3D, avoiding the UV space entirely. Prominent examples include the Non Rigid Iterative Closest Point (NICP) method \citep{Amberg2007}, a generalization of the Iterative Closest Point (ICP) method \citep{Chen1991,Besl1992} that introduces local deformations, or the Coherent Point Drift (CPD) algorithm \citep{Myronenko2007,Myronenko2010}. NICP operates on meshes and solves a non-convex energy minimization problem that encourages the vertices of the registered mesh to be close to the target surface, and the local transformations to be similar for spatially close points. Due to its non-convex nature, NICP is sensitive to initialization, and is most often used in conjunction with sparse annotations (\ie landmarks for which a 1-to-1 correspondence is known a priori). Similarly, CPD also encourages the motion of neighboring points to be similar, but operates on point clouds and frames the registration problem as that of mass matching between probability distributions. As such, it is closely related to optimal transport registration \citep{Feydy2017}. We refer to relevant surveys \citep{VanKaick2011,Tam2013} for a more complete review of non-deep learning based surface registration methods.

\subsubsection{Linear, multilinear, and non-linear morphable models}

Linear morphable models for the human face were first introduced in the seminal work of \cite{Blanz1999}. The authors proposed to model the variability of human facial anatomy by applying Principal Component Analysis (PCA) \citep{Pearson1901,Hotelling1933} to 200 laser scans (100 male and 100 female) of young adults in a neutral pose. Scans were aligned by image registration in the UV space with a regularized form of optical flow. The resulting set of components forms an orthogonal basis of faces that can be manipulated to synthesize new faces. \cite{Amberg2008} extended the PCA approach to handle expressions for expression invariant 3D face recognition, using scans registered directly with NICP \citep{Amberg2007}.
\cite{Patel2009} introduced the widely-used Basel Face Model (BFM), also trained on 200 scans registered with NICP. It is only with the work of \cite{Booth2016,Booth2018} that a morphable model trained on a large heterogeneous population, known as the Large Scale Face Model (LSFM) was made available. The authors use the BFM template and a modification of the NICP algorithm, along with automated pruning strategies, to build a high quality model of the human face from almost 10000 subjects. LSFM is trained on neutral scans only, but can be combined with a bank of facial expressions, such as the popular FaceWarehouse \citep{Cao2014}.

Multilinear extensions of linear morphable models have been considered as early as \cite{Vlasic:2005:FTM:1073204.1073209}  where a tensor factorization was used to model different modes of variation independently (\eg, identity and expression) with applications to face transfer, and refined by \cite{Bolkart2015}. However, the multilinear approach requires every combination of subject and expression to be present exactly once in the dataset, a requirement that can be both hard to satisfy and limiting in practice. \cite{Salazar2014} proposed an explicit decomposition into blendshapes as an alternative. In \cite{Li2017}, the authors propose to combine an articulated jaw with linear blending to obtain a non-linear model of facial expressions.

\subsubsection{Part-based models}
\label{sec:part_based_3dmms}

Besides a global PCA model, \cite{Blanz1999} also presented a part-based morphable model. The authors manually segmented the face into separate regions and trained specialized 3DMMs for each part, that can then be morphed independently. The resulting model is more expressive than a global PCA would be, and is obtained by combining the parts using a modification of the image blending algorithm of \cite{Burt1985}. \cite{DeSmet2011} and \cite{Tena2011} showed manual segmentation may not be optimal, and that better segmentation can be defined by statistical analysis. \cite{Tena2011} designed an interpretable region-based model for facial animation purposes. 

Part-based models also appear when attempting to represent together different distinct parts of the body. \cite{Romero2017} model hands and bodies together by replacing the hand region of SMPL \citep{Loper2015} with a new specialized hand model called MANO. \cite{Joo2018} present the \textit{Frankenstein} model, a morphable model of the whole human body that combines existing specialized models of the face \citep{Cao2014}, body \citep{Loper2015}, and a new artist-generated model for hands. The model's parameters are defined as the concatenation of all the parts' parameters. The final reconstruction is obtained by linear blending of the vertices of the separate parts using a manually-crafted matrix. The final model has fewer vertices than the sum of its parts, and the parts were manually aligned. As per the author's own description, minimal blending is done at the seams.

In \cite{Ploumpis2019a,Ploumpis2019}, a high-definition head and face model is created by blending together the Liverpool-York Head model (LYHM) \citep{Dai2017} and the Large-Scale Face Model (LSFM) \citep{Booth2018}. While LYHM includes a facial region, replacing it with LSFM offers more details. Two approaches are proposed to combine the models smoothly. A regression model learned between the two models' parameter spaces, and a Gaussian Process Morphable Model (GPMM) approach \citep{Luthi2018} where the covariance matrix of a GPMM is carefully crafted from the covariance matrices of its parts using a weighting scheme based on the Euclidean distance of the vertices to the nose tip of the registered meshes (\ie the outputs of the head and face models). A refinement phase involving non-rigid ICP further tunes the covariance matrix of the GPMM.

We refer the interested reader to the recent review of \cite{Egger2019} for more information.

\subsection{Deep Learning on Surfaces}

Deep neural networks now permeate computer vision, but have only become prominent in 3D vision and graphics in the past few years. We review some of the recent algorithmic advances for representation learning on surfaces, surface registration, and morphable models.

\subsubsection{Geometric deep learning on point clouds and meshes}
Recent methods from the field of Geometric Deep Learning \citep{Bronstein2017} have emerged and propose analogues of classical deep learning operations such as convolutions for meshes  and point clouds.

Point cloud processing methods treat the discrete surface as an unordered point set, with no pre-defined notion of intrinsic distances or connectivity. The pioneering work of PointNet \citep{Qi2017a} defines a point set processing layer as a $1 \times 1$ convolution shared among all points, followed by batch normalization, and ReLU activation. The resulting local point-wise features are aggregated into a global representation of the surface by max pooling. In spite of its simplicity, PointNet achieved state of the art result in both 3D object classification and point cloud segmentation tasks, and remains competitive to this day. Follow-up works have explored extending PointNet to enable hierarchical feature learning \citep{Qi2017}, as well as more powerful architectures that attempt to learn the metric of the surface via local kernel functions \citep{xu_spidercnn:_2018,Lei2019,Zhang2019}, or by building a k-NN graph in the feature space \citep{Wang2018_dgcnn}. While these methods obtain higher classification and segmentation accuracy, their computational complexity limits their application to large-scale point clouds, a task for which PointNet is often preferred.

Graph Neural Networks, on the other hand, assume the input to be a graph, which naturally defines connectivity and distances between points. Initial formulations were based on the convolution theorem and defined graph convolutions using the graph Fourier transform, obtained by eigenanalysis of the combinatorial graph Laplacian \citep{Bruna2013b}, and relied on smoothness in the spectral domain to enforce spatial locality. \cite{defferrard_convolutional_2016} accelerated spectral graph CNNs by expanding the filters on the orthogonal basis of Chebyshev polynomials of the graph Laplacian, also providing naturally localized filters. However, the Laplacian is topology-specific which hurts the performance of these methods when a fixed connectivity cannot be guaranteed. \cite{Kipf2017} further simplified graph convolutions by reducing ChebNet to its first order expansion, merging trainable parameters, and removing the reliance on the eigenvalues of the Laplacian. The resulting model, GCN,  has been shown to be equivalent to Laplacian smoothing \citep{Li2018} and has not been successful in shape processing applications.  Attention-based models \citep{Monti2017,Fey2017,Verma2018,Velickovic2017} dynamically compute weighted features of a vertex's neighbours and do not expect a uniform connectivity in the dataset, and generalize the early spatial mesh CNNs that operated on pre-computed geodesic patches \citep{masci_geodesic_2015,boscaini_learning_2016}. Spatial and spectral approaches have both been shown to derive from the more general neural message passing \citep{Gilmer2017} framework. Recently, SpiralNet \citep{DBLP:conf/eccv/LimDCK18}, a specialized operator for meshes, has been introduced based on a consistent sequential enumeration of the neighbors around a vertex. \cite{Gong2019} introduces a refinement of the SpiralNet operator coined SpiralNet++ which simplifies the computation of the spiral patches.

Finally, recent work explored skip connections to help training deep graph neural networks. In Appendix B of \cite{Kipf2017}, the authors propose a residual architecture for deep GCNs. \cite{Hamilton2017} introduce an architecture for inductive learning on graphs based on an aggregation step followed by concatenation of the previous feature map and transformation by a fully-connected layer. \cite{Li2019_deepgcn} study very deep variants of the Dynamic Graph CNN \citep{Wang2018_dgcnn} using residual and dense connections for point cloud processing. Finally, in \cite{Gong2020}, the authors relate graph convolution operators to radial basis functions to propose affine skip connections, and demonstrate improved performance compared to vanilla residuals for a range of operators.

\subsubsection{Registration}

The methods presented in Section \ref{sec:lit_rev_registration} are framed as optimization problems that need to be solved for every surface individually. Although able to produce highly accurate registrations, they can be costly to apply to large datasets, and are based on axiomatic conceptualizations of the registration task. The reliance on sparse annotations to accurately register expressive scans also means the data needs to be manually annotated, a tedious and expensive task. A new class of learning-based surface registration models is therefore emerging that, once passed the initial training effort, promise to reduce the registration of new data to a fast inference pass, and to potentially outperform hand-crafted algorithms. In PointNetLK \citep{Aoki2019}, the authors adapt the image registration of \cite{Lucas1981} to point clouds in a supervised learning setting. A PointNet \citep{Qi2017a} encoder is trained to predict a rigid body transformation $\mathbf{G} \in SE(3)$, with a loss defined between the network's prediction $\mathbf{G}_{est}$ and a ground truth transformation $\mathbf{G}_{gt}$ as $||\mathbf{G}_{est}^{-1} \mathbf{G}_{gt} - \mathbf{I}||_F$, with $||.||_F$ the Frobenius (matrix $\ell_2$) norm. A similar technique is employed in \cite{Wang2019}, where the authors introduce a supervised learning model for rigid registration coined as Deep Closest Point (DCP). DCP learns to predict the parameters of a rigid motion to align two point clouds, and is trained on synthetically generated pairs of point clouds, for which the ground truth parameters are known. The follow-up work of PRNet \citep{NIPS2019_9085} offers a self-supervised approach for learning rigid registration between partial point clouds.  In \cite{Lu2019}, and \cite{Li2019_iterative_marching_points}, supervised learning algorithms are defined for rigid registration, but with losses defined on dense correspondences between points, and on a soft-assigment matrix, respectively. Finally, \cite{Shimada2019} designed a U-Net like architecture on voxel grids for non-rigid point set registration, however, their method is limited by the resolution of the grid and does not build latent representations of the scans, nor does it provide a morphable model.

\subsubsection{Morphable models}

\cite{8354111} train a hybrid encoder-decoder architecture on rendered height maps from 3D face scans using an image CNN encoder and a multilinear decoder. This approach circumvents the need for prior registration of the scans to a template, but the face model itself remains linear.

Concurrently, there has been a surge of interest for deep non-linear morphable models to better capture extreme variations. \cite{bagautdinov_modeling_2018} model facial geometry in UV space with a variational auto-encoder (VAE). \cite{Tran2018} replace the linear bases with fully-connected decoders to model 3D geometry and texture from images, a technique extended in \cite{Tran2019}. \cite{ranjan_generating_2018} introduce a convolutional mesh auto-encoder based on Chebyshev graph convolutions \citep{defferrard_convolutional_2016}. \cite{bouritsas2019neural},  use Spiral Convolutions \citep{DBLP:conf/eccv/LimDCK18} to learn non-linear morphable models of bodies and faces. In both these works, the connectivity of the 3D meshes is assumed to be fixed; that is, the scans have to be registered a priori. The non-linear deep neural network replaces the PCA for dimensionality reduction.

In \cite{Liu_2019_ICCV}, an asymetric autoencoder is proposed. A PointNet encoder is applied to rigidly aligned heterogeneous raw scans, and two fully-connected decoders produce identity and expression blendshapes independently on the BFM face template. Thus, the algorithm produces a registration of the input scan. Mesh convolutional decoders are proposed in \cite{Kolotouros2019_cvpr} for human body reconstruction from single images. In \cite{Kolotouros2019_iccv}, model-fitting is introduced to also produce representations directly on the SMPL model.

\section{State of the Art}
\label{sec:comparison_3dfc}

The autoencoder architecture of \cite{Liu_2019_ICCV} is the current state of the art for the learned registration of 3D face scans. A learning-based approach for registration is desirable since a model that generalizes would be able to register new scans very quickly, thus potentially offsetting the time spent training the model. Other benefits compared to traditional optimization-based registration may include increased robustness to noise in the data. Furthermore, an autoencoder learns an efficient latent representation of the scans, which may later be processed for other applications, while the trained decoder can be used in isolation as a morphable model.

Motivated by the aforementioned potential upsides, we review the approach of \cite{Liu_2019_ICCV} and identify key limitations and areas of improvement. We further evaluate a pre-trained model provided by the authors of \cite{Liu_2019_ICCV} on the same dataset used in the original paper (also provided by the authors). We refer to the provided pre-trained model as \textit{the baseline}.

\subsection{Problem formulation and architecture}

A crop of the mean face of the BFM 2009 model is chosen as a face template on which to register the raw 3D face scans. A \textit{registered} (densely aligned) face is modeled as an identity shape with an additive expression deformation:
\begin{equation}
    \mathbf{S} = \mathbf{S}_{id} + \bm{\Delta}\mathbf{S}_{exp}
\end{equation}
With $\mathbf{S} = [x_1, y_1, z_1 ; \ldots ; x_N, y_N, z_N]$ the concatenated, consistently ordered, Cartesian 3D coordinates of the vertices. For this template, $N = 29495$.

A subset of $N_s$ vertices from a processed input scan (details of the processing below) are sampled at random to obtain a point cloud representation of the scan. A vanilla PointNet encoder without spatial transformers produces a joint embedding $\mathbf{z}_{joint} \in \mathbb{R}^{1024}$. Two fully-connected (FC) layers, without non-linearities, are applied in parallel to obtain identity and expression latent vectors in $\mathbb{R}^{512}$:
\begin{align}
    \mathbf{z}_{id} & = \mathbf{W}_{id} \cdot \mathbf{z}_{joint} + \mathbf{b}_{id} = \FC_{id}(\mathbf{z}_{joint})\\
    \mathbf{z}_{exp} & = \mathbf{W}_{exp} \cdot \mathbf{z}_{joint} + \mathbf{b}_{exp} = \FC_{exp}(\mathbf{z}_{joint}).
\end{align}
Two multi-layer perceptrons consisting of two fully-connected layers with ReLU activations decode the identity and expression blendshapes from their corresponding vectors:
\begin{align}
    \mathbf{S}_{id} & = \FC^2_{id} \left( \xi \left( \FC^1_{id}(z_{id}) \right) \right)\\
                    & = \FC^2_{id} \left( \xi \left( \FC^1_{id} \left( \FC_{id}(z_{joint}) \right) \right) \right)\label{eq:dec_id_full}\\
    \bm{\Delta}\mathbf{S}_{exp} & = \FC^2_{exp} \left( \xi \left( \FC^1_{exp}(z_{exp}) \right) \right)\\
                                & = \FC^2_{exp} \left( \xi \left( \FC^1_{exp} \left( \FC_{exp}(z_{joint}) \right) \right) \right)\label{eq:dec_exp_full}
\end{align} with $\xi(x) = \max(0, x)$ the element-wise ReLU non-linearity.\\
Both decoders are symmetric, with $\FC^1_{(\cdot)} : \mathbb{R}^{512} \rightarrow \mathbb{R}^{1024}$ and $\FC^2_{(\cdot)} : \mathbb{R}^{1024} \rightarrow \mathbb{R}^{3}$.

\subsection{Training data}

\begin{table}[t]
\scriptsize
\centering
\caption{\small Summary of training data - reproduced from \cite{Liu_2019_ICCV}. }

\resizebox{\columnwidth}{!}{
\begin{tabular}{l |c| c  c | c  c }
\toprule
Database &\#Subj. & \#Neu. &\#Sample & \#Exp. & \#Sample\\ %
\hline\hline
BU$3$DFE~\cite{LijunYin2006} &    $100$  &  $100$   & $1{,}000$ & $2{,}400$      & $2{,}400$\\
BU$4$DFE~\cite{Yin2008} &    $101$  &  $>$$101$  & $1{,}010$ & $>$$606$       & $2{,}424$\\
Bosphorus~\cite{Savran2008} & $105$  & $299$    & $1{,}495$ & $2{,}603$      & $2{,}603$\\
FRGC~\cite{Phillips2005} & $577$  &$3{,}308$   & $6{,}616$ & $1{,}642$      & $1{,}642$\\
Texas-$3$D\cite{Gupta2010}  & $116$  &$813$     & $1{,}626$ & $336$        & $336$\\
MICC\cite{Masi_micc}      & $53$   &$103$     & $515$   & $\mathbf{-}$ & $\mathbf{-}$\\
BJUT-$3$D~\cite{Baocai_bjut}   & $500$  &$500$     & $5{,}000$ & $\mathbf{-}$ & $\mathbf{-}$\\
\midrule
Real Data  &$1{,}552$   & $5{,}224$ & $17{,}262$& $7{,}587$      & $9{,}405$   \\      
\hline        
Synthetic Data &$1{,}500$ & $1{,}500$ &$15{,}000$&$9{,}000$&$9{,}000$ \\
\bottomrule
\end{tabular}
}
\label{tab:databases}
\end{table}

The training data is formed from seven publicly available face datasets of subjects from a wide range of ethnic backgrounds, ages, and gender, as well as a set of synthetic 3D faces. Table \ref{tab:databases} summarizes the exact composition of the training set.

\paragraph{Synthetic faces} \cite{Liu_2019_ICCV} use the BFM 2009 morphable model to synthesize neutral faces of 1500 subjects, and the 3DDFA expression model \cite{Zhu2015} to further generate 6 random expressions for each synthetic subject.

\paragraph{Real scans} Both neutral and expressive scans are kept, and the data is unlabeled. The data was processed by first converting the scans to textured meshes using simple processing steps, \eg Delaunay triangulation of the depth images. Automatic keypoint localization was applied on rendered frontal views of the scans to detect facial landmarks. The 2D landmarks were back-projected on the raw textured mesh using the camera parameters. The cropped BFM template was annotated with matching landmarks, such that Procrustes analysis could be applied to find a similarity transformation to align the raw scan with the template.

\paragraph{Pre-processing} In \cite{Liu_2019_ICCV}, the authors applied cropping to remove points outside of the unit sphere originating at the tip of the nose of the subject. The authors also applied mesh subdivision to obtain denser ground-truth meshes, thereby facilitating the sub-sampling of 29495 vertices from scans with insufficient native resolution. Finally, the sampling of points from the  scans for training was done at the pre-processing stage. Data augmentation was carried out by randomly sampling vertices from some scans several times and storing the different point clouds separately.

\subsection{Losses and training procedure}

\begin{figure}[tb]
    \centering
    \includegraphics[align=c,width=41mm]{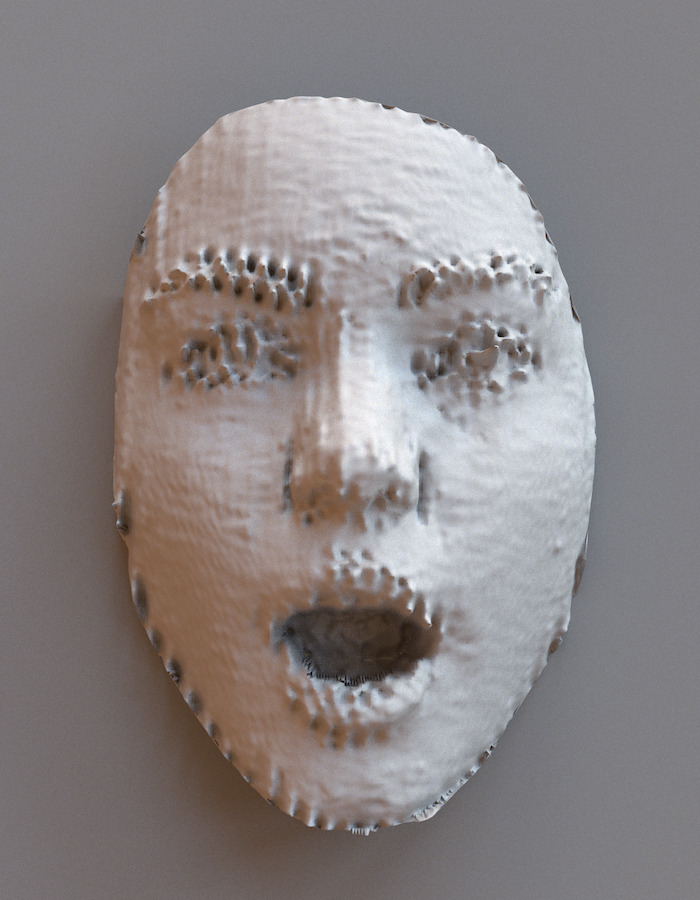}
    \includegraphics[align=c,width=41mm]{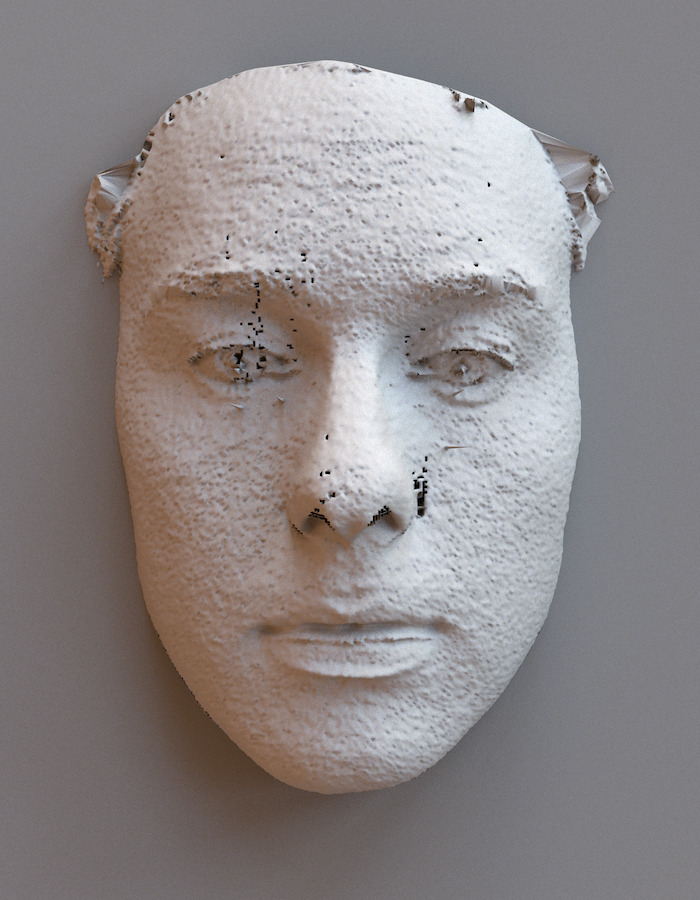}
    \caption{Example sensor noise on the Bosphorus (left) and FRGC (right) datasets. Spikes highlighted on the FRGC scan.}
    \label{fig:sensor_noise}
\end{figure}

\cite{Liu_2019_ICCV} sample $N_s = N = 29495$ vertices from the (subdivided) scans. This number being equal to the number of vertices in the template is a choice, and not a requirement.

Since the synthetic scans are, by nature, in correspondence with the BFM template, \cite{Liu_2019_ICCV} use the element-wise $\ell_1$ norm to train with supervision. For real scans, self-supervised training is carried out to minimize the Chamfer distance between the output $\mathbf{S}$ of the decoder and the potentially subdivided ground-truth scan.

Additional losses are used for synthetic and real scans. Edge-length loss is applied to discourage poor triangulations for the reconstruction. For real scans, the edge-lengths in the output are regularized towards those of the template. For synthetic scans, the edge-length loss is applied as a function of the difference between the edge-length of the input and the output meshes. Normal consistency is used for vertex normals. Due to the presence of noise in the raw scans in the mouth region (points from the inside of the mouth, teeth, or tongue), Laplacian regularization is applied to penalize large changes in curvature in a pre-defined mouth region on the BFM template.

The autoencoder is trained in successive phases. First, only the identity decoder is trained on the synthetic data only, then on a combination of synthetic and real data. After 10 epochs, the identity decoder and the fully-connected layer of the identity branch of the encoder are frozen (\ie backpropagation is disabled) and the expression decoder is trained on synthetic data alone, and then on a mixture of synthetic and real data. Finally, both decoders and encoder branches are trained simultaneously on both synthetic and real scans. We refer the reader to the original work for details.

\subsection{Limitations}

We now study the limitations of the approach.

\subsubsection{Data processing and representation}

\paragraph{Cropping} Although cropping is a simple solution to remove unnecessary parts of the scans, we argue relying on it makes the method less robust. Cropping points outside of the unit sphere centered at the tip of the nose is affected by the quality of the landmark detection. Similarly, choosing the unit sphere centered at the origin of the ambient space will be affected by the location of the scan in $\mathbb{R}^3$. In both cases, even though it is systematic, cropping is inconsistent: as the method is not adaptive, there is no guarantee that the noise (\ie the points that do not contribute to a better face reconstruction and could even degrade the performance) will be discarded. In particular, for range scans such as those from the FRGC \citep{Phillips2005}, Bosphorus \citep{Savran2008} and Texas 3D \citep{Gupta2010} datasets, spikes an irregularities are commonly observed due to sensor noise, as shown in Figure \ref{fig:sensor_noise}. Median filtering has traditionally been applied to the depth images before conversion to 3D surfaces as a means to alleviate this issue \citep{Gupta2010}, but incurs additional human intervention and might cause a loss of details. Cropping would not remove spikes, nor would it discard other irrelevant points if contained within the unit sphere. At the same time, cropping might discard points that would have contributed to the face region.

\begin{figure}[tb]
    \centering
    \includegraphics[width=84mm]{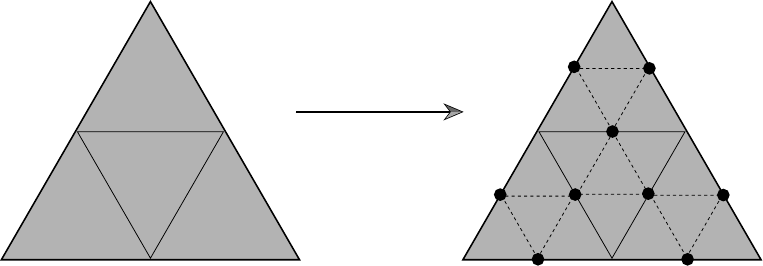}
    \caption{Refinement step of the loop subdivision scheme. Adapted from \cite[Chapter 3.8]{10.5555/3044800}.}
    \label{fig:loop_subdiv}
\end{figure}

\paragraph{Subdivision scheme and vertex subsampling} In \cite{Liu_2019_ICCV}, mesh subdivision was used to improve the accuracy of the dense correspondences (\ie provide more ground truth points for the Chamfer loss), and to enable consistent sampling of 29495 vertices for the input point cloud, even from low-resolution face scans that might not have enough remaining vertices in the facial region after cropping (\eg most scans from the BU-3DFE database \citep{LijunYin2006}). The authors then sampled 29495 vertices at random from the (subdivided) mesh to obtain a point cloud.

Subdivision schemes do not introduce additional details in the scan, but create a denser triangulation from existing triangles. The amount of memory required to store the same geometry is thus largely increased. Figure \ref{fig:loop_subdiv} illustrates the refinement step of the Loop scheme used by \cite{Liu_2019_ICCV}. Assuming we started with one triangle and applied the scheme twice, the figure on the left in Figure \ref{fig:loop_subdiv} shows the result after one subdivision step, and the figure on the right the result after two such steps. We can see that after one step, no vertices were introduced inside of the original triangle: all of the new vertices are located on its edges. After two steps, only 3 vertices have been placed inside the original triangle, yet the number of vertices has been multiplied by 5. In practice, two subdivision steps is the maximum that would be applied due to the rapid increase in memory required to store the subdivided meshes.

It is therefore apparent that a point cloud sampled uniformly at random from the vertices of the mesh cannot - in general - yield a uniform coverage of the surface, even after several mesh subdivision steps. Moreover, using the (subdivided) mesh as a ground truth in the Chamfer loss biases the reconstruction: closest points for vertices of the reconstructed mesh will either never be found inside the triangles of the scan, or in an unfavorable ratio when at least two subdivision steps have been applied. 

\paragraph{Number of point clouds sampled per scan}

\cite{Liu_2019_ICCV} sampled one point cloud per expression scan, and \textit{at most} ten point clouds per neutral scan, per subject. 
As this is done during pre-processing, all samples must be stored individually.
No other data augmentation or transformation (\eg jittering) was used. To avoid overfitting  to a particular sampling of a given surface, we argue that as many different point clouds as possible should be presented to the model for each mesh.

\subsubsection{Architectural limitations and conclusion}

We review the limitations of the two main blocks of the algorithm of \cite{Liu_2019_ICCV}, and conclude the section.

\paragraph{Decoder}
\label{par:soa_decoder}

While MLP decoders are powerful and fully capable of representing details, they do not take advantage of the known template connectivity and geometry. In fact, careful tuning is required to obtain sound shapes: \cite{Liu_2019_ICCV} rely on a strong edge length prior, and use synthetic data extensively during training to condition both the encoders and decoders to respect the geometry of the template.

\begin{figure}[tb]
    \centering
    \includegraphics[width=41mm]{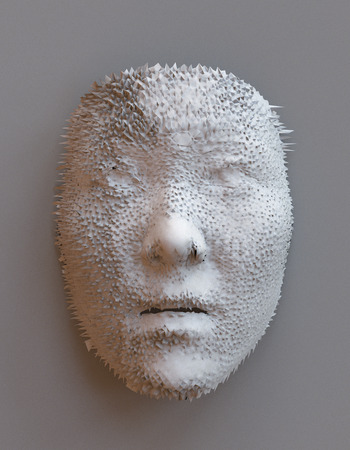}
    \includegraphics[width=41mm]{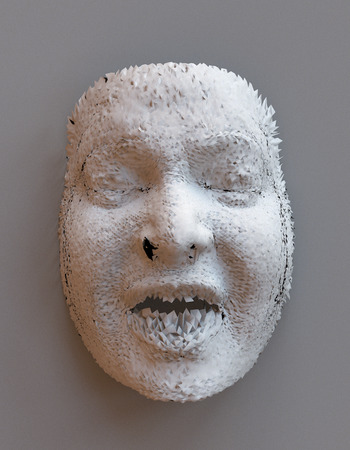}
    \caption{Artifacts obtained with the architecture of \cite{Liu_2019_ICCV}.}
    \label{fig:artifacts_3dfc}
\end{figure}

We observe significant artifacts for a large portion of the input scans, as shown in Figure \ref{fig:artifacts_3dfc}. Notably, we observe tearing-like artifacts and self-intersecting edges, as well as excessive roughness and ragged edges at the boundaries of the shape. In particular, heavy artifacting is present in the mouth region despite the use of the Laplacian loss. Such registrations cannot be exploited for downstream tasks (such as learning from or statistical analysis on the registered scans) without heavy post-processing to correct the artifacts and improve surface fairness.

\paragraph{Encoder}
\label{par:soa_encoder}

\begin{figure}[tbp]
    \centering
    \subfloat[PointNet]{
        \includegraphics[width=.4\linewidth]{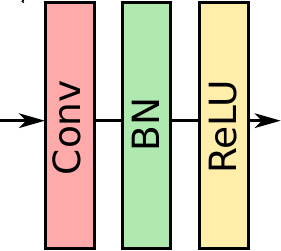}   
        \label{fig:pointnet_block}
    }
    \subfloat[Our block]{
        \includegraphics[width=.4\linewidth]{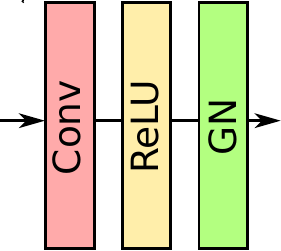}   
        \label{fig:mod_pointnet_block}
    }
    \caption{\textbf{Variants of the PointNet block:} The vanilla PointNet block (a) consists of a $1 \times 1$ convolution followed by batch normalization and ReLU activation. We propose a variant (b) better suited to small batch sizes by replacing batch normalization with group normalization and normalizing the features post-activation.}
    \label{fig:encoder_blocks}
\end{figure}

A vanilla PointNet \citep{Qi2017a} layer consists of a $1 \times 1$ convolution, followed by batch normalization and a ReLU activation, as shown in Figure \ref{fig:pointnet_block}. Choosing $N_s = N$ facilitates mixed batching of synthetic and real scans, but according to \cite{Liu_2019_ICCV}, the optimal batch size for the model was found experimentally to be 1. As batch normalization is known to result in degraded performance for small batch sizes \citep{Wu}, we therefore investigate possible improvements.

\paragraph{Number of parameters}

While the PointNet encoder used in \cite{Liu_2019_ICCV} enables a high degree of weight sharing, the fully-connected decoders use dense fully-connected layers. This design choice results in a high number of parameters (183.6M), which, combined with the limited data augmentation and absence of regularization, promotes overfitting.

\paragraph{Conclusion}
The reliance on subdivision and cropping, the high number of trainable parameters, as well as the training methodology utilised, make the method of \cite{Liu_2019_ICCV} only suitable for in-sample registration, and thus the fast inference time does not fully offset the offline training time. The presence of significant noise and artifacts on registrations of scans from the training set further limits the applicability of the model on its own.

\section{Description of the Method}
\label{sec:method}

We now introduce Shape My Face, our registration and morphable model pipeline. Our approach is based on the idea that registration can be cast as a translation problem, where one seeks to faithfully translate a latent geometric information (the surface) from an arbitrary input modality to a controlled template mesh. It is therefore natural to adopt an autoencoder architecture, with the advantages exposed in Section \ref{sec:comparison_3dfc}. We also wish to ensure our model is compact and performs reliably and satisfyingly on unseen data. The emphasis is, therefore, on robustness and applicability to real-world data, potentially on the edge.

\subsection{Preliminaries and Stochastic Training}

We choose the mean face of the LSFM model to be our template. We manually cropped the same facial region as the template of \cite{Liu_2019_ICCV} from a full-face combined LSFM and FaceWarehouse morphable model, and ensured a 1-to-1 correspondence between vertices. We choose LSFM since it is more representative of the mean human face than the BFM 2009 mean, and to facilitate the prototyping of a mouth model, as explained in Section \ref{sec:mouth_model}.

\begin{figure}[tbp]
    \centering
    \subfloat[Mean $\bm{\mu}$]{
        \includegraphics[width=26mm]{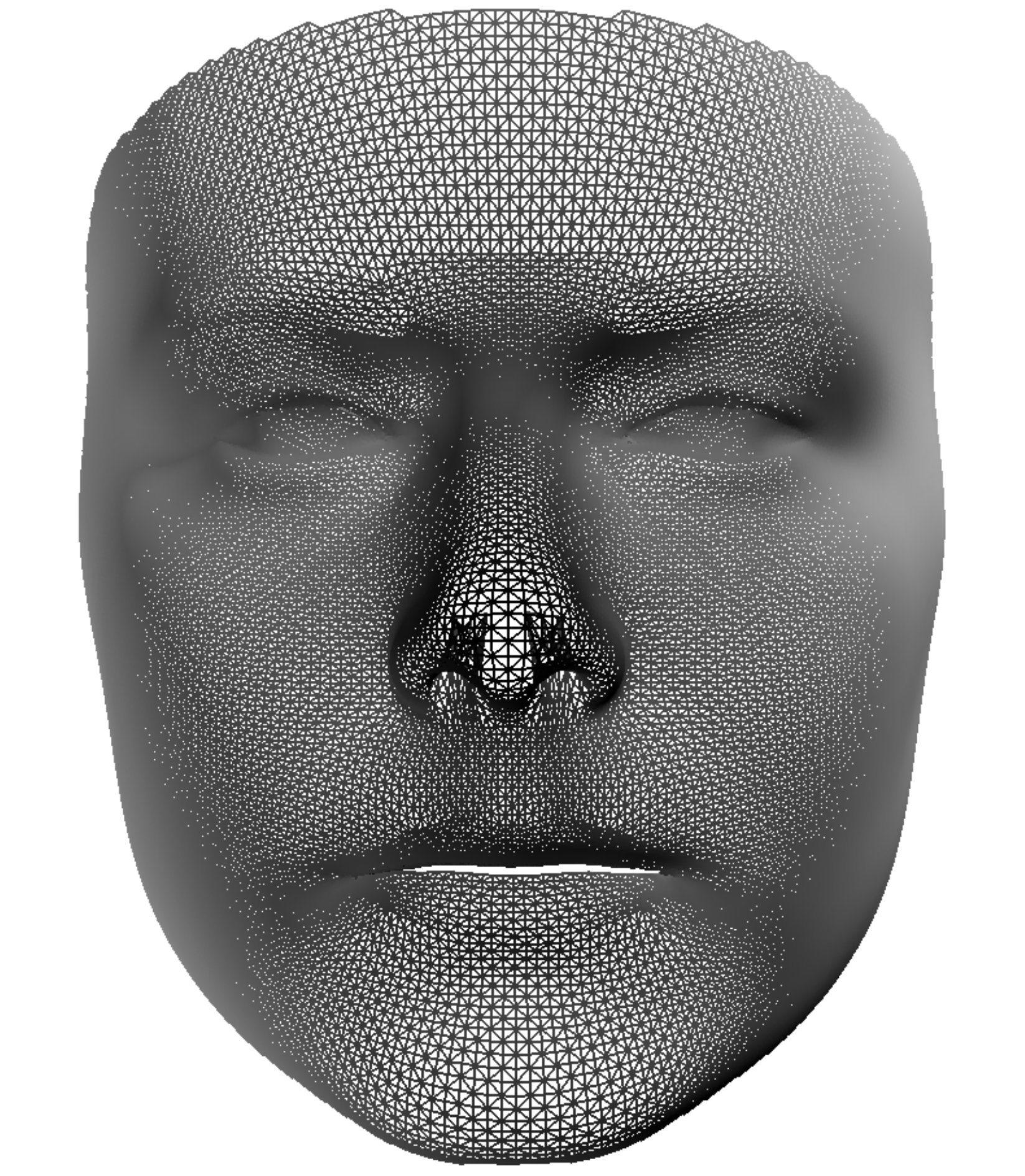}   
        \label{fig:lsfm_mean}
    }
    \subfloat[Boundary]{
        \includegraphics[width=26mm]{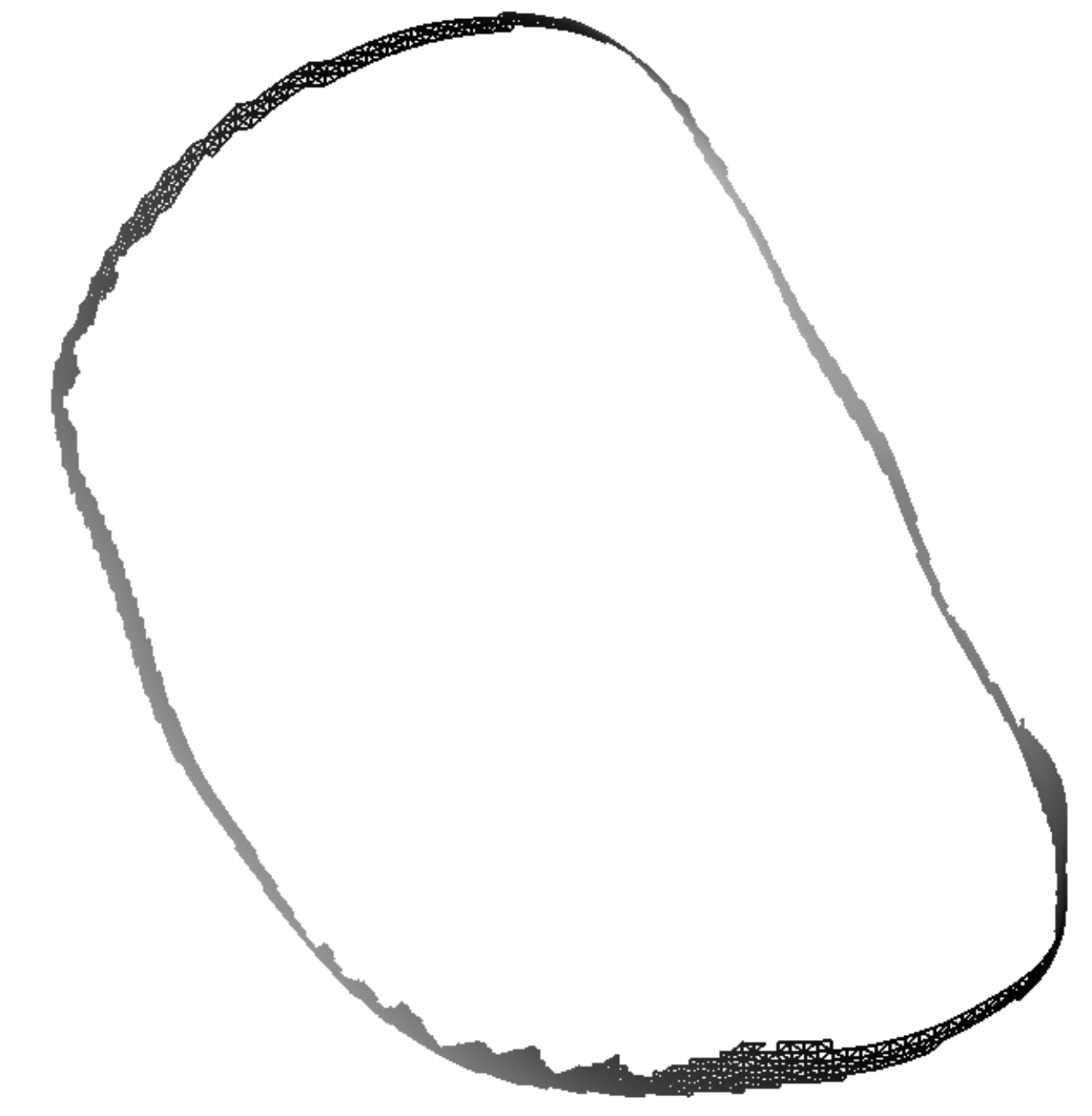}   
        \label{fig:boundary_crop}
    }
    \subfloat[Inner lips]{
        \includegraphics[width=26mm]{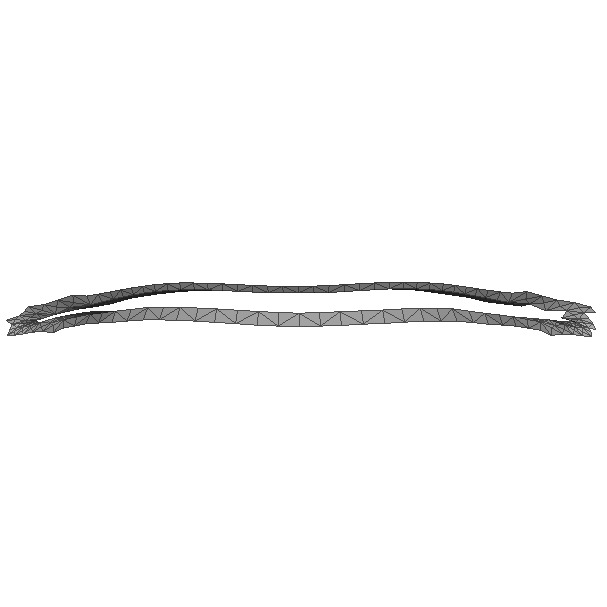}   
        \label{fig:inside_lips_crop}
    }
    \caption{\textbf{Parts of the face model:} We decode shapes by predicting new vertex positions for the mean face of the LSFM model \cite{Booth2016,Booth2018} (a). To avoid ragged boundaries, we encourage a small crop of the boundary (b) of the reconstructions to be close in position and curvature to that of the LSFM mean face. We propose a parameter-free approach for achieving high quality mouth reconstructions by reconstructing a crop of the mouth region on a small mouth-specific PCA model, and blending the reconstruction with the shapes predicted by the decoders using a smooth blending mask derived from the geodesic distance of the vertices in the template to a small crop of the lips (c).}
    \label{fig:face_model_parts}
\end{figure}

We adopt a formulation in terms of blendshapes and define the output of our network to be
\begin{equation}
    \mathbf{S} = \bm{\mu} + \bm{\Delta}\mathbf{S}_{id} + \bm{\Delta}\mathbf{S}_{exp}
\end{equation}
Where $\bm{\mu}$ is the template mean face shown in Figure \ref{fig:lsfm_mean}, and $\bm{\Delta}_{id}$ and $\bm{\Delta}_{exp}$ are identity and expression deformation fields, respectively, defined on the vertices of $\bm{\mu}$. We motivate this choice to encourage better disentanglement by modeling both identity and expression as additive deformations of a plausible mean human face.

\begin{figure*}[t!]
    \centering
    \includegraphics[width=174mm]{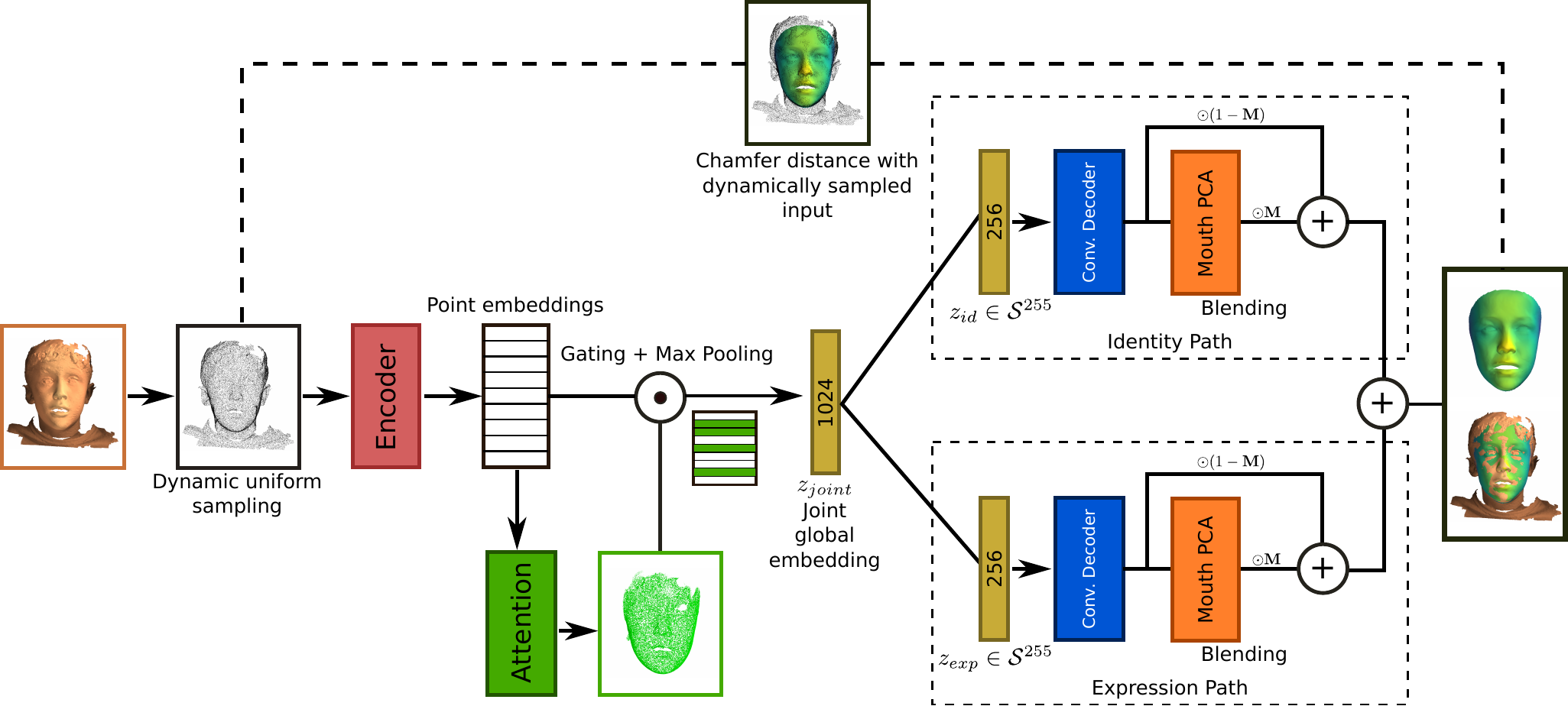}
    \caption{\textbf{Flow-chart representation of our approach:} We sample $2^{16}$ points uniformly at random on the surface of the scan to register. A modified PointNet encoder computes features and an attention score for each point, from which a global embedding $\mathbf{z}_{joint}$ is obtained. We produce two hyperpsherical embeddings $\mathbf{z}_{id}$ and $\mathbf{z}_{exp}$ from $\mathbf{z}_{joint}$, and apply \textit{mesh inception} decoders to output corresponding identity and expression blendshapes. To improve denoising, we smoothly blend the mouth region in a blendshape with its projection on a specialized PCA mouth model. During training (dotted lines), we measure the fit of the registration between the output of the network and the dynamically sampled input point cloud. This ensures vertices of the reconstruction can be matched to points anywhere on the surface of the scan, and not only to the vertices.}
    \label{fig:flowchart}
\end{figure*}

We follow an encoder-decoder architecture using a point cloud encoder and two symmetric non-linear decoders for the identity and expression blendshapes. As we will develop further, we propose a novel approach to avoid mouth artifacts by blending the non-linear blendshapes smoothly with linear blendshapes of the mouth region (defined based on the geodesic radius from the inside of the mouth). The flowchart of the method is presented in Figure \ref{fig:flowchart}.

\paragraph{Input shape representations} At inference time, our method only requires that we may randomly sample points on the surface of the scan. At training time, we optionally use the normal vectors at the sampled points (see Section \ref{sec:losses}). Therefore, any input modality that satisfies these requirements is suitable for training and inference.

In this work, we deal with training datasets of raw scans represented as meshes rigidly aligned (with scaling) with the template. Contrary to \cite{Liu_2019_ICCV}, we do not apply any further processing on the 3D scans after rigid alignment. In particular, no surface subdivision and no offline sampling for data augmentation are done. We will also demonstrate inference on raw point clouds directly (Section \ref{sec:in_the_wild}).

We dynamically sample $N_s = 2^{16} = 65536$ points uniformly at random on the surface of the input mesh using a triangle weighting scheme. Furthermore, we use the sampled point cloud as ground truth in the Chamfer loss. This ensures the vertices of the registration can be matched to points anywhere on the input surface, including inside triangles where the true projection of the vertices of the registration are more likely to lie.

We denote the triangulated raw input scan by the tuple $(\mathbf{S}_{in}, \mathbf{T}_{in})$, where $\mathbf{S}_{in}$ is the set of vertices of the mesh, and $\mathbf{T}_{in}$ the triangles. We write $\mathbf{P}_{in}$ the point cloud dynamically sampled on the surface of $(\mathbf{S}_{in}, \mathbf{T}_{in})$, and $\mathbf{N}_{in}$ the associated sampled point normals.

We use both synthetic and real scans in training. The training procedure is detailed in Section \ref{sec:implem_details}.

\subsection{Encoder and attention}

In PointNet \citep{Qi2017a}, the authors introduce one of the first CNN architectures for point clouds. A PointNet layer consists of a $1 \times 1$ convolution followed by batch normalization and a ReLU activation, as shown in Figure \ref{fig:pointnet_block}. PointNet showed high performance on classification and segmentation tasks using moderately dense point clouds as input (2048 points for the ModelNet40 meshes). In this work, we sample $2^{16} = 65536$ points from the input scans, which limits the batch sizes that can be accommodated with a single GPU implementation. As mentioned in Section \ref{par:soa_encoder}, batch normalization is known to be ineffective for small batches \citep{Wu}, as the sample estimators of the feature mean and standard deviation become noisy. We therefore propose modified PointNet layers with group normalization \citep{Wu}, that we choose to apply after the ReLU non-linearity. Our modified PointNet layers are illustrated in Figure \ref{fig:mod_pointnet_block}. We denote by $\mathrm{PN}(f_{in}, f_{out}, g)$ the block consisting of a $1 \times 1$ convolution with $f_{in}$ input features and $f_{out}$ output features, followed by one ReLU activation, and group normalization with group size $g$. The sequence of point convolutional layers in our encoder can thus be written $E(\mathbf{\cdot}) = \mathrm{PN}(3, 64, 32) \rightarrow \mathrm{PN}(64, 64, 32) \rightarrow \mathrm{PN}(64, 64, 32)$ $\rightarrow \mathrm{PN}(64, 128, 32) \rightarrow \mathrm{PN}(128, 1024, 32)$.

\paragraph{Visual attention}
\label{sec:attention}

To improve the robustness of our method to noise and variations in the physical extent of the scans, we introduce a novel visual attention mechanism implemented as a binary-classification PointNet sub-network applied to the features of the last PointNet layer and before the max-pooling operation. This can be seen as a form of region-proposal \citep{He2017} or segmentation sub-network followed by a gating mechanism. We use our modified PointNet layers and obtain the following sequence of operations 
$\mathrm{PN}(1024, 128, 4) \rightarrow \mathrm{PN}(128, 32, 4) \rightarrow \text{Conv} 1\times1 (32, 1)$. We use a smaller group size of 4 for group normalization to discourage excessive correlation in the features. The logits obtained as output of the attention sub-network are converted to a smooth mask by applying the sigmoid function and used as gating values to the max pooling operation - controlling which points are used to build the global latent representation $\mathbf{z}_{joint} \in \mathbb{R}^{1024}$ for the scan.

\paragraph{Hyperspherical embeddings}
Two dense layers predict separate identity and expression embeddings from $\mathbf{z}_{joint}$. \newline We choose $\mathbf{z}_{id}, \mathbf{z}_{exp} \in \mathbb{R}^{256}$. Contrary to \cite{Liu_2019_ICCV}, the mapping is non-linear: we normalize the identity and expression vectors, such that they lie on the hypersphere $\mathcal{S}^{255}$. Hyperspherical embeddings have been successful in image-based face recognition \cite{Wang2018_cosface,Deng2019} and shown to improve clusterability \citep{Aytekin2018}. Additionnally, we found the normalization to improve numerical stability during training.

The full encoder can be summarized as follows:

\begin{align}
& \mathbf{\tilde{Z}} = E(\mathbf{P}_{in}) \\
& \mathbf{A} = \text{Attention}(\mathbf{\tilde{Z}}) \\
& \mathbf{z}_{joint} = \text{MaxPool}(\sigma(\mathbf{A}) \odot \mathbf{\tilde{Z}}) \\
& \mathbf{z}_{id} = \text{Normalize}(\text{FC}_{1024, 256}(\mathbf{z}_{Joint}))\\
& \mathbf{z}_{exp} = \text{Normalize}(\text{FC}_{1024, 256}(\mathbf{z}_{Joint}))
\end{align} where $\odot$ denotes the element-wise (Hadamard) product and $\sigma(x) = \frac{1}{1 + e^{-x}}$ is the sigmoid function applied element-wise.

\subsection{Mesh convolution decoders}
\label{sec:conv_decoders}

\begin{figure}[t]
    \centering
    \includegraphics[width=84mm]{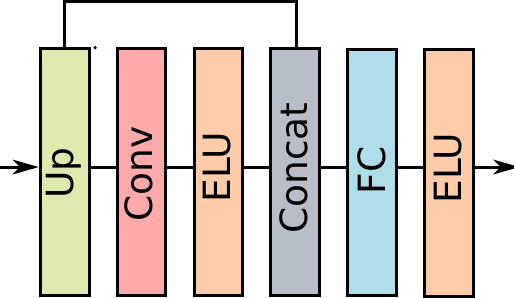}
    \caption{\textbf{One Mesh Inception block:} Our mesh convolution block offers two paths for the information to flow from one resolution to the next. We concatenate the activated feature map of the current convolution layer with the upsampled feature map of the previous layer. The features are combined in a learnable way by a fully connected layer followed by another ELU activation.}
    \label{fig:decoder_block}
\end{figure}

As developed in Section \ref{par:soa_decoder}, the fully-connected decoders used in \cite{Liu_2019_ICCV} suffer from two main challenges. First, they employ a high number of parameters, which promotes overfitting. Second, they do not leverage the known template geometry, and therefore require heavy tuning and regularization to produce sound shapes without abrupt changes in curvature and triangle geometry.%

We propose non-linear decoders based on mesh convolutions. Our method is applicable to any intrinsic convolution operator on meshes. In this particular implementation, we use the SpiralNet++ operator. Denoting $\vx^{(k)}_i$ the features of vertex $i$ at layer $k$, we have:
\begin{equation}
    \label{eq:spiralnetpp}
    \vx^{(k)}_i = \gamma^{(k)} \left( \concat_{j \in S(i, M)} \vx_j^{(k-1)} \right)
\end{equation} with $\gamma^{(k)}$ an MLP, $\concat$ the concatenation, and $S(i, M)$ the spiral sequence of neighbors of $i$ of length (\ie kernel size) $M$.

We observed training was difficult with the vanilla operators. As some operators such as SpiralNet++ and ChebNet already have a form of residual connections built-in (the independent weights given to the center vertex of the neighborhood), vanilla residuals or the recently-proposed affine skip connections \citep{Gong2020} would be redundant. We instead propose a block reminiscent of the inception block in images \citep{Szegedy2014} that can benefit any graph convolution operator. We concatenate the output of the previous upsampled feature map with the output of the convolution after an ELU non-linearity \citep{Clevert2016}. The concatenated feature maps are combined and transformed to the desired output dimension using an FC layer followed by another ELU non-linearity, as illustrated in Figure \ref{fig:decoder_block}.

We found this technique to drastically improve convergence and details in the reconstructed shapes. The technique is comparable to GraphSAGE \citep{Hamilton2017}, using graph convolutions followed by ELU as the $\textsc{Aggregate}_k$ function in \citep[Algorithm 1]{Hamilton2017}, and ELU non-linearities. We refer to our block as \textit{Mesh Inception}.

For upsampling, we follow the approach of \cite{ranjan_generating_2018}. We decimate the template four times using the Qslim method \citep{garland_surface_1997} and build sparse upsampling matrices using barycentric coordinates. We set the kernel sizes of our convolution layers to 32, 16, 8, and 4, starting from the coarsest decimation of the template.

\subsection{Mouth model and blending}
\label{sec:mouth_model}

\begin{figure}[b]
    \centering
    \subfloat[Raw]{
        \includegraphics[width=.31\linewidth]{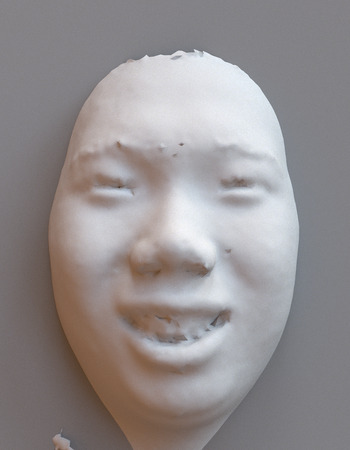}
        \label{fig:lap_pca_gt}
    }
    \subfloat[PCA]{
        \includegraphics[width=.31\linewidth]{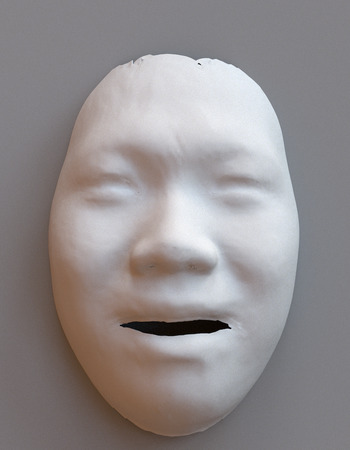}
        \label{fig:lap_pca_pca}
    }
    \subfloat[Laplacian]{
        \includegraphics[width=.31\linewidth]{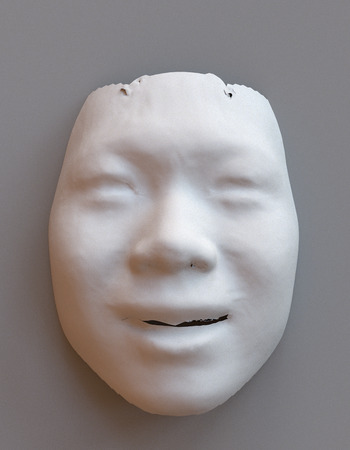}
        \label{fig:lap_pca_lap}
    }
    \caption{\textbf{Laplacian loss and statistical mouth model:} Laplacian loss (c) limits the expressivity of the scans but does not eliminate the artifacts completely (sample from the BU-3DFE dataset).}
    \label{fig:laplacian_issue}
\end{figure}

\begin{figure}[t]
    \centering
    \subfloat[Geodesic dist.]{
        \includegraphics[width=.31\linewidth]{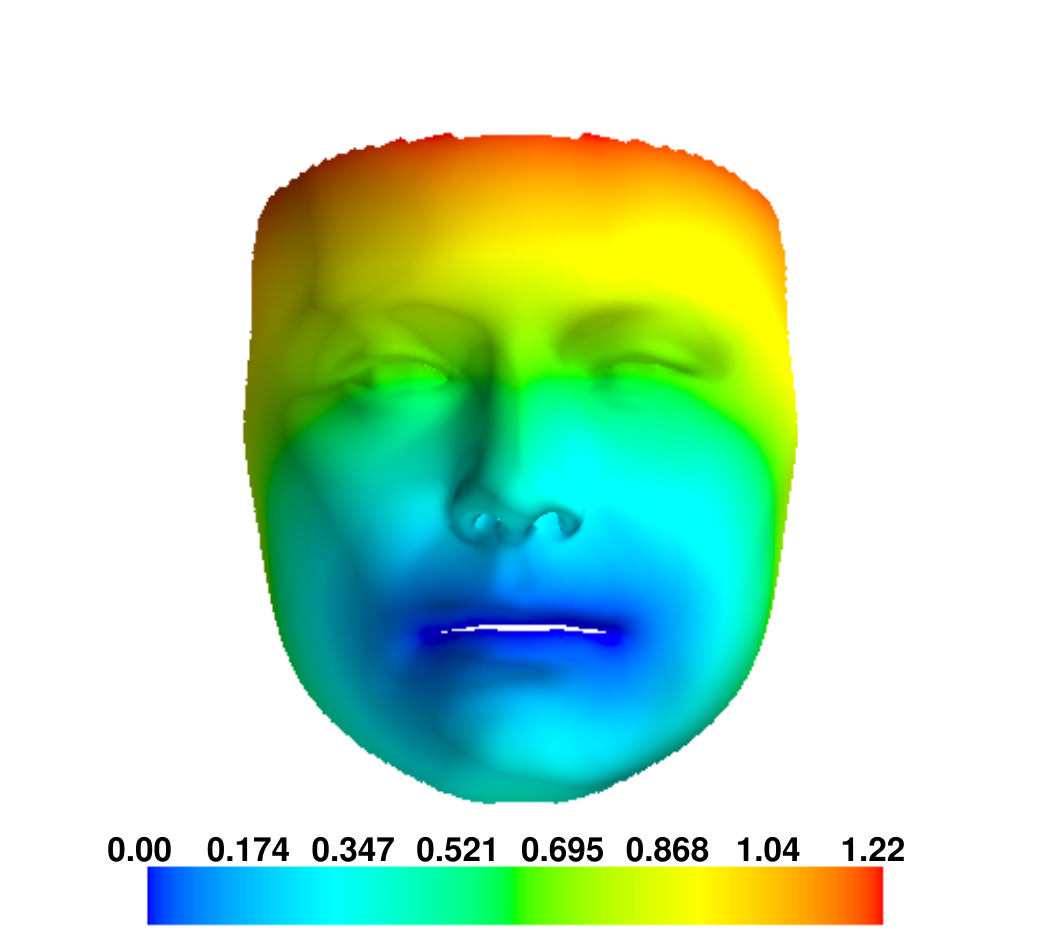}
        \label{fig:geodesic_lsfm}
    }
    \subfloat[Mask]{
        \includegraphics[width=.31\linewidth]{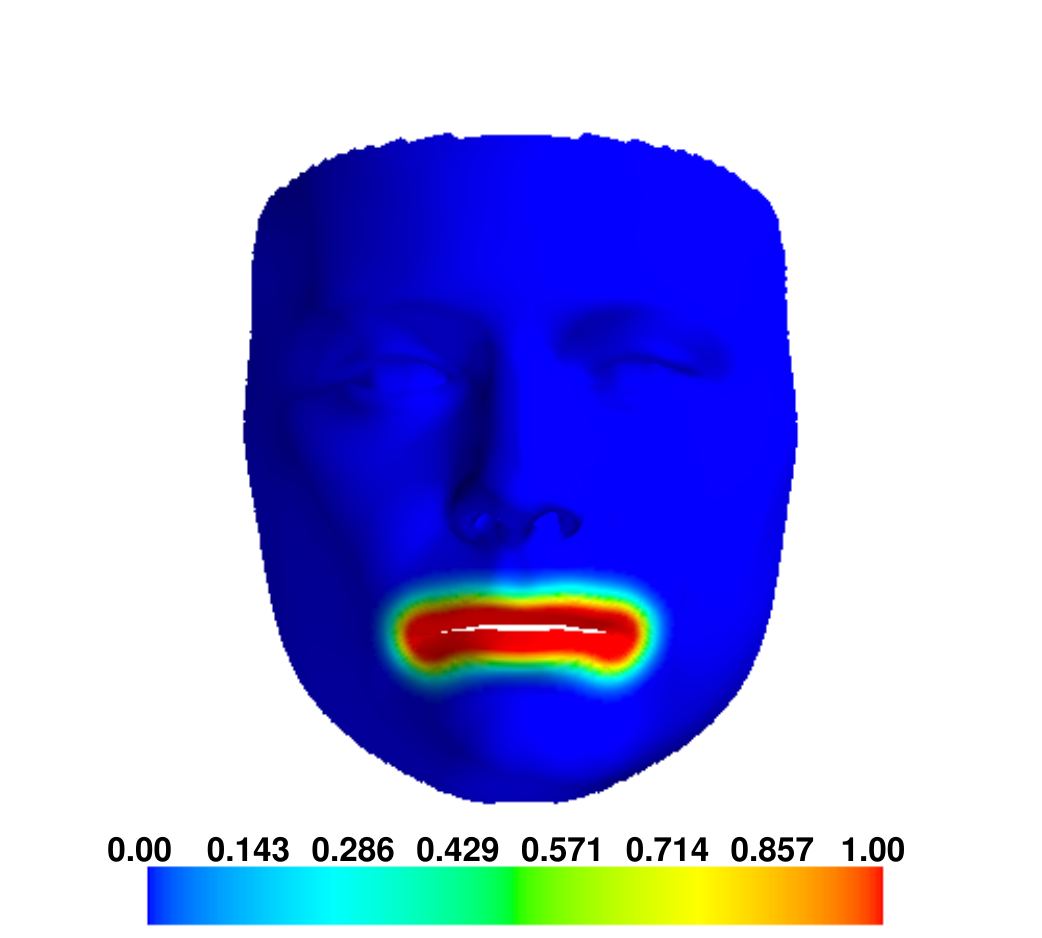}
        \label{fig:bm_on_mean}
    }
    \subfloat[Mouth]{
        \includegraphics[width=.31\linewidth]{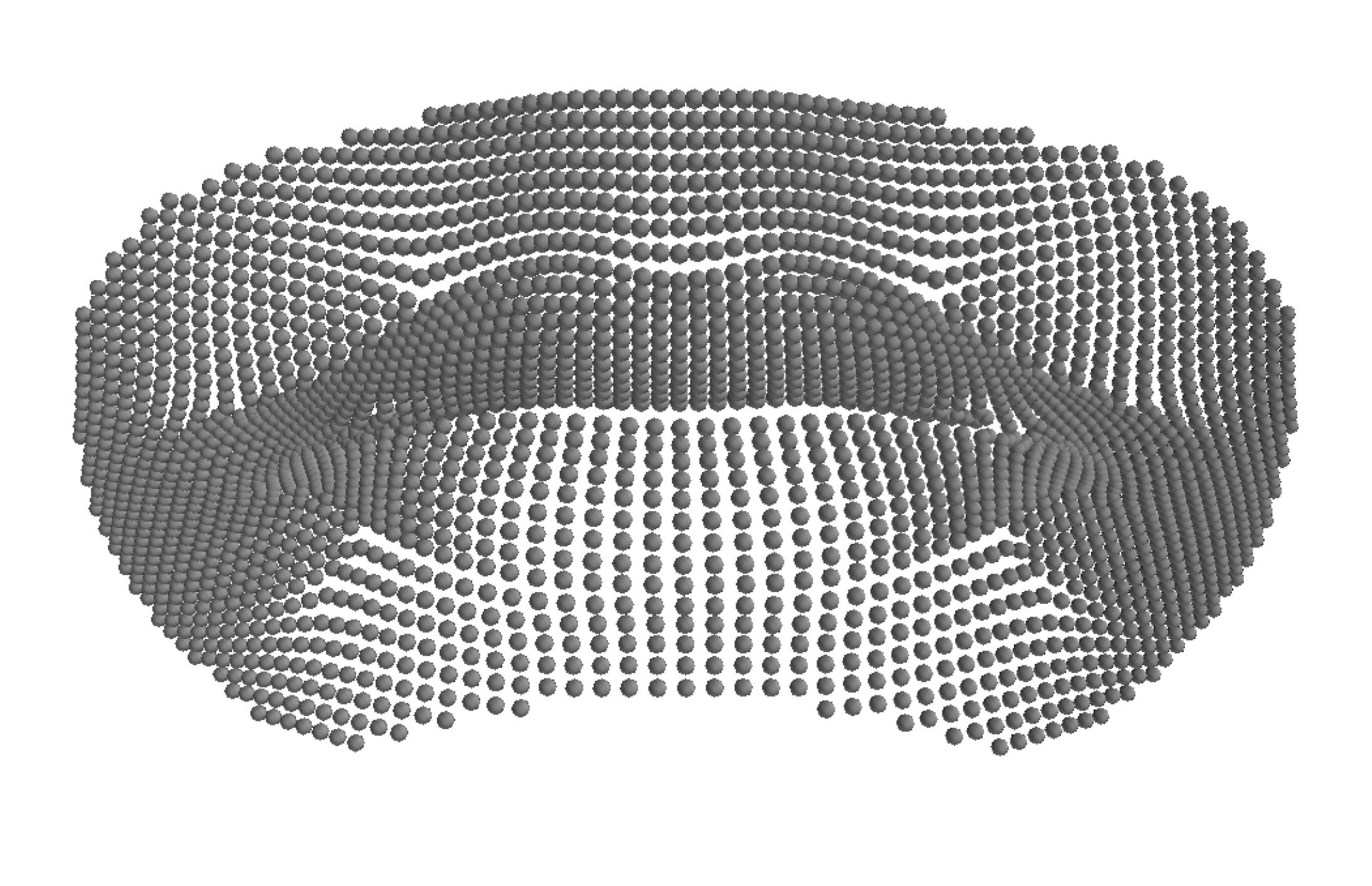}   
        \label{fig:mouth_region}
    }
    \caption{\textbf{Mouth region and blending:} From the small crop of the lips of Figure \ref{fig:inside_lips_crop}, we compute the geodesic distance of all vertices of the template to the vertices of the crop $S_{inner}$ (a). We define the mouth region as the vertices within a chosen geodesic radius of $S_{inner}$ (c). We define the blending mask as a function of the geodesic distance, shown as a heatmap in (b).}
    \label{fig:mask_on_mean}
\end{figure}

Though the raw scans are rigidly aligned with the template on 5 facial landmarks that include the two corners of the mouth \citep{Liu_2019_ICCV}, the mouth expressions introduce a high level of variability in the position of the lips. Additionally, numerous expressive scans include points captured from the tongue, the teeth, or the inside of the mouth. This noise and variability in the dataset makes finding good correspondences for the mouth region difficult and leads to severe artifacting in the form of vertices from the lips being pulled towards the center of the mouth. In \cite{Liu_2019_ICCV}, the authors advocate for the use of Laplacian regularization to prevent extreme deformations by penalizing the average mean curvature over a pre-defined mouth region, controlled by a weight $\lambda_{Lap}$. While this shows some success, we experimentally observed that, for small to moderate values of $\lambda_{Lap}$, artifacts remained. As shown in Figure \ref{fig:laplacian_issue}, while artifacts were reduced for large values of $\lambda_{Lap}$, so was the range of expressions.

In this work, we introduce a new approach based on blending a specialized linear morphable model with the non-linear face model. We first isolate a small set of vertices, $S_{inner}$, from the innermost part of the lips of the cropped LSFM mean face, as shown in Figure \ref{fig:inside_lips_crop}. We then compute the geodesic distance from $S_{inner}$ to all vertices of the template using the heat method with intrinsic Delaunay triangulation \citep{Crane:2017:HMD}, which is visualised in Figure \ref{fig:geodesic_lsfm}. We redefine the mouth region to be the set of vertices $S_{mouth}$ within a given geodesic radius $d$ from $S_{inner}$. By visual inspection, we choose $d = 0.15$. The resulting mouth region is shown as a point cloud in Figure \ref{fig:mouth_region}.

To obtain a linear morphable model of this mouth region, we cropped the PCA components of the full face LSFM and FaceWarehouse model whose mean we used to obtain our face template. We keep only a subset, $\mathbf{W}_{id}$, of 30 identity components (from LSFM) and a subset, $\mathbf{W}_{exp}$, of 20 expression components (from FaceWarehouse). While it is well known that computing PCA on the cropped region of the raw data leads to more compact bases \citep{Blanz1999, Tena2011}, re-using the LSFM and FaceWarehouse bases enabled efficient prototyping. There is a trade-off between representation power and clean noise-free reconstructions: the model needs to be powerful enough to represent a wide range of expressions but restrictive enough that it does not represent the unnatural artifacts.

We project the mouth region of the blendshapes on the PCA mouth model during training and blend them smoothly with their respective source blendshapes, \ie, we project the mouth region of $\mathbf{S}_{id}$ on $\mathbf{W}_{id}$ and the mouth region of $\mathbf{S}_{exp}$ on $\mathbf{W}_{exp}$. Blending should be seamless, but - equally importantly - should also remove artifacts. We propose to define a blending mask intrinsically as a Gaussian kernel of the geodesic distance from $S_{inner}$:
\begin{equation}
    \label{eq:bm}
    b(r, c, \tau) = \begin{cases}
        \exp^{(-(r - c)^2 / \tau^2)}, & \text{if} \; r \geq c\\
        1, & \text{otherwise.}
    \end{cases}
\end{equation}
Where $c$ and $\tau$ control the geodesic radius for which the PCA model is given a weight of $1$, and the rate of decay, respectively. Compared to exponential decay, the squared ratio $((r-c) / \tau)^2$ allows us to favor more strongly the PCA model when $r-c \leq \tau$ and decay faster for $r-c > \tau$. Enforcing weights of $1$ within a certain radius helps ensure the artifacts are entirely removed.

The mouth region of the blendshape $\mathbf{S}_{(.)}$ is redefined as:
\begin{align}
    \mathbf{S}_{(.), mouth} = \mathbf{M} \odot \left( \mathbf{P}_{(.)} \mathbf{Y}_{(.), mouth} \right) + (\mathbf{1} - \mathbf{M}) \odot \mathbf{Y}_{(.), mouth}
\end{align}
With $\mathbf{M}$ the blending mask, $\mathbf{Y}_{(.), mouth}$ the mouth region in the output of the mesh convolutions, and $\mathbf{P}_{(.)}$ the projection matrix on the matching PCA basis.

\begin{figure}[t]
    \centering
    \includegraphics[width=84mm]{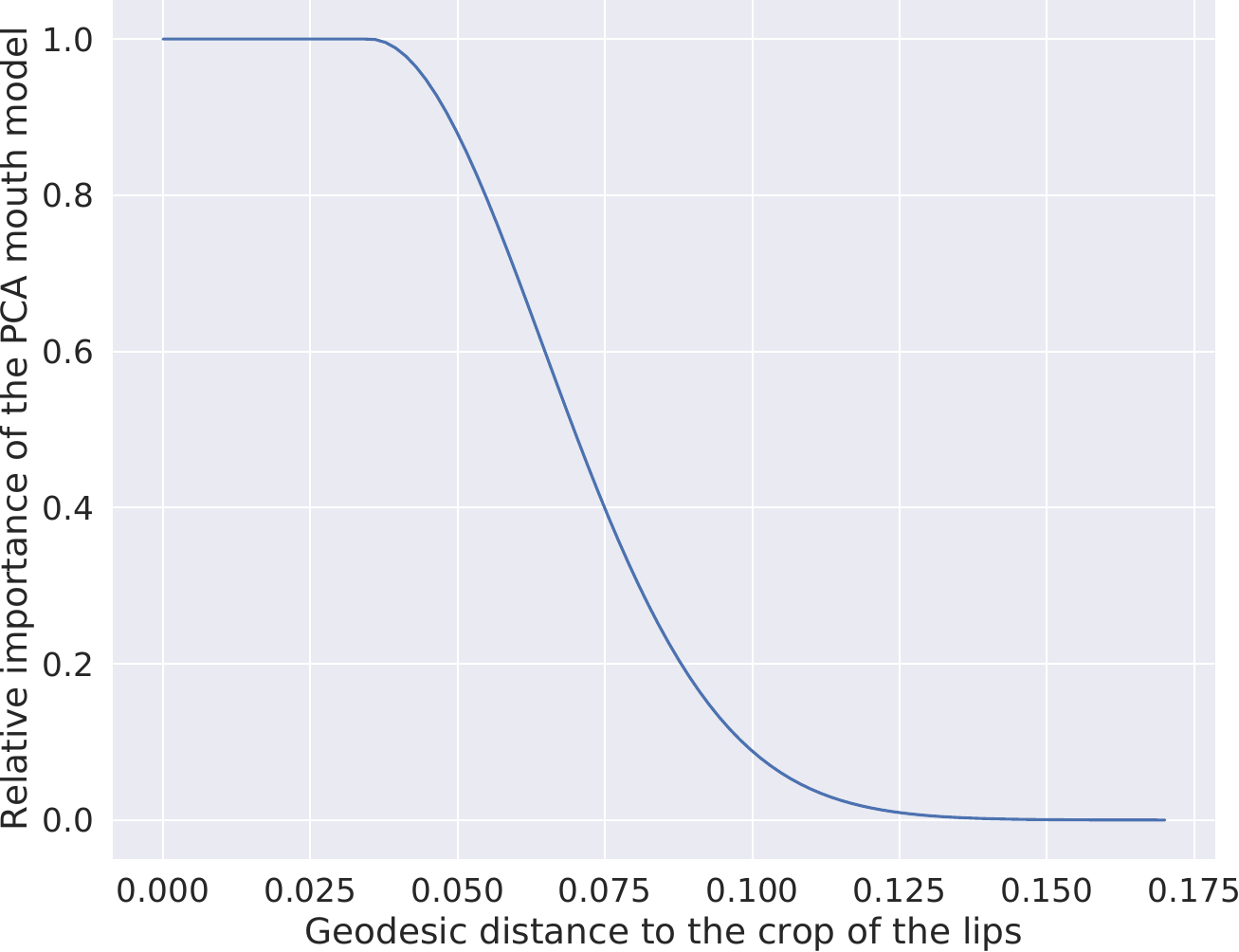}
    \caption{\textbf{Blending function:} Plot of $b(r, c, \tau)$ for the values of $c$ and $\tau$ used in this paper. We enforce a weight of $1$ on the PCA model for the vertices within geodesic distance $c$ of $S_{inner}$. We choose the rate of decay $\tau$ to enforce a weight close to $0$ on the PCA model at the edges of the mouth region.}
    \label{fig:blending_mask_profile}
\end{figure}

We choose $c$ experimentally. As $c$ varies, we adapt $\tau$ to ensure the contribution of the PCA model to the reconstruction of the mouth region is low at the edges of the crop, and avoid seams. For a desired weight $\epsilon << 1$ at distance $r$ and given $c$, we compute 
\begin{equation}
    \tau(r, c, \epsilon) = \frac{r - c}{\sqrt{-\log(\epsilon)}}.
\end{equation}
In practice, we choose $c = 3.5e-2$ and $\epsilon = 5e-4$. We plot the resulting $b(\cdot, c, \tau)$ in Figure \ref{fig:blending_mask_profile}.

In this work, we fixed $c$ and $\tau$ for all shapes, on the assumption that the geodesic distance from the inner lips does not vary excessively in the dataset. However, it is perfectly reasonable to consider both parameters to be trainable, or to predict them from the latent vectors $\mathbf{z}_{joint}$, $\mathbf{z}_{id}$ or $\mathbf{z}_{exp}$ to obtain shape or blendshape-specific blending masks.

\subsection{Losses}
\label{sec:losses}

For synthetic scans, we define 
\begin{equation}
    L_{vertex}(\mathbf{S}, \mathbf{S}_{in}) = ||\mathbf{S}_{in} - \mathbf{S}||_1.
\end{equation}
For real scans, we use the Chamfer distance
\begin{equation}
    L_{vertex}(\mathbf{S}, \mathbf{P}_{in}) = \sum_{p \in \mathbf{S}} \min_{q \in \mathbf{P}_{in}} || p - q ||_2^2 + \sum_{q \in \mathbf{P}_{in}} \min_{p \in \mathbf{S}} || p - q ||_2^2.
    \label{eq:chamfer_loss}
\end{equation}
As in \cite{Liu_2019_ICCV}, we discard $q$ from the error if 
\begin{equation}
\min_{q \in \mathbf{P}_{in}} || p - q ||_2^2 > \sigma \quad \text{or} \quad \min_{p \in \mathbf{S}} || p - q ||_2^2 > \sigma.
\end{equation} We set $\sigma = 5e-4$.

For synthetic scans, we let $\mathbf{n}(p)$ be the normal vector at vertex $p \in \mathbf{S}$, and $\mathbf{n}_{in}(p)$  be the normal in the synthetic scan, and define the normal loss as:
\begin{equation}
    L_{normal} = \frac{1}{N} \sum_{p \in \mathbf{S}} (1 - <\mathbf{n}(p), \mathbf{n}_{in}(p)>).
\end{equation}
For real scans, we use
\begin{equation}
    L_{normal} = \frac{1}{N} \sum_{p \in \mathbf{S}} (1 - <\mathbf{n}(p), \mathbf{N}_{in}(q)>),
\end{equation} where $q$ is the closest point in $\mathbf{P}_{in}$ found by Eq. \ref{eq:chamfer_loss}. In both cases, we set a weight of $\lambda_{norm} = 1e-4$.

Mesh convolutions are aware of the template connectivity and geometry, and do not require as much regularization as MLPs, we therefore use a weight of $\lambda_{edge} = 5e-5$ for the edge-loss, whose formulation is identical to \cite{Liu_2019_ICCV}.

To regularize the attention mechanism during the initial supervised training steps, we assume all points sampled from the synthetic faces are equally fully important and none should be removed. We encourage the attention mask for the points sampled from synthetic scans to be $1$ everywhere, using the binary cross entropy loss with a weight $\lambda_{att} = 1e-4$.

Finally, we enforce both an edge loss and $\ell_1$ loss regularization between the reconstruction and the template in a small crop of the boundary, shown in Figure \ref{fig:boundary_crop}, to eliminate tearing artifacts. We let $\lambda_{bnd} = 1e-3$.

\subsection{Training, models, and implementation details}
\label{sec:implem_details}

\paragraph{Training data}

As previously exposed, we use the same raw aligned data as the baseline model of \cite{Liu_2019_ICCV}, but do not apply any further pre-processing, including data augmentation. To keep the ratio of identity and expression scans identical, we simply sample from the same scan as many times as required in a given training epoch.

In addition to the seven datasets of Table \ref{tab:databases}, we further add two large-scale databases of 3D human facial scans. The MeIn3D \citep{Booth2017,Booth2018,bouritsas2019neural} database contains 9647 neutral face scans of people of diverse age and ethic background. We also select 17750 scans from the 4DFAB \citep{Cheng2018} database. 4DFAB contains neutral and expressive scans of 180 subjects captured in 4 sessions spanning a period of 5 years. Each session comprises up to 7 tasks, consisting of either utterances, voluntary, or spontaneous expressions. %

\begin{table}[t]
\scriptsize
\centering
\caption{\textbf{A very large scale morphable model:} Summary of the additional databases used to train SMF+. }
\resizebox{\columnwidth}{!}{
\begin{tabular}{l |c| c  c | c  c }
\toprule
Database &\#Subj. & \#Neu. &\#Sample & \#Exp. & \#Sample\\ %
\hline\hline
MeIn3D~\cite{Booth2017} &    $9{,}647$  &  $9{,}647$   & $9{,}647$ & $0$      & $0$\\
4DFAB~\cite{Cheng2018} &    $180$  &  $6{,}449$  & $6{,}449$ & $11{,}301$       & $11{,}301$\\
\midrule
Real Data (additional) & $9{,}727$ & $16{,}096$ & $16{,}096$ & $11{,}301$ & $11{,}301$\\
\midrule
Real Data (total) & $11{,}379$   & $21{,}320$ & $33{,}358$ & $18{,}888$      & $20{,}706$   \\      
\hline        
Synthetic Data (total) &$1{,}500$ & $1{,}500$ &$15{,}000$&$9{,}000$&$9{,}000$ \\
\bottomrule
\end{tabular}
}
\label{tab:databases_plus}

\end{table}

For a given subject in the 4DFAB database, we select the first frame of all sequences in the first two tasks as neutral scans. We select the middle frame of every sequence of the first two tasks as expressive scans for the six basic expressions (happy, sad, surprised, angry, disgust, and fear) and utterances. For tasks 3, 4, and 5, we select the frames at $1/3$ and $2/3$ of the sequence. For task 6, we select the frames at $1/3$ and $2/3$ of the sequence for the first two sessions, and the middle frame otherwise. We pick the middle frame for all other sequences.

In this work, we evaluate two models. We call SMF our model trained on the same dataset as the baseline. Our model trained with the addition of the MeIn3D and 4DFAB datasets is denoted by SMF+. The breakdown of the dataset for SMF+ is presented in Table \ref{tab:databases_plus}. %

\paragraph{Training procedure}

The BFM 2009 model was trained on a sample size of 200 subjects, and offers a limited representation of the diversity of human facial anatomy. We found the synthetic data to hinder the performance of the model, and to limit the realistic nature of the reconstructions. Mesh convolution operators learn to represent signals on the desired template and can readily exploit its connectivity and learn local geometric properties, we therefore drastically reduce the reliance on synthetic data to only the very first stages of training to condition the attention mechanism.

We first train the encoder and the identity decoder on synthetic data only for 5 epochs; and then on real neutral scans only for a further 10 epochs. We repeat this procedure for the expression decoder by freezing the identity decoder and the identity branch of the encoder, using only expressive scans. We then train both decoders jointly and the encoder for 10 epochs on the entire set of real scans. Finally, we change the batch size to 1 and refine the complete model for 15 epochs on the entire set of real scans.

We set the initial batch size to 2 and 8 for SMF and SMF+, respectively. We train the models with the Adam optimizer \citep{Kingma2014}, with a learning rate of $1e-4$, and automatically decay the learning rate by a factor of $0.5$ every 5 epochs. No additional regularization is used.

\renewcommand\tabularxcolumn[1]{m{#1}}
\newcolumntype{Y}{>{\centering\arraybackslash}X}

\begin{table}[t]
    \caption{\textbf{Summary:} Side by side comparison of SMF and the baseline of \cite{Liu_2019_ICCV}}
    \small
    \centering
    \begin{tabularx}{\linewidth}{ @{}c|Y|Y@{} }
                 & \textbf{Baseline}         &  \textbf{SMF} \\
         \toprule
        Encoder  & Vanilla PointNet &  Modified PointNet\\
        $\mathbf{z}_{joint}$ space & $\mathbb{R}^{1024}$ & $\mathbb{R}^{1024}$\\
        $\mathbf{z}_{id}$ space & $\mathbb{R}^{512}$ & $\mathcal{S}^{255} \subset \mathbb{R}^{256}$\\
        $\mathbf{z}_{exp}$ space & $\mathbb{R}^{512}$ & $\mathcal{S}^{255} \subset \mathbb{R}^{256}$\\
        \# input points & 29495 & 65536\\
        \midrule
        Template & BFM 2009 & LSFM\\
        Decoders & 2-layer MLPs & Mesh inception\\
        \midrule
        Preprocessing & Cropping, Subdivision, Data augmentation & None\\
        \midrule
        Input & Pre-computed & Stochastic\\
        Ground truth & Subdivided mesh & Stochastic\\
        \midrule
        Losses & $\ell_1$/ Chamfer, normal, edge, Laplacian & $\ell_1$/ Chamfer, normal, edge, boundary, attention\\
        \midrule
        Additional features & None & Visual attention\\
        \midrule
        Trained model file size & 701MB (\texttt{float32}) & 179MB (\texttt{float32})\\
        \# Trainable params. & 183.6 millions (100\%) & 15.5 millions (8.8\%)\\
        \bottomrule
    \end{tabularx}
    \label{tab:side_by_side}
\end{table}

\paragraph{Software implementation and hardware}

Our model is implemented with Pytorch. We use the CGAL library for the computation of the geodesic distance using the heat method \citep{cgal:cvf-hm3-20a}, implemented in C++ as a Pytorch extension. We render figures using the Mitsuba 2 renderer \citep{NimierDavidVicini2019Mitsuba2}.

All models were trained on a single Nvidia TITAN RTX, in a desktop workstation with an AMD Threadripper 2950X CPU and 128GB of DDR4 2133MHz memory.

\paragraph{Side by side comparison}

We summarize the differences between SMF and the baseline in Table \ref{tab:side_by_side}.

\begin{table*}[t]
    \renewcommand\arraystretch{0.96}
    \centering
    \caption{\textbf{Semantic landmarks error on BU-3DFE:} Comparison of the mean and standard deviation for semantic landmark error ($mm$) for BU-3DFE using the \textit{BU-3DFE $83$} facial landmark set. Landmark regions are as described in \cite{Salazar2014}. \textit{L} and \textit{R} are shorthand for \textit{Left} and \textit{Right} respectively. \textit{Avg Face} is the average for all inner face landmarks, and therefore excludes \textit{Chin}, \textit{L Face}, and \textit{R Face}. Baseline is \cite{Liu_2019_ICCV}, GMCO is \cite{Bolkart2015}, FAEIFC is \cite{Salazar2014}, and GPMM is \cite{8373814}.}
    \resizebox{\linewidth}{!}{
    \begin{tabular}{l|ccccccccc}
\toprule
Region      & NICP              & GMCO             & FAEIFC           & GPMM             & Baseline \textbf{In} & Baseline \textbf{Out} & SMF \textbf{In}    & SMF \textbf{Out}   & SMF+ \textbf{In}\\
\hline\hline
L Eyebrow   & $4.59 {\pm} 1.34$ & $6.28{\pm}3.30$  & $4.69{\pm}4.64$  & $6.25{\pm}2.58$  & $7.98 {\pm} 2.77$    & $19.75 {\pm} 4.09$    & $7.20 {\pm} 2.15$  & $6.59 {\pm} 1.99$  & $7.23 {\pm} 2.24$\\
R Eyebrow   & $4.37 {\pm} 1.32$ & $6.75{\pm}3.51$  & $5.35{\pm}4.69$  & $4.57{\pm}3.03$  & $6.84 {\pm} 2.51$    & $20.86 {\pm} 4.25$    & $6.70 {\pm} 2.23$  & $6.48 {\pm} 2.19$  & $6.81 {\pm} 2.31$\\
L Eye       & $3.28 {\pm} 0.98$ & $3.25{\pm}1.84$  & $3.10{\pm}3.43$  & $2.00{\pm}1.32$  & $5.08 {\pm} 1.65$    & $11.75 {\pm} 1.66$    & $3.50 {\pm} 1.10$  & $3.40 {\pm} 1.09$  & $3.92 {\pm} 1.22$\\
R Eye       & $3.09 {\pm} 0.98$ & $3.81{\pm}2.06$  & $3.33{\pm}3.53$  & $2.88{\pm}1.29$  & $3.80 {\pm} 1.35$    & $10.27 {\pm} 1.59$    & $4.82 {\pm} 1.44$  & $5.09 {\pm} 1.45$  & $4.92 {\pm} 1.46$\\
Nose        & $3.74 {\pm} 0.87$ & $3.96{\pm}2.22$  & $3.94{\pm}2.58$  & $4.33{\pm}1.24$  & $4.90 {\pm} 1.24$    & $9.14 {\pm} 1.56$     & $4.62 {\pm} 1.20$  & $4.64 {\pm} 1.17$  & $4.48 {\pm} 1.19$\\
Mouth       & $4.82 {\pm} 2.33$ & $5.69{\pm}4.45$  & $3.66{\pm}3.13$  & $4.45{\pm}2.02$  & $5.32 {\pm} 2.28$    & $8.56 {\pm} 2.40$     & $6.25 {\pm} 2.39$  & $6.00 {\pm} 2.36$  & $6.33 {\pm} 2.39$\\
\hline
Chin        & $7.75 {\pm} 2.79$ & $7.22{\pm}4.73$  & $11.37{\pm}5.85$ & $7.47{\pm}3.01$  & $11.39 {\pm} 4.78$   & $25.69 {\pm} 6.93$    & $38.73 {\pm} 9.18$ & $38.08 {\pm} 8.47$ & $38.96 {\pm} 9.88$\\
L Face      & $8.14 {\pm} 2.34$ & $18.48{\pm}8.52$ & $12.52{\pm}6.04$ & $12.10{\pm}4.06$ & $15.63 {\pm} 6.09$   & $29.30 {\pm} 7.35$    & $14.08 {\pm} 3.64$ & $12.97 {\pm} 3.50$ & $14.31 {\pm} 3.79$\\
R Face      & $7.68 {\pm} 1.91$ & $17.36{\pm}9.17$ & $10.76{\pm}5.34$ & $13.17{\pm}4.54$ & $11.93 {\pm} 4.02$   & $27.38 {\pm} 5.11$    & $20.61 {\pm} 5.19$ & $19.59 {\pm} 5.23$ & $20.64 {\pm} 5.34$\\
\hline
Avg Face    & $4.12 {\pm} 0.83$ &   -              &   -           &    -             & $5.79 {\pm} 1.34$    & $13.40 {\pm} 1.89$    & $5.70 {\pm} 1.11$  & $5.52 {\pm} 1.09$  & $5.77 {\pm} 1.12$\\
Avg         & $4.91 {\pm} 0.80$ & $8.49{\pm}4.29$  & $8.09{\pm}5.75$  & $6.52{\pm}3.86$  & $7.37 {\pm} 1.57$    & $16.45 {\pm} 2.19$    & $9.07 {\pm} 1.28$  & $8.72 {\pm} 1.24$  & $9.15 {\pm} 1.36$\\
\bottomrule
    \end{tabular}
    }
    \label{tab:bu3d_landmark_correspondence}
\end{table*}

\begin{table*}[]
    \renewcommand\arraystretch{0.96}
    \centering
    \caption{\textbf{Semantic landmarks error on 3DMD:} Comparison of the mean and standard deviation for semantic landmark error ($mm$) for 3DMD using the \textit{ibug $68$} facial landmark set. \textit{L} and \textit{R} are shorthand for \textit{Left} and \textit{Right} respectively. \textit{Avg} is the average for all inner face landmarks.}
    
    \begin{tabular}{l|cccccc}
\toprule
Region & NICP & \cite{Liu_2019_ICCV} & SMF & SMF+ & SMF + NICP & SMF+ + NICP\\
\hline\hline
L Eyebrow & $5.94 {\pm} 2.16$ & $6.23 {\pm} 1.54$ & $5.57 {\pm} 1.66$ & $5.87 {\pm} 1.84$ & $5.54 {\pm} 1.70$ & $5.84 {\pm} 1.86$\\
R Eyebrow & $5.27 {\pm} 2.05$ & $6.14 {\pm} 1.74$ & $6.32 {\pm} 2.20$ & $6.70 {\pm} 2.18$ & $6.27 {\pm} 2.21$ & $6.68 {\pm} 2.19$\\
L Eye & $4.29 {\pm} 1.26$ & $3.83 {\pm} 1.17$ & $4.27 {\pm} 0.99$ & $4.75 {\pm} 0.99$ & $4.25 {\pm} 1.00$ & $4.79 {\pm} 0.99$\\
R Eye & $4.02 {\pm} 1.37$ & $3.79 {\pm} 1.27$ & $4.03 {\pm} 1.18$ & $4.08 {\pm} 1.15$ & $3.98 {\pm} 1.19$ & $4.01 {\pm} 1.16$\\
Nose & $4.56 {\pm} 0.96$ & $4.94 {\pm} 1.17$ & $5.30 {\pm} 0.85$ & $5.42 {\pm} 0.87$ & $5.22 {\pm} 0.83$ & $5.32 {\pm} 0.86$\\
Mouth & $3.96 {\pm} 1.70$ & $4.73 {\pm} 1.65$ & $6.36 {\pm} 1.16$ & $6.38 {\pm} 1.15$ & $6.31 {\pm} 1.16$ & $6.25 {\pm} 1.14$\\
\hline
Jaw & $24.58 {\pm} 4.69$ & $35.76 {\pm} 5.59$ & $24.91 {\pm} 5.67$ & $25.25 {\pm} 5.75$ & $24.87 {\pm} 5.73$ & $25.22 {\pm} 5.76$\\
\hline
Avg Face & $4.43 {\pm} 0.95$ & $4.84 {\pm} 1.05$ & $5.57 {\pm} 0.74$ & $5.73 {\pm} 0.76$ & $5.52 {\pm} 0.73$ & $5.65 {\pm} 0.74$\\
Avg & $9.47 {\pm} 1.59$ & $12.57 {\pm} 1.76$ & $10.41 {\pm} 1.75$ & $10.61 {\pm} 1.78$ & $10.36 {\pm} 1.76$ & $10.55 {\pm} 1.77$\\
\bottomrule
    \end{tabular}
    \label{tab:3dmd_landmark_correspondence}
\end{table*}

\section{Experimental evaluation: Registration}
\label{sec:registration_experiments}

We first evaluate SMF and SMF+ on surface registration tasks. In addition to the original data from \cite{Liu_2019_ICCV}, we test the generalisability of our method on a previously unseen dataset, 3DMD. 3DMD is a high resolution dataset containing in excess of $24,000$ scans captured from more than $3,000$ individuals. The dataset contains subjects from a wide range of ethnicities and age groups, each expressing a variety of facial expressions including neutral, happy, sad, angry and surprised. As stated in Section \ref{sec:comparison_3dfc}, we obtained a pre-trained model from \cite{Liu_2019_ICCV} trained on the entire dataset, which we use as a baseline.

\subsection{Landmark localization}
\label{sec:landmark_error_xp}

To reproduce the experiment of \cite{Liu_2019_ICCV}, we first evaluate the methods on the BU-3DFE database. We train SMF on the whole training set, as well as on the training set without BU-3DFE. We also re-trained a model using the methodology described in \cite{Liu_2019_ICCV}, excluding BU-3DFE from the training set. We include SMF+ (in sample) for comparison. We also report the performance of Non-Rigid ICP (NICP) initialized with landmarks and with additional stiffness weights to regularize deformations of the boundary, and use the values reported in \cite{Liu_2019_ICCV} for the algorithms of \cite{Bolkart2015}, \cite{Salazar2014}, and \cite{8373814}. For 3DMD, we report the performance of NICP initialized with landmarks, the pre-trained model of \cite{Liu_2019_ICCV}, SMF, and SMF+. For the sake of completeness, we also report the results of initializing NICP with the registration provided by SMF and SMF+ in place of the LSFM mean face and without landmarks information or stiffness weights. Given manual annotations on the raw scans, grouped by semantic label $\left(l^*_i\right)_{i=1}^k$ (\eg left eye, or nose), we compute the semantic landmark error per landmark group as $\frac{1}{k}\sum_{i=1}^k ||\hat{l}_i - l_i^*||$, with $\hat{l}_i$ the corresponding landmark in the registration. 

We report the mean and standard deviation of the error within each group. Table \ref{tab:bu3d_landmark_correspondence} summarizes the results for the BU-3DFE dataset, and Table \ref{tab:3dmd_landmark_correspondence} the results on 3DMD. We note that applying NICP did not significantly change the landmark error, which is likely due to the reconstructions output by SMF and SMF+ being already sufficiently close to the ground truth surface. There is, however, an advantage in using SMF to initialize NICP compared to landmarks: the typical runtime of the public implementation of NICP we used (from the publicly available LSFM code \citep{Booth2018}) with landmarks initialization was between 45 and 60 seconds per scan, while the initialization with SMF achieved equally detailed registrations in around 20 seconds per scan.

We note the high error values for the jaw landmarks on both datasets. The landmarks for the chin and jaw are at the boundary of the template. Since our method is trained on point clouds sampled from the raw scans with no manual cropping, the closest points for the boundary of the template are not at the edges of a tight crop of the face, and therefore the vertices of the boundary get pulled farther than where jaw landmarks are manually annotated. This results in large error values for these landmarks. Increasing the weight of our boundary loss regularization may help mitigate this phenomenon.

\subsection{Surface error}
\label{sec:surface_error_xp}

While a low landmark localization error suggests key facial points are faithfully placed in the registration, it does not paint the whole picture and does not indicate the general reconstruction fidelity. In particular, it is affected by the inevitable imprecision of manual labeling, and the error is measured on a small number of points.

To further assess the performance of the models, we measure the surface reconstruction error between the registrations and the ground-truth raw scans. We randomly select a sample of 5000 training scans and a sample of 5000 test scans (from the 3DMD dataset) and measure the distance of the vertices of the registrations to their closest point anywhere on the ground-truth surface (\ie, not the closest vertex). We summarize each scan by its mean surface error. 

\paragraph{Training and test set}

\begin{figure}[t]
    \centering
    
    \includegraphics[width=84mm]{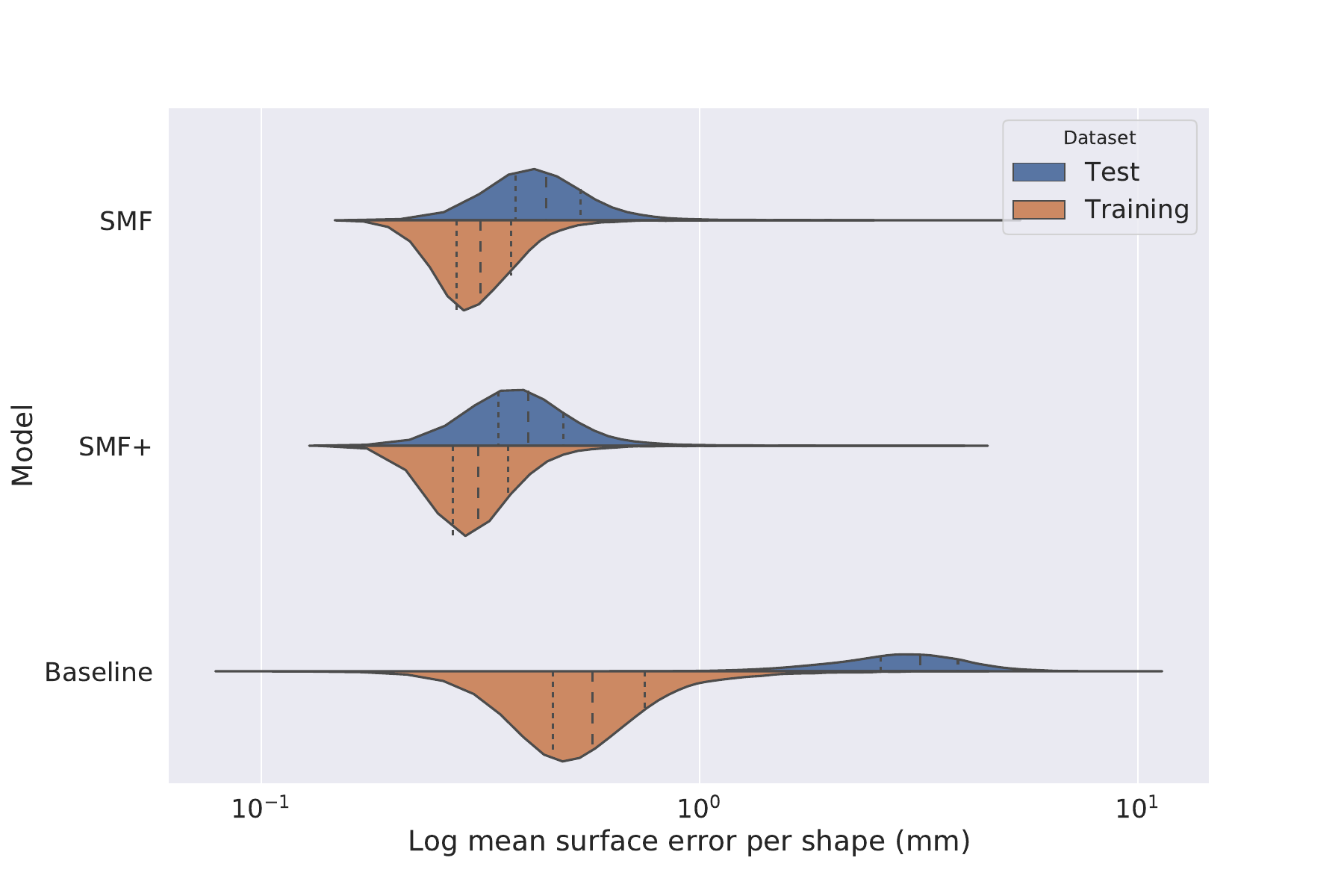}
    
    \caption{\textbf{Visualizing the error mean distribution on the training and test sets:} violin plots of the mean (per scan) surface error on the training and test sets for SMF, SMF+, and the baseline, plotted on a log scale. A violin plot represents the range of the data along with a kernel density estimation of the distribution. We split the plots to help compare the error distribution on the two datasets. Vertical dotted lines represent the quartiles of the distribution.}
    \label{fig:boxplots}
\end{figure}

We visualize the error distribution on the subsets of the training and test sets in Figures \ref{fig:boxplots}. 
On both the training and test sets, the models exhibit typically low error, with a pronounced skew of the mean towards lower values ($0.306\mathrm{mm}$ for SMF and $0.297\mathrm{mm}$ for SMF+). On the training set, the mean (per scan) error distributions of SMF and SMF+ appear very similar, with a slight advantage to SMF+. On the test set, however, SMF+ displays significantly lower values at the quartiles and a tighter distribution, suggesting the addition of the MeIn3D and 4DFAB datasets was effective in reducing the generalization gap and the variance of the model.

\paragraph{BU-3DFE and 3DMD}

\begin{figure}[t]
    \centering
    \includegraphics[width=84mm]{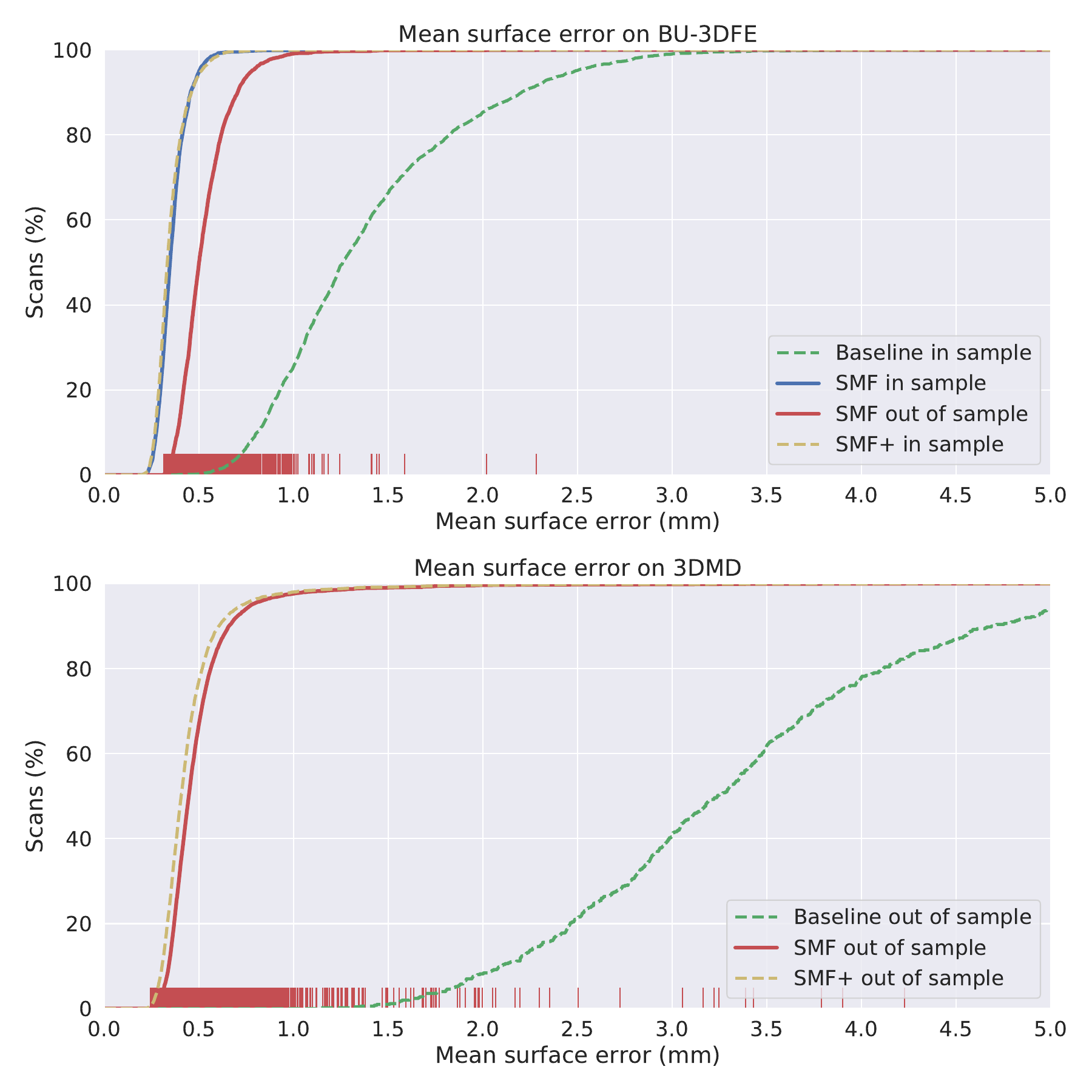}
    \caption{\textbf{Cumulative error:} Cumulative distribution function for the mean (per scan) error on the training and test sets for the models evaluated for the semantic landmark accuracy experiment. Even though the semantic landmark error of the baseline was not atypical, the distribution of the surface error reveals that the registration accuracy is actually much lower than that of SMF and SMF+. The rug plot (red bars at the bottom) visualize the distribution of the samples in terms of mean error for SMF evaluated out of sample. On 5000 sample test scans, few outliers had high mean surface error. SMF+ performs comparably with SMF in sample but has distinctly lower generalization error.}
    \label{fig:cdf_bu3d_3dmd}
\end{figure}

To complete the evaluation on BU3D and 3DMD, we produce in Figure \ref{fig:cdf_bu3d_3dmd} the cumulative distribution function (CDF) of the surface error for the entire BU3D dataset, and for the aforementioned sample of 5000 test scans, for the same models as in Section \ref{sec:landmark_error_xp}. To help visualize the counts of extreme values, we provide a rug plot for SMF evaluated out of sample.  As evidenced by the plots, SMF and SMF+ performed very similarly, while SMF trained without BU-3D had slightly lower performance. The baseline model, on the other hand, had significantly worse error distribution. Table \ref{tab:error_values_bu3d} provides numerical values for the 25\%, 50\%, 75\%, and 99\% quantiles for the models we plotted. We omit the baseline evaluated out of sample on BU-3DFE due to the very high landmark localization error. On our separate test set, a similar development unfolds.

\begin{table}[t]
    \centering
    \caption{\textbf{Mean surface error quantiles:} on BU-3DFE (left) and 3DMD (right) in mm.}
    \resizebox{\linewidth}{!}{
    \begin{tabular}{c|c|c|c|c}
         \textbf{BU-3DFE} & 25\% & 50\% & 75\% & 99\% \\
         \hline
         SMF In & 0.306 & 0.347 & 0.396 & 0.597\\
         SMF+ In & 0.297 & 0.333 & 0.387 & 0.617\\
         SMF Out & 0.434 & 0.501 & 0.594 & 0.983\\
         Baseline In & 0.995 & 1.261 & 1.683 & 2.989\\
    \end{tabular} \begin{tabular}{c|c|c|c|c}
         \textbf{3DMD} & 25\% & 50\% & 75\% & 99\% \\
         \hline
         SMF Out & 0.381 & 0.447 & 0.535 & 1.489\\
         SMF+ Out & 0.347 & 0.407 & 0.490 & 1.326\\
         - & - & - & - & -\\
         Baseline Out & 2.605 & 3.239 & 3.894 & 6.497\\
    \end{tabular}
    }
    \label{tab:error_values_bu3d}
\end{table}

The difference between the error distributions of SMF+ and SMF is small but significant, with SMF+ outperforming the model trained on less data. Out of sample, the baseline model's performance is significantly degraded, with the bottom 25\% of the surface error already reaching 2.60mm.

\begin{figure*}[p]
    \centering
    \setlength\tabcolsep{1.5pt}
    \small{
    \begin{tabular}{ccccc|@{\hskip 2.5mm}ccccc}
        Raw scan & Baseline & Att. & Reconst. & Error & Raw scan & Baseline & Att. & Reconst. & Error \\
        \includegraphics[width=.086\linewidth]{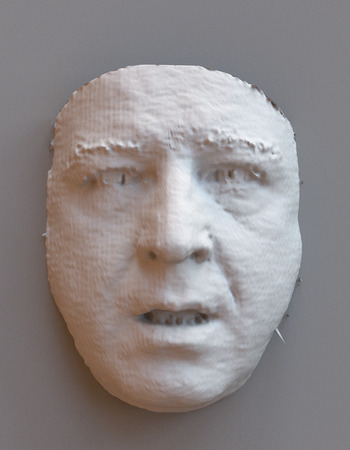} &
        \includegraphics[width=.086\linewidth]{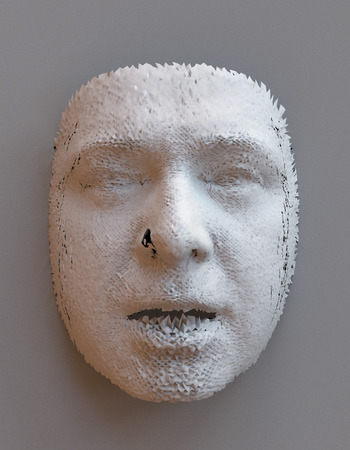} &
        \includegraphics[width=.086\linewidth]{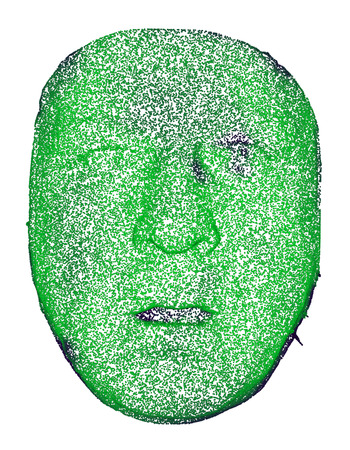} &
        \includegraphics[width=.086\linewidth]{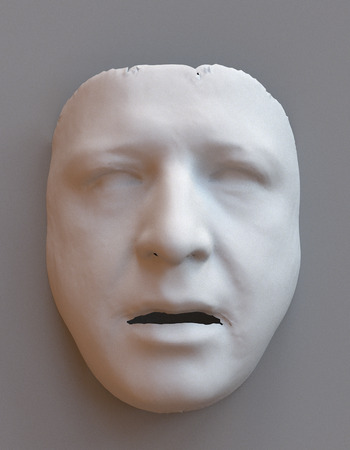} &
        \includegraphics[width=.086\linewidth]{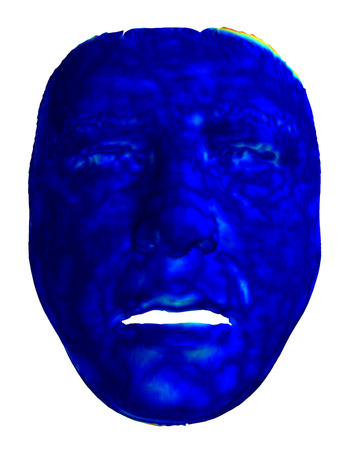} &
        \includegraphics[width=.086\linewidth]{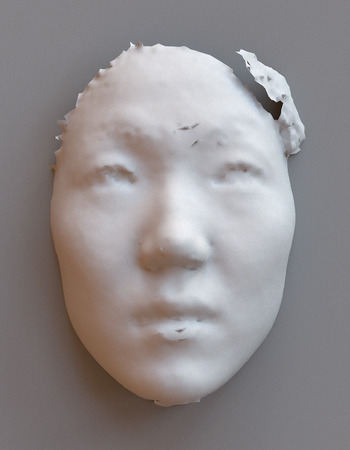} &
        \includegraphics[width=.086\linewidth]{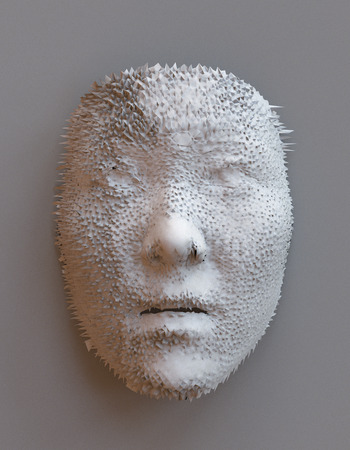} &
        \includegraphics[width=.086\linewidth]{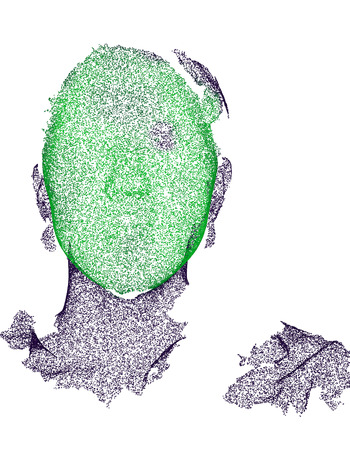} &
        \includegraphics[width=.086\linewidth]{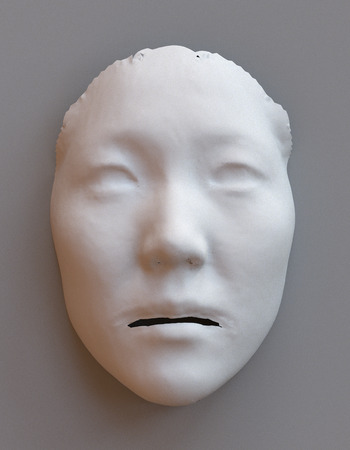} &
        \includegraphics[width=.086\linewidth]{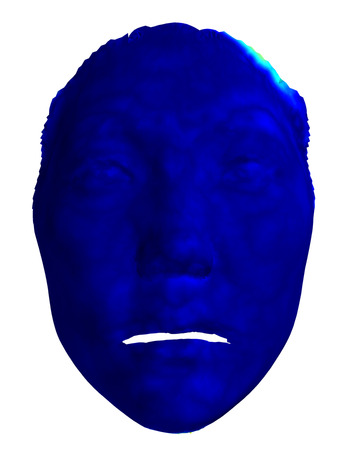}\\

        \includegraphics[width=.086\linewidth]{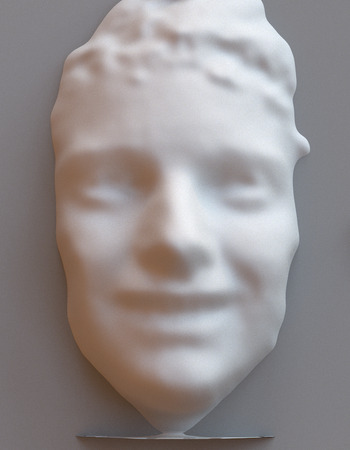} &
        \includegraphics[width=.086\linewidth]{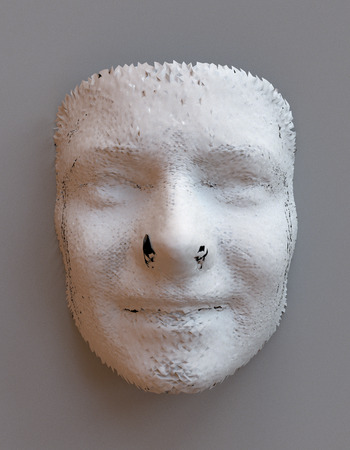} &
        \includegraphics[width=.086\linewidth]{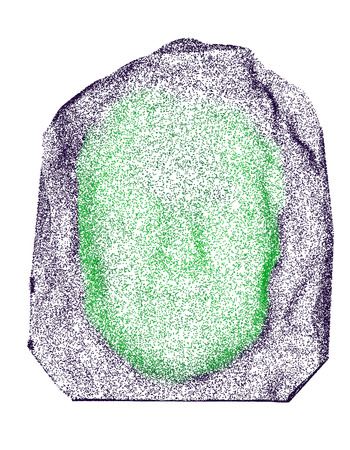} &
        \includegraphics[width=.086\linewidth]{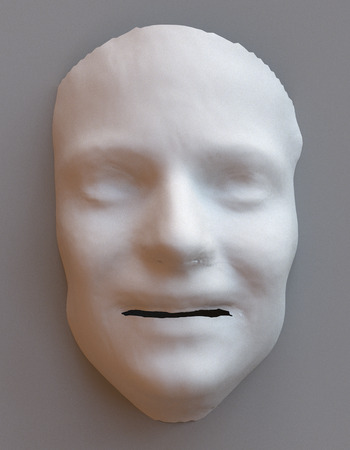} &
        \includegraphics[width=.086\linewidth]{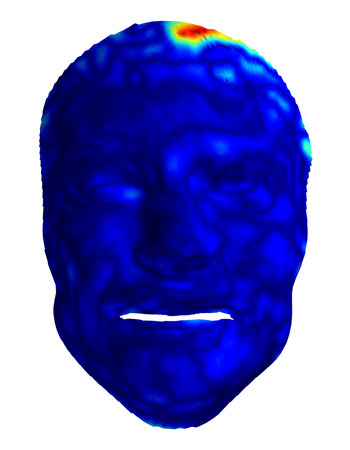} &
        \includegraphics[width=.086\linewidth]{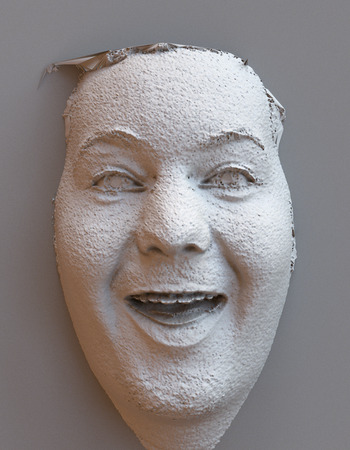} &
        \includegraphics[width=.086\linewidth]{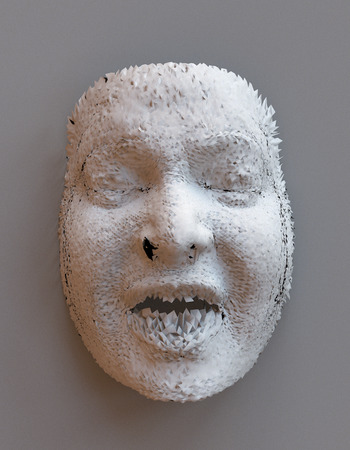} &
        \includegraphics[width=.086\linewidth]{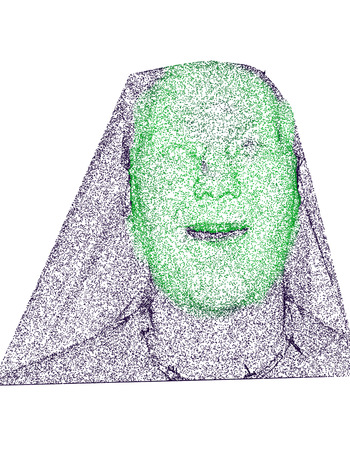} &
        \includegraphics[width=.086\linewidth]{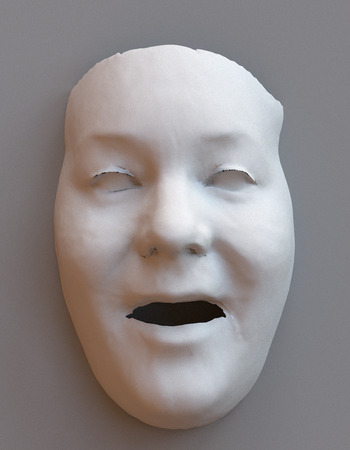} &
        \includegraphics[width=.086\linewidth]{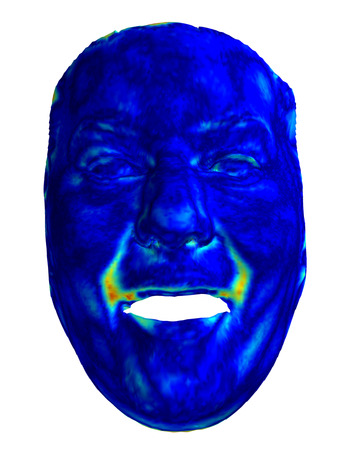}\\

        \includegraphics[width=.086\linewidth]{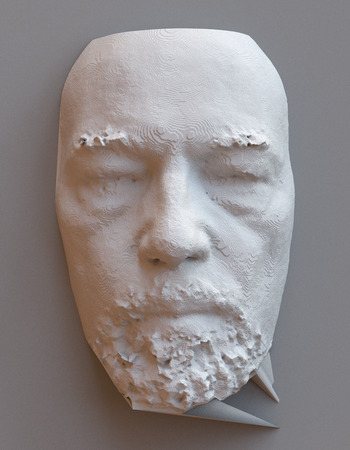} &
        \includegraphics[width=.086\linewidth]{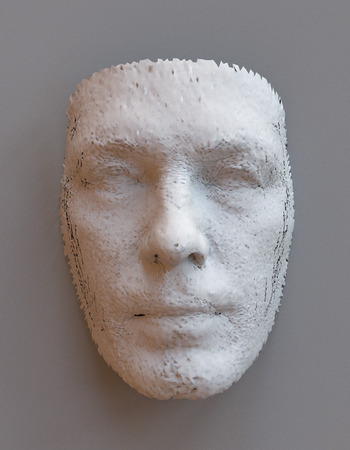} &
        \includegraphics[width=.086\linewidth]{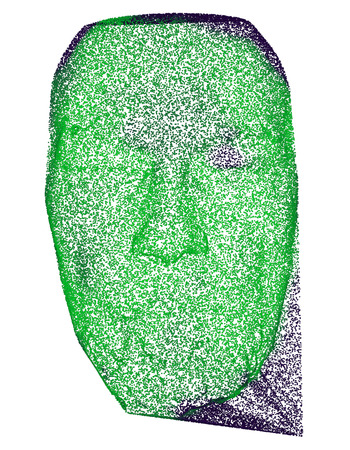} &
        \includegraphics[width=.086\linewidth]{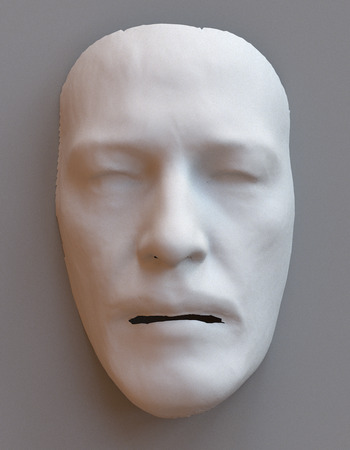} &
        \includegraphics[width=.086\linewidth]{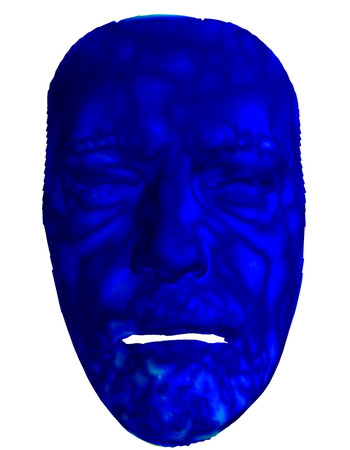} &
        \includegraphics[width=.086\linewidth]{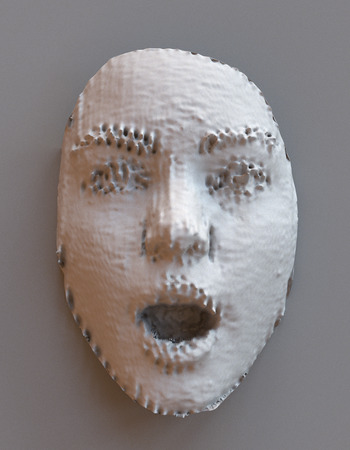} &
        \includegraphics[width=.086\linewidth]{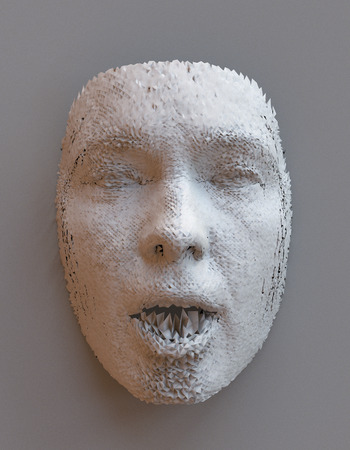} &
        \includegraphics[width=.086\linewidth]{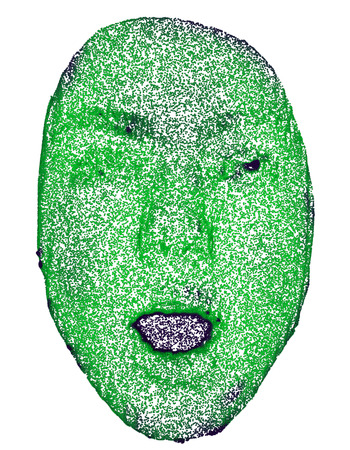} &
        \includegraphics[width=.086\linewidth]{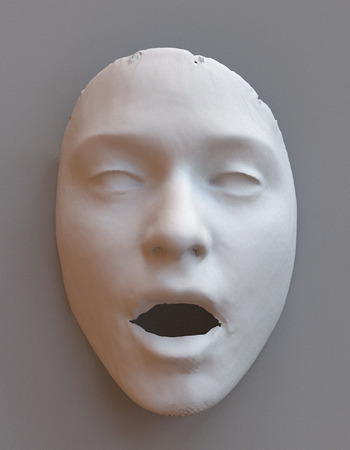} &
        \includegraphics[width=.086\linewidth]{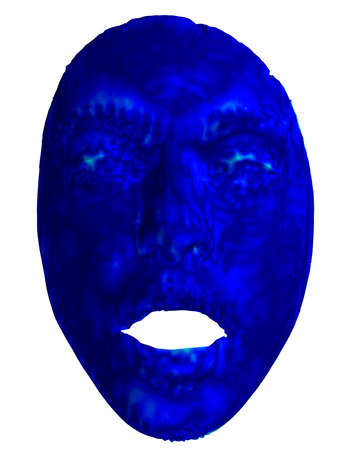}\\

        \includegraphics[width=.086\linewidth]{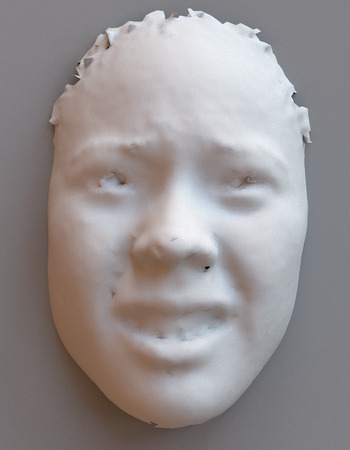} &
        \includegraphics[width=.086\linewidth]{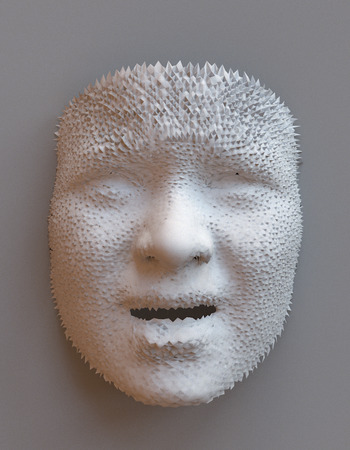} &
        \includegraphics[width=.086\linewidth]{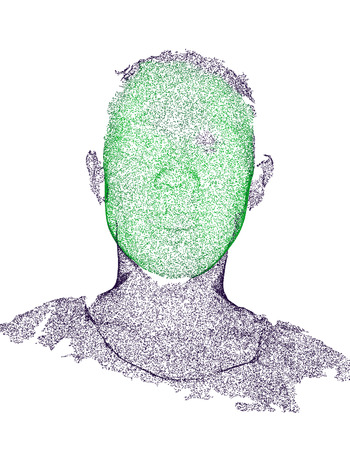} &
        \includegraphics[width=.086\linewidth]{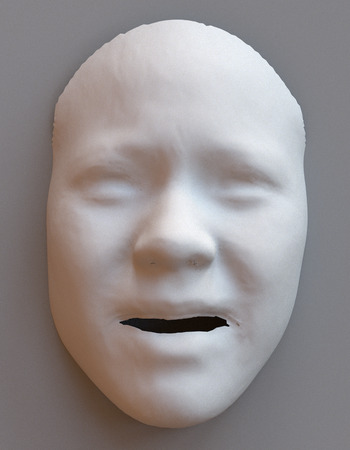} &
        \includegraphics[width=.086\linewidth]{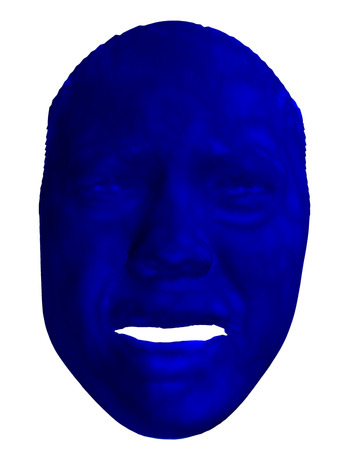} &
        \includegraphics[width=.086\linewidth]{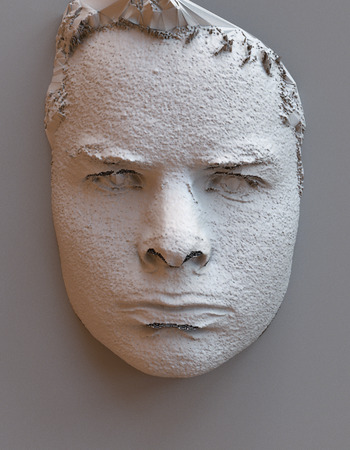} &
        \includegraphics[width=.086\linewidth]{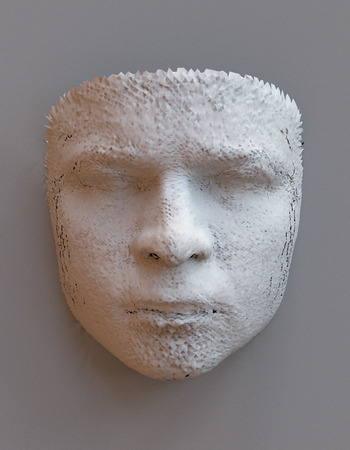} &
        \includegraphics[width=.086\linewidth]{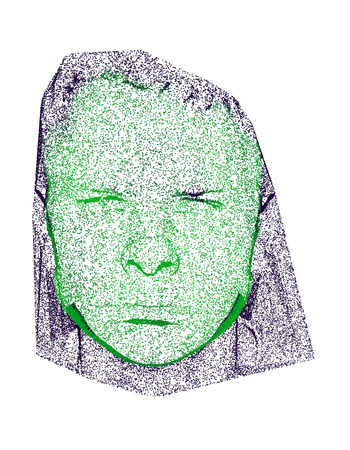} &
        \includegraphics[width=.086\linewidth]{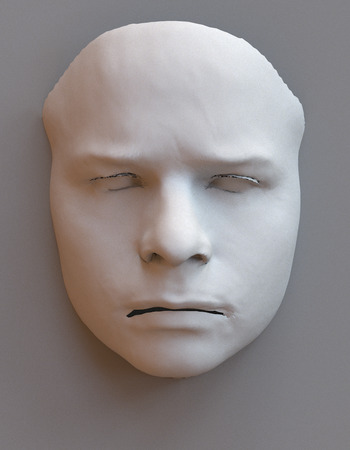} &
        \includegraphics[width=.086\linewidth]{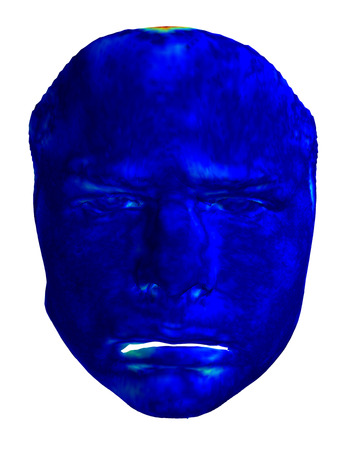}\\

        \includegraphics[width=.086\linewidth]{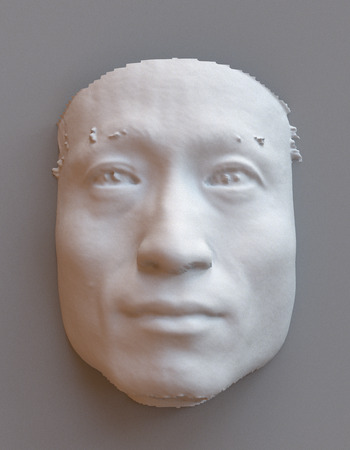} &
        \includegraphics[width=.086\linewidth]{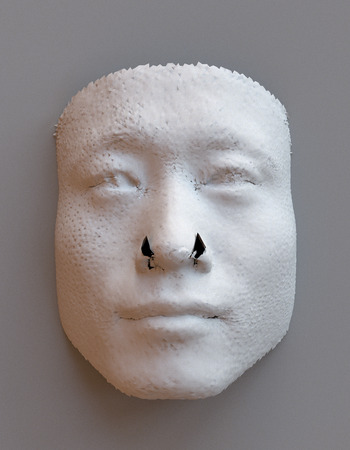} &
        \includegraphics[width=.086\linewidth]{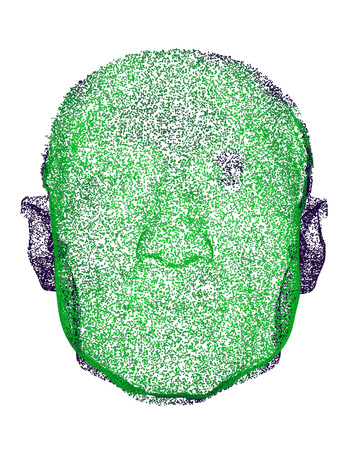} &
        \includegraphics[width=.086\linewidth]{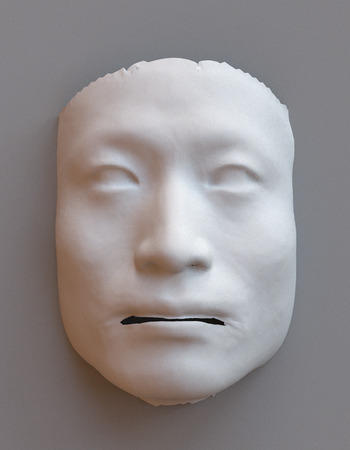} &
        \includegraphics[width=.086\linewidth]{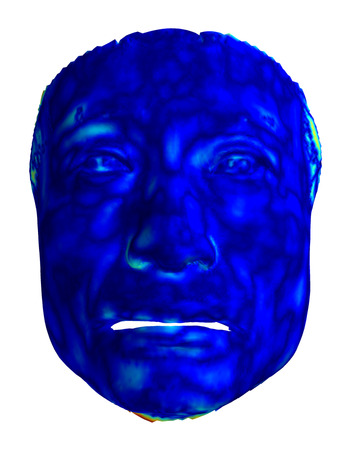} &
        \includegraphics[width=.086\linewidth]{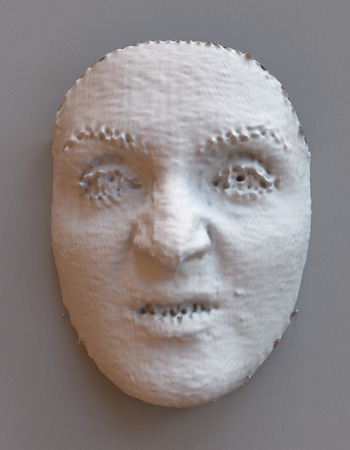} &
        \includegraphics[width=.086\linewidth]{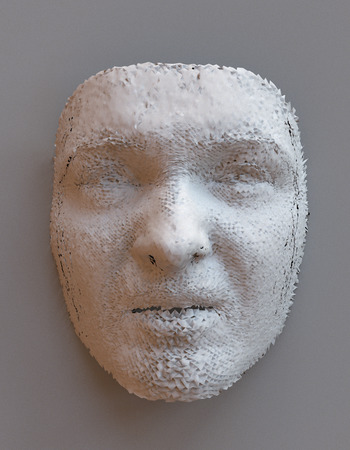} &
        \includegraphics[width=.086\linewidth]{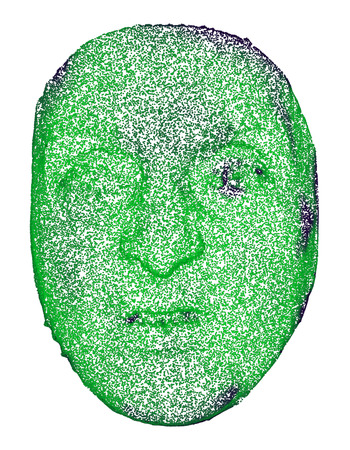} &
        \includegraphics[width=.086\linewidth]{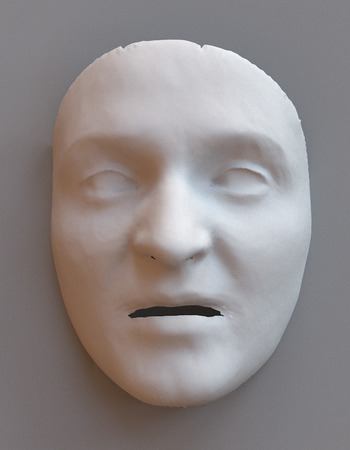} &
        \includegraphics[width=.086\linewidth]{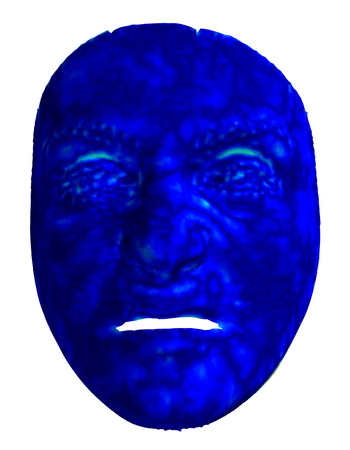}\\

        \includegraphics[width=.086\linewidth]{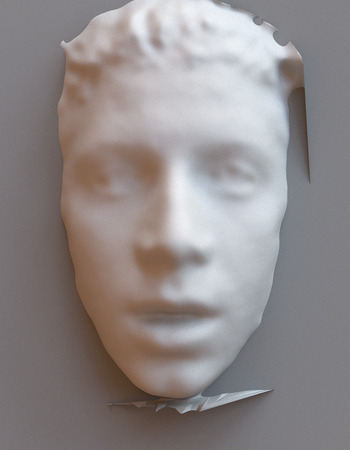} &
        \includegraphics[width=.086\linewidth]{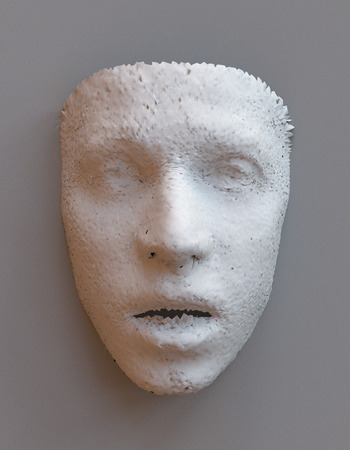} &
        \includegraphics[width=.086\linewidth]{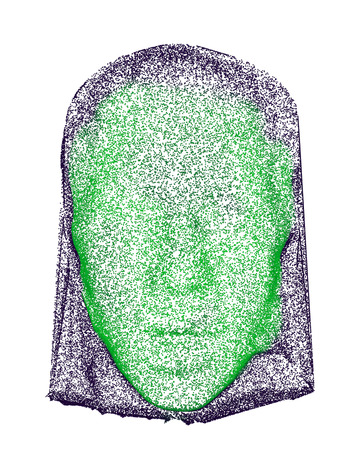} &
        \includegraphics[width=.086\linewidth]{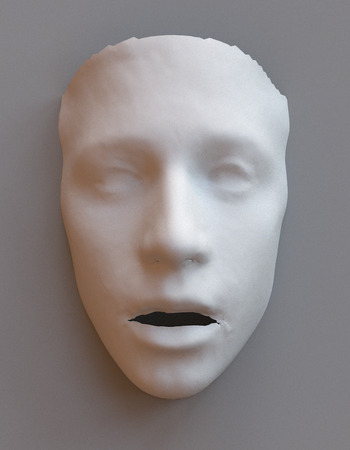} &
        \includegraphics[width=.086\linewidth]{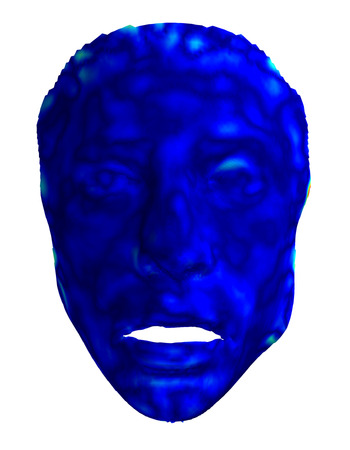} &
        \includegraphics[width=.086\linewidth]{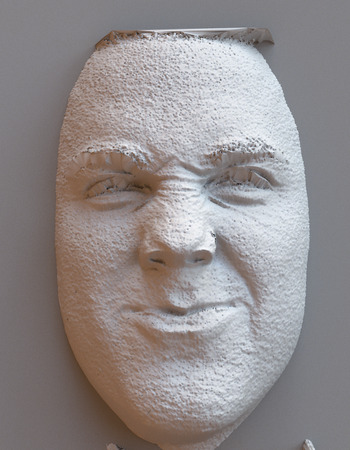} &
        \includegraphics[width=.086\linewidth]{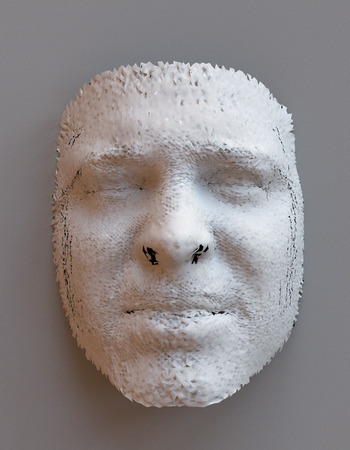} &
        \includegraphics[width=.086\linewidth]{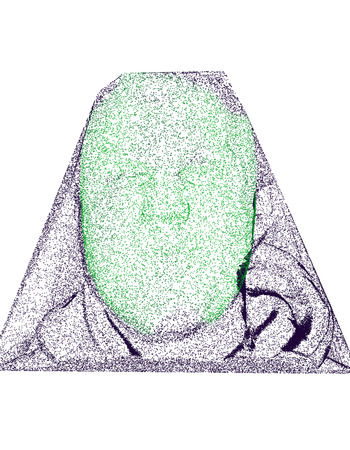} &
        \includegraphics[width=.086\linewidth]{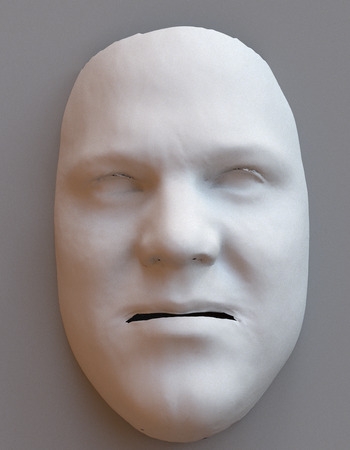} &
        \includegraphics[width=.086\linewidth]{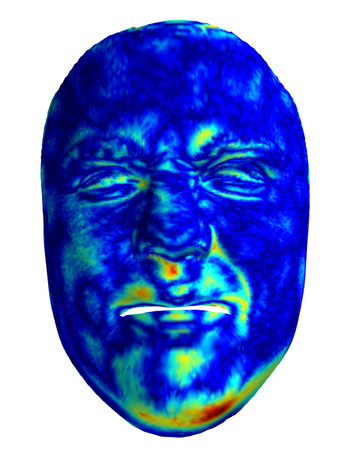}\\

        \includegraphics[width=.086\linewidth]{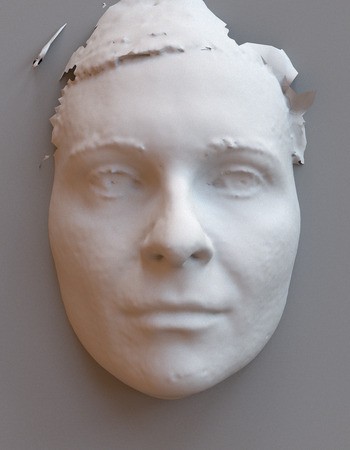} &
        \includegraphics[width=.086\linewidth]{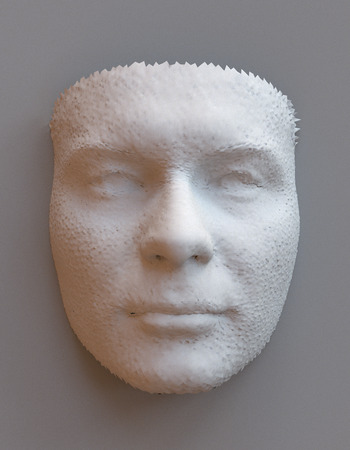} &
        \includegraphics[width=.086\linewidth]{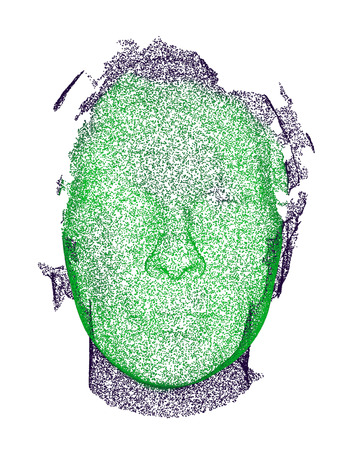} &
        \includegraphics[width=.086\linewidth]{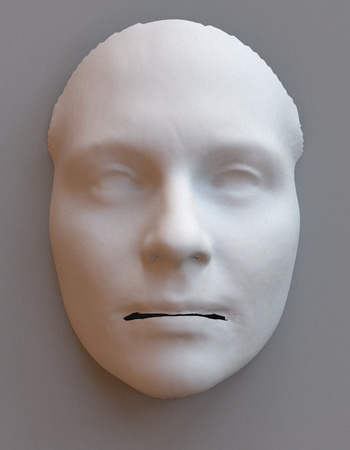} &
        \includegraphics[width=.086\linewidth]{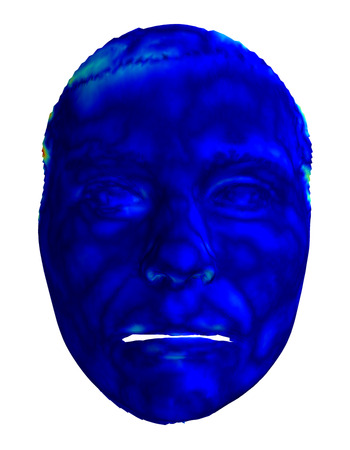} &
        \includegraphics[width=.086\linewidth]{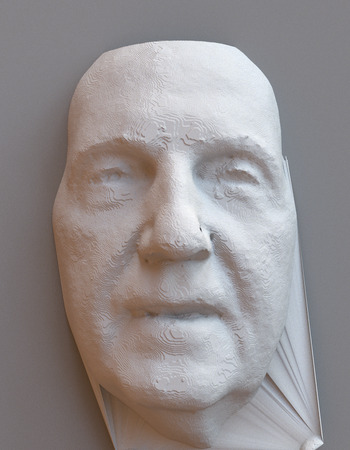} &
        \includegraphics[width=.086\linewidth]{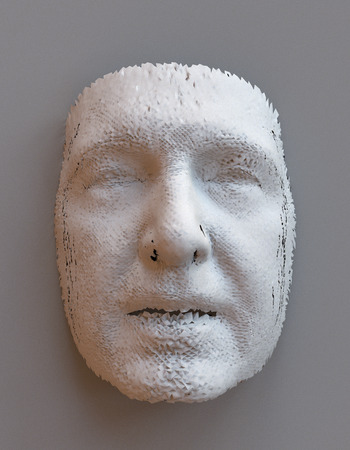} &
        \includegraphics[width=.086\linewidth]{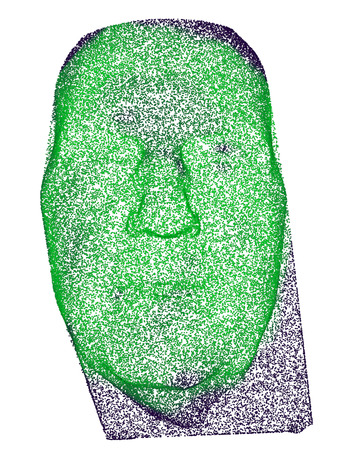} &
        \includegraphics[width=.086\linewidth]{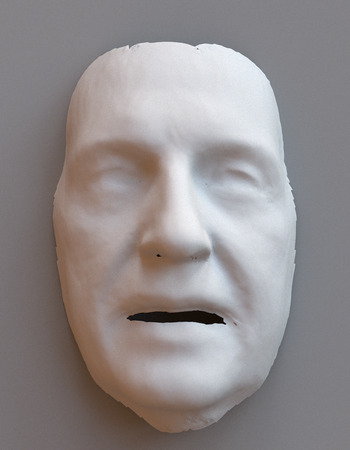} &
        \includegraphics[width=.086\linewidth]{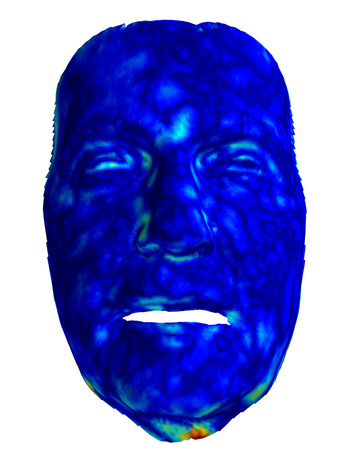}\\

        \includegraphics[width=.086\linewidth]{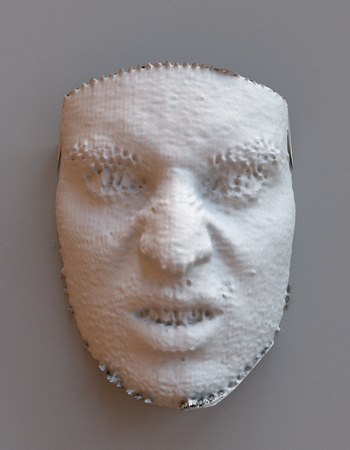} &
        \includegraphics[width=.086\linewidth]{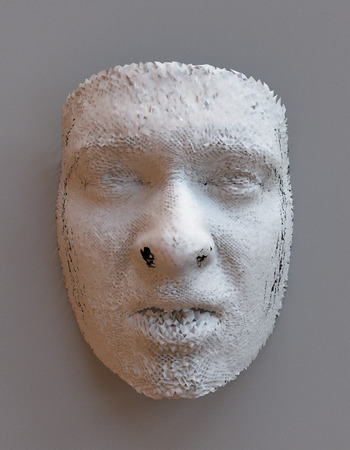} &
        \includegraphics[width=.086\linewidth]{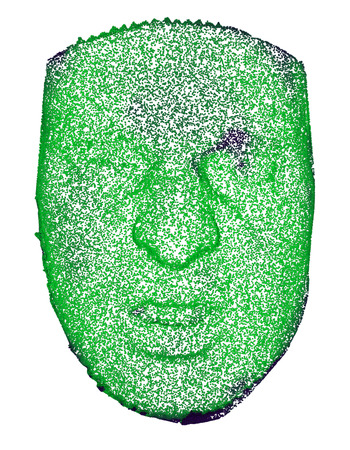} &
        \includegraphics[width=.086\linewidth]{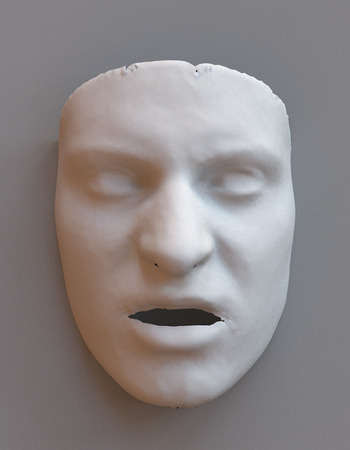} &
        \includegraphics[width=.086\linewidth]{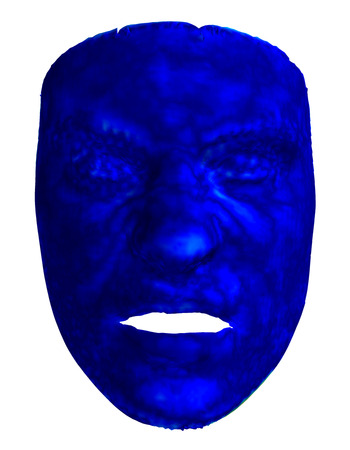}&
        \includegraphics[width=.086\linewidth]{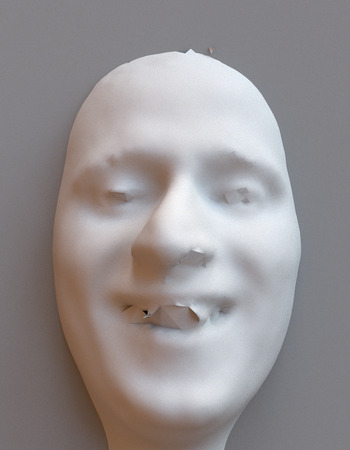} &
        \includegraphics[width=.086\linewidth]{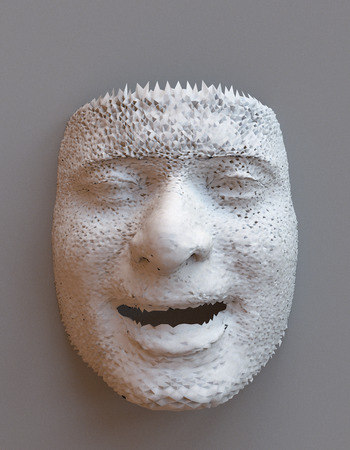} &
        \includegraphics[width=.086\linewidth]{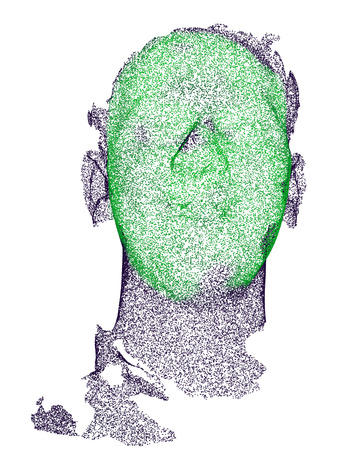} &
        \includegraphics[width=.086\linewidth]{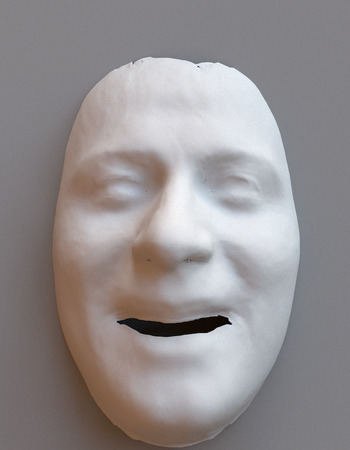} &
        \includegraphics[width=.086\linewidth]{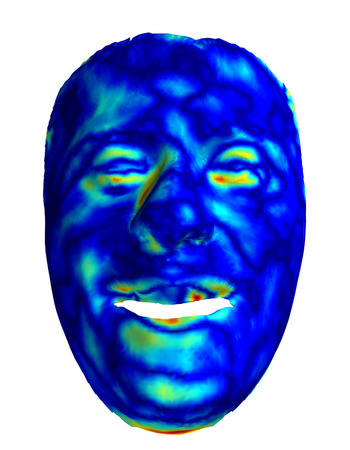}\\

        \includegraphics[width=.086\linewidth]{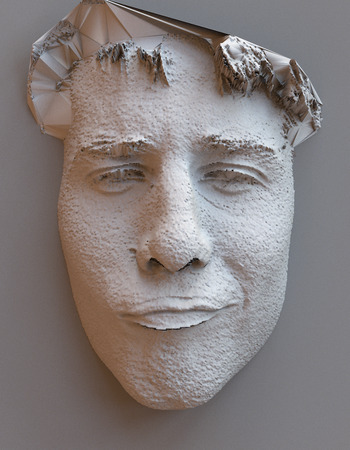} &
        \includegraphics[width=.086\linewidth]{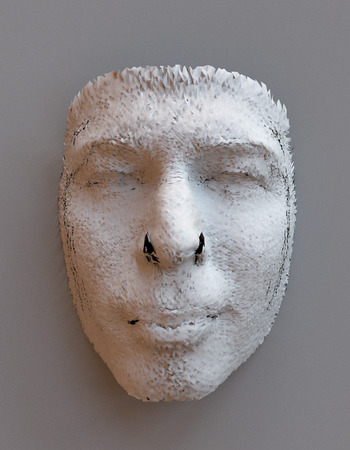} &
        \includegraphics[width=.086\linewidth]{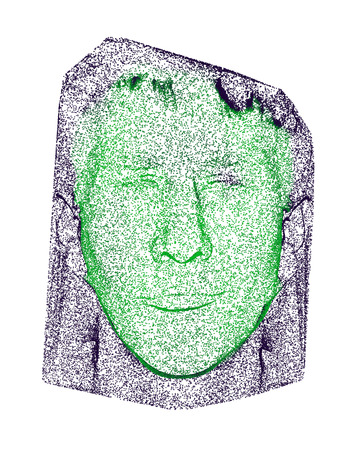} &
        \includegraphics[width=.086\linewidth]{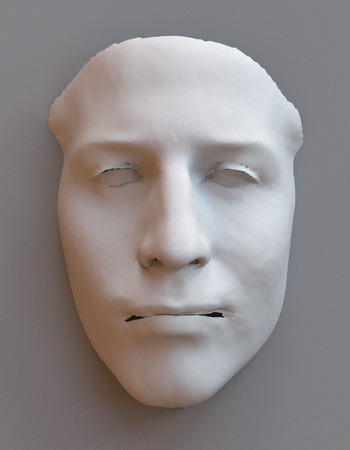} &
        \includegraphics[width=.086\linewidth]{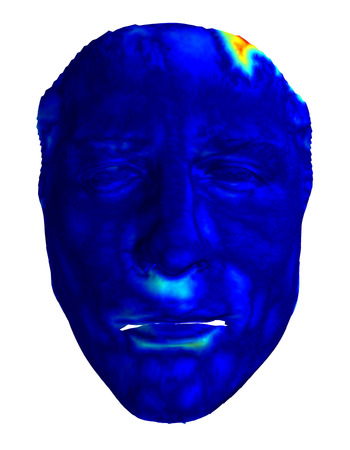} &
        \includegraphics[width=.086\linewidth]{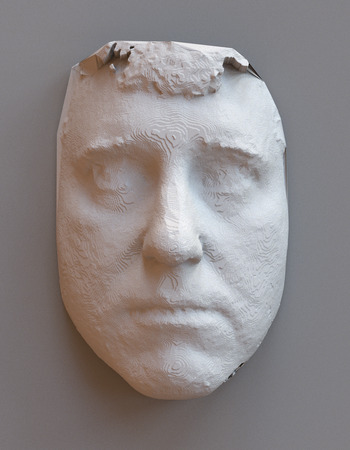} &
        \includegraphics[width=.086\linewidth]{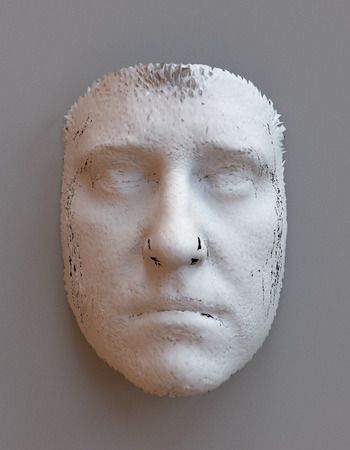} &
        \includegraphics[width=.086\linewidth]{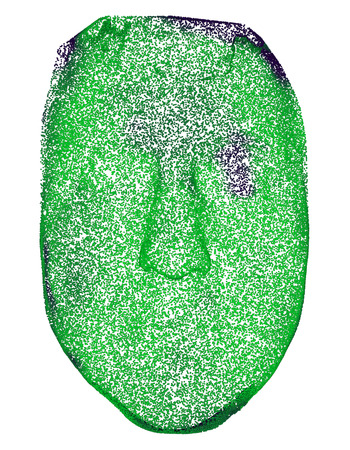} &
        \includegraphics[width=.086\linewidth]{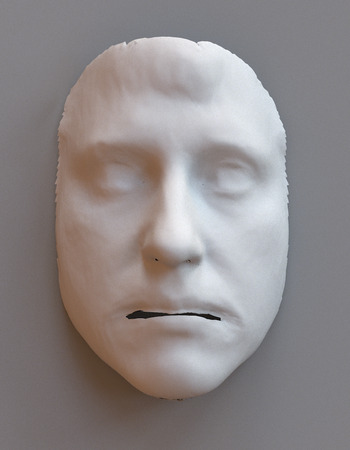} &
        \includegraphics[width=.086\linewidth]{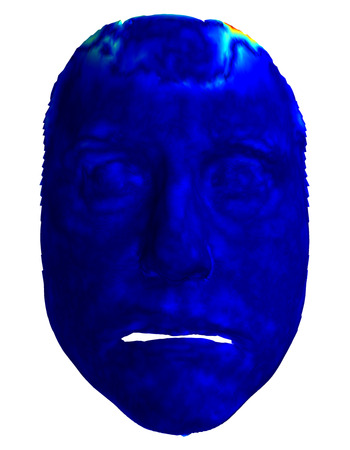}\\
        \end{tabular}
    }
    \includegraphics[width=.5\linewidth]{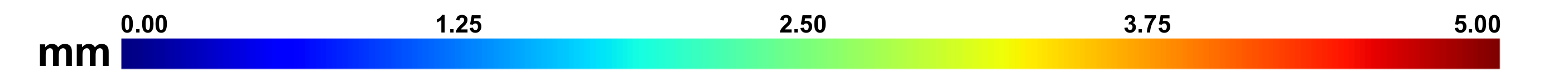}
    \caption{\textbf{Sample reconstructions on the training set for SMF:} arranged in two columns. From left to right: raw scan, output of the baseline, point cloud sampled on the scan by SMF and predicted attention mask, output of SMF, and surface reconstruction error visualized as a texture on the output of SMF. We can see SMF markedly outperforms the baseline and provides accurate natural-looking reconstructions with uniformly low error in the facial region and accurate representation of both identity and expression.}
    \label{fig:sample_reconst_train}
\end{figure*}

\begin{figure*}[p]
    \centering
    \setlength\tabcolsep{1.5pt}
    \small{
    \begin{tabular}{ccccc|@{\hskip 2.5mm}ccccc}
        Raw scan & Baseline & Att. & Reconst. & Error & Raw scan & Baseline & Att. & Reconst. & Error \\
        \includegraphics[width=.086\linewidth]{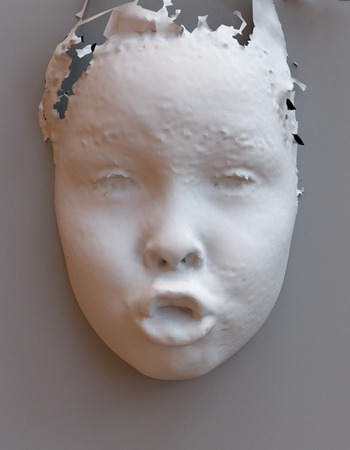} &
        \includegraphics[width=.086\linewidth]{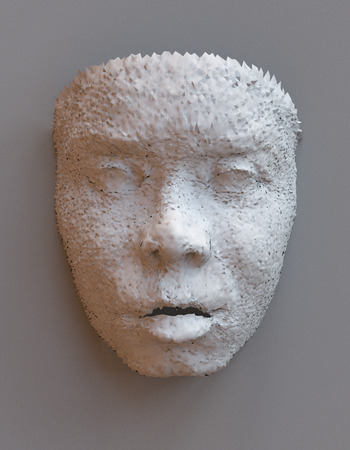} &
        \includegraphics[width=.086\linewidth]{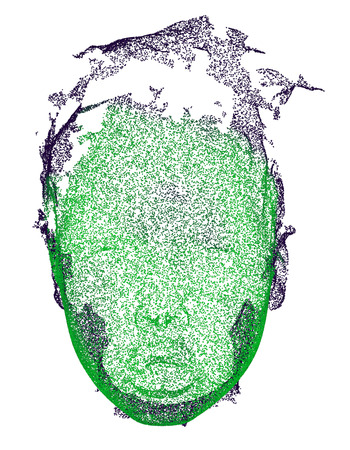} &
        \includegraphics[width=.086\linewidth]{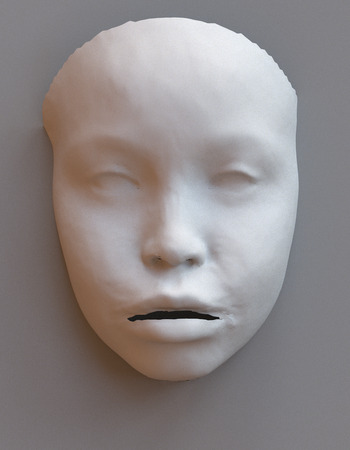} &
        \includegraphics[width=.086\linewidth]{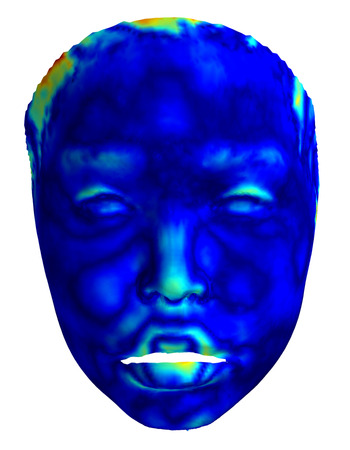} &
        \includegraphics[width=.086\linewidth]{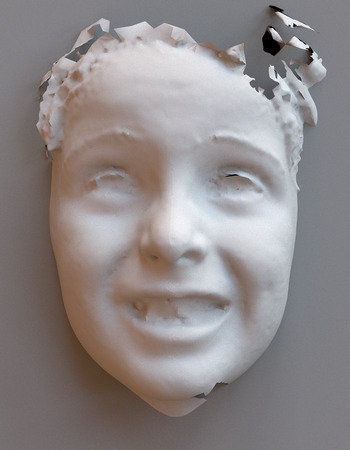} &
        \includegraphics[width=.086\linewidth]{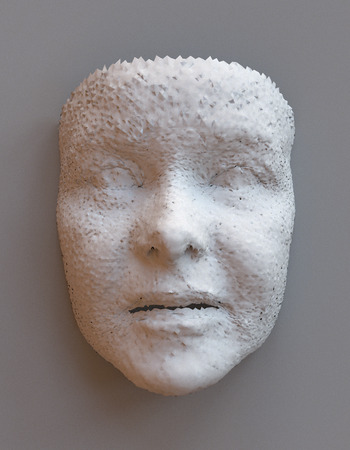} &
        \includegraphics[width=.086\linewidth]{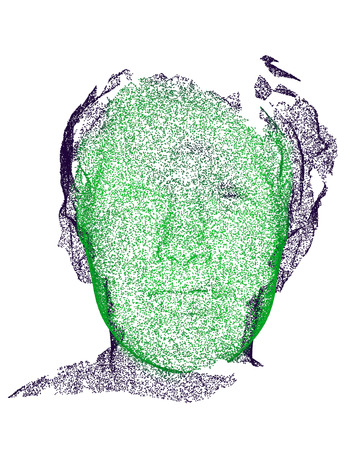} &
        \includegraphics[width=.086\linewidth]{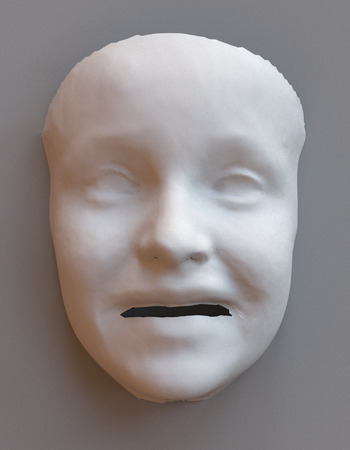} &
        \includegraphics[width=.086\linewidth]{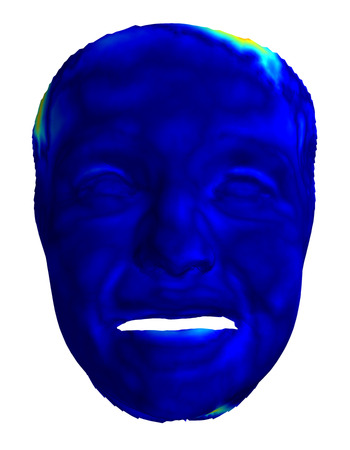}\\

        \includegraphics[width=.086\linewidth]{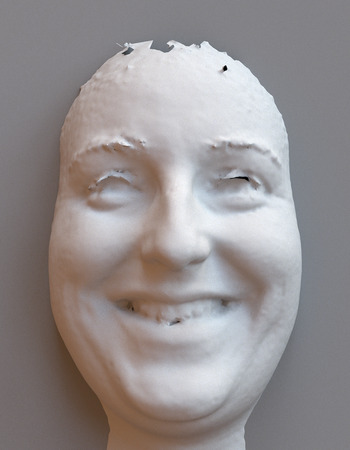} &
        \includegraphics[width=.086\linewidth]{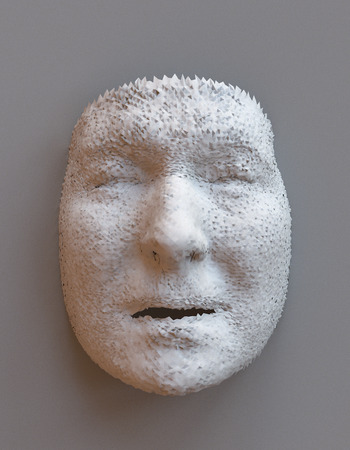} &
        \includegraphics[width=.086\linewidth]{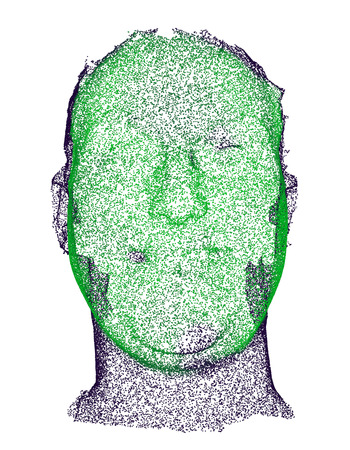} &
        \includegraphics[width=.086\linewidth]{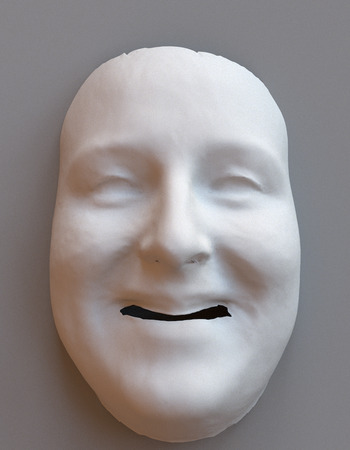} &
        \includegraphics[width=.086\linewidth]{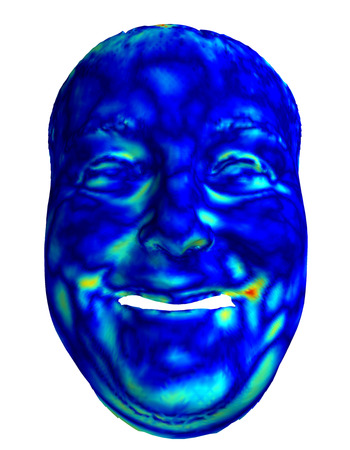} &
        \includegraphics[width=.086\linewidth]{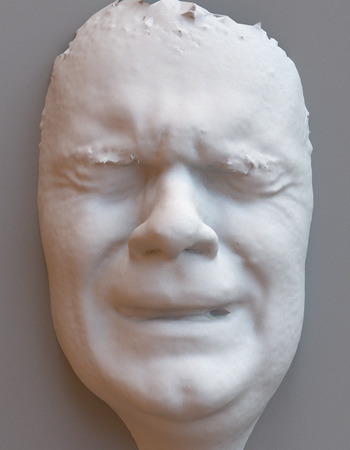} &
        \includegraphics[width=.086\linewidth]{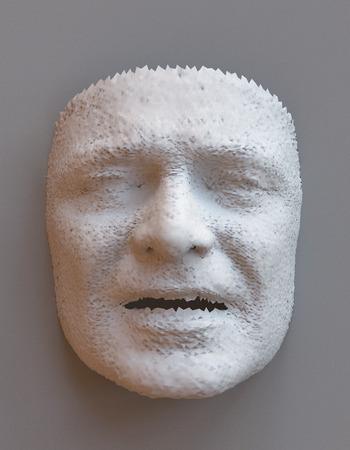} &
        \includegraphics[width=.086\linewidth]{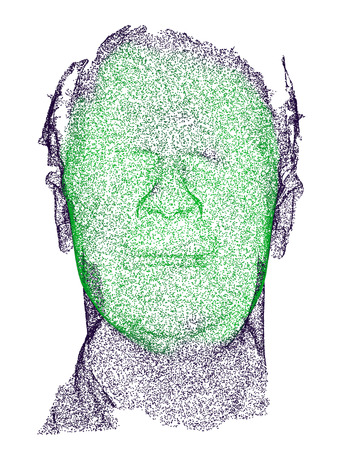} &
        \includegraphics[width=.086\linewidth]{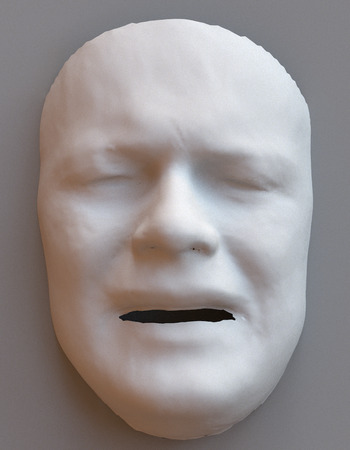} &
        \includegraphics[width=.086\linewidth]{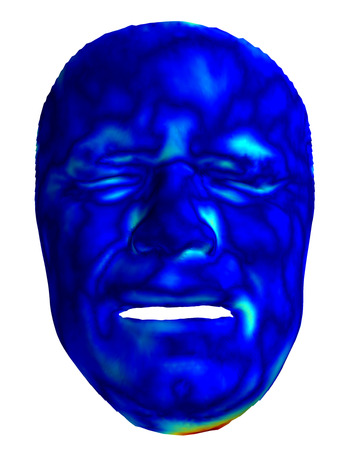}\\

        \includegraphics[width=.086\linewidth]{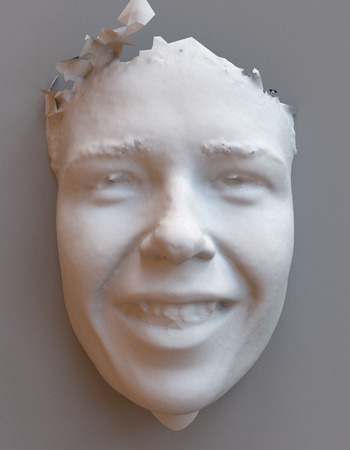} &
        \includegraphics[width=.086\linewidth]{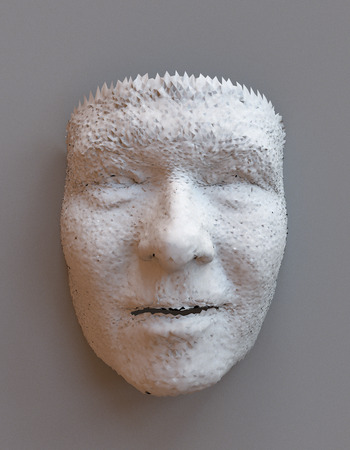} &
        \includegraphics[width=.086\linewidth]{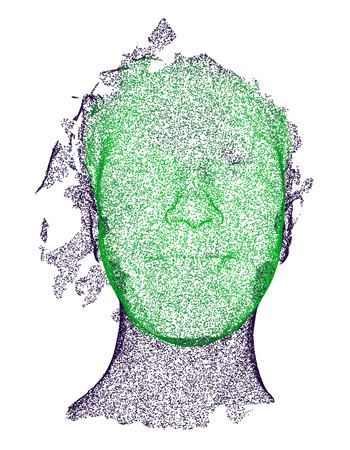} &
        \includegraphics[width=.086\linewidth]{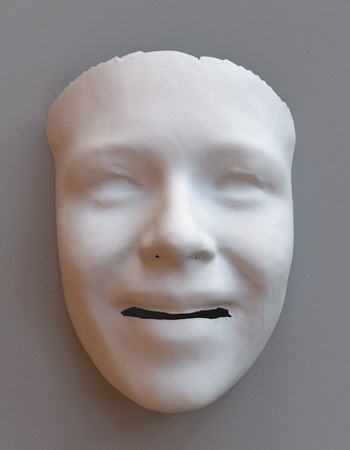} &
        \includegraphics[width=.086\linewidth]{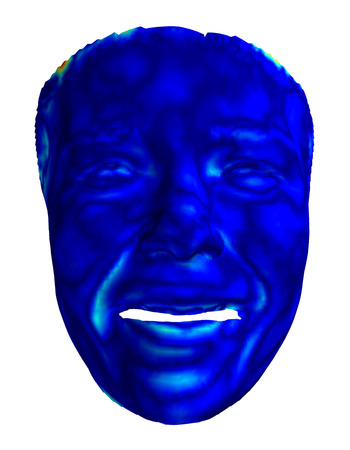} &
        \includegraphics[width=.086\linewidth]{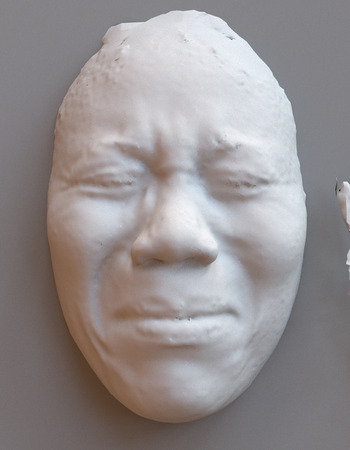} &
        \includegraphics[width=.086\linewidth]{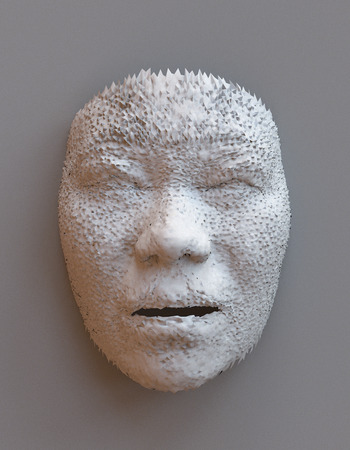} &
        \includegraphics[width=.086\linewidth]{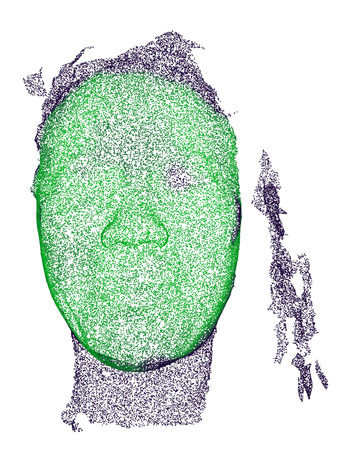} &
        \includegraphics[width=.086\linewidth]{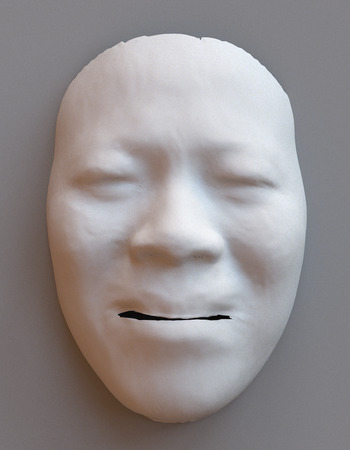} &
        \includegraphics[width=.086\linewidth]{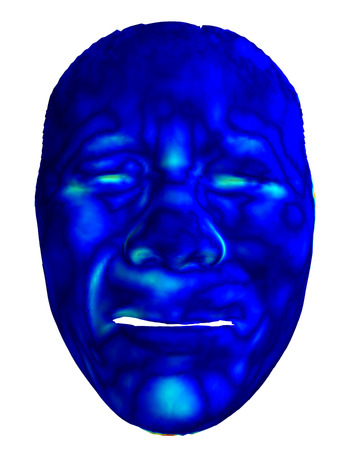}\\

        \includegraphics[width=.086\linewidth]{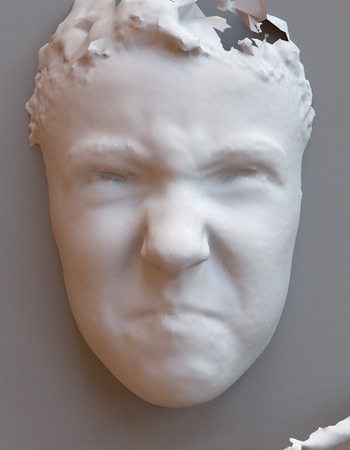} &
        \includegraphics[width=.086\linewidth]{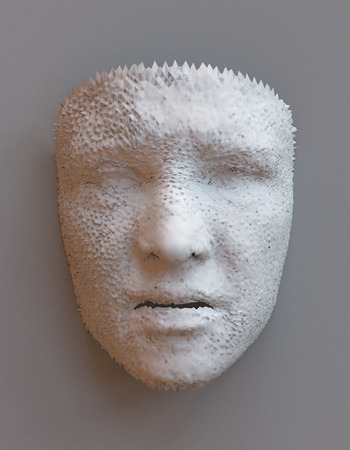} &
        \includegraphics[width=.086\linewidth]{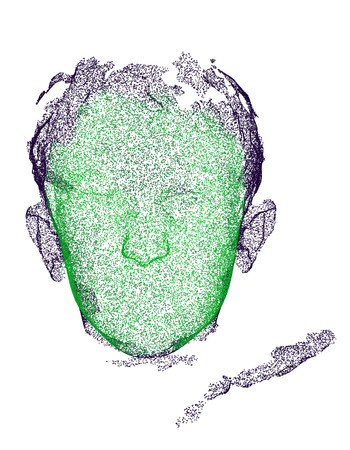} &
        \includegraphics[width=.086\linewidth]{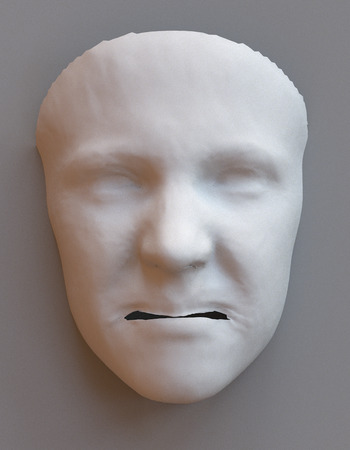} &
        \includegraphics[width=.086\linewidth]{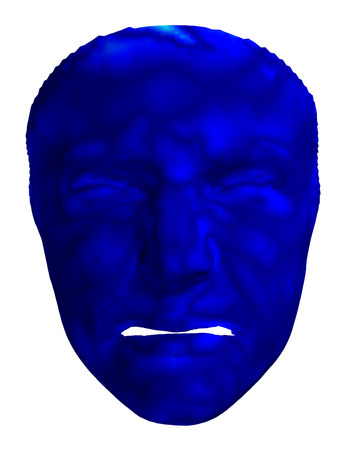} &
        \includegraphics[width=.086\linewidth]{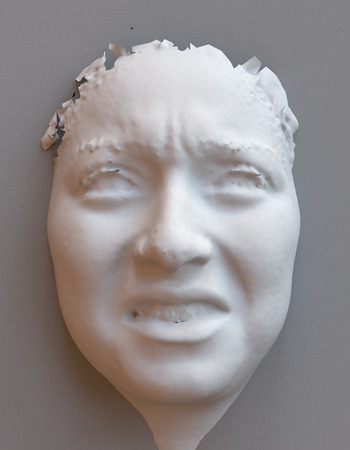} &
        \includegraphics[width=.086\linewidth]{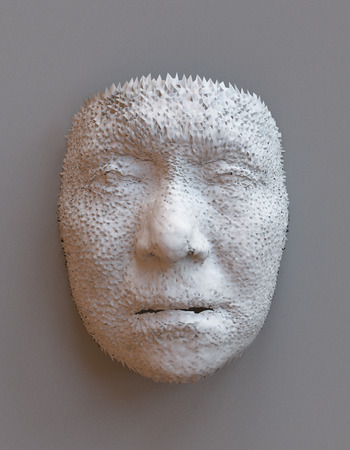} &
        \includegraphics[width=.086\linewidth]{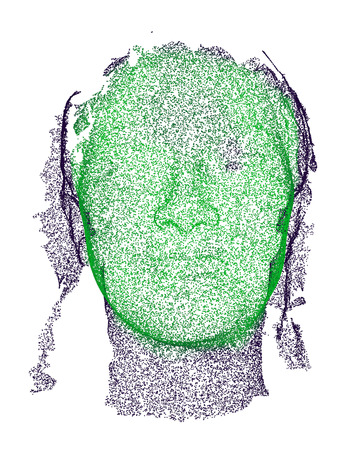} &
        \includegraphics[width=.086\linewidth]{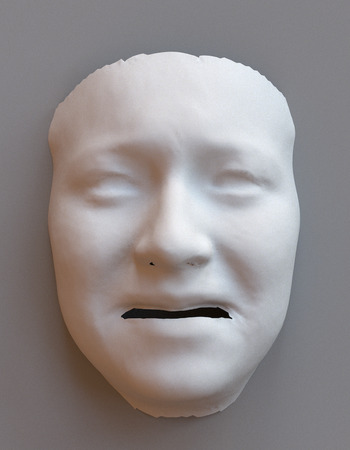} &
        \includegraphics[width=.086\linewidth]{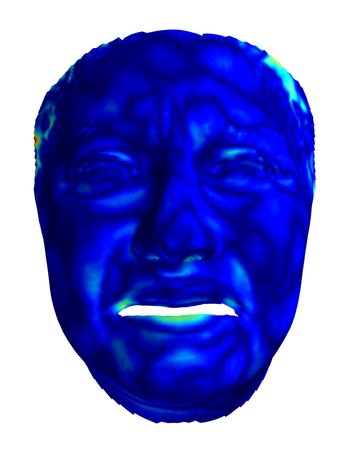}\\

        \includegraphics[width=.086\linewidth]{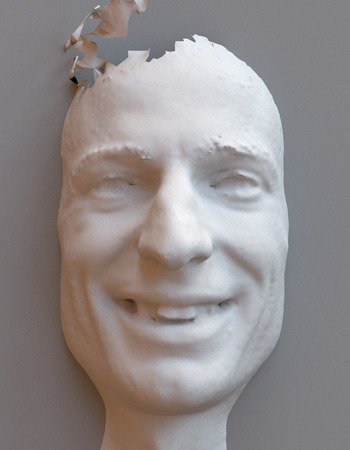} &
        \includegraphics[width=.086\linewidth]{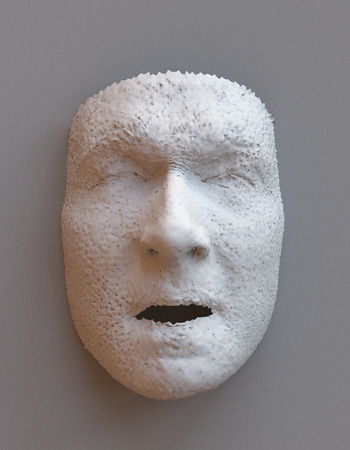} &
        \includegraphics[width=.086\linewidth]{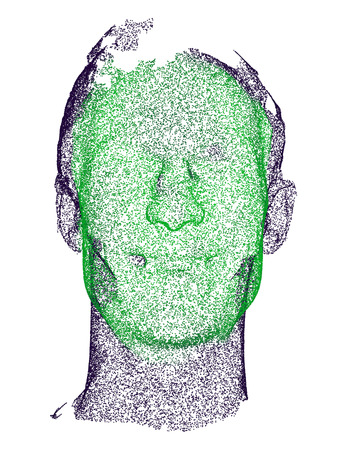} &
        \includegraphics[width=.086\linewidth]{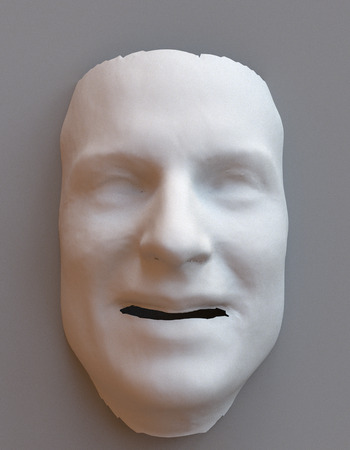} &
        \includegraphics[width=.086\linewidth]{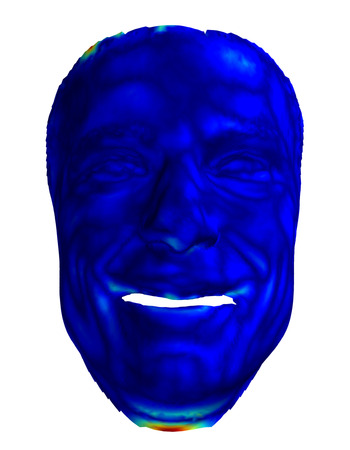} &
        \includegraphics[width=.086\linewidth]{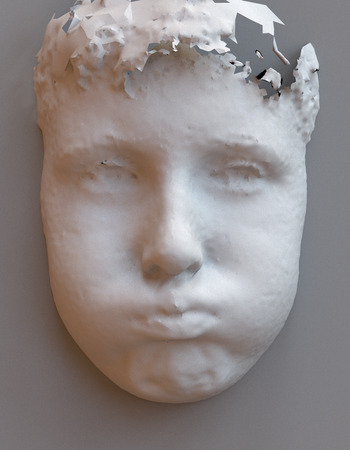} &
        \includegraphics[width=.086\linewidth]{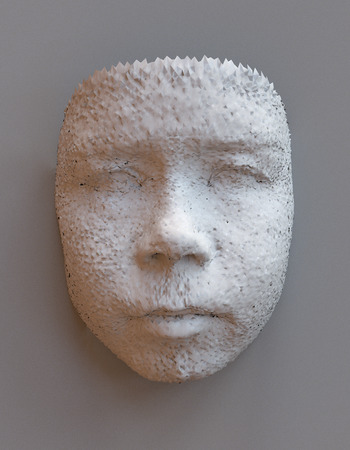} &
        \includegraphics[width=.086\linewidth]{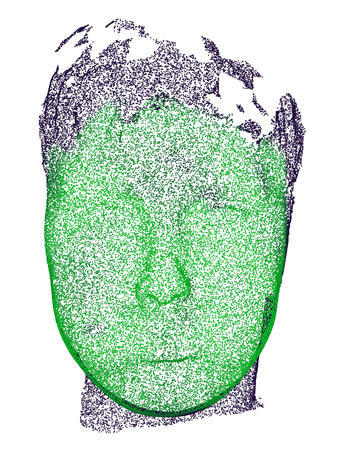} &
        \includegraphics[width=.086\linewidth]{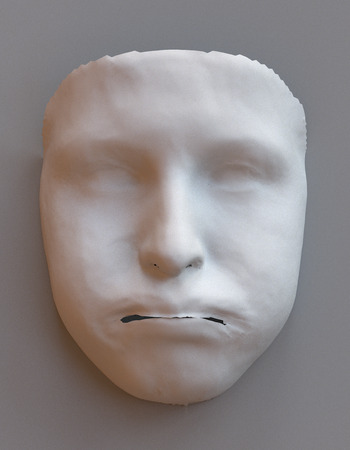} &
        \includegraphics[width=.086\linewidth]{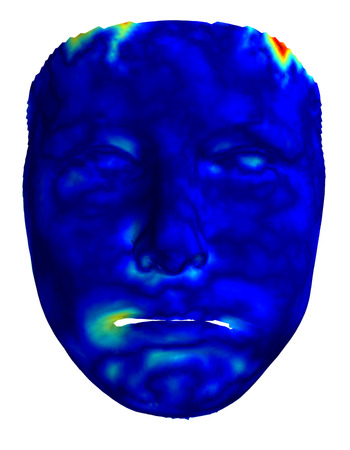}\\

        \includegraphics[width=.086\linewidth]{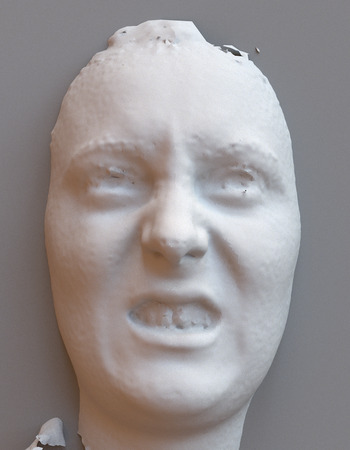} &
        \includegraphics[width=.086\linewidth]{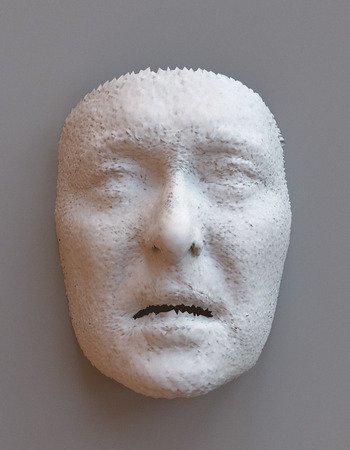} &
        \includegraphics[width=.086\linewidth]{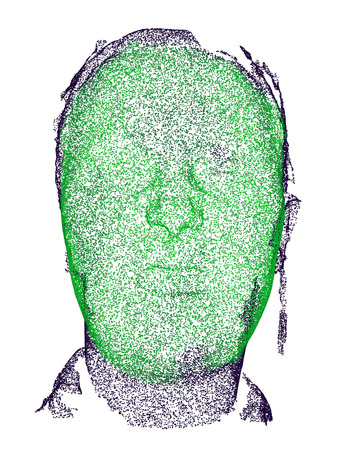} &
        \includegraphics[width=.086\linewidth]{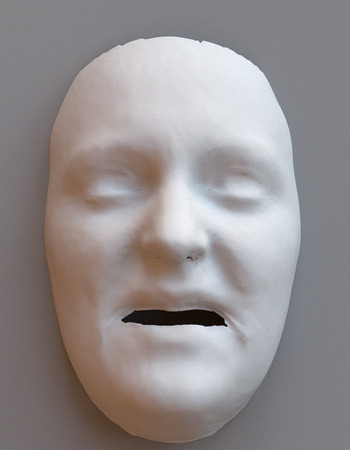} &
        \includegraphics[width=.086\linewidth]{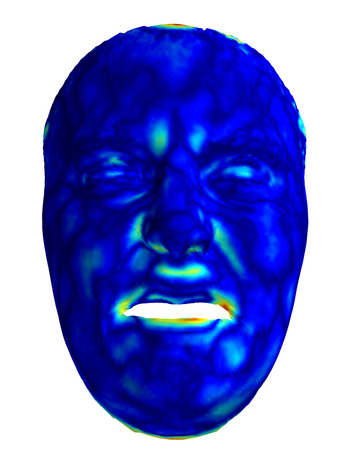} &
        \includegraphics[width=.086\linewidth]{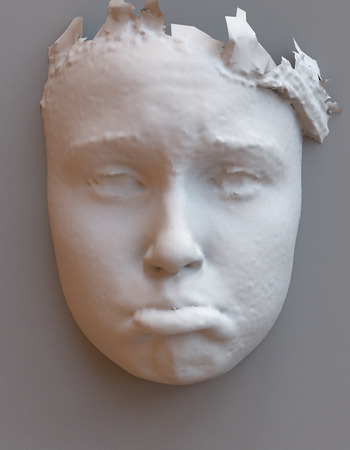} &
        \includegraphics[width=.086\linewidth]{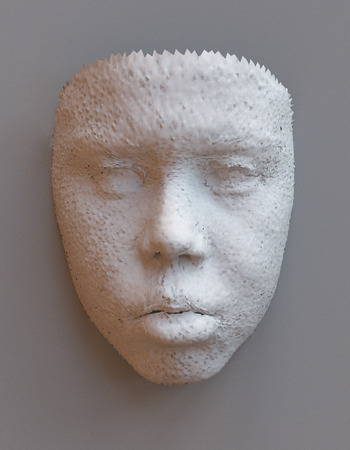} &
        \includegraphics[width=.086\linewidth]{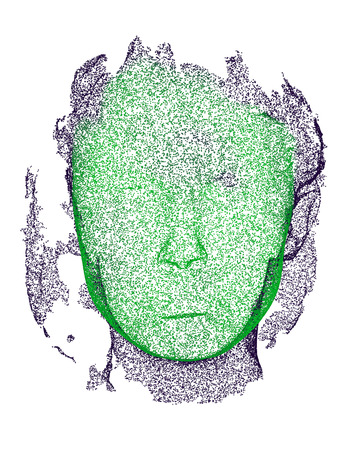} &
        \includegraphics[width=.086\linewidth]{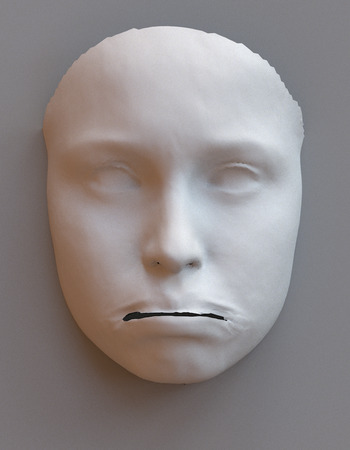} &
        \includegraphics[width=.086\linewidth]{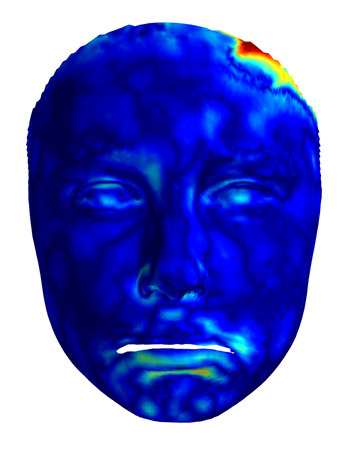}\\
    \end{tabular}
    }
    \includegraphics[width=.5\linewidth]{Fig_colourbar.pdf}
    \caption{\textbf{Sample reconstructions on the test set for SMF:} arranged in two columns. From left to right: raw scan, output of the baseline, point cloud sampled on the scan and predicted attention mask, output of SMF, and surface reconstruction error visualized as a texture on the output of SMF. The test reconstructions look comparable to the training reconstructions for SMF, with high quality registrations across gender, age and ethnicity, even for extreme facial expressions.}
    \label{fig:sample_renders_3dmd}
\end{figure*}

\begin{figure*}[p]
    \centering
    \setlength\tabcolsep{1.5pt}
    \small{
    \begin{tabular}{cccc|@{\hskip 2.5mm}cccc}
        Raw scan & Att. & Reconst. & Error & Raw scan & Att. & Reconst. & Error \\
        \includegraphics[width=.086\linewidth]{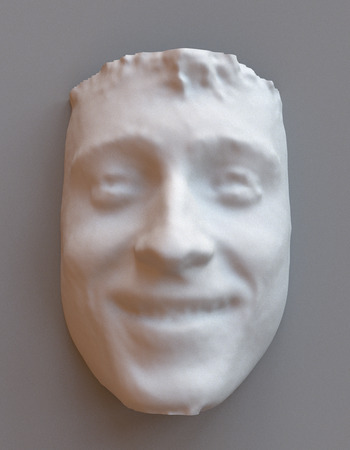} &
        \includegraphics[width=.086\linewidth]{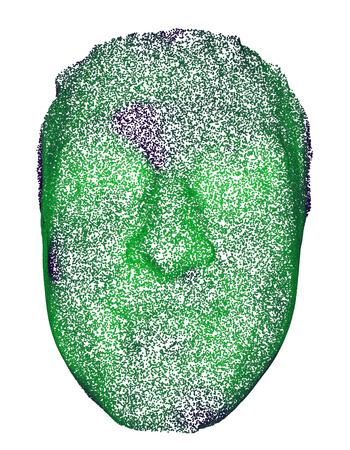} &
        \includegraphics[width=.086\linewidth]{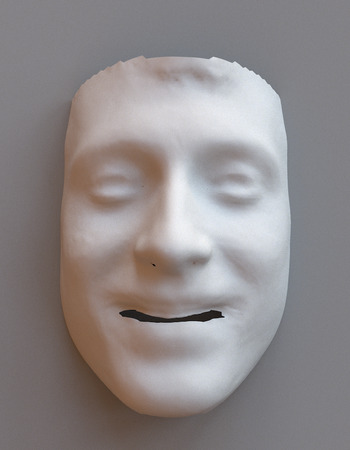} &
        \includegraphics[width=.086\linewidth]{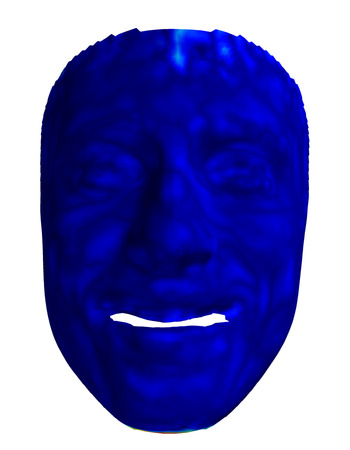} &
        \includegraphics[width=.086\linewidth]{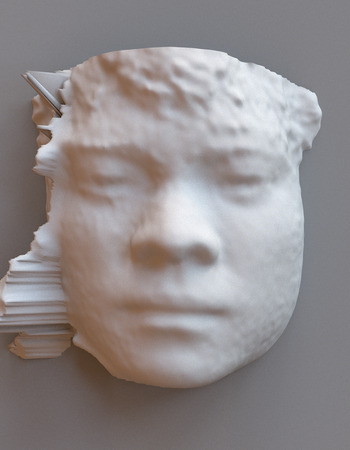} &
        \includegraphics[width=.086\linewidth]{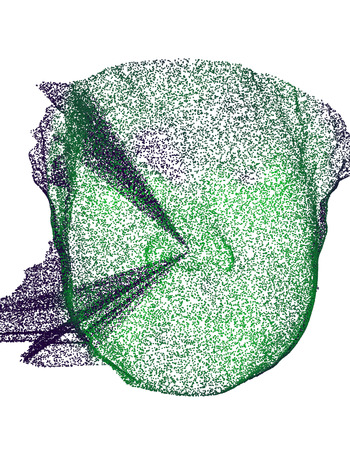} &
        \includegraphics[width=.086\linewidth]{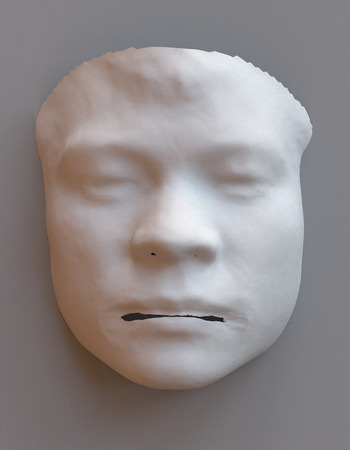} &
        \includegraphics[width=.086\linewidth]{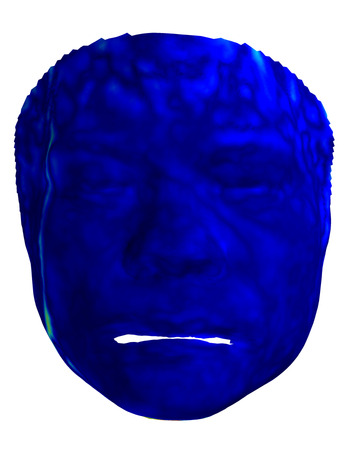}\\

        \includegraphics[width=.086\linewidth]{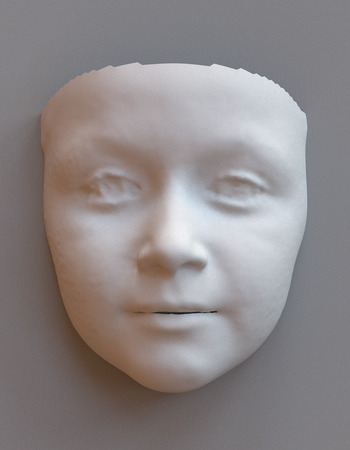} &
        \includegraphics[width=.086\linewidth]{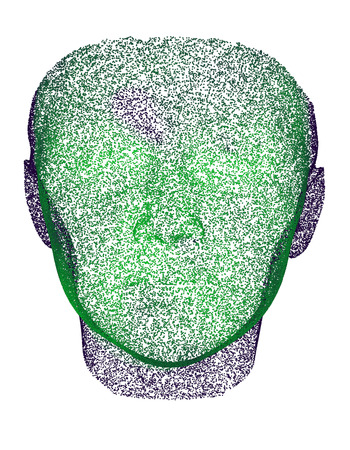} &
        \includegraphics[width=.086\linewidth]{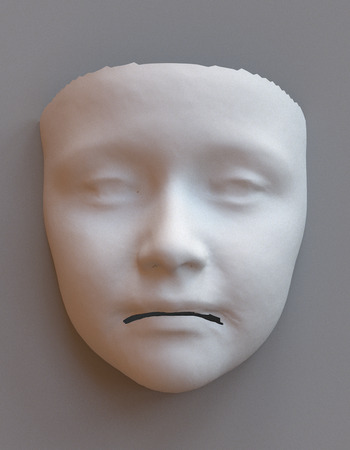} &
        \includegraphics[width=.086\linewidth]{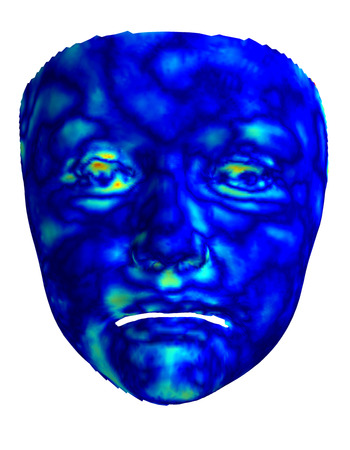} &
        \includegraphics[width=.086\linewidth]{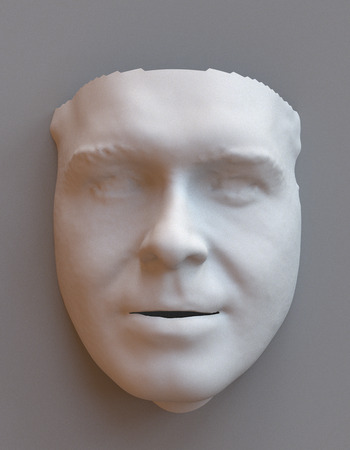} &
        \includegraphics[width=.086\linewidth]{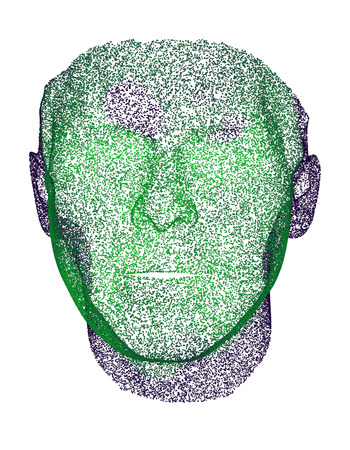} &
        \includegraphics[width=.086\linewidth]{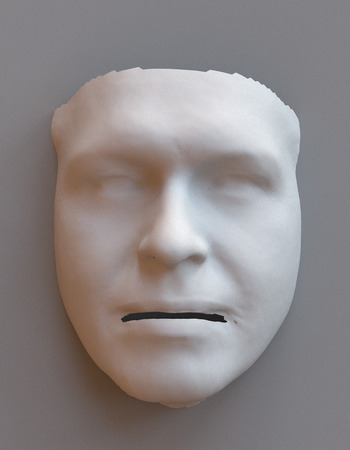} &
        \includegraphics[width=.086\linewidth]{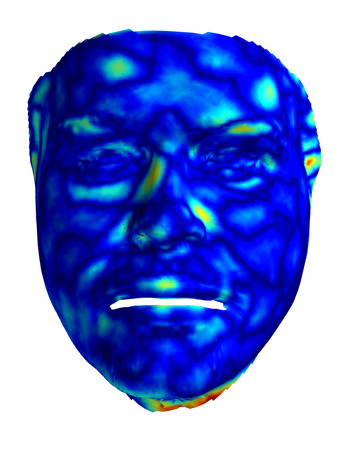}\\
    \end{tabular}
    }
    \includegraphics[width=.5\linewidth]{Fig_colourbar.pdf}
    \caption{\textbf{Sample reconstructions on additional training data for SMF+:} arranged in two columns. From left to right: raw scan, point cloud sampled on the scan by SMF+ and predicted attention mask, output of SMF+, and surface reconstruction error visualized as a texture on the output of SMF+. Top row: 4DFAB, bottom row: MeIn3D.}
    \label{fig:reconst_train_smf_plus}
\end{figure*}

\paragraph{Training and test reconstructions}

We visualize sample reconstructions from SMF on the training and test sets. For each scan, we render the input point cloud sampled on the mesh, and the attention score predicted by SMF for every point as a heatmap, with bright green denoting attention scores close to $1$, and black denoting attention scores close to $0$. We also render the reconstruction produced by SMF, and the heatmap of the surface error as a texture on the registration. We render the reconstruction produced by the baseline for comparison. Figure \ref{fig:sample_reconst_train} provides visualizations for 18 training samples arranged in two columns. Figure \ref{fig:sample_renders_3dmd} shows the comparative performance of the baseline and our model for 12 test scans arranged in two columns. We show sample reconstructions from SMF+ on MeIn3D and 4DFAB in Figure \ref{fig:reconst_train_smf_plus}.

Visual inspection correlates strongly with the numerical evaluation. Our SMF model consistently produces registrations that are smooth and detailed, with very low surface error. The attention mechanism appears to successfully segment the face, eliminating gross corruption, and discarding points from the tongue and teeth for several scans. Our model faithfully represents both the identity and expression, even for extreme expressions on the test set.

In particular, factors such as age, ethnicity, and gender are accurately captured. Non-linear deformations of the nose, cheeks, and mouth are well preserved across a wide range of identities and expressions. Finally, despite the inclusion of points from the teeth and tongue in the raw scans, SMF produces artifact-free and expressive mouth reconstructions with seamless blending in the vast majority of cases.

\subsection{Stability to resampling}
\label{sec:robustness}

Given the stochastic nature of the method, we evaluate the stability of the reconstructions to resampling of the input scans. We then focus on evaluating the attention mechanism.

\begin{figure}[t]
    \centering
    \includegraphics[width=84mm]{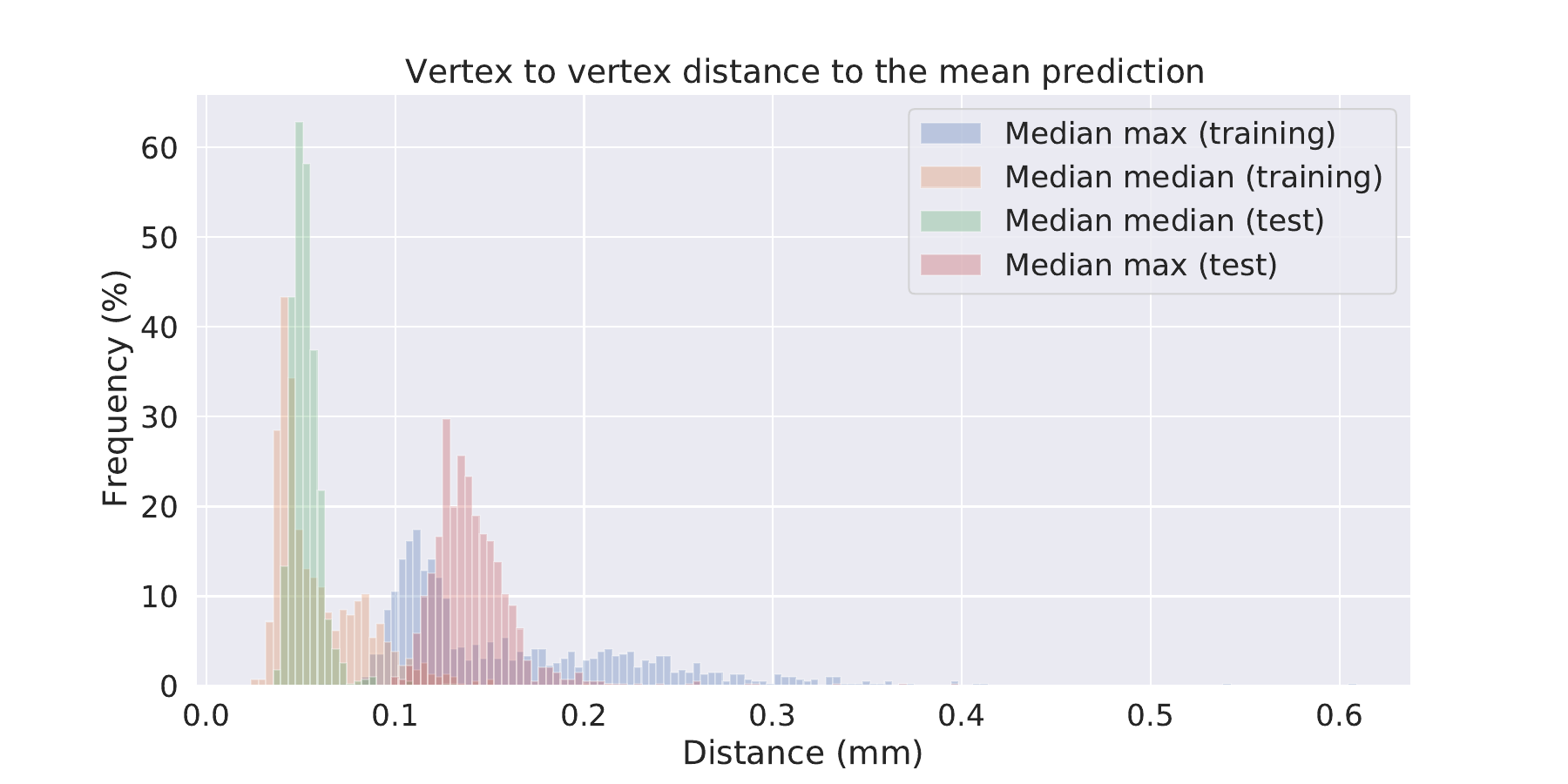}
    \caption{\textbf{Per vertex distance to the mean prediction:} We sample 100 different point clouds for 1000 training and test scans and compute, for each vertex in each registration, its median Euclidean distance to the matching vertex in the average reconstruction. We present histograms of the max and median values (across vertices) per scan to show our method is stable to resampling the same input surface.}%
    \label{fig:histogram_dist_to_mean_prediction}
\end{figure}

\begin{figure}[t]
    \centering
    \includegraphics[width=.35\linewidth]{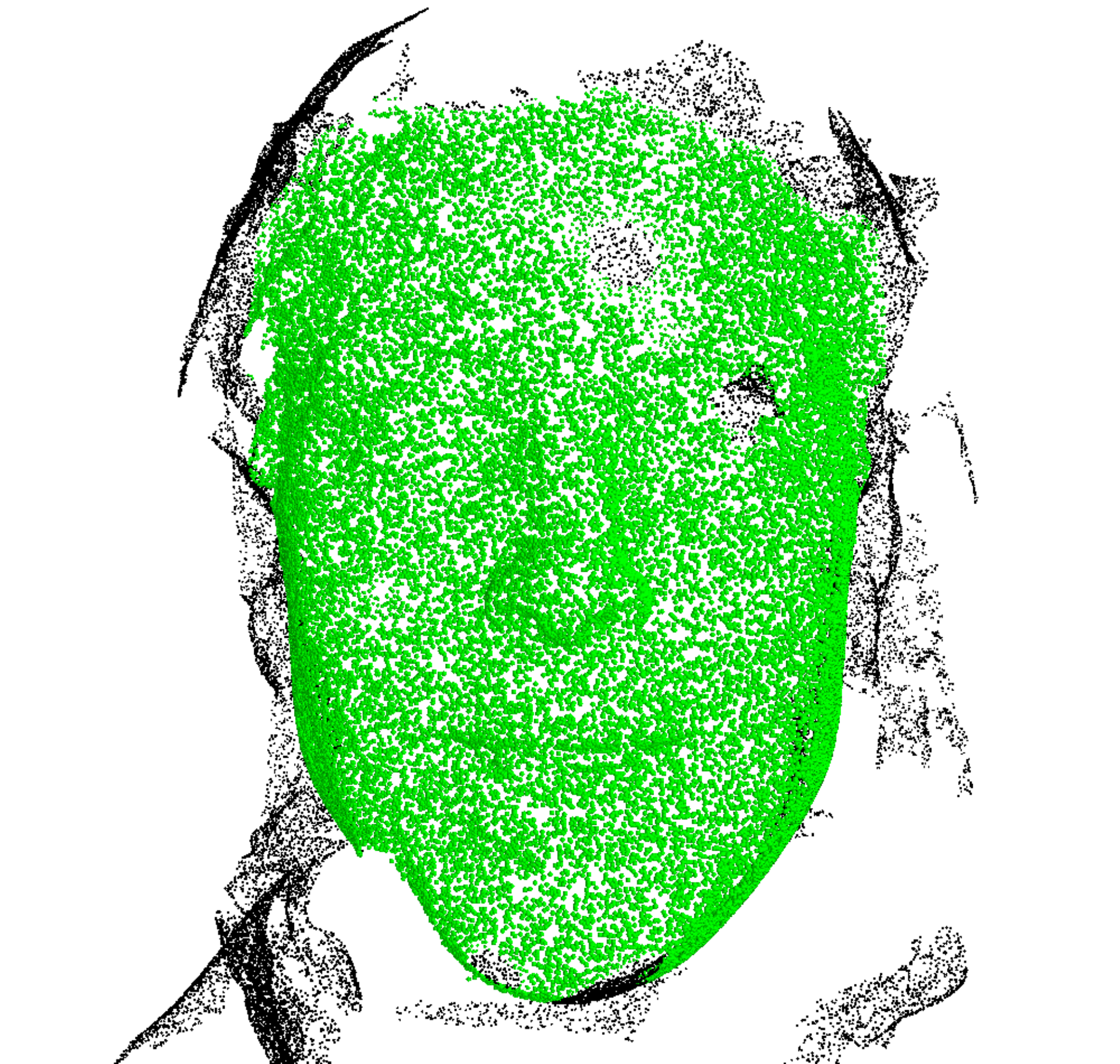}
    \includegraphics[width=.35\linewidth]{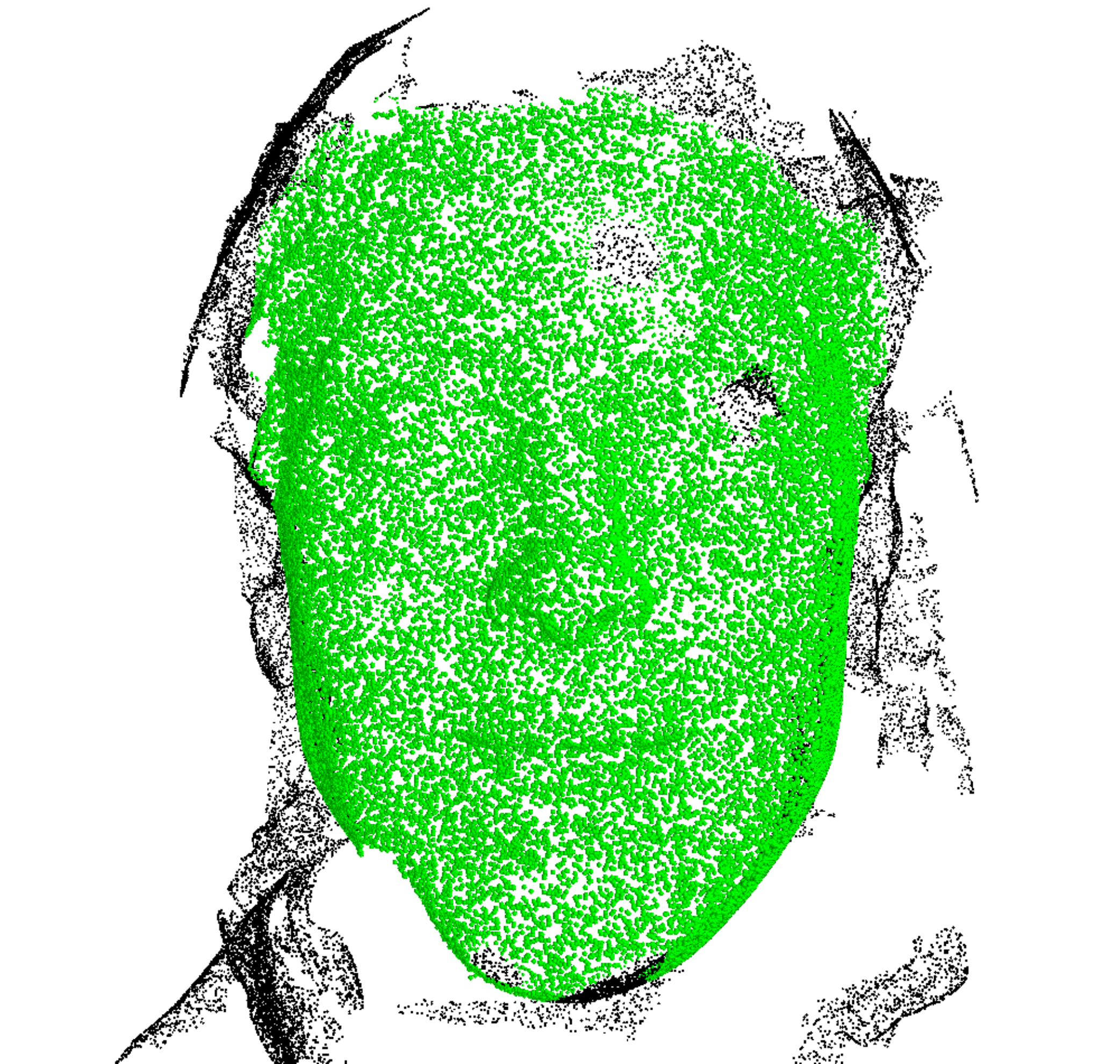}
    \caption{\textbf{Attention mask:} Attention mask for two point clouds sampled from the same \textit{test} shape (3DMD). It can be seen the attention mechanism excludes the points inside of the mouth and outside of the face area. The mask is also stable to resampling of the scan.}
    \label{fig:attentionrobust}
\end{figure}

We select a subset of 1000 scans each of the training and test sets and produce 100 reconstructions with SMF, randomly sampling a new point cloud on the surface of the scan at each iteration. For each scan, we compute the mean reconstruction. For each point of the 100 reconstructions, we compute its Euclidean distance to the matching point in the mean reconstruction for that scan. We then take the median and max of these distances for every point in the scan and compute their median across the scan, denoted by "median median" and "median max", as indications of the typical typical-case and typical worst-case variations. We collect both values for each of the 1000 training and 1000 test scans, and plot their histograms in Figure \ref{fig:histogram_dist_to_mean_prediction}. The results show our method is stable with respect to resampling, the median median variations, in particular, are concentrated below $0.1\mathrm{mm}$ with a typical maximum variation in position from the mean below $0.2\mathrm{mm}$ per vertex. Interestingly, we observe less spread on the test set than on the training set, but slightly higher typical maximum displacement per vertex, still below $0.2\textrm{mm}$ per vertex. Figure \ref{fig:attentionrobust} illustrates that the attention mechanism is also stable.

\subsection{Ablation study on the decoder}
\label{sec:ablation_decoder}

We now study different variations of SMF by changing the decoder. Figure \ref{fig:ablation_decoders} presents the comparative performance of SMF, SMF+ and the ablations measured by average surface reconstruction error and ordered by test error. We also report the landmark localization errors of some of the variants in Table \ref{tab:ablation_lm_bu3d} for BU-3DFE and Table \ref{tab:ablation_lm_3dmd} for 3DMD.

\begin{table*}[t]
    \renewcommand\arraystretch{0.96}
    \centering
    \caption{\textbf{Semantic landmarks error on BU-3DFE for the ablations compared:} Comparison of the mean and standard deviation for semantic landmark error ($mm$) for BU-3DFE using the \textit{BU-3DFE $83$} facial landmark set. Landmark regions are as described in \cite{Salazar2014}. \textit{L} and \textit{R} are shorthand for \textit{Left} and \textit{Right} respectively. \textit{Avg Face} is the average for all inner face landmarks, and therefore excludes \textit{Chin}, \textit{L Face}, and \textit{R Face}. No s.c. is short for no skip connections in the mesh convolutional decoder. Models indicated as "no m.m." have no PCA mouth model or any form of regularization for the mouth region. S.d. stands for single decoder. }
    \resizebox{\linewidth}{!}{
    \begin{tabular}{l|c|c|c|c|c|c|c|c|c|c}
\toprule
Region & SMF & SMF no s.c. & SMF fc & SMF fc' & SMF fc' no m.m. & SMF fc lap=1e-3 & SMF no m.m. & SMF lap=1e-3 & SMF s.d. & SMF 512 s.d.\\
\hline\hline
L Eyebrow & $7.20 {\pm} 2.15$ & $8.56 {\pm} 2.50$ & $7.79 {\pm} 2.14$ & $6.89 {\pm} 2.19$ & $7.14 {\pm} 2.28$ & $8.27 {\pm} 2.41$ & $6.98 {\pm} 2.03$ & $7.93 {\pm} 2.24$ & $7.13 {\pm} 2.18$ & $6.57 {\pm} 2.02$\\
R Eyebrow & $6.70 {\pm} 2.23$ & $8.63 {\pm} 2.60$ & $8.46 {\pm} 2.51$ & $7.00 {\pm} 2.38$ & $7.16 {\pm} 2.35$ & $6.85 {\pm} 2.19$ & $6.47 {\pm} 2.13$ & $6.26 {\pm} 1.96$ & $6.81 {\pm} 2.32$ & $6.42 {\pm} 2.21$\\
L Eye & $3.50 {\pm} 1.10$ & $8.15 {\pm} 1.56$ & $4.97 {\pm} 1.35$ & $3.80 {\pm} 1.24$ & $3.73 {\pm} 1.23$ & $4.71 {\pm} 1.37$ & $3.63 {\pm} 1.11$ & $4.77 {\pm} 1.25$ & $3.46 {\pm} 1.13$ & $3.38 {\pm} 1.10$\\
R Eye & $4.82 {\pm} 1.44$ & $8.11 {\pm} 1.62$ & $6.59 {\pm} 1.69$ & $5.06 {\pm} 1.58$ & $4.97 {\pm} 1.59$ & $5.83 {\pm} 1.64$ & $5.05 {\pm} 1.43$ & $4.36 {\pm} 1.24$ & $5.76 {\pm} 1.57$ & $4.04 {\pm} 1.32$\\
Nose        & $4.62 {\pm} 1.20$ & $5.93 {\pm} 1.41$ & $5.84 {\pm} 1.16$ & $4.38 {\pm} 1.01$ & $4.59 {\pm} 1.01$ & $5.13 {\pm} 1.16$ & $4.51 {\pm} 1.18$ & $4.51 {\pm} 1.21$ & $4.84 {\pm} 1.18$ & $4.73 {\pm} 1.11$\\
Mouth       & $6.25 {\pm} 2.39$ & $6.27 {\pm} 2.13$ & $5.62 {\pm} 2.22$ & $5.17 {\pm} 2.37$ & $5.84 {\pm} 2.35$ & $5.67 {\pm} 2.20$ & $9.54 {\pm} 2.38$ & $6.62 {\pm} 2.29$ & $5.96 {\pm} 2.39$ & $6.09 {\pm} 2.42$\\
\hline
Chin        & $38.73 {\pm} 9.18$ & $38.57 {\pm} 8.67$ & $33.38 {\pm} 8.39$ & $27.87 {\pm} 8.18$ & $28.27 {\pm} 8.11$ & $16.40 {\pm} 6.84$ & $37.24 {\pm} 9.14$ & $17.84 {\pm} 6.61$ & $31.74 {\pm} 8.70$ & $34.01 {\pm} 8.82$\\
L Face & $14.08 {\pm} 3.64$ & $13.81 {\pm} 3.41$ & $13.94 {\pm} 5.01$ & $12.00 {\pm} 4.45$ & $11.73 {\pm} 4.38$ & $13.57 {\pm} 5.23$ & $13.96 {\pm} 3.63$ & $11.99 {\pm} 4.25$ & $12.83 {\pm} 3.57$ & $13.43 {\pm} 3.66$\\
R Face & $20.61 {\pm} 5.19$ & $18.92 {\pm} 4.75$ & $14.23 {\pm} 3.39$ & $13.33 {\pm} 3.86$ & $13.44 {\pm} 3.88$ & $12.51 {\pm} 3.65$ & $19.80 {\pm} 5.10$ & $13.20 {\pm} 3.86$ & $17.97 {\pm} 4.90$ & $19.35 {\pm} 5.04$\\
\hline
Avg Face & $5.70 {\pm} 1.11$ & $7.43 {\pm} 1.35$ & $6.60 {\pm} 1.12$ & $5.48 {\pm} 1.10$ & $5.72 {\pm} 1.08$ & $6.15 {\pm} 1.09$ & $6.34 {\pm} 1.08$ & $5.90 {\pm} 1.07$ & $5.79 {\pm} 1.13$ & $5.43 {\pm} 1.09$\\
Avg         & $9.07 {\pm} 1.28$ & $10.25 {\pm} 1.37$ & $9.00 {\pm} 1.21$ & $7.64 {\pm} 1.22$ & $7.83 {\pm} 1.22$ & $7.74 {\pm} 1.26$ & $9.43 {\pm} 1.30$ & $7.53 {\pm} 1.20$ & $8.51 {\pm} 1.28$ & $8.49 {\pm} 1.30$\\
\bottomrule
    \end{tabular}
    }
    \label{tab:ablation_lm_bu3d}
\end{table*}

\begin{table*}[t]
    \renewcommand\arraystretch{0.96}
    \centering
    \caption{\textbf{Semantic landmarks error on 3DMD for the ablations compared:} Comparison of the mean and standard deviation for semantic landmark error ($mm$) for 3DMD using the \textit{ibug $68$} facial landmark set. \textit{L} and \textit{R} are shorthand for \textit{Left} and \textit{Right} respectively. \textit{Avg} is the average for all inner face landmarks.  No s.c. is short for no skip connections in the mesh convolutional decoder. Models indicated as "no m.m." have no PCA mouth model or any form of regularization for the mouth region. S.d. stands for single decoder.}
    \resizebox{\linewidth}{!}{
    \begin{tabular}{l|c|c|c|c|c|c|c|c|c|c}
\toprule
Region & SMF & SMF no s.c. & SMF fc & SMF fc' & SMF fc' no m.m. & SMF fc lap=1e-3 & SMF no m.m. & SMF lap=1e-3 & SMF s.d. & SMF 512 s.d.\\
\hline\hline
L Eyebrow & $5.55 {\pm} 1.76$ & $5.28 {\pm} 1.75$ & $5.12 {\pm} 1.53$ & $6.45 {\pm} 1.52$ & $6.03 {\pm} 1.58$ & $5.49 {\pm} 1.87$ & $5.45 {\pm} 1.44$ & $4.87 {\pm} 1.57$ & $4.91 {\pm} 1.44$ & $6.19 {\pm} 1.47$\\
R Eyebrow & $6.29 {\pm} 2.32$ & $5.93 {\pm} 2.16$ & $4.84 {\pm} 1.79$ & $6.57 {\pm} 1.91$ & $6.42 {\pm} 1.83$ & $6.09 {\pm} 2.26$ & $6.19 {\pm} 1.93$ & $4.79 {\pm} 1.81$ & $4.80 {\pm} 1.77$ & $5.67 {\pm} 1.68$\\
L Eye & $4.26 {\pm} 1.07$ & $4.37 {\pm} 1.10$ & $3.69 {\pm} 1.14$ & $5.97 {\pm} 1.41$ & $6.13 {\pm} 1.33$ & $4.43 {\pm} 1.07$ & $3.99 {\pm} 1.06$ & $3.75 {\pm} 1.14$ & $3.67 {\pm} 1.15$ & $6.06 {\pm} 1.44$\\
R Eye & $4.03 {\pm} 1.31$ & $3.76 {\pm} 1.20$ & $3.52 {\pm} 1.26$ & $5.75 {\pm} 1.50$ & $5.52 {\pm} 1.39$ & $4.38 {\pm} 1.26$ & $3.04 {\pm} 1.11$ & $3.58 {\pm} 1.27$ & $3.61 {\pm} 1.31$ & $6.28 {\pm} 1.48$\\
Nose & $5.26 {\pm} 0.87$ & $4.77 {\pm} 0.78$ & $3.84 {\pm} 0.81$ & $5.66 {\pm} 0.86$ & $4.84 {\pm} 0.81$ & $5.29 {\pm} 0.87$ & $5.19 {\pm} 0.86$ & $4.24 {\pm} 0.89$ & $3.99 {\pm} 0.82$ & $5.25 {\pm} 0.88$\\
Mouth & $6.31 {\pm} 1.19$ & $6.09 {\pm} 1.23$ & $4.87 {\pm} 1.23$ & $4.96 {\pm} 1.32$ & $5.10 {\pm} 1.27$ & $9.91 {\pm} 1.24$ & $6.71 {\pm} 1.17$ & $4.98 {\pm} 1.23$ & $5.88 {\pm} 1.19$ & $4.64 {\pm} 1.29$\\
\hline
Jaw & $24.74 {\pm} 5.95$ & $25.23 {\pm} 6.05$ & $27.87 {\pm} 5.77$ & $27.34 {\pm} 5.78$ & $27.15 {\pm} 5.77$ & $24.57 {\pm} 5.98$ & $28.46 {\pm} 5.75$ & $27.91 {\pm} 5.77$ & $27.48 {\pm} 5.78$ & $26.68 {\pm} 5.79$\\
\hline
Avg Face & $5.54 {\pm} 0.85$ & $5.29 {\pm} 0.83$ & $4.41 {\pm} 0.81$ & $5.60 {\pm} 0.89$ & $5.45 {\pm} 0.85$ & $7.00 {\pm} 0.87$ & $5.52 {\pm} 0.81$ & $4.51 {\pm} 0.83$ & $4.82 {\pm} 0.79$ & $5.36 {\pm} 0.86$\\
Avg & $10.34 {\pm} 1.92$ & $10.27 {\pm} 1.94$ & $10.28 {\pm} 1.83$ & $11.04 {\pm} 1.88$ & $10.87 {\pm} 1.86$ & $11.39 {\pm} 1.94$ & $11.25 {\pm} 1.84$ & $10.36 {\pm} 1.84$ & $10.48 {\pm} 1.83$ & $10.69 {\pm} 1.86$\\
\bottomrule
    \end{tabular}
    }
    \label{tab:ablation_lm_3dmd}
\end{table*}

\subsubsection{Single decoder}

While two or more decoders can be used to promote separation between factors of variation in the data, and ties to the morphable model and generative model aspect of our work, our registration framework is equally applicable to single decoders (abbreviated s.d.). We keep the architecture of Section \ref{sec:conv_decoders} and the mouth models of Section \ref{sec:mouth_model}, and set the dimension of the latent space to 256 ("SMF s.d.") and 512 ("SMF 512 s.d."). As can be seen in Figure \ref{fig:ablation_decoders}, SMF and SMF 512 s.d. have similar average error and error variance, with SMF 512 s.d. slightly outperforming SMF, while SMF s.d. shows a slightly greater drop in performance. These results are expected: training a single decoder is no harder than training two and using a single mouth model with both identity and expression bases is also simpler, but the single latent space of dimension 256 in SMF s.d. further constraints the model compared to SMF. Sample reconstructions are presented in Figure \ref{fig:ablation_decoders_renders}.

\subsubsection{Fully-connected decoders}

We investigate whether the improvements in the encoder and training methodology enable generalization with fully-connected decoders, and how the performance of such decoders compares to that of our mesh convolutional decoders. We follow the same architecture as \cite{Liu_2019_ICCV} for the decoders. The models compared are: SMF fc, obtained by substituting the mesh inception decoders with MLPs, keeping the dimension of the identity and expression latent spaces to 256 and all other hyperparameters identical and SMF 512 fc with latent spaces of dimension 512.

\begin{figure}[t]
    \centering
    \includegraphics[width=84mm]{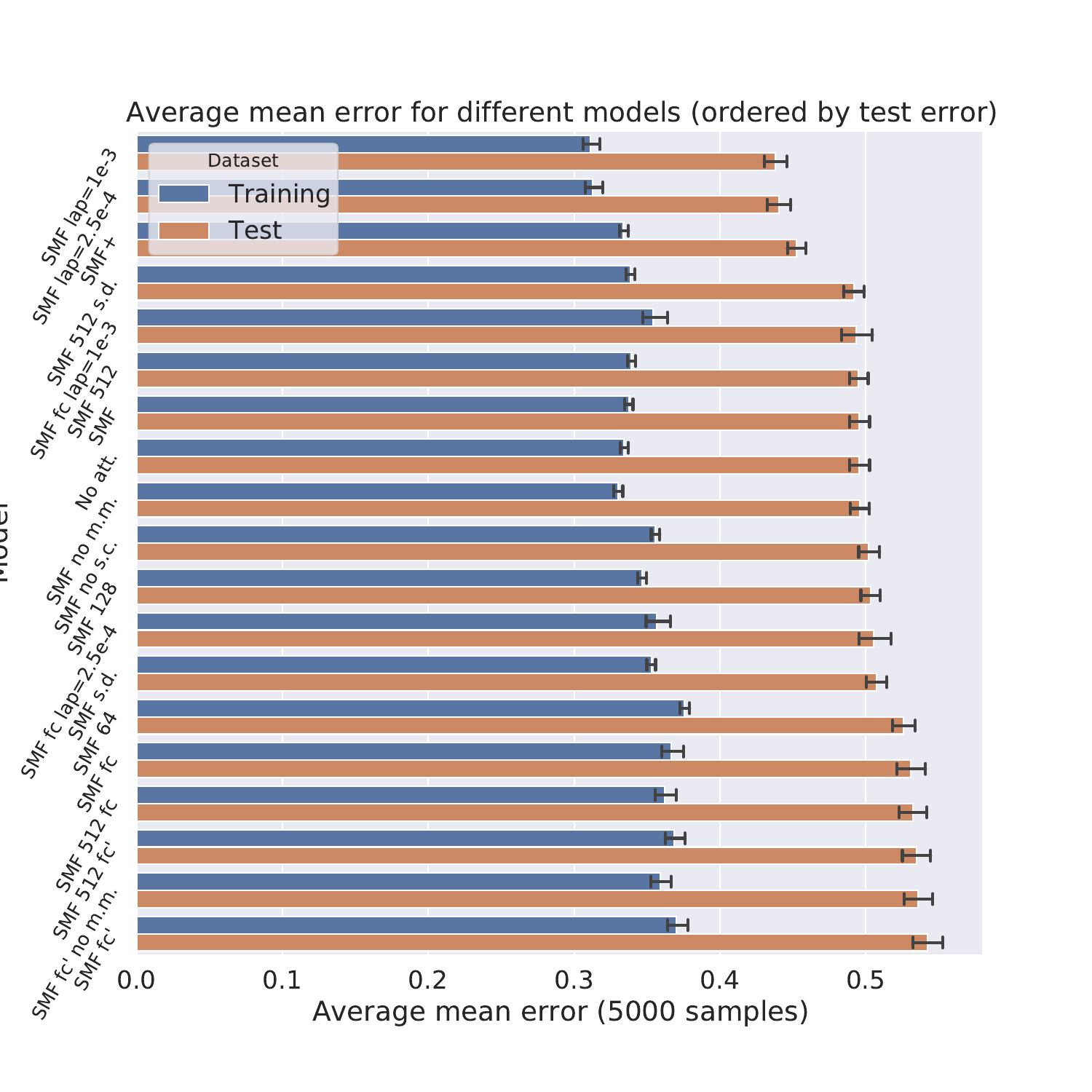}
    \caption{\textbf{Comparison of the average mean (per scan) surface fitting error} for different choices of decoders, on 5000 random training scans and 5000 random test scans, ordered by average test error.}
    \label{fig:ablation_decoders}
\end{figure}

As reported in Figure \ref{fig:ablation_decoders}, SMF outperforms both variants in terms of surface error, while the fc variants performed better in terms of landmark error. Visualizing the reconstructions in Figure \ref{fig:ablation_decoders_renders}, however, shows heavy noise. We therefore increased the edge-length regularization to $\lambda_{edge} = 2e-4$ and re-evaluated the models (now SMF fc' and SMF 512 fc'). The models with increased edge-length regularization provided smoother reconstructions, but still suffered from artifacting and also performed worse in both surface error and landmark error. This ablation confirms that, in order to obtain reconstructions that are free from noise and large variations in curvature with fully-connected decoders, increased regularization is required, at some cost in accuracy. It is also apparent that some error metrics, such as landmark localization error, favor models that fit the positions of individual vertices at the expense of surface fairness.

It is worth noting that SMF with fully-connected decoders generalizes well to the test set, and that fully-connected decoders may in some cases provide finer details, albeit with additional noise. Our mesh inception decoders, however, achieve comparable performance, with no noise and with a fraction of the trainable parameters. We also note that the variance of the mean surface error is higher with fully-connected decoders, as indicated by the wider error bars in Figure \ref{fig:ablation_decoders}.

\subsubsection{Skip connections}

We now compare our mesh inception decoders with standard SpiralNet++ decoders, keeping all hyperparameters equal, and report the performance of "SMF no s.c.". The model without skip connections performed noticeably worse than SMF in terms of surface error, and landmark error on BU-3DFE, but was slightly better on the landmark localization task on 3DMD. Visual inspection in Figure \ref{fig:ablation_decoders_renders} reveals the presence of artifacts, especially around the mouth area.

\subsubsection{Mouth model}

As stated in Section \ref{sec:mouth_model}, the purpose of introducing a constrained PCA model for the mouth is to produce reconstructions that do not display unnatural deformations of the mouth in the presence of noisy points from the teeth or tongue, by finding a trade-off between model expressivity and robustness. Thus, it is expected that using the PCA model may come at a loss of precision. 

\begin{figure*}[t]
    \centering
    \setlength\tabcolsep{1.5pt}
    {\small
    \begin{tabular}{c|ccccc@{\hskip 1mm}|@{\hskip 1mm}ccccc}

& \multicolumn{5}{c}{Training} & \multicolumn{5}{c}{Test}\\

\rotatebox[origin=c]{90}{Raw} &
\includegraphics[align=c,width=.09\linewidth]{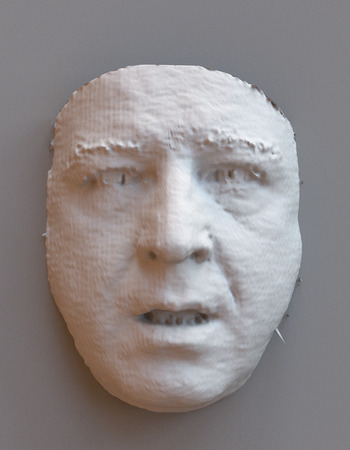} &
\includegraphics[align=c,width=.09\linewidth]{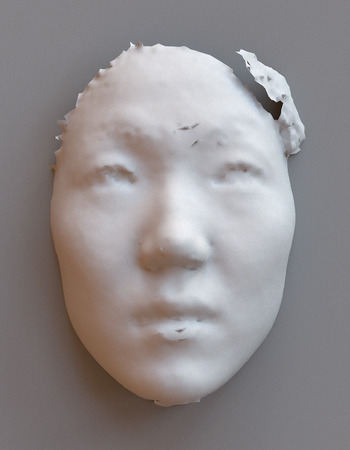} &
\includegraphics[align=c,width=.09\linewidth]{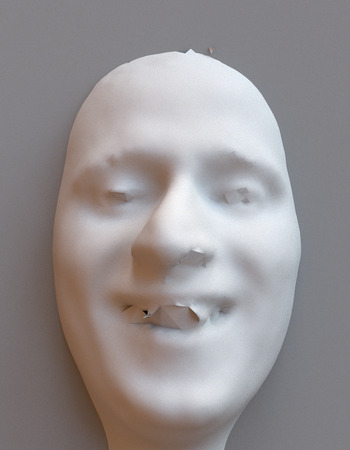} &
\includegraphics[align=c,width=.09\linewidth]{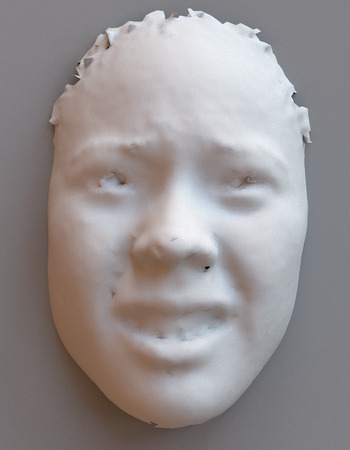} &
\includegraphics[align=c,width=.09\linewidth]{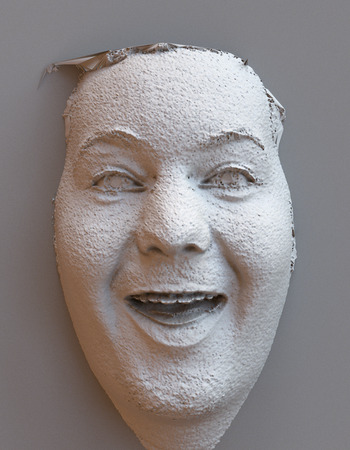} &
\includegraphics[align=c,width=.09\linewidth]{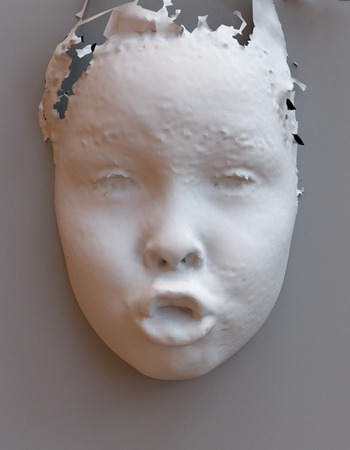} &
\includegraphics[align=c,width=.09\linewidth]{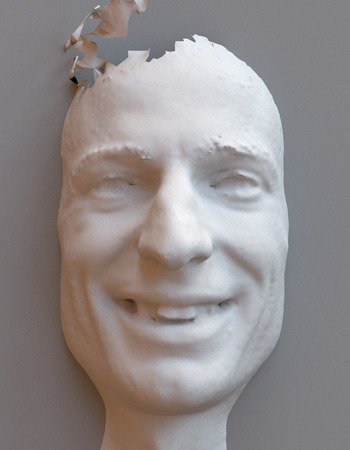} &
\includegraphics[align=c,width=.09\linewidth]{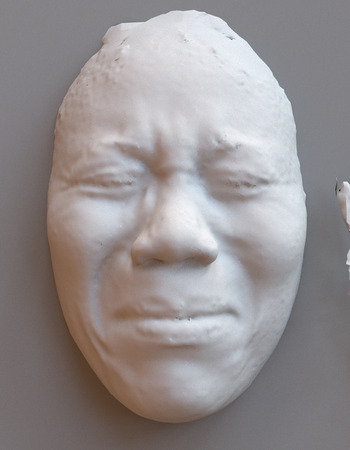} &
\includegraphics[align=c,width=.09\linewidth]{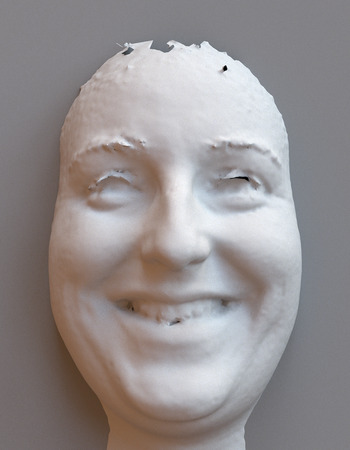} &
\includegraphics[align=c,width=.09\linewidth]{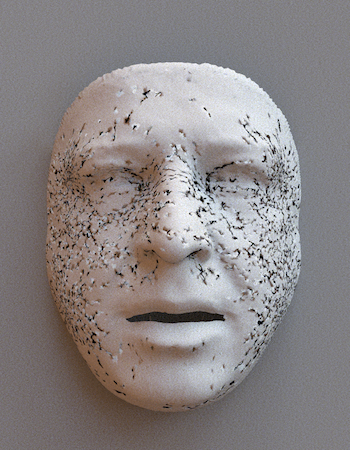}\\

\rotatebox[origin=c]{90}{SMF fc} &
\includegraphics[align=c,width=.09\linewidth]{Fig20_10.jpg} &
\includegraphics[align=c,width=.09\linewidth]{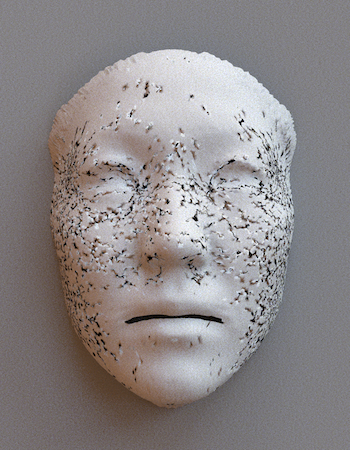} &
\includegraphics[align=c,width=.09\linewidth]{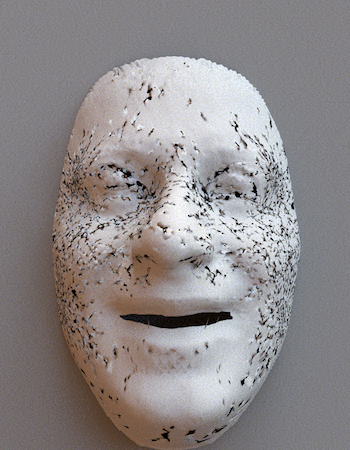} &
\includegraphics[align=c,width=.09\linewidth]{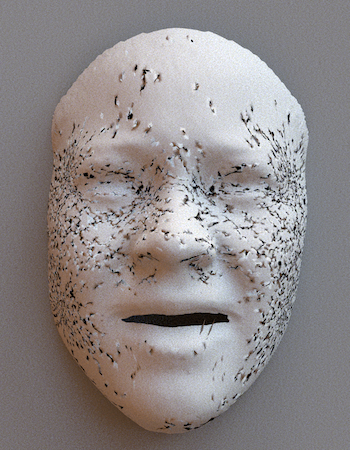} &
\includegraphics[align=c,width=.09\linewidth]{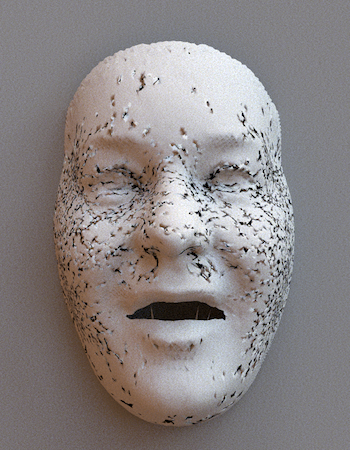} &
\includegraphics[align=c,width=.09\linewidth]{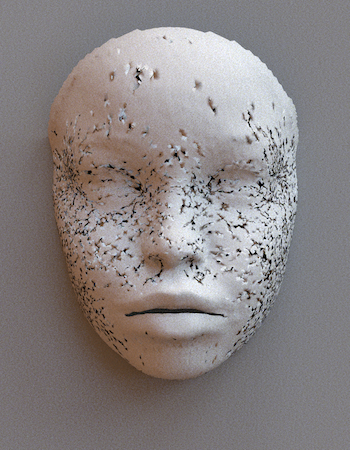} &
\includegraphics[align=c,width=.09\linewidth]{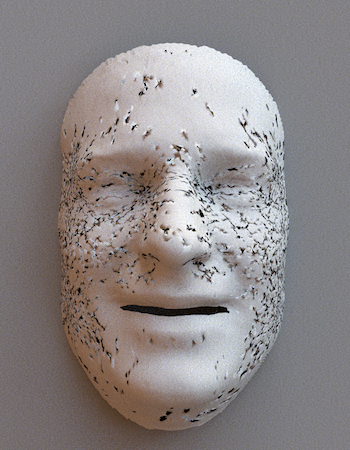} &
\includegraphics[align=c,width=.09\linewidth]{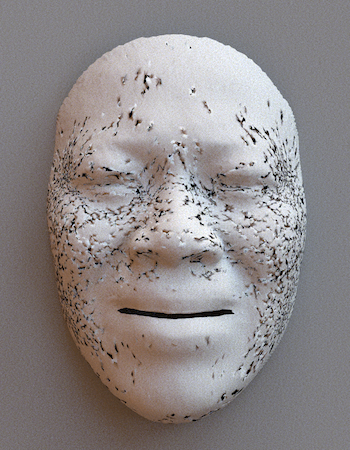} &
\includegraphics[align=c,width=.09\linewidth]{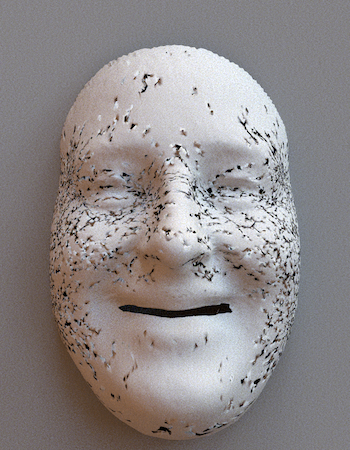} &
\includegraphics[align=c,width=.09\linewidth]{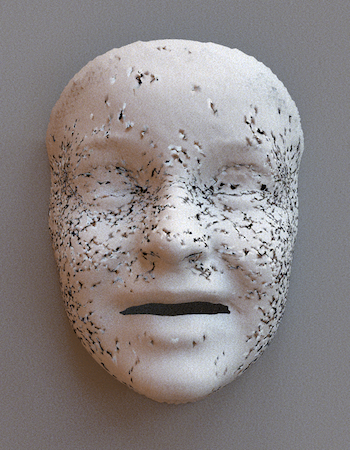}\\

\rotatebox[origin=c]{90}{SMF fc'} &
\includegraphics[align=c,width=.09\linewidth]{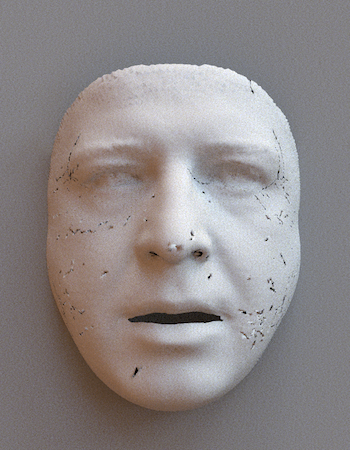} &
\includegraphics[align=c,width=.09\linewidth]{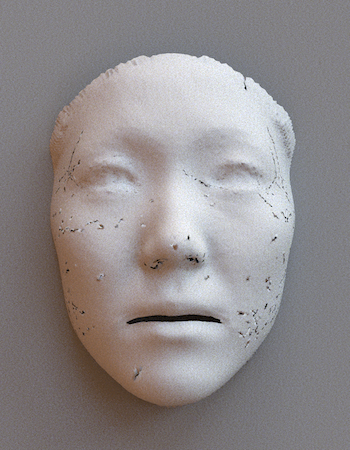} &
\includegraphics[align=c,width=.09\linewidth]{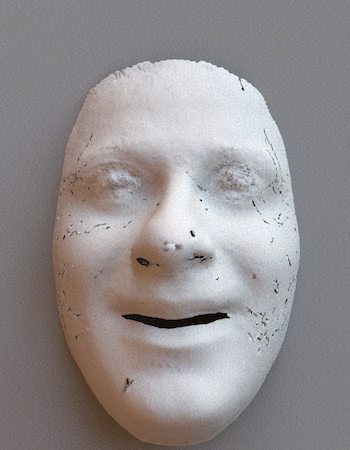} &
\includegraphics[align=c,width=.09\linewidth]{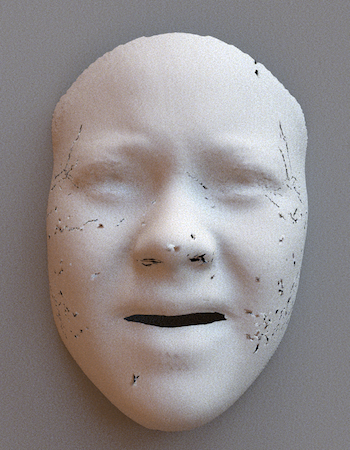} &
\includegraphics[align=c,width=.09\linewidth]{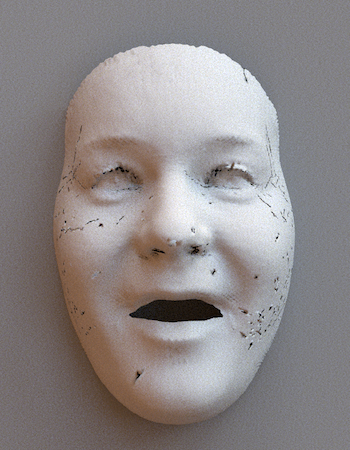} &
\includegraphics[align=c,width=.09\linewidth]{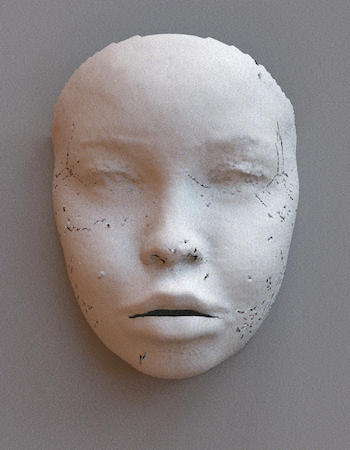} &
\includegraphics[align=c,width=.09\linewidth]{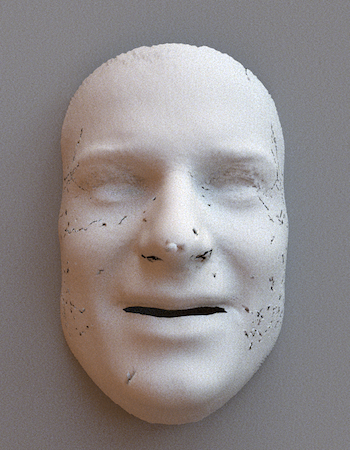} &
\includegraphics[align=c,width=.09\linewidth]{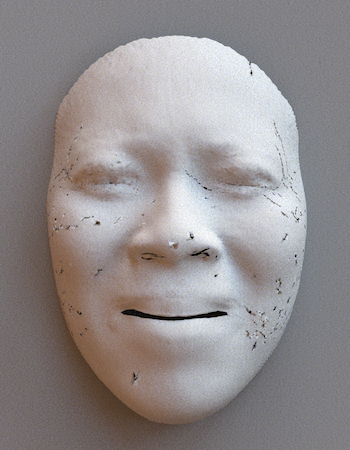} &
\includegraphics[align=c,width=.09\linewidth]{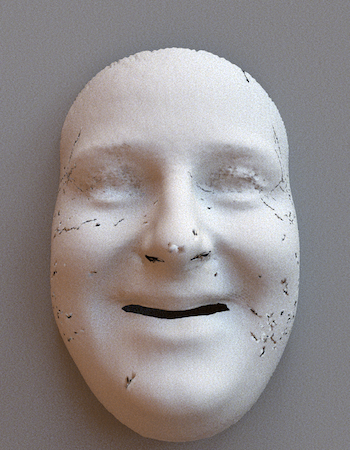} &
\includegraphics[align=c,width=.09\linewidth]{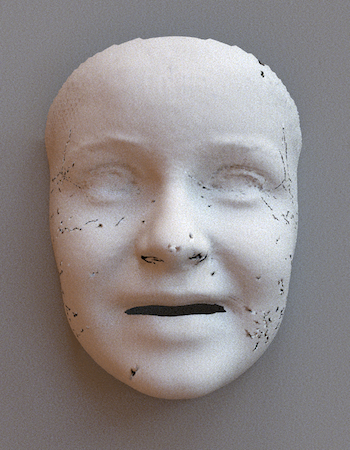}\\

\rotatebox[origin=c]{90}{SMF lap=1e-3} &
\includegraphics[align=c,width=.09\linewidth]{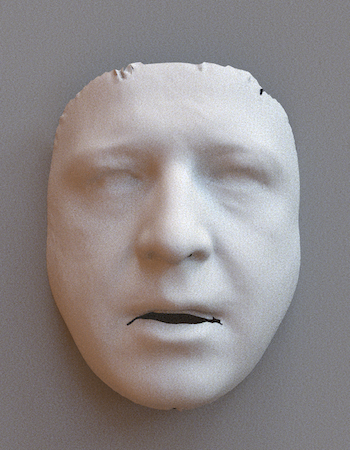} &
\includegraphics[align=c,width=.09\linewidth]{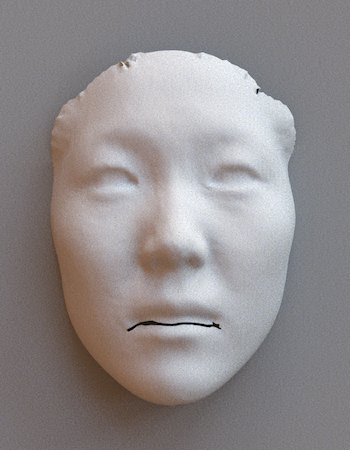} &
\includegraphics[align=c,width=.09\linewidth]{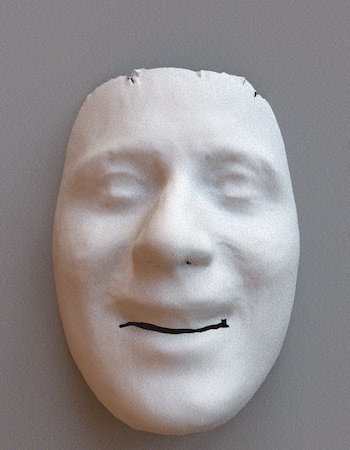} &
\includegraphics[align=c,width=.09\linewidth]{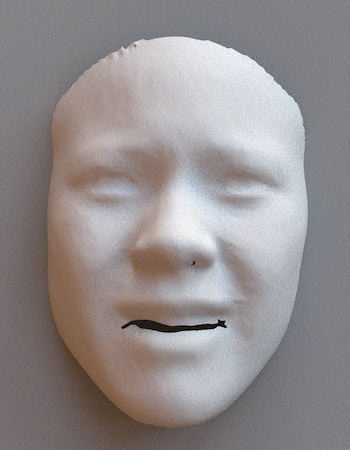} &
\includegraphics[align=c,width=.09\linewidth]{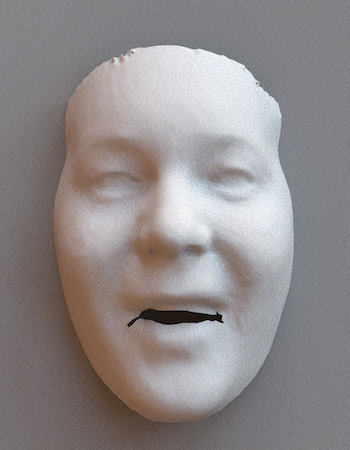} &
\includegraphics[align=c,width=.09\linewidth]{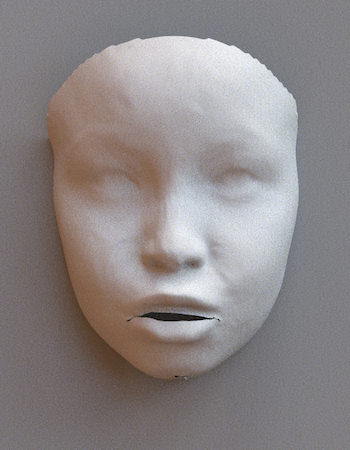} &
\includegraphics[align=c,width=.09\linewidth]{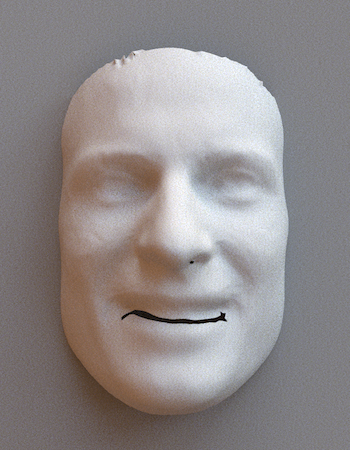} &
\includegraphics[align=c,width=.09\linewidth]{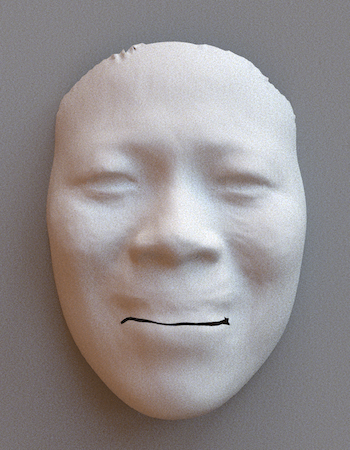} &
\includegraphics[align=c,width=.09\linewidth]{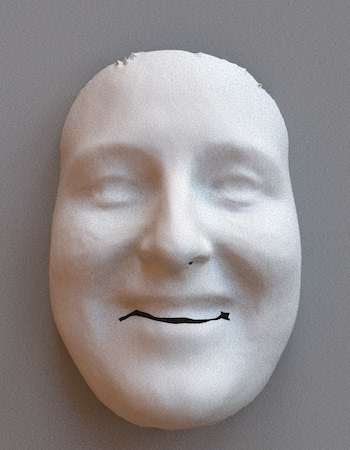} &
\includegraphics[align=c,width=.09\linewidth]{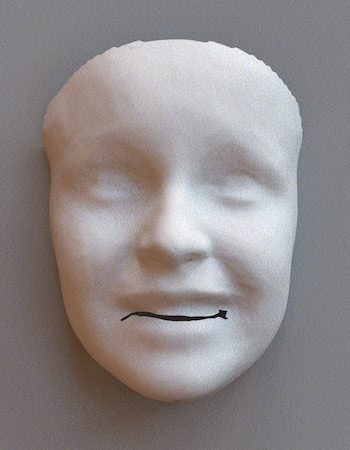}\\

\rotatebox[origin=c]{90}{SMF fc lap=1e-3} &
\includegraphics[align=c,width=.09\linewidth]{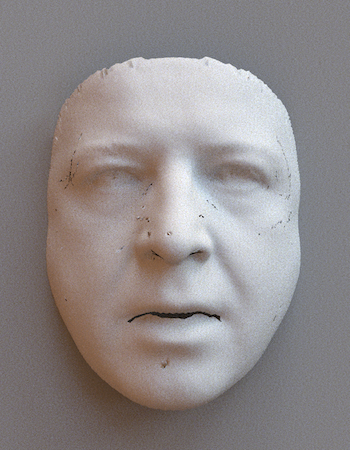} &
\includegraphics[align=c,width=.09\linewidth]{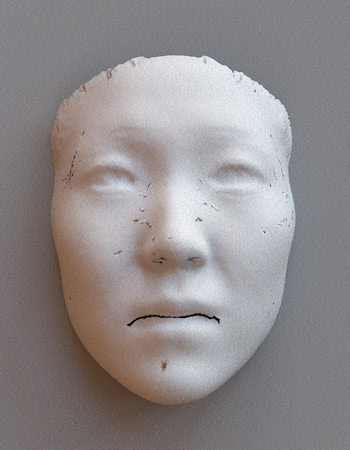} &
\includegraphics[align=c,width=.09\linewidth]{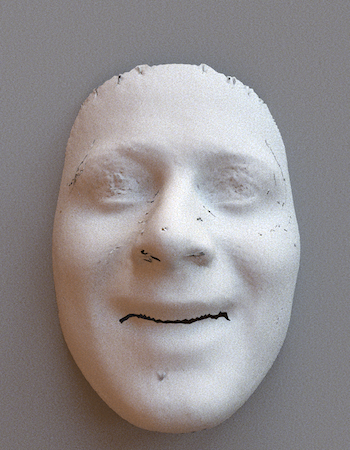} &
\includegraphics[align=c,width=.09\linewidth]{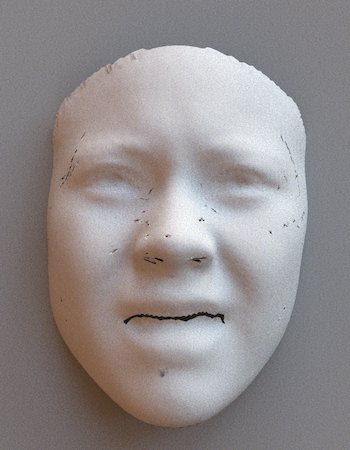} &
\includegraphics[align=c,width=.09\linewidth]{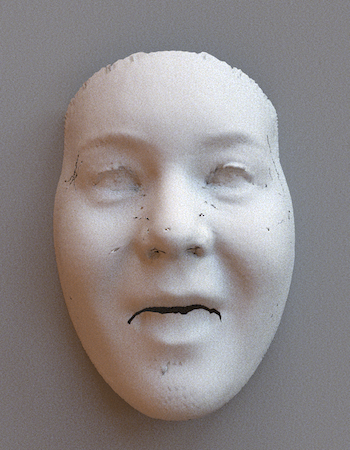} &
\includegraphics[align=c,width=.09\linewidth]{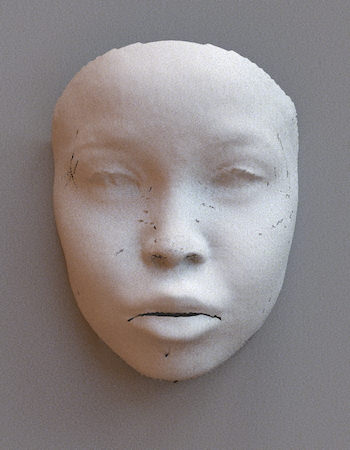} &
\includegraphics[align=c,width=.09\linewidth]{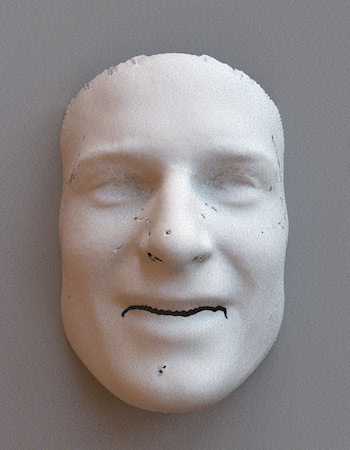} &
\includegraphics[align=c,width=.09\linewidth]{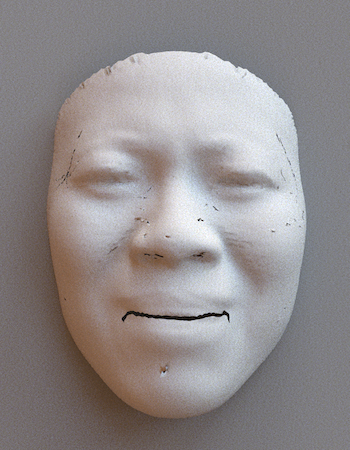} &
\includegraphics[align=c,width=.09\linewidth]{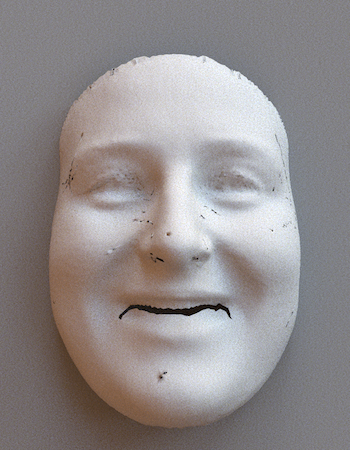} &
\includegraphics[align=c,width=.09\linewidth]{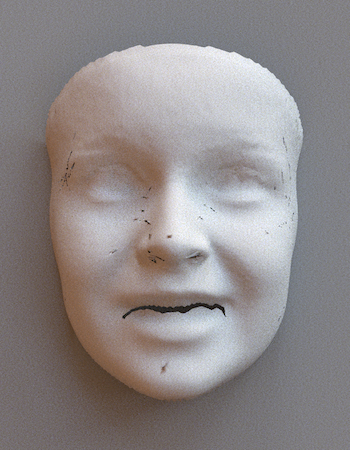}\\

\rotatebox[origin=c]{90}{SMF 512 s.d.} &
\includegraphics[align=c,width=.09\linewidth]{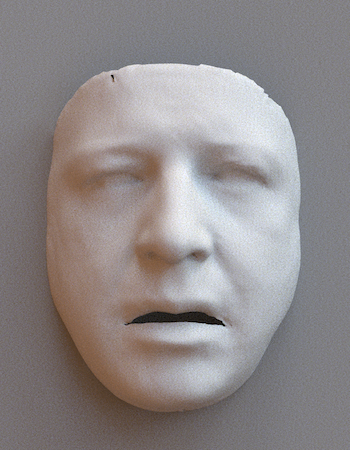} &
\includegraphics[align=c,width=.09\linewidth]{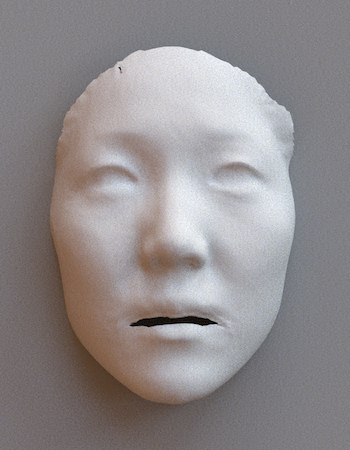} &
\includegraphics[align=c,width=.09\linewidth]{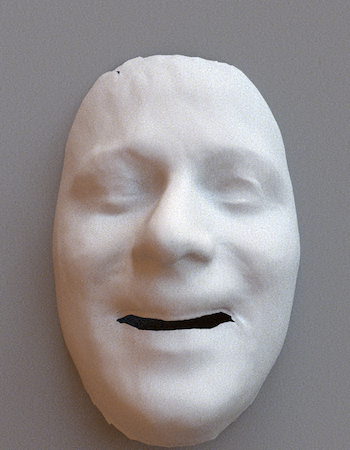} &
\includegraphics[align=c,width=.09\linewidth]{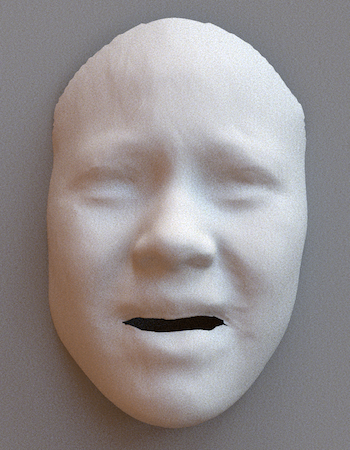} &
\includegraphics[align=c,width=.09\linewidth]{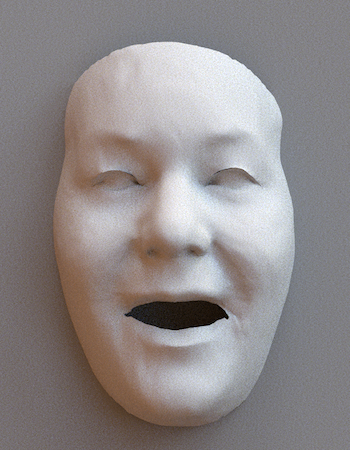} &
\includegraphics[align=c,width=.09\linewidth]{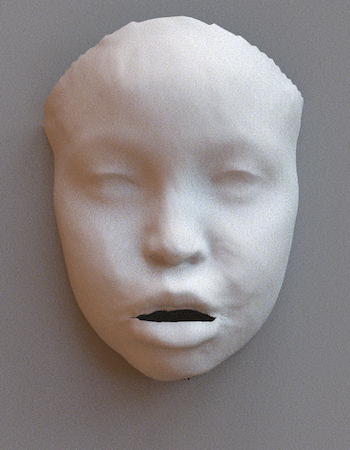} &
\includegraphics[align=c,width=.09\linewidth]{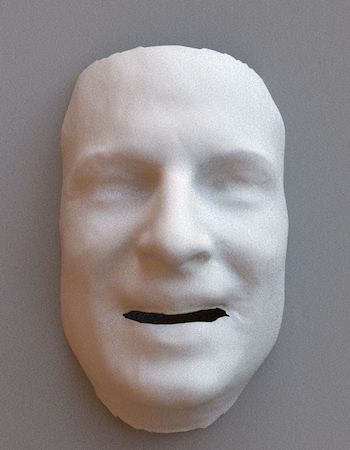} &
\includegraphics[align=c,width=.09\linewidth]{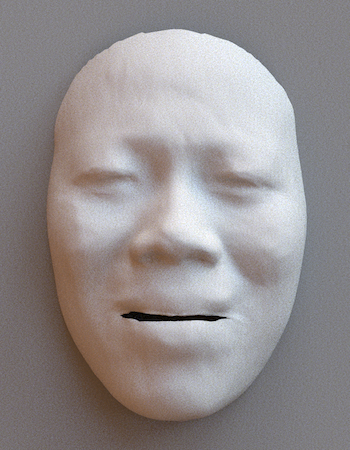} &
\includegraphics[align=c,width=.09\linewidth]{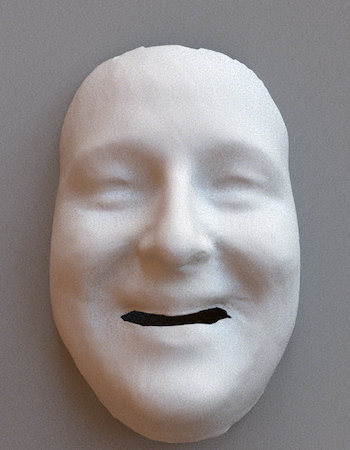} &
\includegraphics[align=c,width=.09\linewidth]{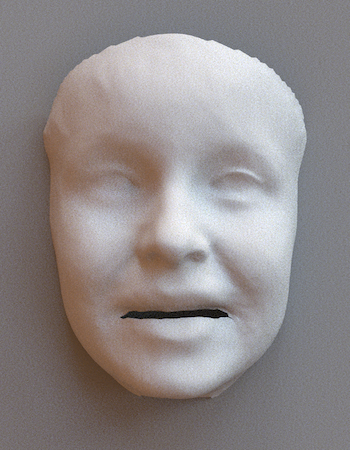}\\

\rotatebox[origin=c]{90}{SMF no m.m.} &
\includegraphics[align=c,width=.09\linewidth]{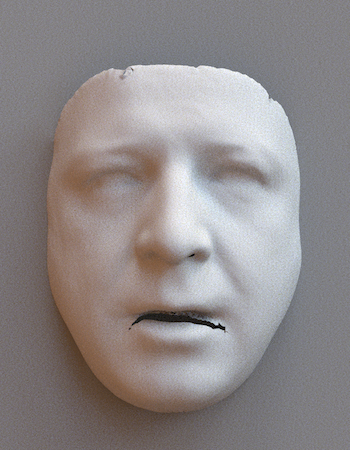} &
\includegraphics[align=c,width=.09\linewidth]{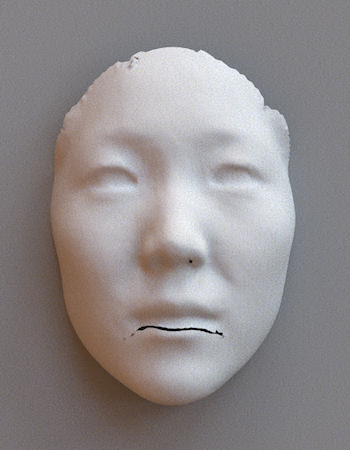} &
\includegraphics[align=c,width=.09\linewidth]{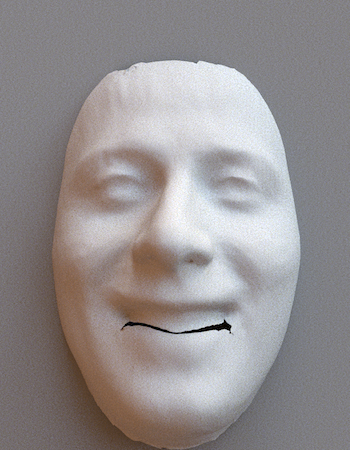} &
\includegraphics[align=c,width=.09\linewidth]{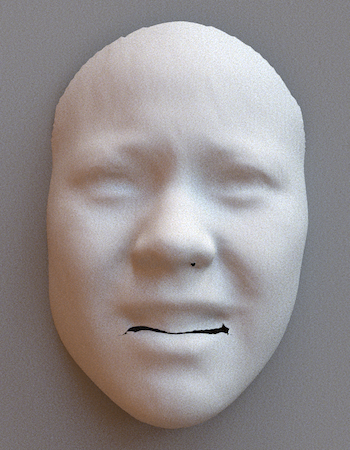} &
\includegraphics[align=c,width=.09\linewidth]{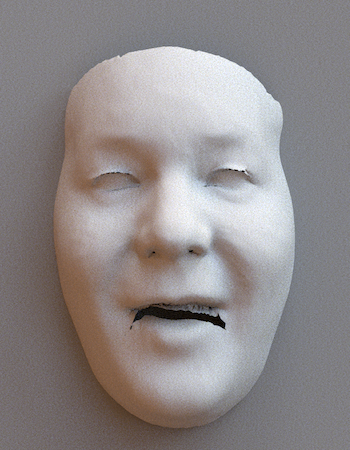} &
\includegraphics[align=c,width=.09\linewidth]{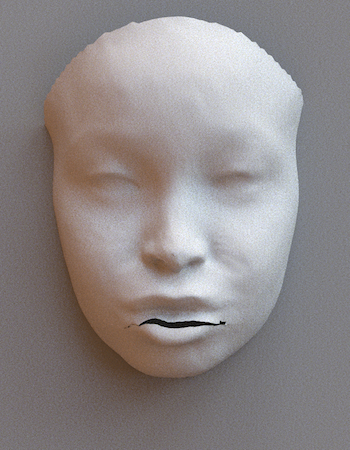} &
\includegraphics[align=c,width=.09\linewidth]{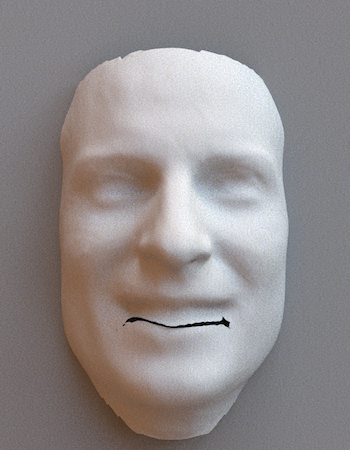} &
\includegraphics[align=c,width=.09\linewidth]{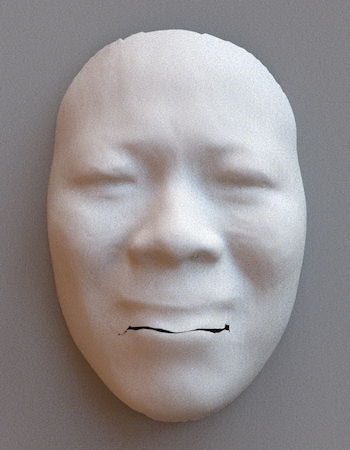} &
\includegraphics[align=c,width=.09\linewidth]{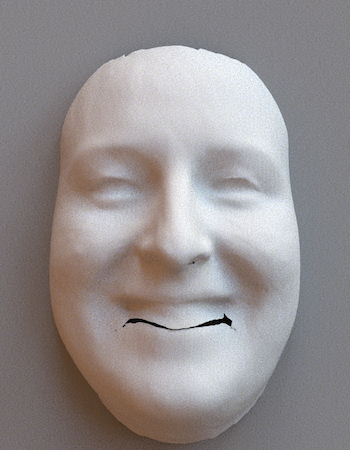} &
\includegraphics[align=c,width=.09\linewidth]{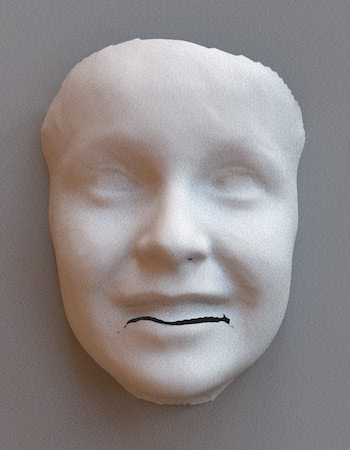}\\

\rotatebox[origin=c]{90}{SMF fc' no m.m.} &
\includegraphics[align=c,width=.09\linewidth]{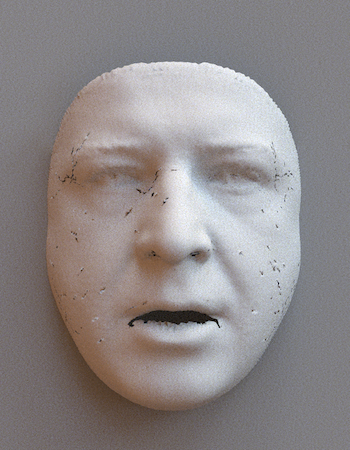} &
\includegraphics[align=c,width=.09\linewidth]{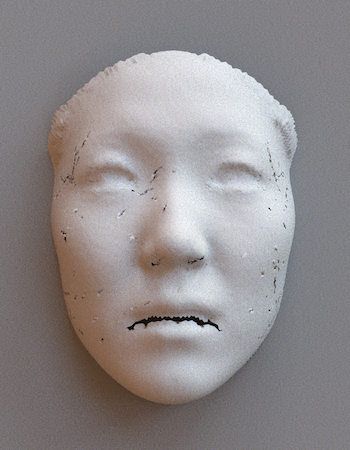} &
\includegraphics[align=c,width=.09\linewidth]{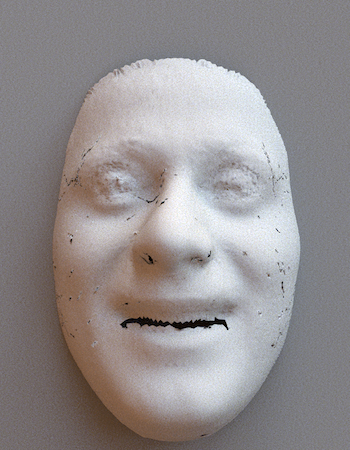} &
\includegraphics[align=c,width=.09\linewidth]{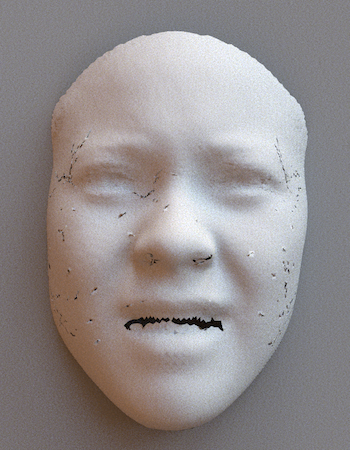} &
\includegraphics[align=c,width=.09\linewidth]{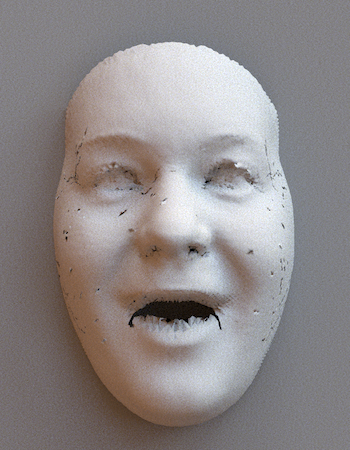} &
\includegraphics[align=c,width=.09\linewidth]{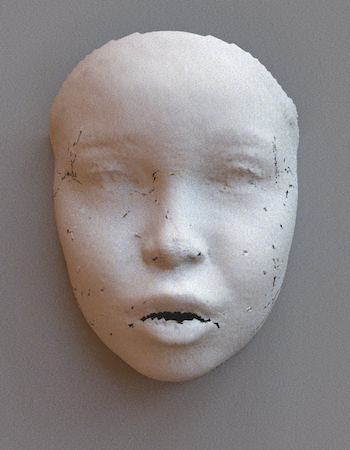} &
\includegraphics[align=c,width=.09\linewidth]{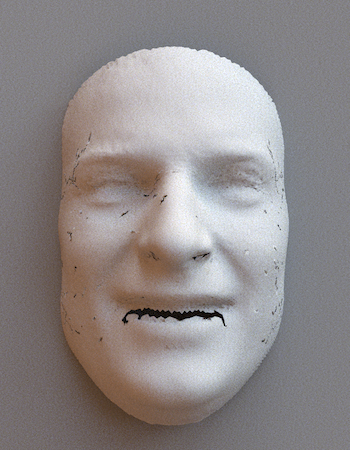} &
\includegraphics[align=c,width=.09\linewidth]{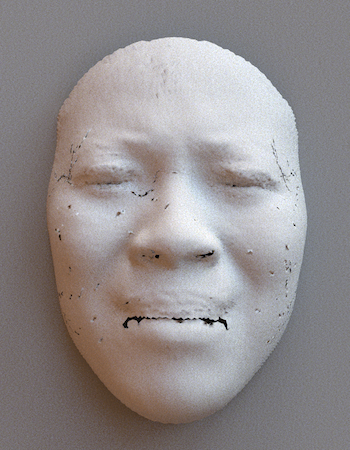} &
\includegraphics[align=c,width=.09\linewidth]{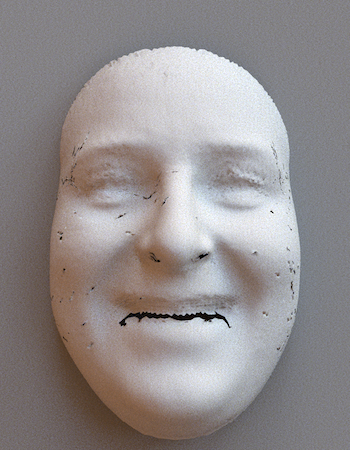} &
\includegraphics[align=c,width=.09\linewidth]{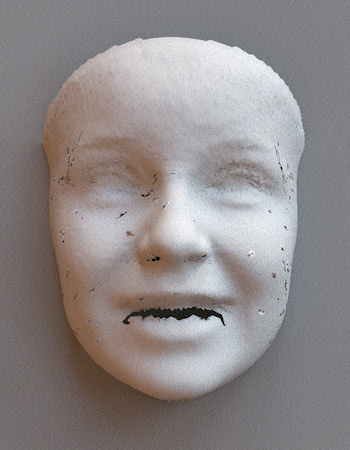}\\

\rotatebox[origin=c]{90}{SMF no s.c.} &
\includegraphics[align=c,width=.09\linewidth]{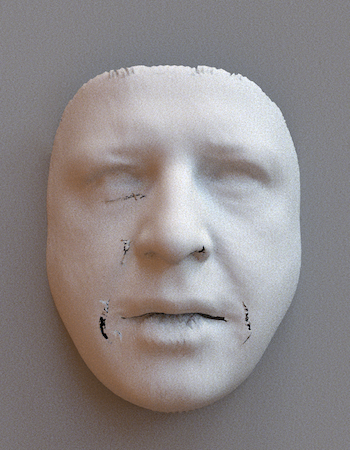} &
\includegraphics[align=c,width=.09\linewidth]{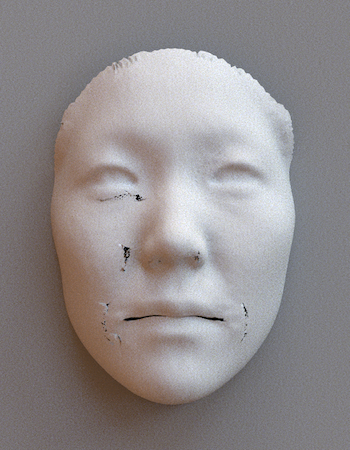} &
\includegraphics[align=c,width=.09\linewidth]{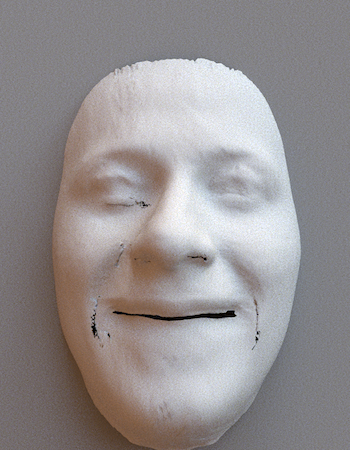} &
\includegraphics[align=c,width=.09\linewidth]{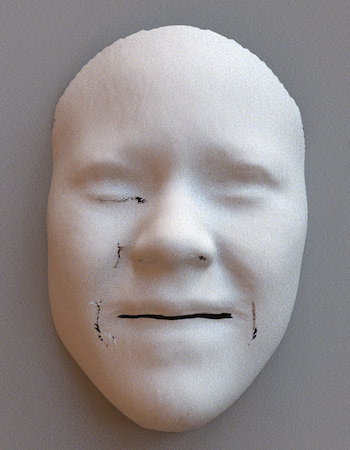} &
\includegraphics[align=c,width=.09\linewidth]{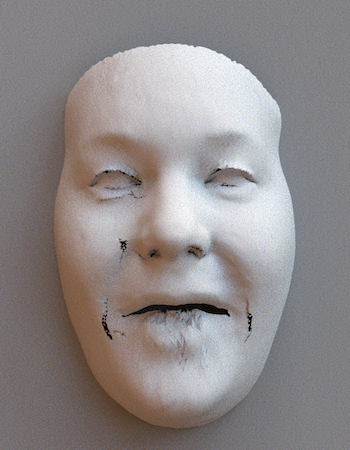} &
\includegraphics[align=c,width=.09\linewidth]{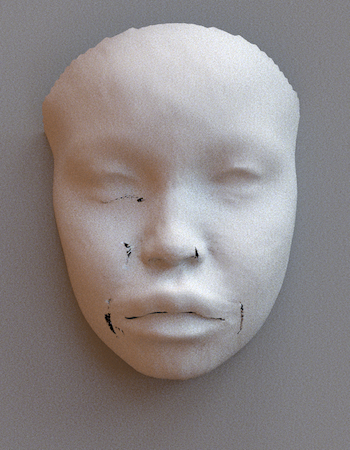} &
\includegraphics[align=c,width=.09\linewidth]{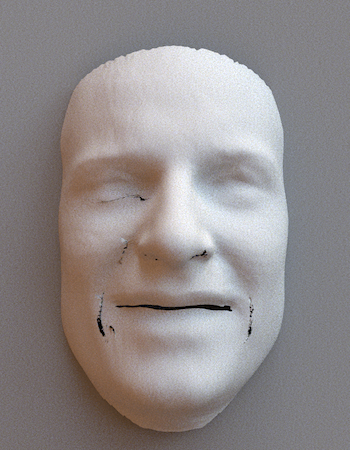} &
\includegraphics[align=c,width=.09\linewidth]{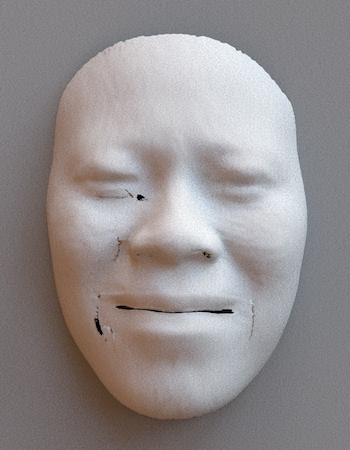} &
\includegraphics[align=c,width=.09\linewidth]{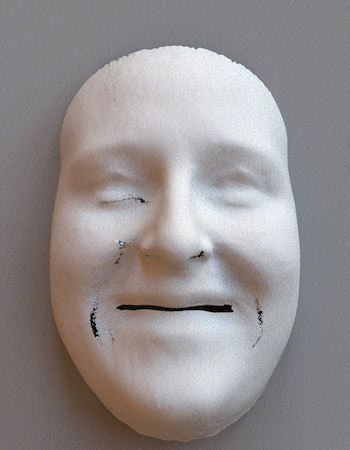} &
\includegraphics[align=c,width=.09\linewidth]{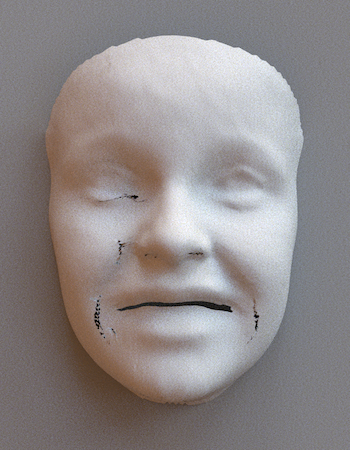}\\

    \end{tabular}}

    \caption{\textbf{Visual comparison of different ablations:} We selected 5 training scans and 5 test scans from Figures \ref{fig:sample_reconst_train} and \ref{fig:sample_renders_3dmd} and produced their registration with various choices of decoders compared in our ablation study.}
    \label{fig:ablation_decoders_renders}
\end{figure*}

We compare the models with mesh inception and fully-connected decoders in four scenarios: PCA mouth model (SMF, SMF fc'), no mouth model and no regularization (SMF no m.m., SMF fc' no m.m.), and Laplacian regularization with weight $\lambda_{lap} = 2.5e-4$ and $\lambda_{lap} = 1e-3$. We implemented Laplacian regularization using uniform weighting, as we found cotangent weights to be highly numerically unstable, leading to severe artifacting in the mouth region. We report the average surface reconstruction errors of the models in Figure \ref{fig:ablation_decoders} as well as their landmark localization errors on BU3D and 3DMD in Table \ref{tab:ablation_lm_bu3d} and Table \ref{tab:ablation_lm_3dmd}. We further provide sample reconstructions in Figure \ref{fig:ablation_decoders_renders}.

Numerically, the models with no mouth regularization showed higher test surface error and lower training surface error, for both fully-connected an mesh inception decoders, with the convolutional decoders markedly outperforming the dense layers. This is explained by the fact that not constraining the vertices in the mouth region enables the model to match them at a low cost (in terms of chamfer distance) with points from the teeth or inside of the mouth, thus lowering the error measured. Laplacian regularization behaves similarly, as visualized in Figure \ref{fig:ablation_decoders_renders}, where the mouth reconstructions of the models that use Laplacian loss are in-between the non-regularized models and the PCA-neural network hybrids in terms of deformations induced by noisy points from the inside of the mouth. On the other hand, the hybrid models produced noise-free reconstructions in all cases. %

We note that relying only on the neural network, with an additional Laplacian loss, improved the surface fairness of the registrations produced by the fully-connected decoders. This is expected, as the hybrid models ought to be harder to optimize. Naturally, the non-linear models are also more powerful and expressive than the PCA mouth models (which is the reason why we use the latter to constrain the former and perform denoising), and should therefore be favored when training on curated noise-free data.

We conclude that, for collections of raw noisy scans, our proposed approach of building hybrid models is effective.

\subsection{Ablation study on the encoder}
\label{sec:ablation_encoder}

We now evaluate the contribution of the improvements we made to the PointNet encoder (attention mechanism, group normalization)by carrying-out an ablation study. We train SMF with our modified PointNet encoder without attention (No att.) and with a vanilla PointNet encoder.  As a reminder, the baseline is evaluated on the processed (cropped, subdivided) data.

\subsubsection{Distribution of the surface error}

\begin{figure}[t]
    \centering
    \includegraphics[width=84mm]{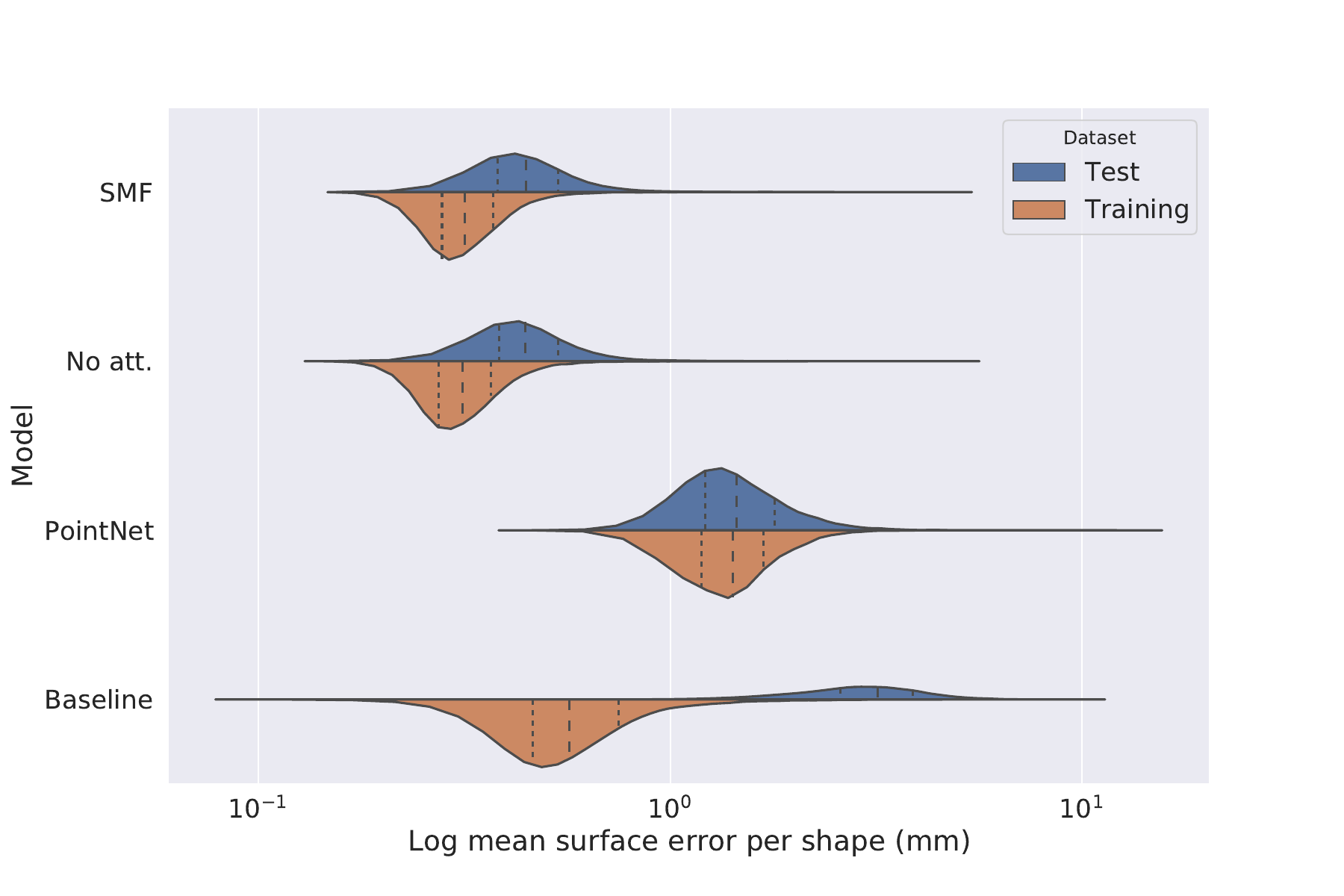}
    \caption{\textbf{Violin plot:} of the training and test error for the model trained without attention compared to SMF, SMF with a vanilla PointNet encoder, and the baseline.}
    \label{fig:dist_attention}
\end{figure}

We visualize the distribution of the surface error on the 5000 training and test scans in Figure \ref{fig:dist_attention}, as well as that of the baseline. 

\begin{figure}[t]
    \centering
    \setlength\tabcolsep{1.5pt}
    {\small
    \begin{tabular}{c|ccc}
        & FRGC & FRGC & 3DMD\\
        \rotatebox[origin=c]{90}{Ground truth} &
        \includegraphics[align=c,width=.25\linewidth]{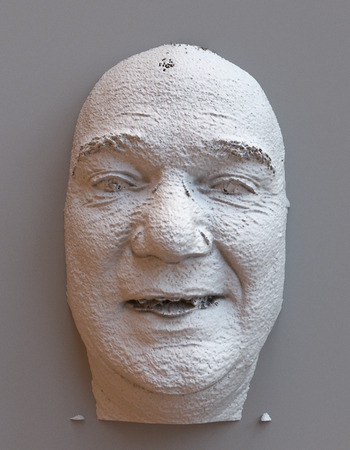} &
        \includegraphics[align=c,width=.25\linewidth]{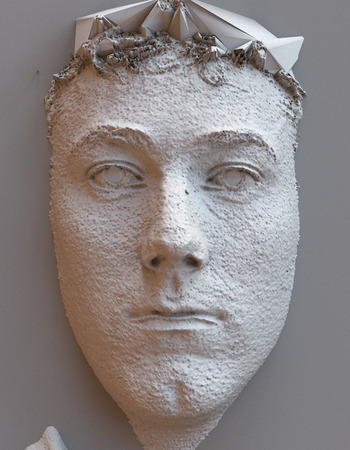} &
        \includegraphics[align=c,width=.25\linewidth]{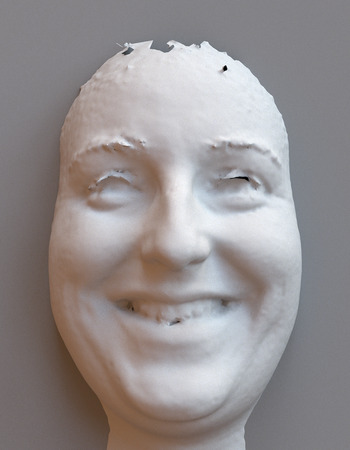}\\
        \rotatebox[origin=c]{90}{Vanilla PointNet} &
        \includegraphics[align=c,width=.25\linewidth]{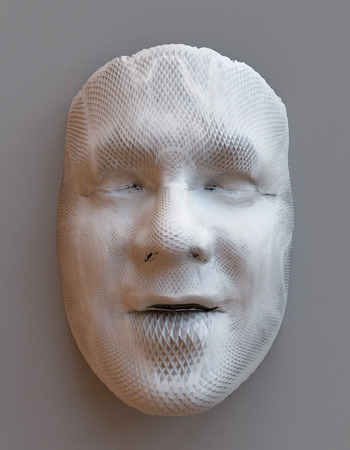} &
        \includegraphics[align=c,width=.25\linewidth]{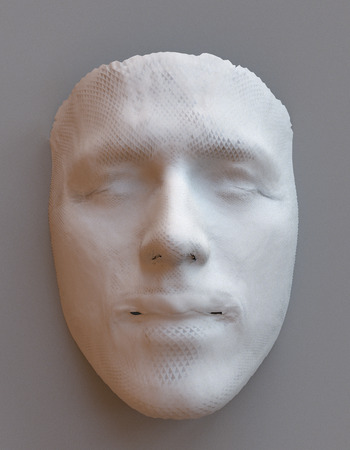} &
        \includegraphics[align=c,width=.25\linewidth]{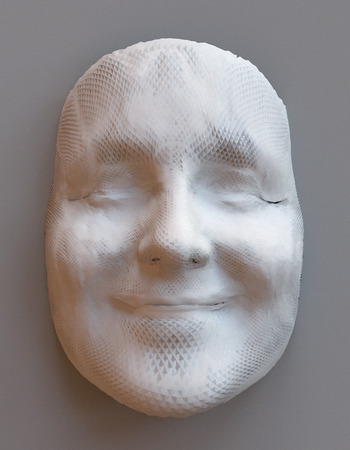}\\
        \rotatebox[origin=c]{90}{No att.} &
        \includegraphics[align=c,width=.25\linewidth]{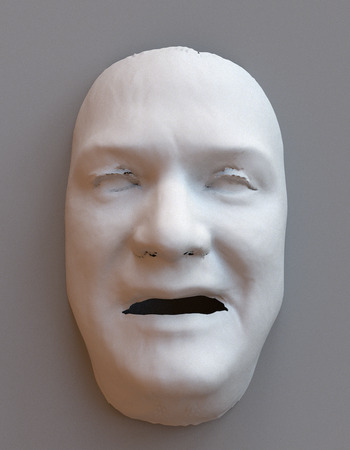} &
        \includegraphics[align=c,width=.25\linewidth]{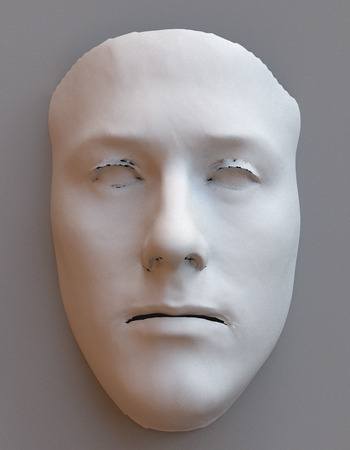} &
        \includegraphics[align=c,width=.25\linewidth]{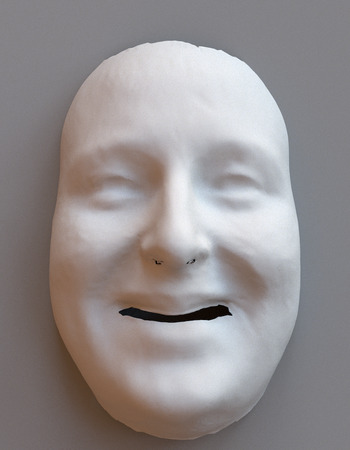}\\
        \rotatebox[origin=c]{90}{SMF} &
        \includegraphics[align=c,width=.25\linewidth]{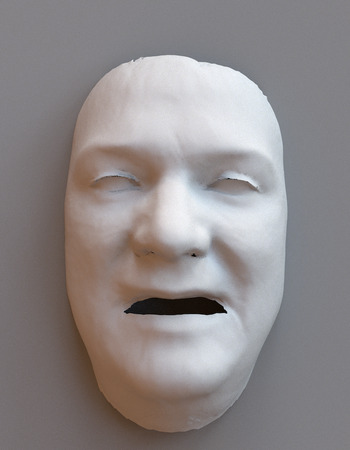} &
        \includegraphics[align=c,width=.25\linewidth]{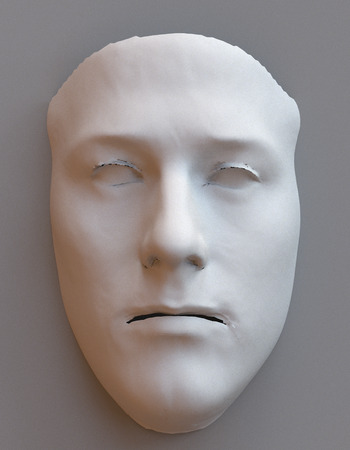} &
        \includegraphics[align=c,width=.25\linewidth]{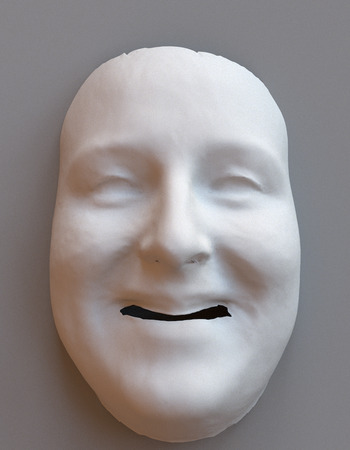}\\
    \end{tabular}
    }
    \caption{\textbf{Ablation study on the attention mechanism:} The attention mechanism helps reduce noise and improve details on out of sample registrations.}
    \label{fig:ablation_renders}
\end{figure}

\begin{figure}[t]
    \centering
    \begin{tabular}{ccc}
    \includegraphics[align=c, width=.25\linewidth]{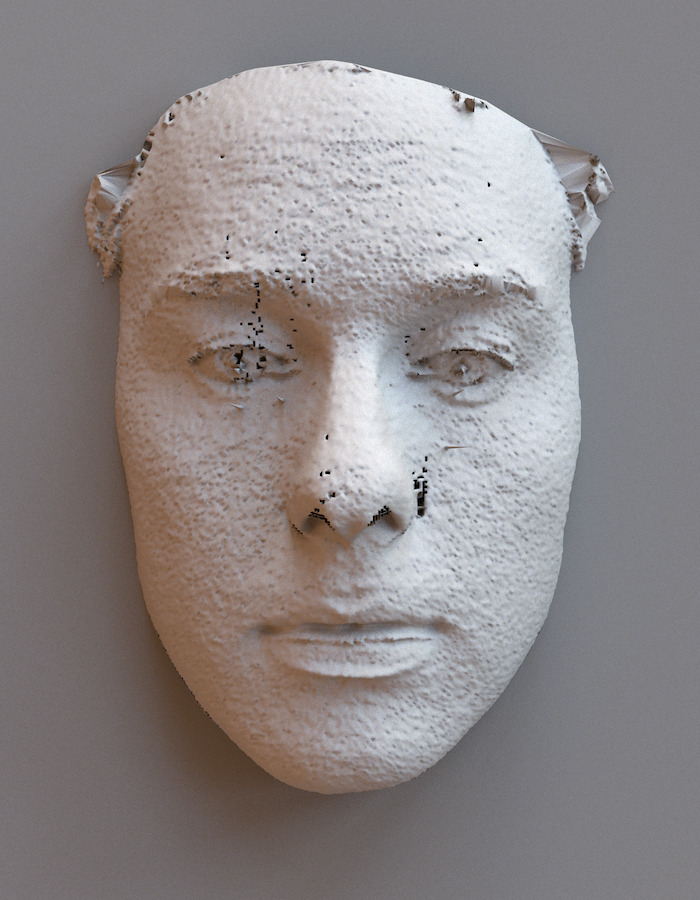} &
    \includegraphics[align=c, width=.25\linewidth]{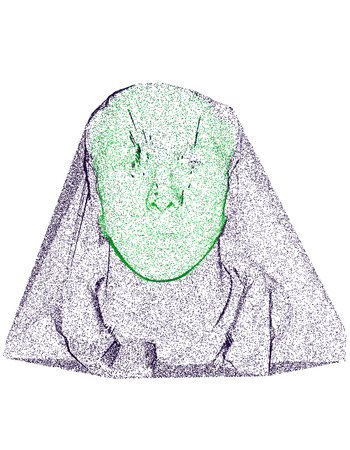} &
    \includegraphics[align=c, width=.25\linewidth]{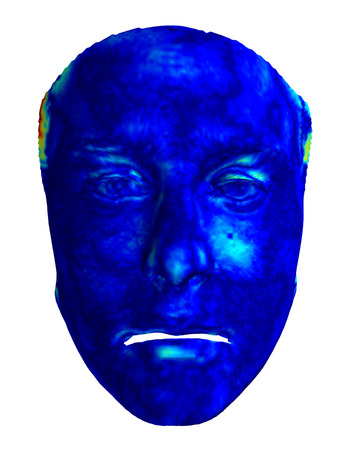}
    \end{tabular}
    \includegraphics[width=.5\linewidth]{Fig_colourbar.pdf}
    \caption{\textbf{Revisiting the example of Figure \ref{fig:sensor_noise}:} the attention mechanism is able to discard noisy points in badly-triangulated range scans.}
    \label{fig:revisit_sensor_noise}
\end{figure}

As can be seen in Figure \ref{fig:dist_attention}, SMF with vanilla PointNet has lower training set performance than the baseline, which used a vanilla PointNet trained on cropped scans, but does not overfit contrary to the baseline. The distributions of the training error of SMF with and without attention are extremely close, with the no attention variant actually showing marginally lower error. As shown in \cite{Qi2017a}, PointNet summarizes the input point cloud with a few (at most as many as the output dimension of the max pooling layer) points from the input. This property makes PointNet naturally robust to noise \textit{to some extent}. When looking at the generalization gap for the models, we can see the surface error increased less for SMF than for the model without attention, as can be further verified in Figure \ref{fig:ablation_decoders}. These observations suggest our changes all contribute to improved performance and improved generalization. We verify the contribution of the attention mechanism visually in Figure \ref{fig:ablation_renders}. We can see SMF without attention performs well, but reconstructions are noisier for the faceted scans from FRGC, and less details are present in the test 3DMD scan. Revisiting the examples of Figure \ref{fig:sensor_noise}, we can also see the attention mechanism helps discard sensor noise in Figure \ref{fig:revisit_sensor_noise}, and in line 3, col. 2 of Figure \ref{fig:sample_reconst_train}, in which points inside the mouth also received attention scores close to $0$.

\subsubsection{Ambient noise}

\begin{figure}[t]
    \centering
    \includegraphics[width=84mm]{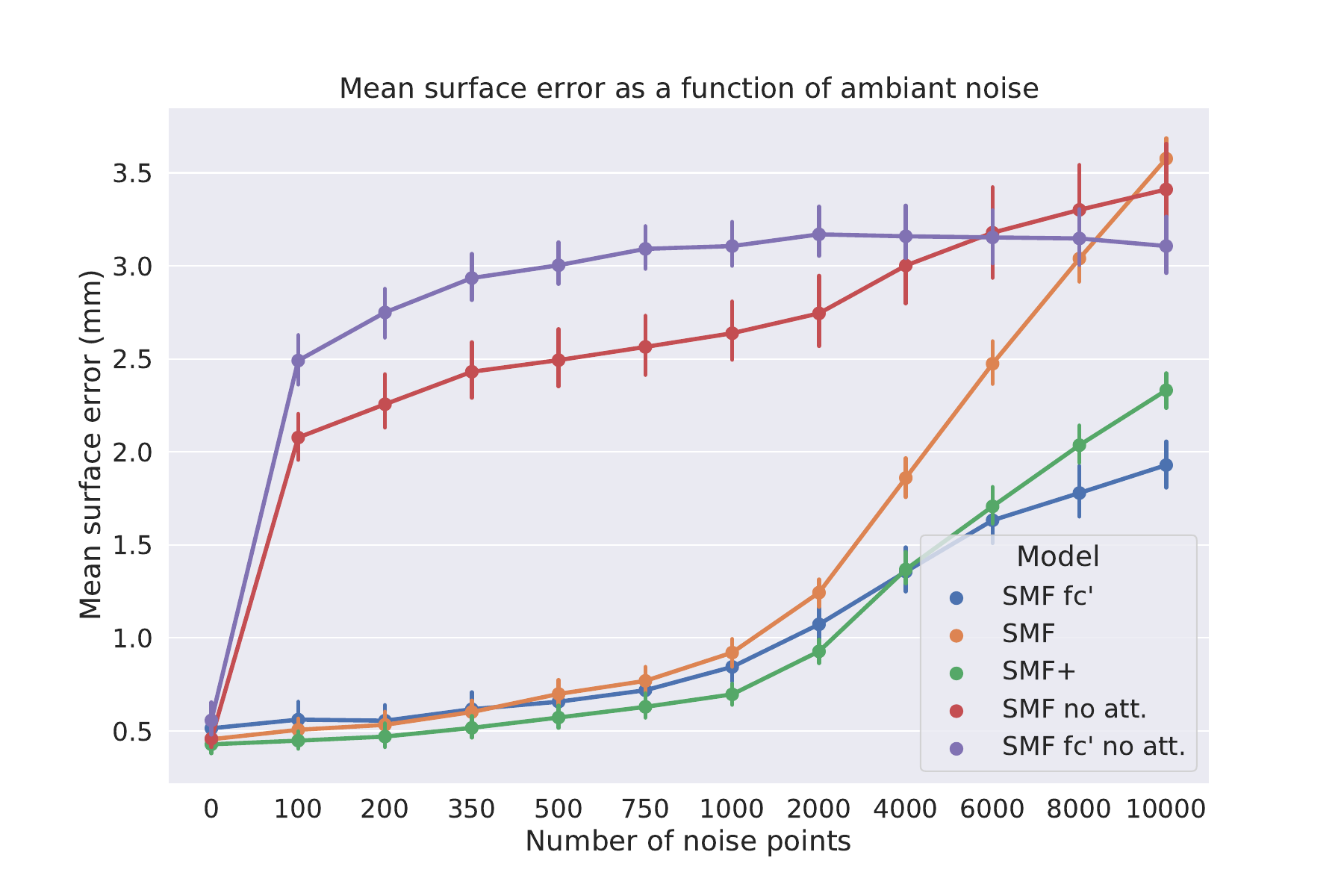}
    \caption{\textbf{Average mean surface error for increasing levels of noise} measured on 100 randomly selected test scans from the 3DMD dataset. Models trained without our attention mechanism are very sensitive to random perturbations of their input, as shown by the sharp increase in mean surface error and the large variance of the surface error, even for low noise levels. Our models trained with attention are, on the other hand, more resilient to corruption.}
    \label{fig:attention_ambient_noise}
\end{figure}

\begin{figure}[t]
    \centering
    \setlength\tabcolsep{1.5pt}
    {\small
    \begin{tabular}{c|cccccccc}
         & SMF no att. & SMF & SMF fc'\\
        \rotatebox[origin=c]{90}{Attention} &
        N/A &
        \includegraphics[align=c,width=.31\linewidth]{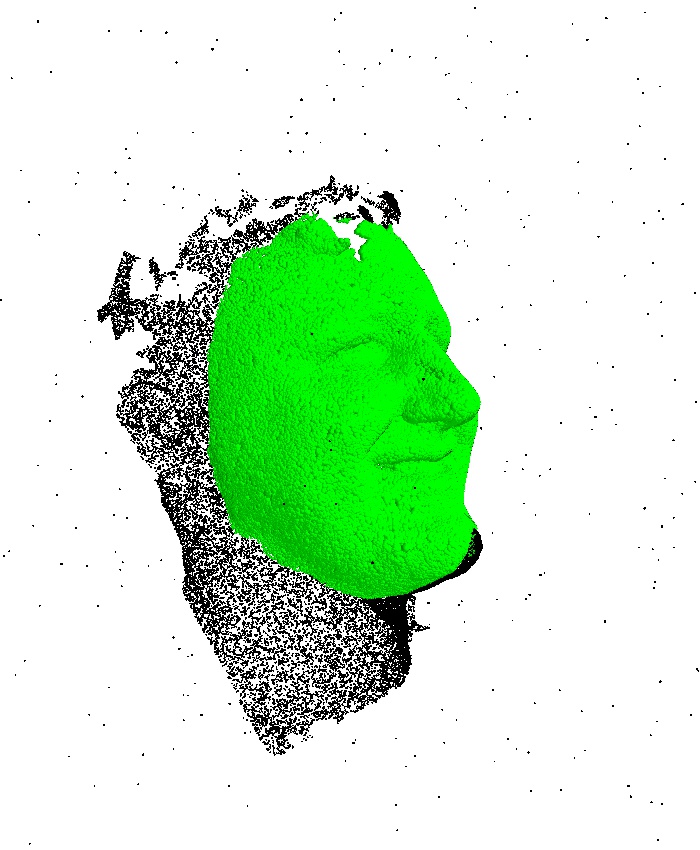} &
        \includegraphics[align=c,width=.31\linewidth]{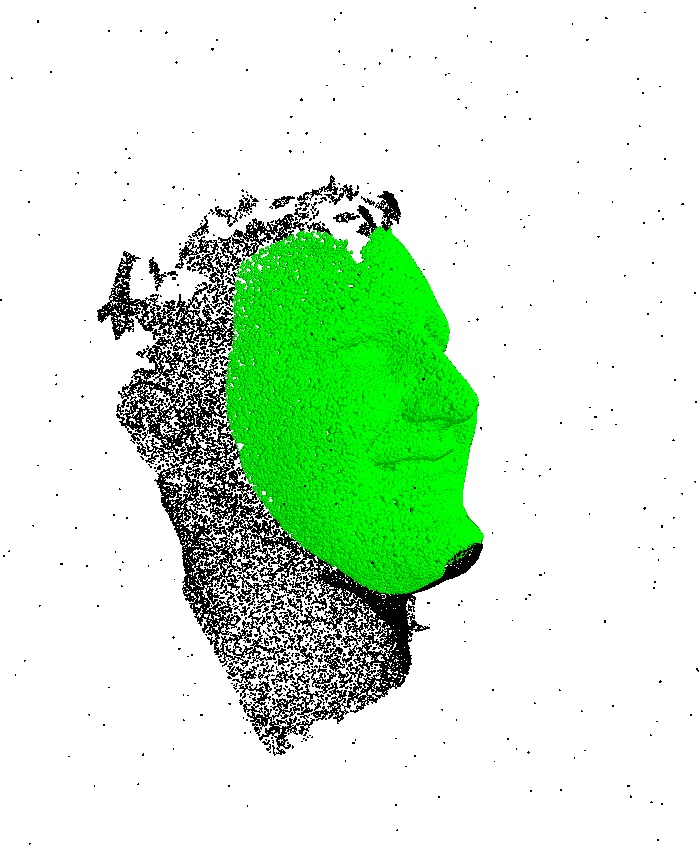}\\
        
        \rotatebox[origin=c]{90}{Reconst.} &
        \includegraphics[align=c,width=.31\linewidth]{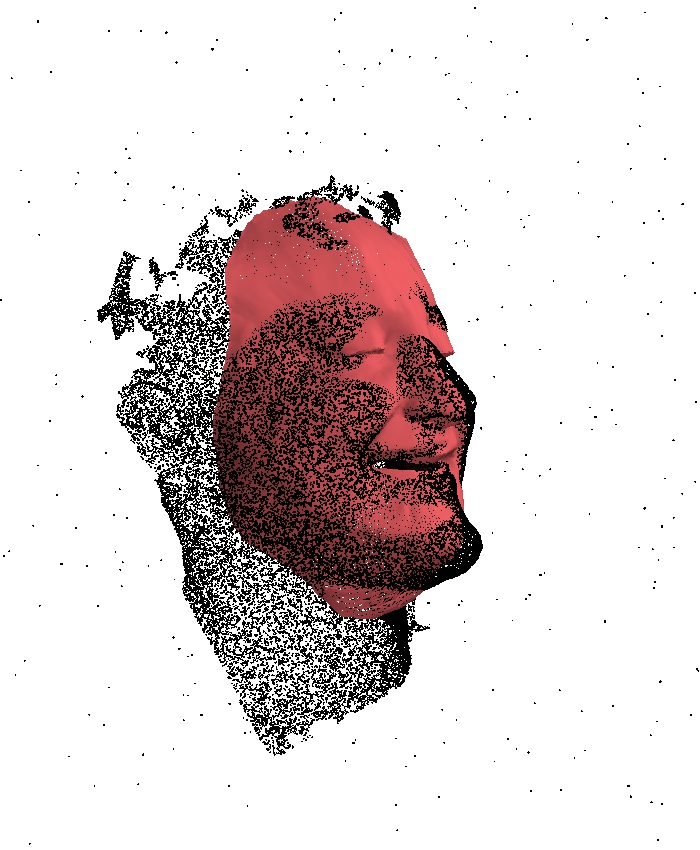} &
        \includegraphics[align=c,width=.31\linewidth]{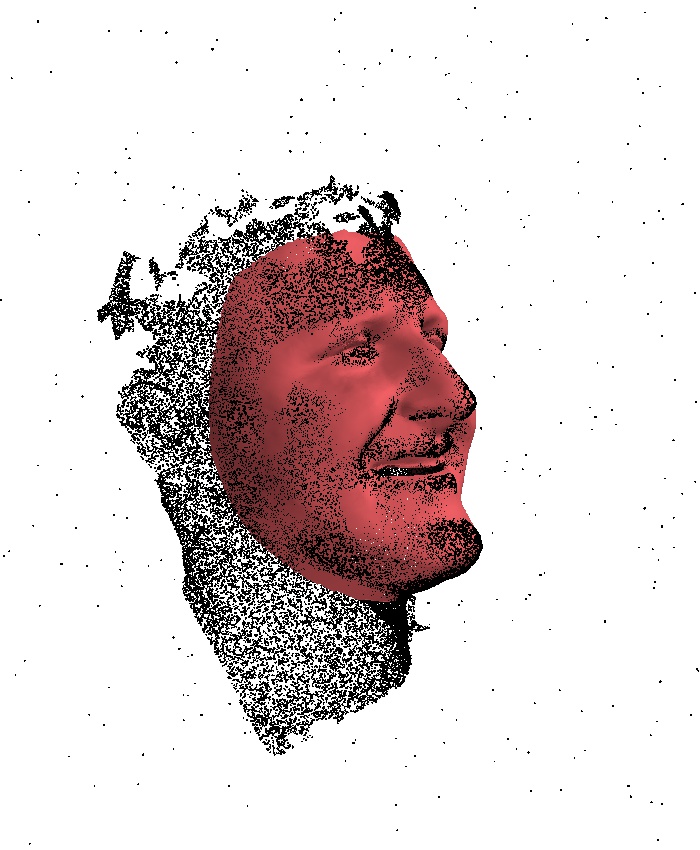} &
        \includegraphics[align=c,width=.31\linewidth]{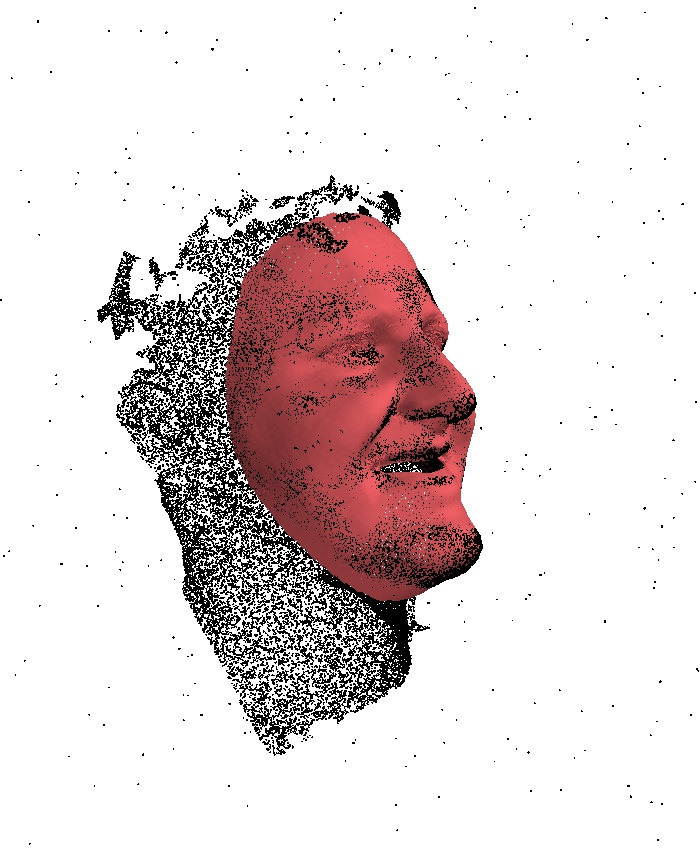}\\
    \end{tabular}}

    \caption{\textbf{Noisy input, attention mask (green) and registration} for three models trained with attention, and 500 noisy points added prior to sampling the input point cloud. Clear segmentations are obtained in all three cases, with noise points receiving a low attention score even for points close to the actual scan. This results in markedly more robust registrations.}
    \label{fig:attention_ablation_pc_and_masks}
\end{figure}

To better showcase the contribution of the attention mechanism, we now evaluate our trained models on 3DMD scans with additional artificial noise added. Our experimental setting is as follows: for a given 3DMD scan, we sample a set $\mathcal{P}$ of $2^{16}$ points at random on the scan. A second set $\mathcal{U}$ of $N$ points is then drawn uniformly at random in an cubic volume containing the scan. Finally, our input point cloud $\mathbf{X}$ consists of $2^{16}$ points drawn uniformly at random  and without replacement from $\mathcal{P} \cup \mathcal{U}$. Examples of resulting point clouds are shown in Figure \ref{fig:attention_ablation_pc_and_masks} for $N = 500$.

We apply SMF, SMF+, SMF fc' as well as SMF and SMF fc' without attention to $\mathbf{X}$ and measure the mean surface error between the reconstructions and the raw scan. In total, we repeated this procedure for 100 scans and 11 noise levels ranging from no noise ($N = 0$) to substantial noise ($N = 10000$). We report the results in Figure \ref{fig:attention_ambient_noise}.

\begin{figure*}[t!]
    \centering
    \setlength\tabcolsep{1.5pt}
    {\small
    \begin{tabular}{c|cccccccc}
        \# & Raw & SMF no att. & SMF fc' no att. & SMF & SMF+ & SMF fc' & SMF lap & SMF fc lap\\
        0 &
        \includegraphics[align=c,width=.09\linewidth]{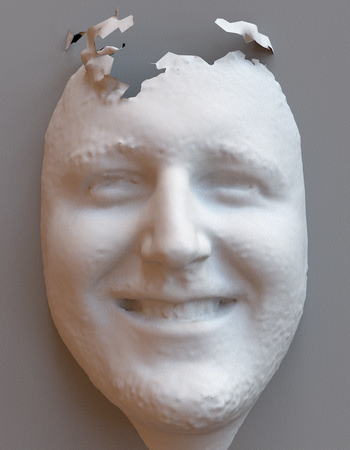} &
        \includegraphics[align=c,width=.09\linewidth]{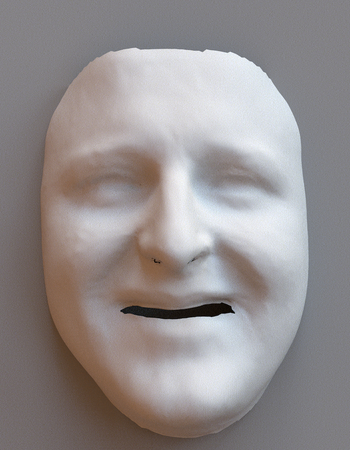} &
        \includegraphics[align=c,width=.09\linewidth]{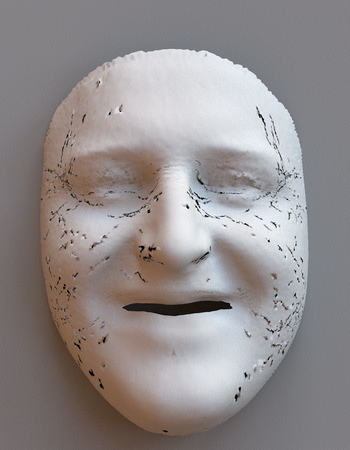} &
        \includegraphics[align=c,width=.09\linewidth]{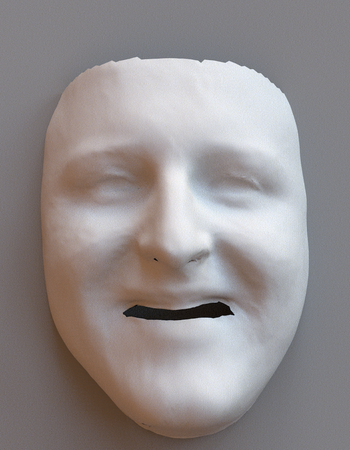} &
        \includegraphics[align=c,width=.09\linewidth]{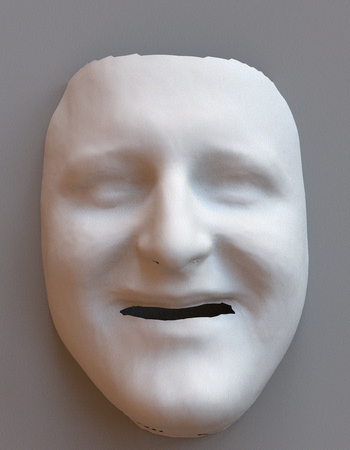} &
        \includegraphics[align=c,width=.09\linewidth]{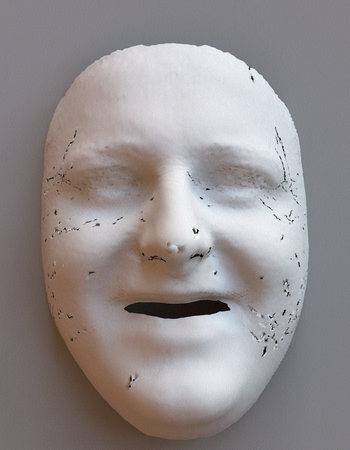} &
        \includegraphics[align=c,width=.09\linewidth]{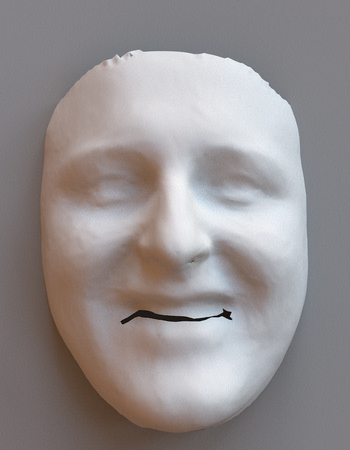} &
        \includegraphics[align=c,width=.09\linewidth]{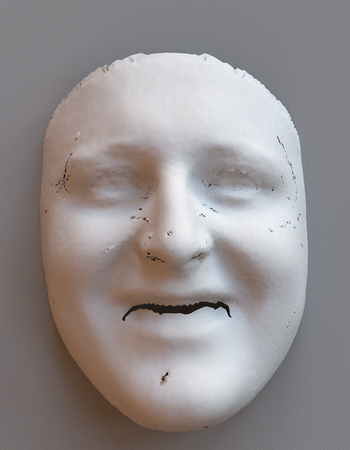}\\
        100 &
        \includegraphics[align=c,width=.09\linewidth]{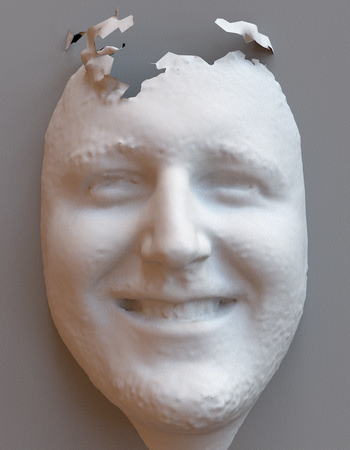} &
        \includegraphics[align=c,width=.09\linewidth]{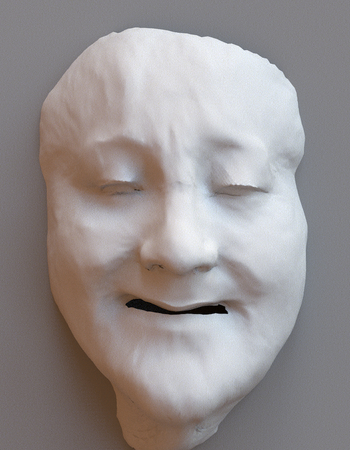} &
        \includegraphics[align=c,width=.09\linewidth]{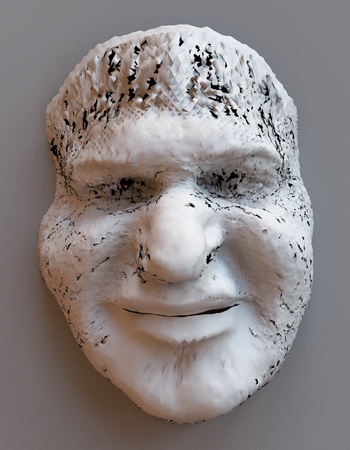} &
        \includegraphics[align=c,width=.09\linewidth]{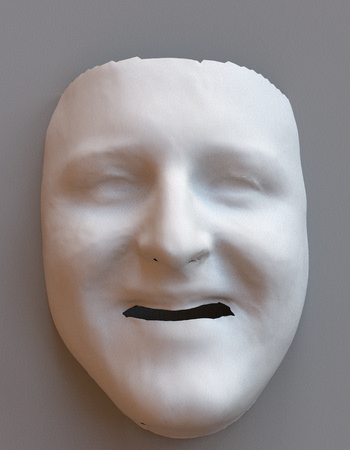} &
        \includegraphics[align=c,width=.09\linewidth]{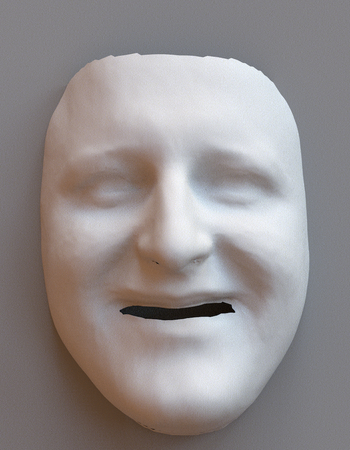} &
        \includegraphics[align=c,width=.09\linewidth]{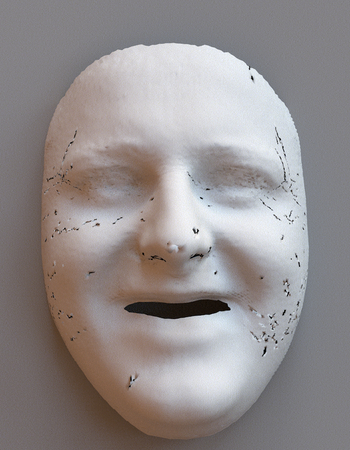} &
        \includegraphics[align=c,width=.09\linewidth]{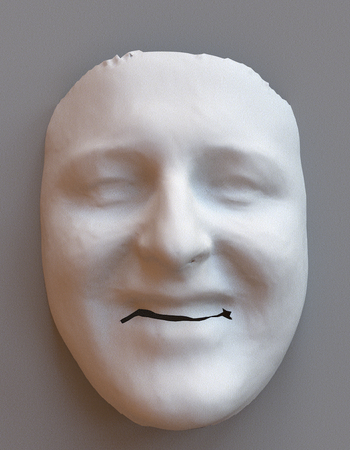} &
        \includegraphics[align=c,width=.09\linewidth]{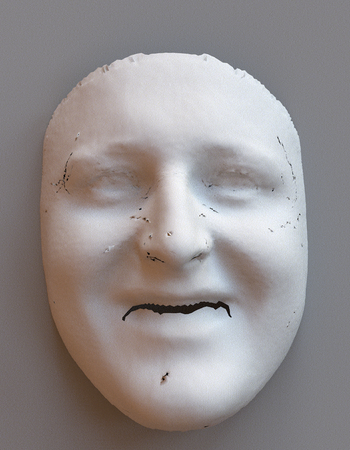}\\
        500 &
        \includegraphics[align=c,width=.09\linewidth]{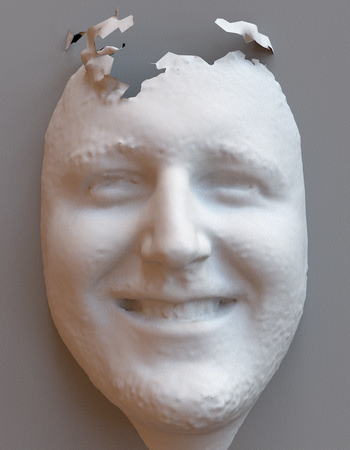} &
        \includegraphics[align=c,width=.09\linewidth]{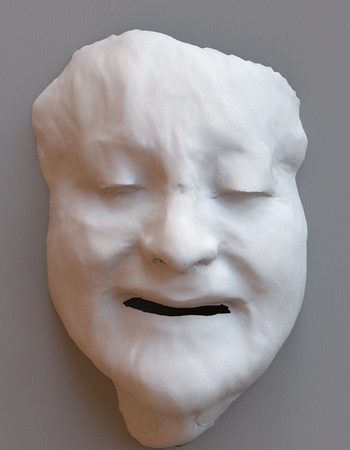} &
        \includegraphics[align=c,width=.09\linewidth]{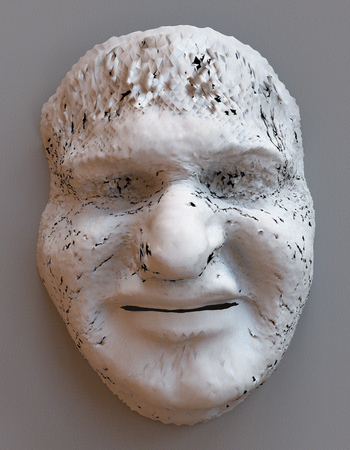} &
        \includegraphics[align=c,width=.09\linewidth]{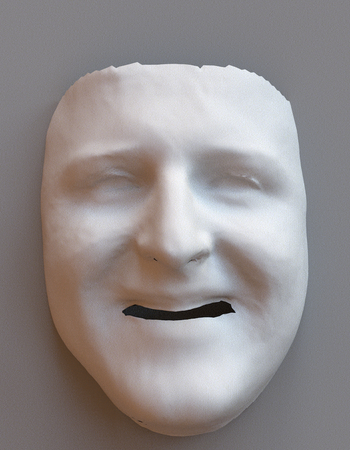} &
        \includegraphics[align=c,width=.09\linewidth]{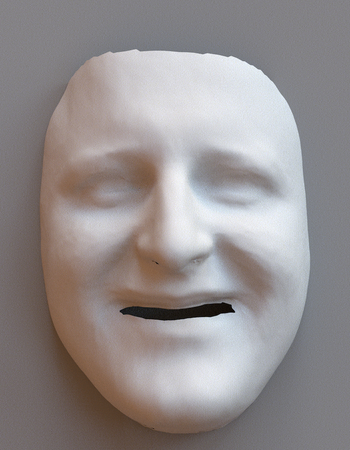} &
        \includegraphics[align=c,width=.09\linewidth]{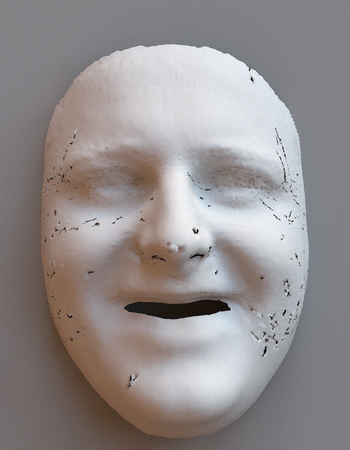} &
        \includegraphics[align=c,width=.09\linewidth]{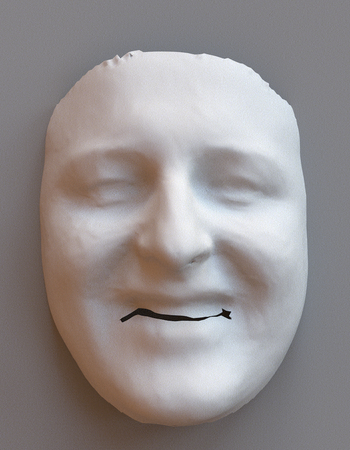} &
        \includegraphics[align=c,width=.09\linewidth]{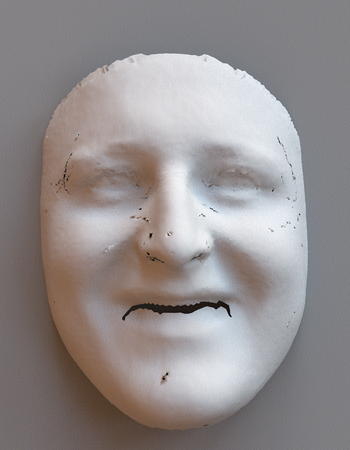}\\
    \end{tabular}}

    \caption{\textbf{Sample registrations for artificial ambient noise} for several choices of decoders and mouth regularization. $\lambda_{edge}=2e-4$ for SMF fc' and SMF fc' no att., and $\lambda_{edge}=5e-5$ for SMF fc lap. We set $\lambda_{lap} = 1e-3$.}
    \label{fig:attention_ablation_render_grid}
\end{figure*}

As can be seen from Figure \ref{fig:attention_ambient_noise}, the models that do not have an attention mechanism are very sensitive to noise. As little as 100 random points prior to sampling the input point cloud lead to significant deformations of the output, regardless of the choice of decoder. This is apparent when visualizing the registrations, \eg for a test subject from the 3DMD dataset in Figure \ref{fig:attention_ablation_render_grid}. On the other hand, the models trained with attention are more robust: the surface error increases slower, and has lower variance as indicated by the shorter error bars. Visually, the reconstructions we obtain from the noisy inputs are indistinguishable from the noise-free inputs for low noise levels.  We note, however, that not all models with attention learn equally good segmentations of the input point cloud. In this particular case, our SMF model was more susceptible to noise than SMF+ and SMF fc. We compare the attention masks of the noisy point clouds of some models in Figure \ref{fig:attention_ablation_pc_and_masks}, and verify that the segmentations isolate the most relevant points.

We will further verify that the attention mechanism improves the quality of reconstructions on noisy out of sample scans in Section \ref{sec:in_the_wild}.

\subsection{Overview}

In Section \ref{sec:landmark_error_xp} and Section \ref{sec:surface_error_xp}, we showed SMF (and SMF+) systematically outperforms the baseline on landmark localization error and offers performance competitive with NICP. Test set performance, in particular, was markedly higher than the current state of the art, and remained very close to the training set error. Direct evaluation of the mean surface registration error offers a more complete picture of the registration quality and leads to similar conclusions. Visual inspection of the reconstructions confirms the quantitative analysis: contrary to the baseline, SMF provides noise-free registrations which closely match the raw scans in both identity and expression. We showed re-sampling the scans typically lead to minor variations in their registrations in Section \ref{sec:robustness}. In Section \ref{sec:ablation_decoder}, we compared our default architecture of two mesh inception convolutional decoders and PCA mouth models with different variations, namely fully-connected decoders, using a single decoder, not using skip connections, not using any mouth regularization, or using uniform Laplacian regularization of the mouth region. We showed our contributions provide tangible benefits in reconstruction accuracy and robustness for noisy raw scans data, while our framework is flexible enough to accommodate various substitutions while preserving the ability of the models to generalize well to unseen data. Finally, we evaluated the contributions of our modifications of the PointNet encoder in Section \ref{sec:ablation_encoder}. In particular, we demonstrated that our attention mechanism markedly improves the models' robustness to random perturbations of their input in the form of ambient noise, regardless of our choice of decoder. This demonstrates that our contributions to the encoding and decoding stages are both orthogonal and complementary.

\section{A large scale hybrid morphable model}
\label{sec:experiments_morphable}

\nocite{tange_ole_2018_1146014}

In this section, we assess the morphable model aspects of SMF. We first study the influence of the dimension of the identity and expression latent spaces on surface reconstruction error both in sample and out of sample. We then show SMF can be used to quickly generate realistic synthetic faces. In Section \ref{sec:interpolation_training}, we evaluate SMF on shape-to-shape translation applications, namely identity and expression tranfer, and morphing. We conclude by showing SMF can be used successfully for registration and translation fully in the wild.

\begin{figure}[b]
    \centering
    \includegraphics[width=84mm]{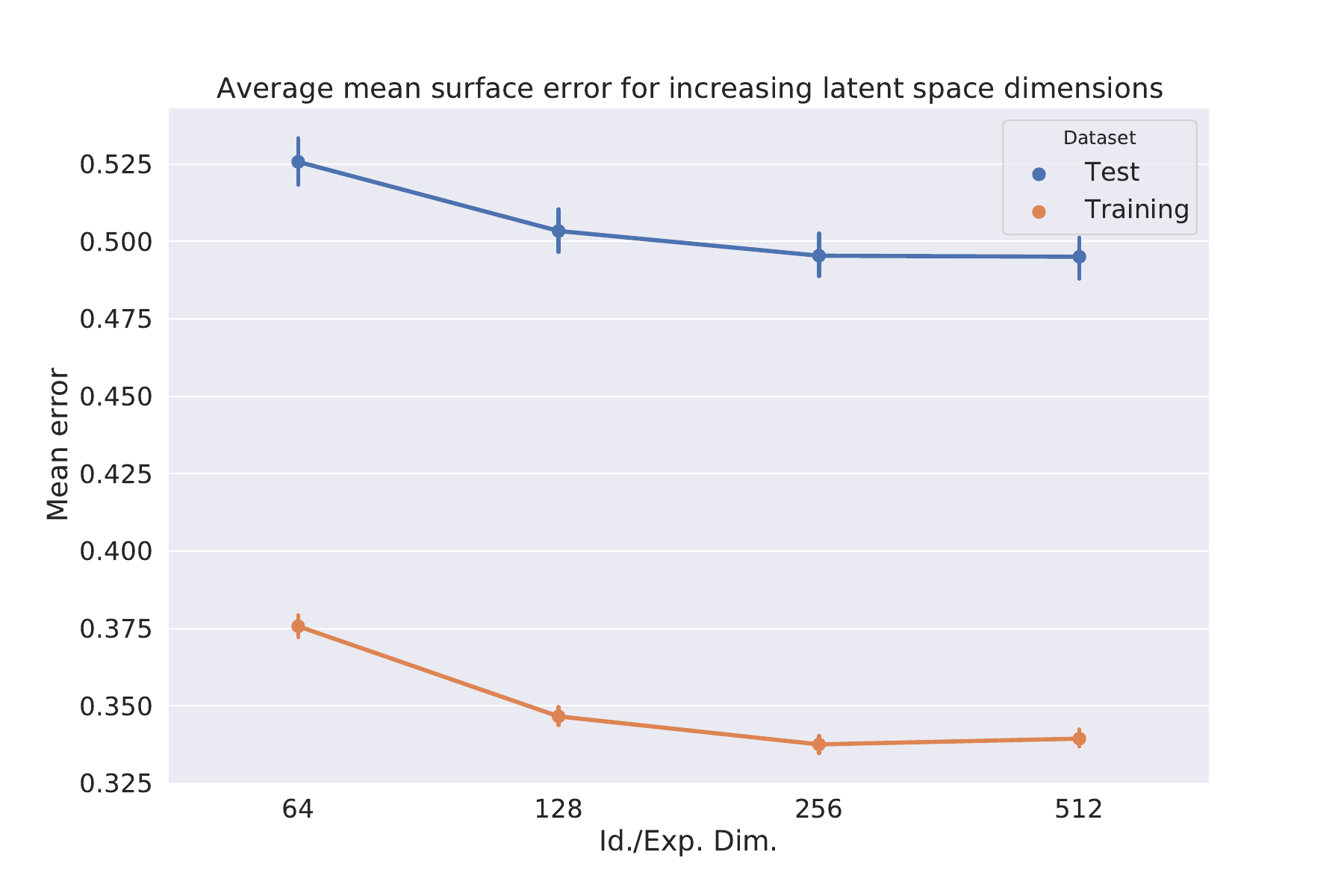}
    \caption{\textbf{Compactness and generalisation:} Training and test error for increasing number of latent dimensions.}
    \label{fig:compactness}
\end{figure}

\subsection{Dimension of the latent spaces}

The classical linear morphable models literature typically reports three main metrics. \textit{Specificity} is evaluated in Section \ref{sec:specificity}. \textit{Compactness} is the proportion of the variance retained for increasing numbers of principal components -  a direct correlate of the training error for PCA models. \textit{Generalization} measures the reconstruction error on the test set for increasing numbers of principal components. Since our model is not linear, we instead report the training and test performance for increasing identity and expression dimensions. We choose symmetric decoders with $\mathbf{z}_{id}$ and $\mathbf{z}_{exp}$ of equal dimension $d$. We vary $d \in \{64, 128, 256, 512\}$. We measure the mean (per scan) surface reconstruction error on the same subsets of 5000 training and 5000 test scans used in Section \ref{sec:registration_experiments}. We plot the mean error across the 5000 scans along with its 95\% confidence interval obtained by bootstrapping in Figure \ref{fig:compactness}.

As expected, both the training and test error decrease steadily up to $d = 256$. For $d = 512$, our data shows increased training and test error compared to $d = 256$. This shows there is diminishing return in increasing the model complexity, and bolsters our choice of $d  = 256$ for SMF.

\subsection{Generating synthetic faces}

We now evaluate the generative ability of our SMF+ model.

\subsubsection{Specificity error}
\label{sec:specificity}

We follow the literature and measure the specificity error as follows: we sample $10{,}000$ shapes at random from the joint latent space; since our model is not explicitly trained as a generative model, no particular structure is to be expected on the latent space and we therefore model the empirical distribution of the joint latent vectors of the training set with a multivariate Gaussian distribution. We estimate the empirical mean and covariance matrix of the $\approx54{,}000$ joint latent vectors and generate $10{,}000$ Gaussian random vectors. We apply the pre-trained decoder to obtain generated faces\footnote{Generating all $10{,}000$ random faces took 55s on a single consumer-grade GPU.}.

\begin{figure}[t]
    \centering
    \includegraphics[width=84mm]{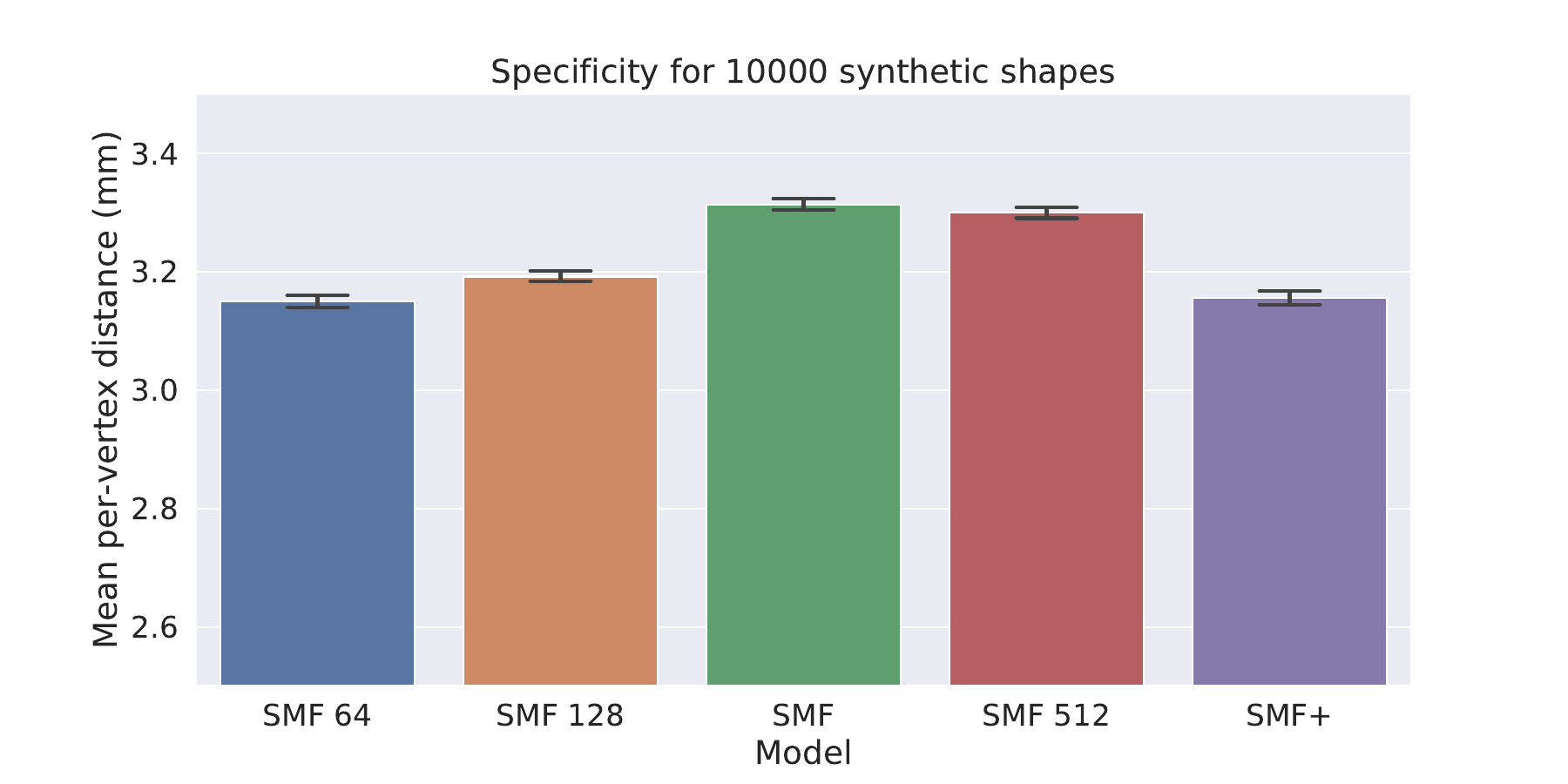}
    \caption{\textbf{Specificity error:} for variants of SMF and SMF+. The specificity error is the mean distance of the sampled scans to their projection on the registered training set.}
    \label{fig:specificity}
\end{figure}

For each of the $10{,}000$ random faces, we find its closest point in the training set in terms of minimum (over all $54{\,}000$ training registrations) of the average (over the 29495 points in the template) vertex-to-vertex Euclidean distance. The mean of these $10{,}000$ distances is the specificity error of the model. For the sake of completeness, we repeated the experiment with the variants of SMF evaluated in Figure \ref{fig:compactness}. We plot the specificity error and its 95\% confidence interval computed by bootstrapping in Figure \ref{fig:specificity}. Both SMF and SMF+ offer low specificity error, suggesting realistic-looking samples can be obtained. SMF+, in particular, has markedly lower specificity error than SMF for the same latent space dimensions, which confirms the benefits of training our very large scale model on the extended training set.

\begin{figure*}[t]
    \centering
    \setlength\tabcolsep{1.5pt}
    {\small
    \begin{tabular}{c|cccccccccc}
        \rotatebox[origin=c]{90}{Sample} & 
        \includegraphics[align=c,width=.09\linewidth]{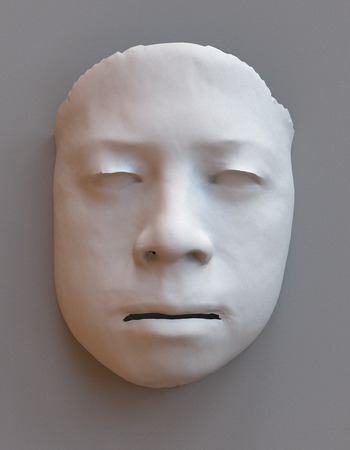} &
        \includegraphics[align=c,width=.09\linewidth]{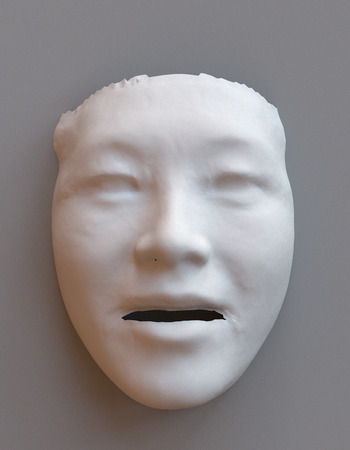} &
        \includegraphics[align=c,width=.09\linewidth]{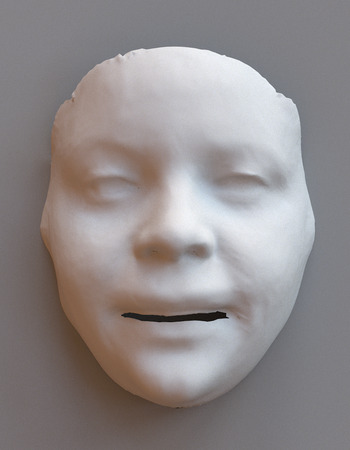} &
        \includegraphics[align=c,width=.09\linewidth]{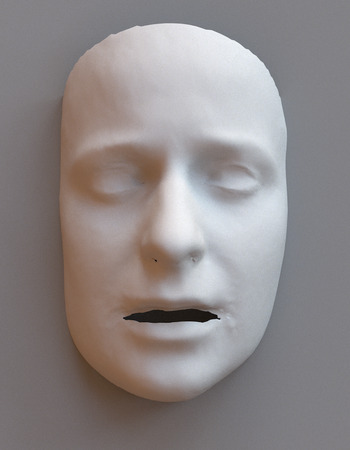} &
        \includegraphics[align=c,width=.09\linewidth]{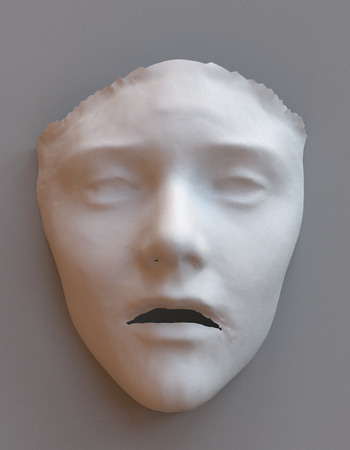} &
        \includegraphics[align=c,width=.09\linewidth]{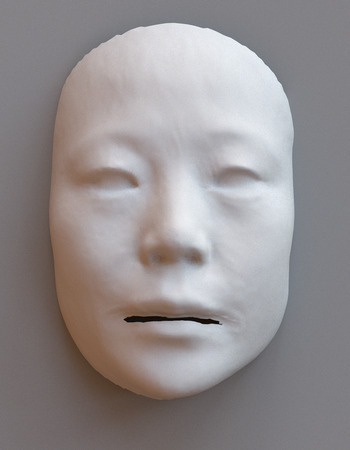} &
        \includegraphics[align=c,width=.09\linewidth]{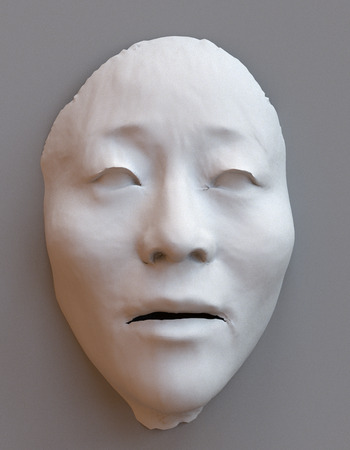} &
        \includegraphics[align=c,width=.09\linewidth]{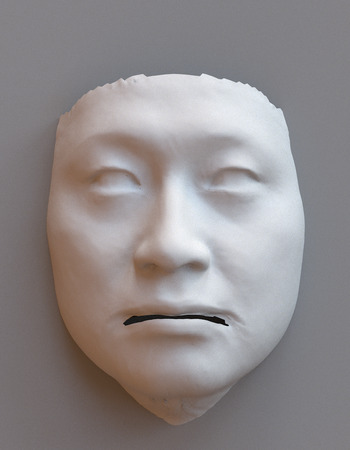} &
        \includegraphics[align=c,width=.09\linewidth]{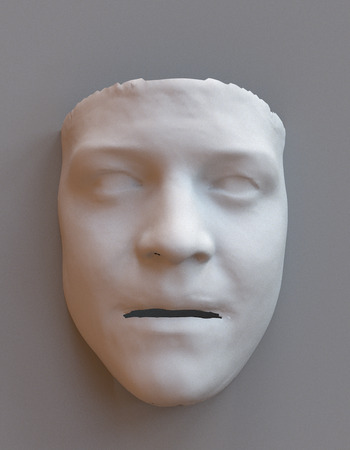} &
        \includegraphics[align=c,width=.09\linewidth]{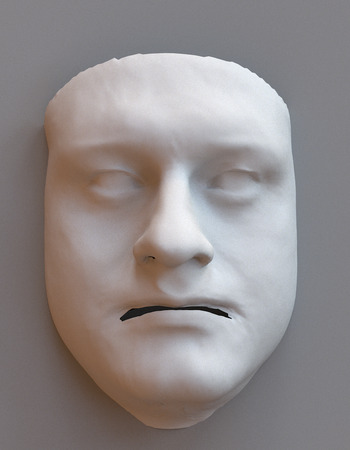}\\

        \rotatebox[origin=c]{90}{Closest reg.} & 
        \includegraphics[align=c,width=.09\linewidth]{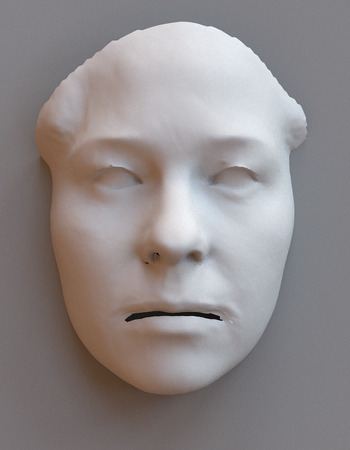} &
        \includegraphics[align=c,width=.09\linewidth]{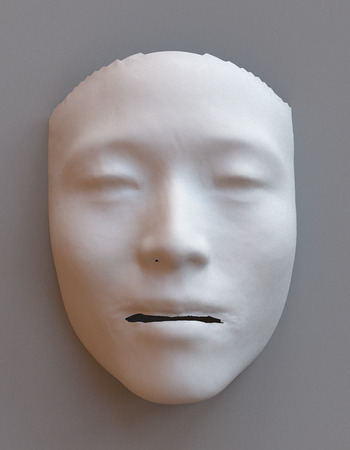} &
        \includegraphics[align=c,width=.09\linewidth]{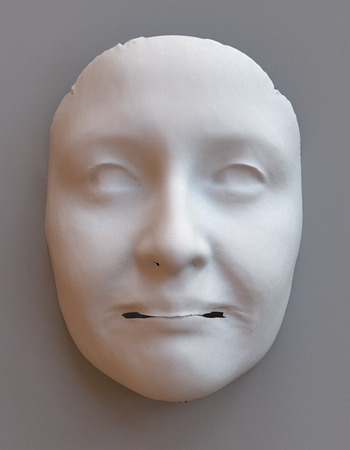} &
        \includegraphics[align=c,width=.09\linewidth]{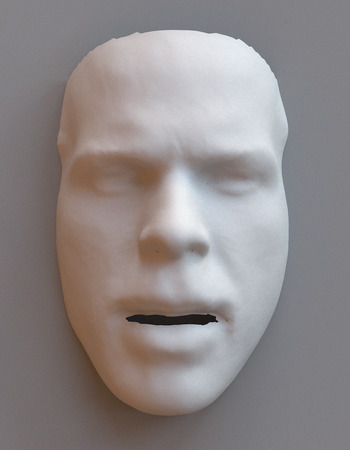} &
        \includegraphics[align=c,width=.09\linewidth]{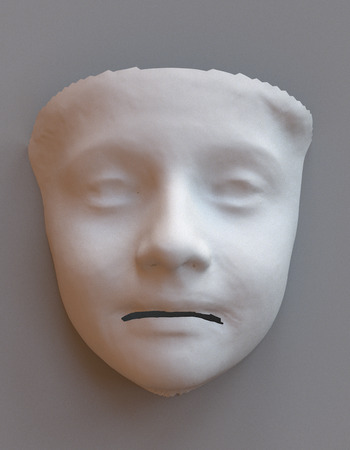} &
        \includegraphics[align=c,width=.09\linewidth]{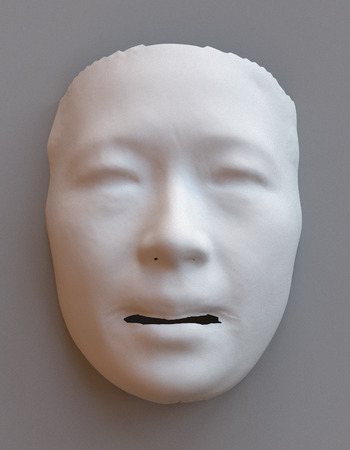} &
        \includegraphics[align=c,width=.09\linewidth]{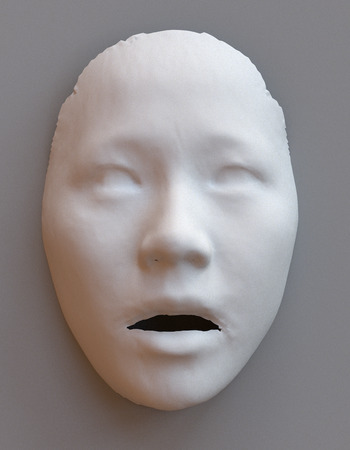} &
        \includegraphics[align=c,width=.09\linewidth]{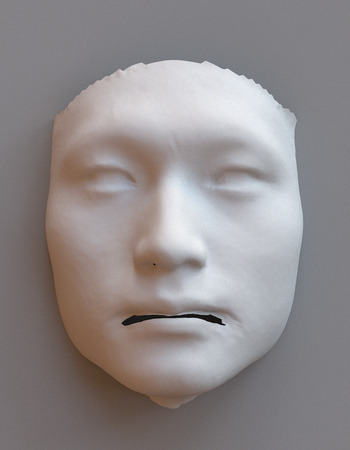} &
        \includegraphics[align=c,width=.09\linewidth]{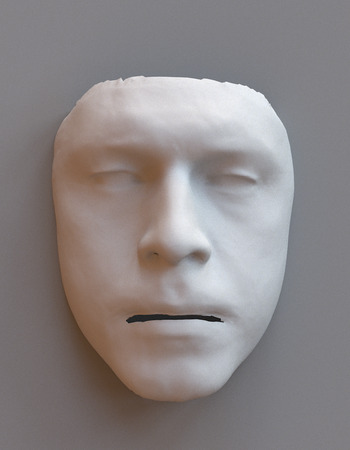} &
        \includegraphics[align=c,width=.09\linewidth]{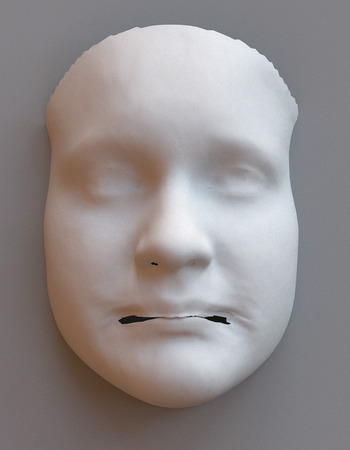}\\

        \vspace{1em}
        \rotatebox[origin=c]{90}{Closest raw} & 
        \includegraphics[align=c,width=.09\linewidth]{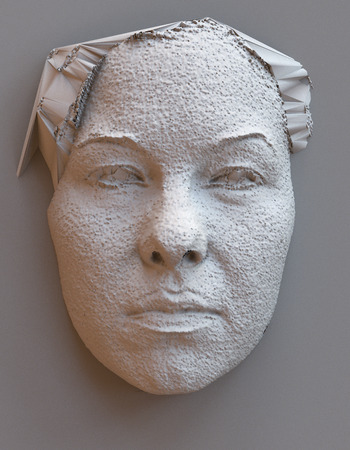} &
        \includegraphics[align=c,width=.09\linewidth]{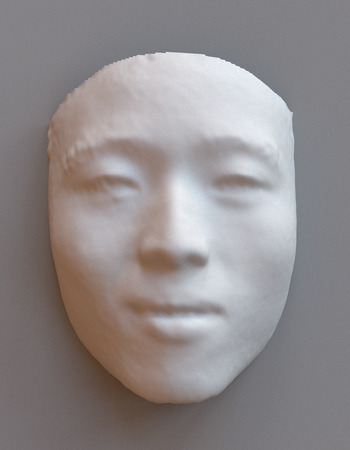} &
        \includegraphics[align=c,width=.09\linewidth]{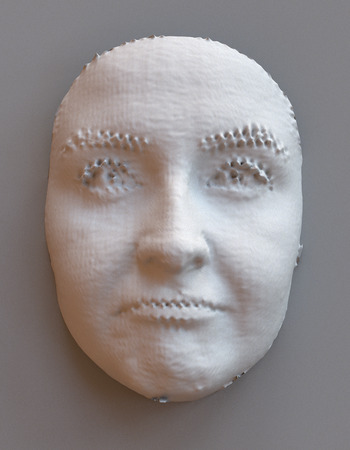} &
        \includegraphics[align=c,width=.09\linewidth]{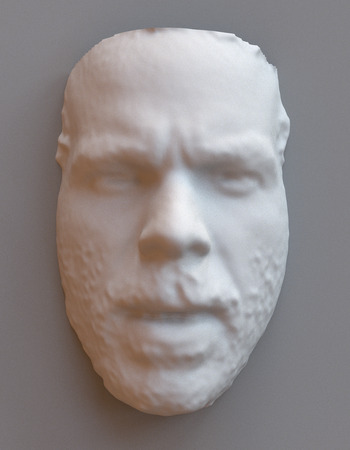} &
        \includegraphics[align=c,width=.09\linewidth]{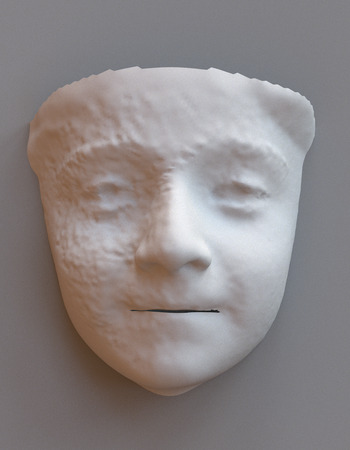} &
        \includegraphics[align=c,width=.09\linewidth]{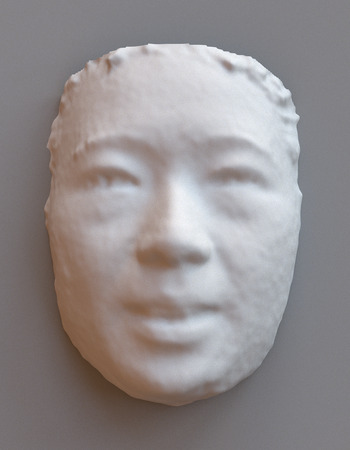} &
        \includegraphics[align=c,width=.09\linewidth]{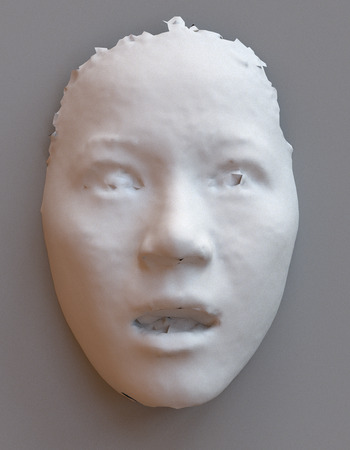} &
        \includegraphics[align=c,width=.09\linewidth]{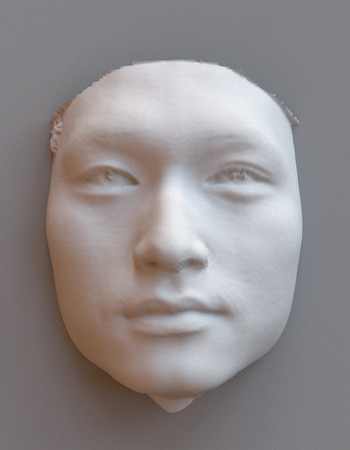} &
        \includegraphics[align=c,width=.09\linewidth]{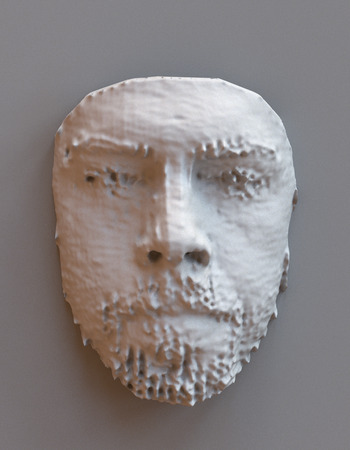} &
        \includegraphics[align=c,width=.09\linewidth]{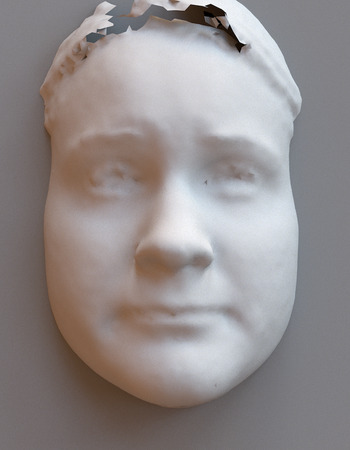}\\

        \rotatebox[origin=c]{90}{Sample} & 
        \includegraphics[align=c,width=.09\linewidth]{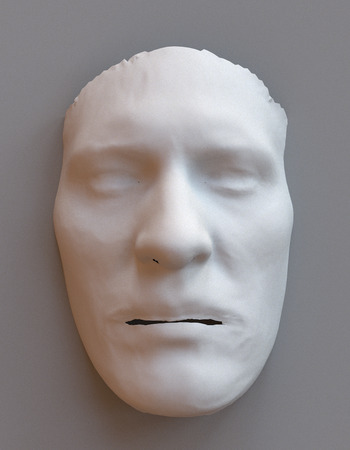} &
        \includegraphics[align=c,width=.09\linewidth]{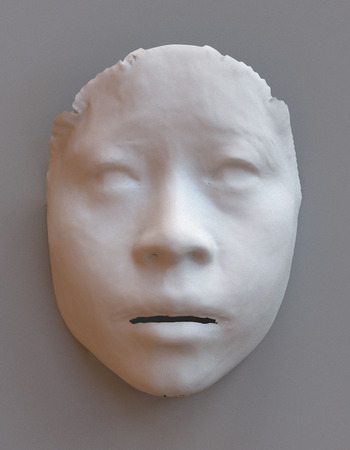} &
        \includegraphics[align=c,width=.09\linewidth]{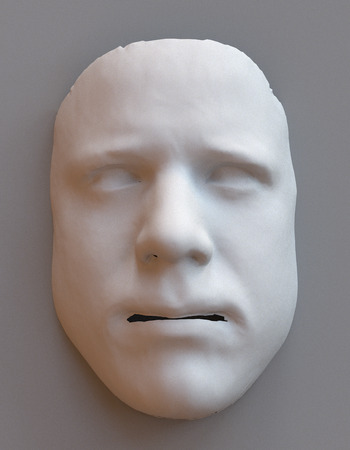} &
        \includegraphics[align=c,width=.09\linewidth]{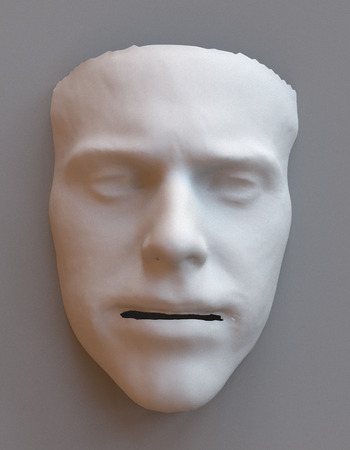} &
        \includegraphics[align=c,width=.09\linewidth]{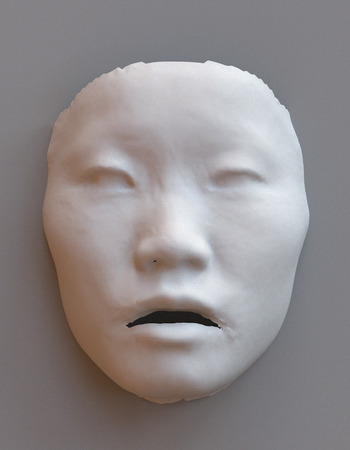} &
        \includegraphics[align=c,width=.09\linewidth]{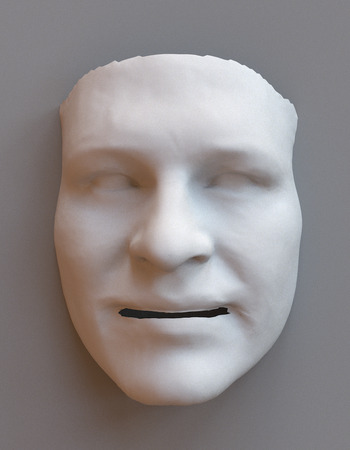} &
        \includegraphics[align=c,width=.09\linewidth]{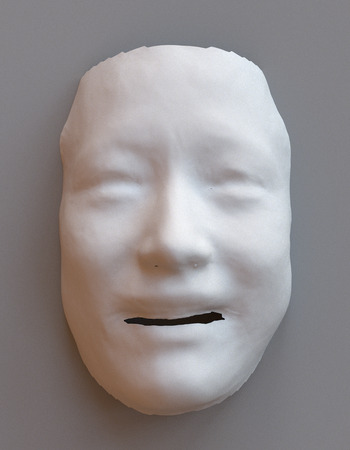} &
        \includegraphics[align=c,width=.09\linewidth]{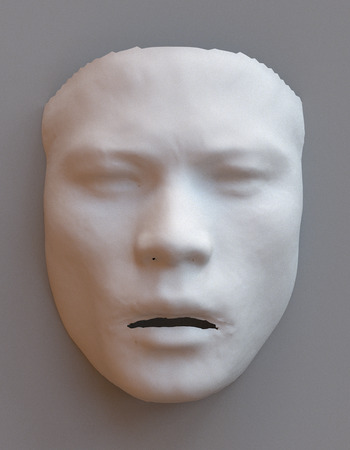} &
        \includegraphics[align=c,width=.09\linewidth]{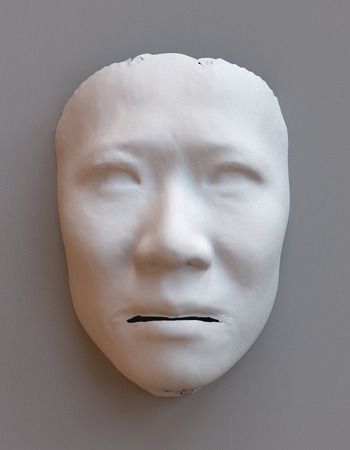} &
        \includegraphics[align=c,width=.09\linewidth]{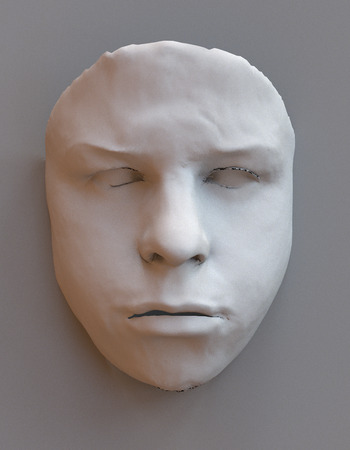}\\
    
        \rotatebox[origin=c]{90}{Closest reg.} & 
        \includegraphics[align=c,width=.09\linewidth]{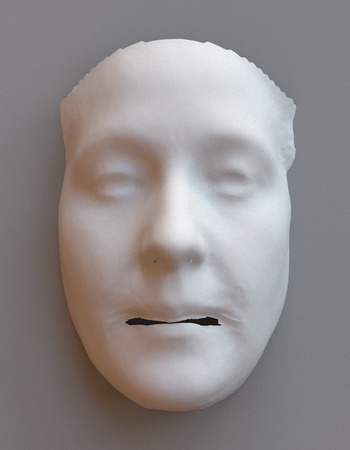} &
        \includegraphics[align=c,width=.09\linewidth]{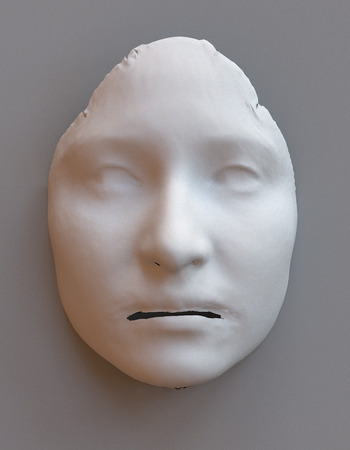} &
        \includegraphics[align=c,width=.09\linewidth]{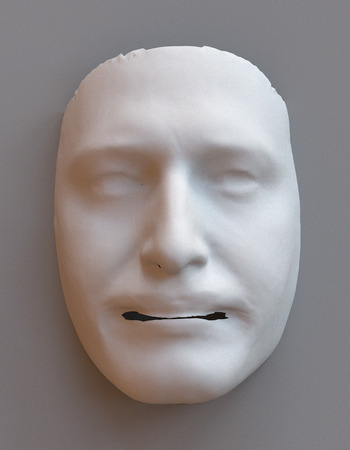} &
        \includegraphics[align=c,width=.09\linewidth]{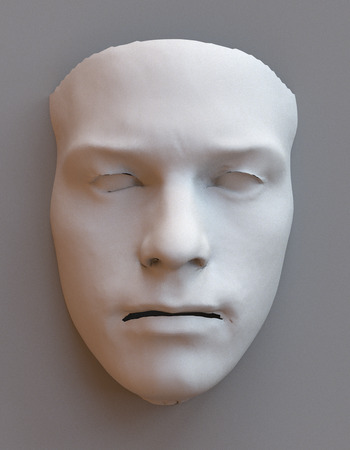} &
        \includegraphics[align=c,width=.09\linewidth]{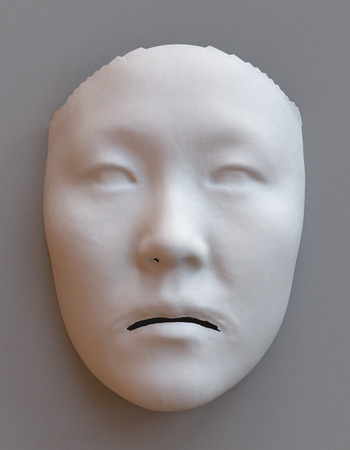} &
        \includegraphics[align=c,width=.09\linewidth]{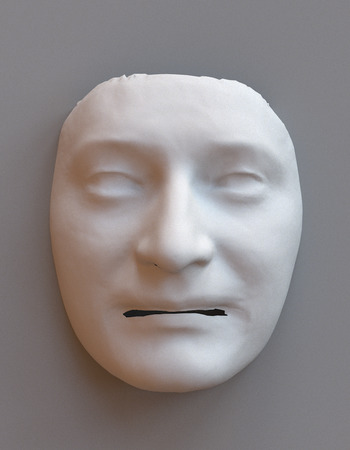} &
        \includegraphics[align=c,width=.09\linewidth]{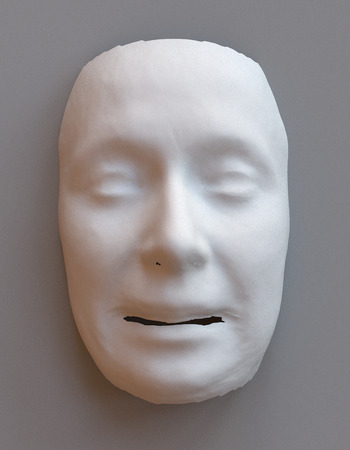} &
        \includegraphics[align=c,width=.09\linewidth]{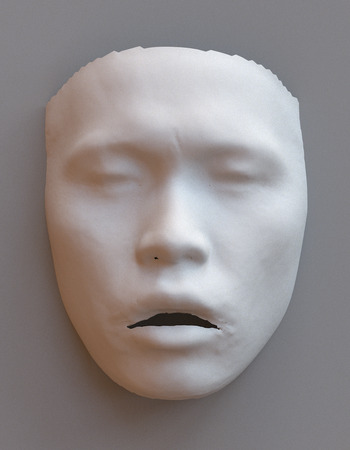} &
        \includegraphics[align=c,width=.09\linewidth]{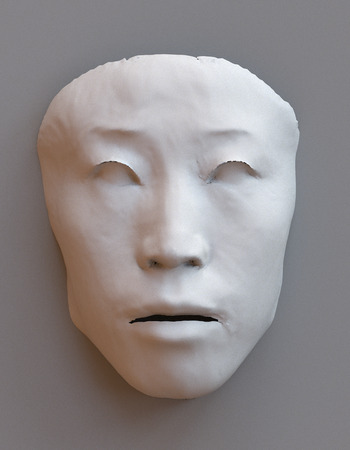} &
        \includegraphics[align=c,width=.09\linewidth]{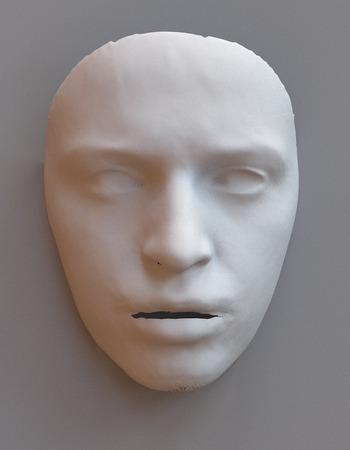}\\
    
        \rotatebox[origin=c]{90}{Closest raw} & 
        \includegraphics[align=c,width=.09\linewidth]{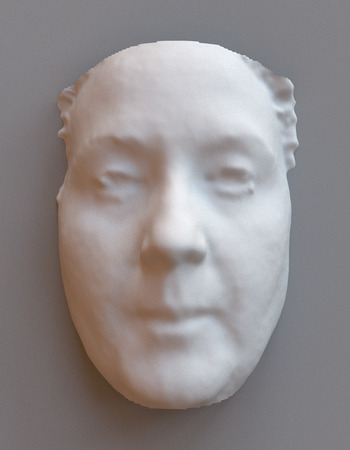} &
        \includegraphics[align=c,width=.09\linewidth]{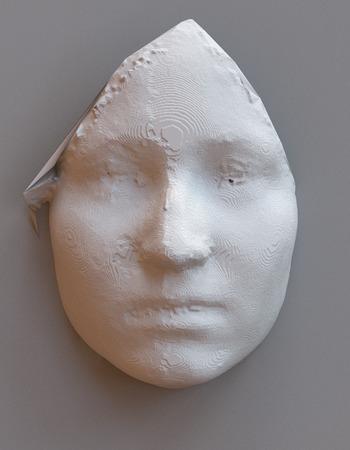} &
        \includegraphics[align=c,width=.09\linewidth]{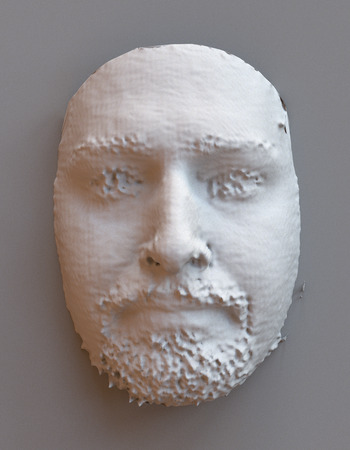} &
        \includegraphics[align=c,width=.09\linewidth]{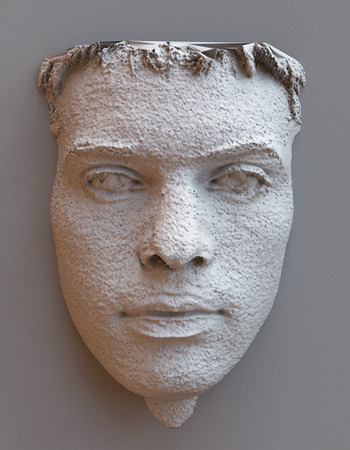} &
        \includegraphics[align=c,width=.09\linewidth]{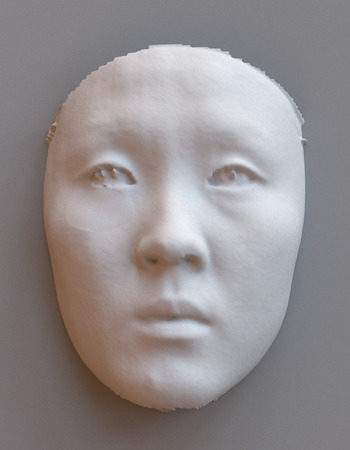} &
        \includegraphics[align=c,width=.09\linewidth]{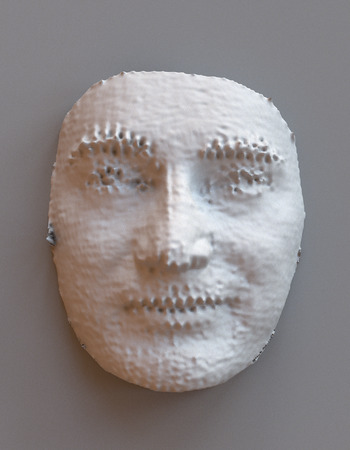} &
        \includegraphics[align=c,width=.09\linewidth]{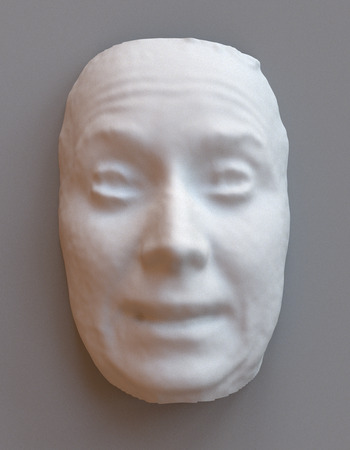} &
        \includegraphics[align=c,width=.09\linewidth]{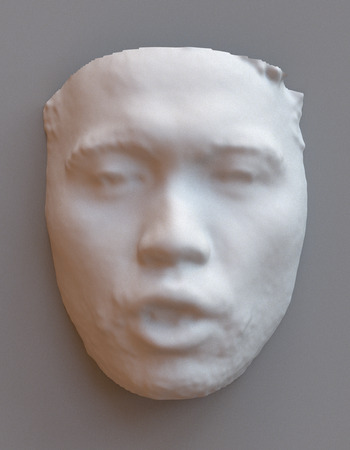} &
        \includegraphics[align=c,width=.09\linewidth]{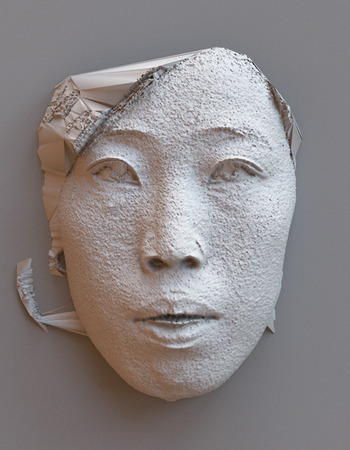} &
        \includegraphics[align=c,width=.09\linewidth]{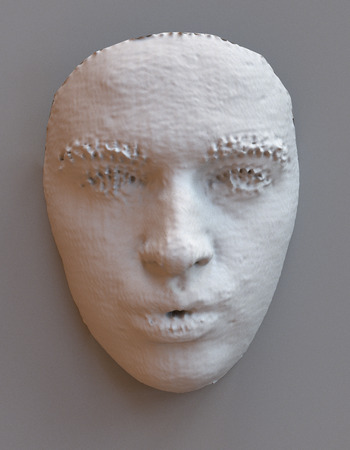}\\
    \end{tabular}}

    \caption{\textbf{Samples from SMF+:} First row: Generated face obtained by sampling a random joint vector. Second row: Closest registration in the training set. Third row: Raw scan from which the closest registration was obtained.}
    \label{fig:smf_plus_samples}
\end{figure*}

\subsubsection{Visualization of the samples}

We now inspect a random subset of the $10{,}000$ samples in Figure \ref{fig:smf_plus_samples}. We render each random sample, its closest point in the registered training set, and the raw scan from which the registration was obtained. We can see the samples generated by SMF+ are highly diverse and realistic-looking: they are close to the registrations of the training set without displaying mode collapse. SMF+ generates detailed faces with sharp features across a wide range of identity, age, ethnic background, and expression, including extreme face and mouth expressions. We further note the absence of artifacts and the seamless blending of the mouth with the rest of the face.

\subsection{Interpolation in the latent space}
\label{sec:interpolation_training}

\begin{figure}[t]
    \centering
    \setlength\tabcolsep{1.5pt}
    {\small
    \begin{tabular}{c|ccccc}
        & $t = 0$ & $t = 0.25$ & $t = 0.5$ & $t = 0.75$ & $t = 1$\\
        \rotatebox[origin=c]{90}{Simultaneous} & 
        \includegraphics[align=c,width=.17\linewidth]{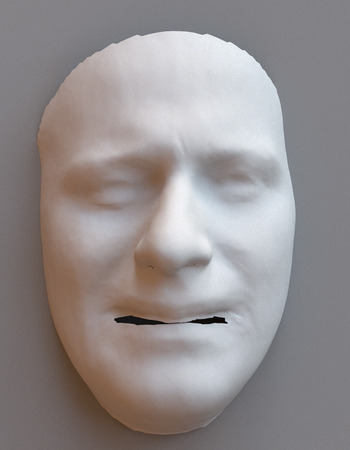} &
        \includegraphics[align=c,width=.17\linewidth]{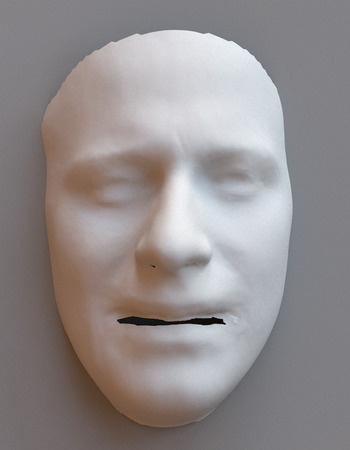} &
        \includegraphics[align=c,width=.17\linewidth]{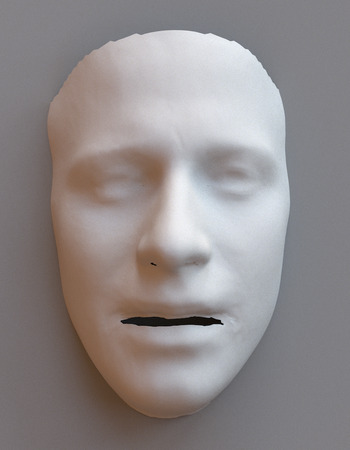} &
        \includegraphics[align=c,width=.17\linewidth]{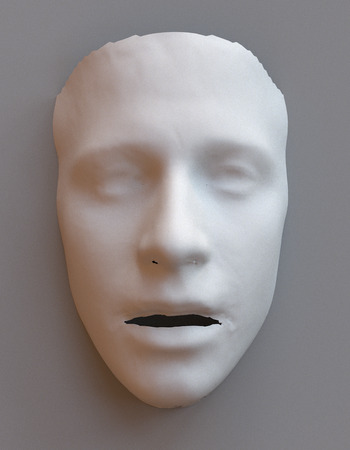} &
        \includegraphics[align=c,width=.17\linewidth]{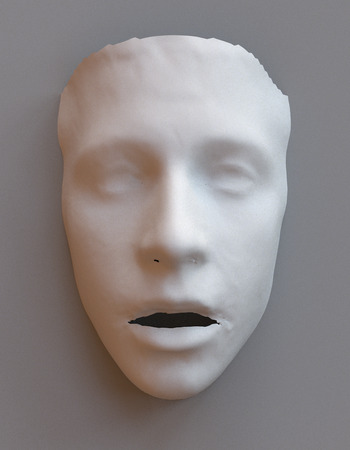}\\

        \rotatebox[origin=c]{90}{Identity only} & 
        \includegraphics[align=c,width=.17\linewidth]{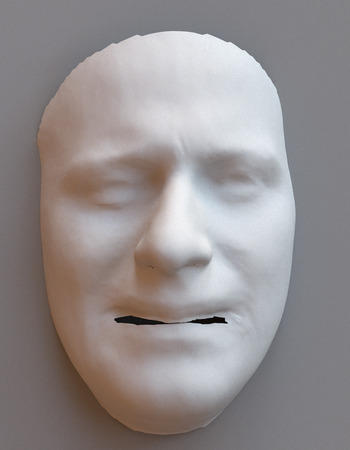} &
        \includegraphics[align=c,width=.17\linewidth]{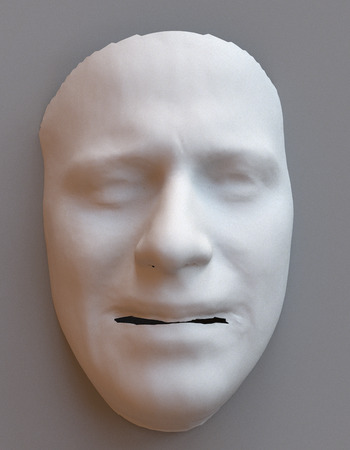} &
        \includegraphics[align=c,width=.17\linewidth]{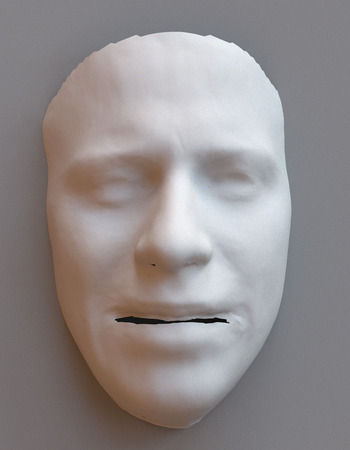} &
        \includegraphics[align=c,width=.17\linewidth]{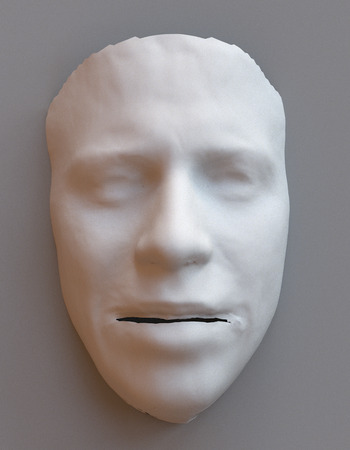} &
        \includegraphics[align=c,width=.17\linewidth]{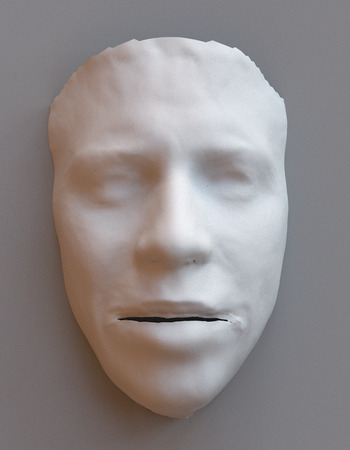}\\

        \rotatebox[origin=c]{90}{Expression only} & 
        \includegraphics[align=c,width=.17\linewidth]{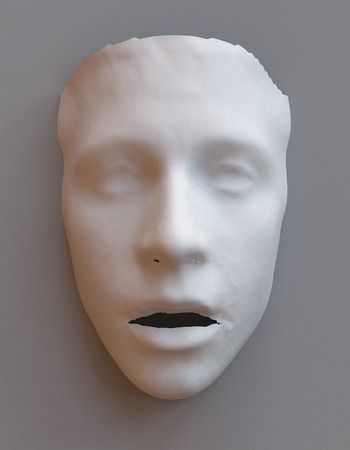} &
        \includegraphics[align=c,width=.17\linewidth]{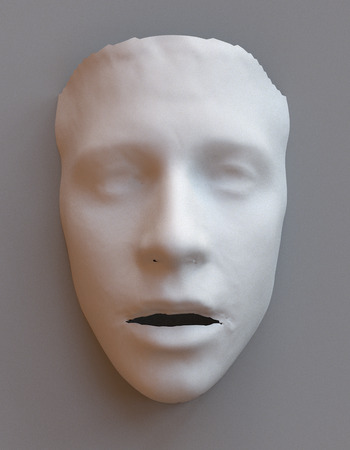} &
        \includegraphics[align=c,width=.17\linewidth]{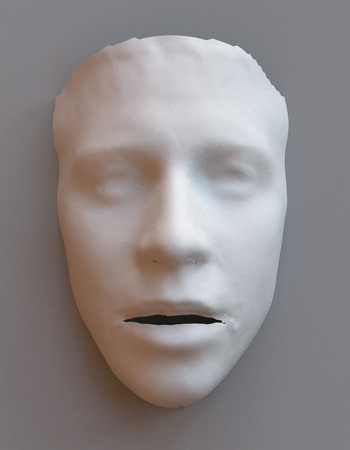} &
        \includegraphics[align=c,width=.17\linewidth]{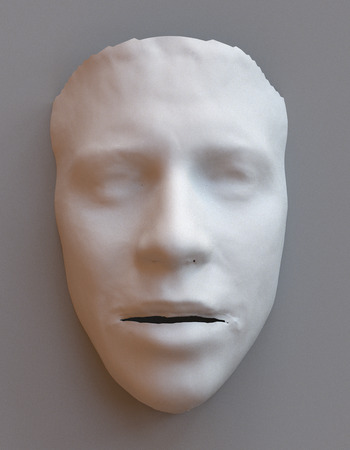} &
        \includegraphics[align=c,width=.17\linewidth]{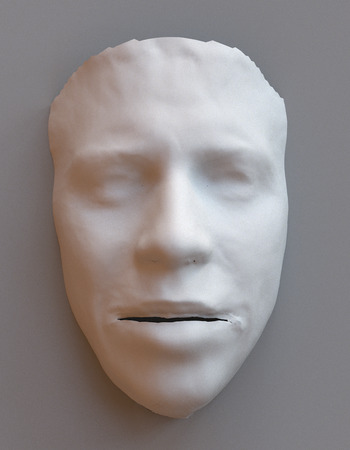}\\
    \end{tabular}
    }

    \caption{\textbf{Interpolation on the training set:} joint interpolation of identity and expression, and interpolation over one factor with the other factor fixed.}
    \label{fig:interpolation}
\end{figure}

We now present a surface-to-surface translation experiment on the training set by showing the results of expression transfer and identity and expression interpolation in the latent spaces of SMF+. Since the latent vectors are hyperspherical, care must be taken to interpolate along the geodesics on the manifold. We therefore interpolate between two latent vectors $\mathbf{z}_1$ and $\mathbf{z}_2$ and $t \in [0, 1]$ as
\begin{equation}
    \mathbf{z}_{i} = \frac{\mathbf{z}_1 + t (\mathbf{z}_1 - \mathbf{z_2})}{||\mathbf{z}_1 + t (\mathbf{z_2} - \mathbf{z}_1)||_2}.
    \label{eq:geodesic_sphere}
\end{equation}

We select two expressive scans of two different subjects, referred to as S1 and S2, from two different databases (BU-3DFE and BU-4DFE) displaying distinct expressions (disgust and surprise). We study three cases: simultaneous interpolation of identity and expression, interpolation of identity for a fixed expression, and interpolation of expression for a fixed identity. We render points along the trajectory defined by Equation \ref{eq:geodesic_sphere} at $t \in \{0, 0.25, 0.5, 0.75, 1\}$. The results of the interpolation are presented in Figure \ref{fig:interpolation}.

We observe smooth interpolation in all three cases. For simultaneous interpolation, we obtain a continuous morphing of the first expressive scan ($t=0$) into the second expressive scan ($t=1$). In particular, we note that the midpoint resembles what would be the neutral scan of a subject presenting physical traits of both the source (nose, forehead) and destination (eyes, jawline) subjects. The interpolation of the identity vector for the fixed expression of S1 shows a smooth transition towards S2 while keeping the correct expression. Conversely, interpolation between S2 and S2 with the expression of S1 shows the overall identity is recognizable and the expression displays a smooth evolution from surprise to disgust. These results show our model can be used for expression transfer and smooth interpolation on the training set. In Section \ref{sec:in_the_wild}, we evaluate SMF on surface-to-surface translation tasks in the wild.

\begin{figure*}[t]
    \centering
    \setlength\tabcolsep{1.5pt}
    {\small
    \begin{tabular}{c|cccccccc}
         & A (body) & A (body) & A (neutral) & A (complex) & A (happy) & A (surprise) & B (happy) & C (neutral)\\
        \rotatebox[origin=c]{90}{Raw scan} &
        \includegraphics[align=c,width=.09\linewidth]{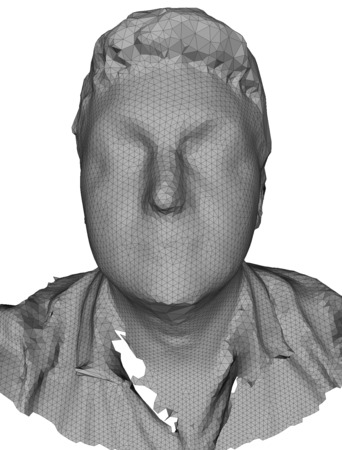} &
        \includegraphics[align=c,width=.09\linewidth]{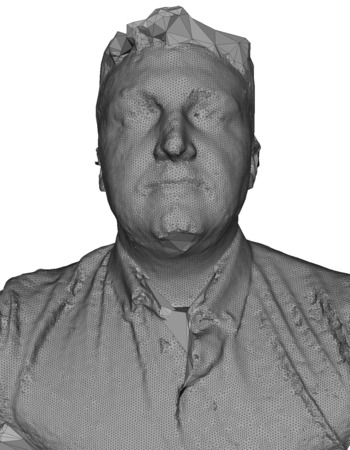} &
        \includegraphics[align=c,width=.09\linewidth]{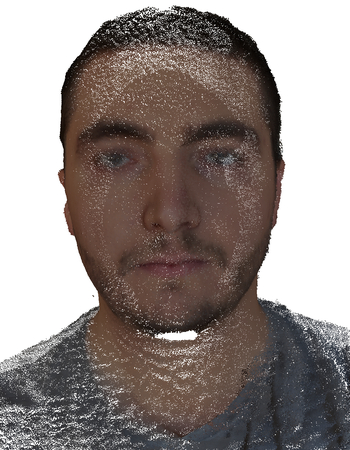} &
        \includegraphics[align=c,width=.09\linewidth]{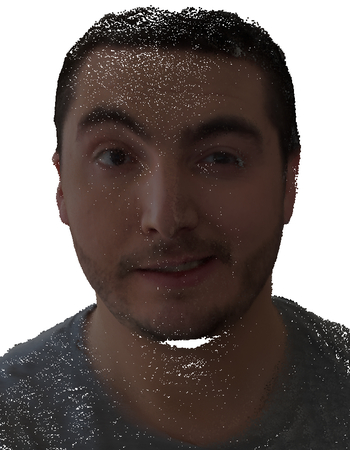} &
        \includegraphics[align=c,width=.09\linewidth]{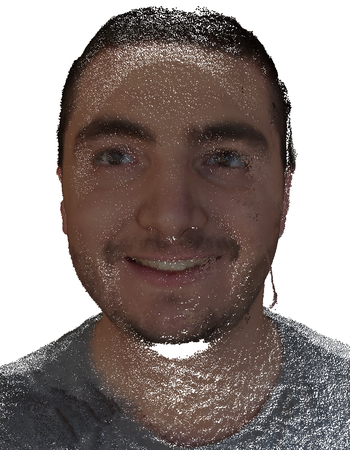} &
        \includegraphics[align=c,width=.09\linewidth]{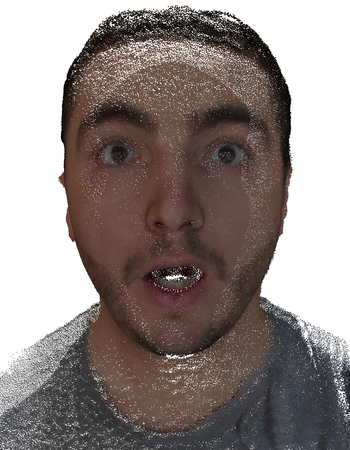} &
        \includegraphics[align=c,width=.09\linewidth]{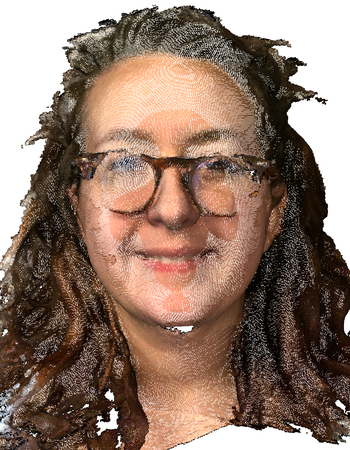} &
        \includegraphics[align=c,width=.09\linewidth]{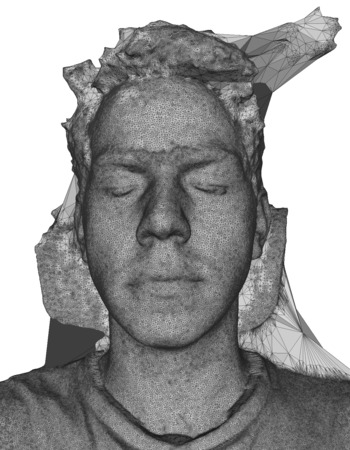}\\
        \hline
        \vspace{.1em}

        \rotatebox[origin=c]{90}{SMF (att. mask)} &
        \includegraphics[align=c,width=.09\linewidth]{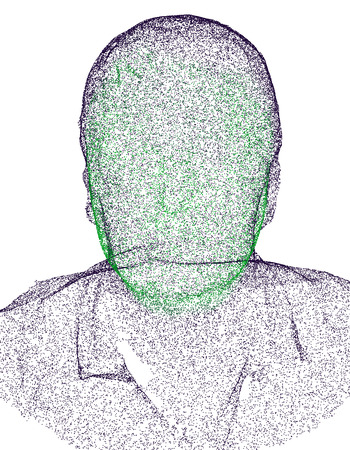} &
        \includegraphics[align=c,width=.09\linewidth]{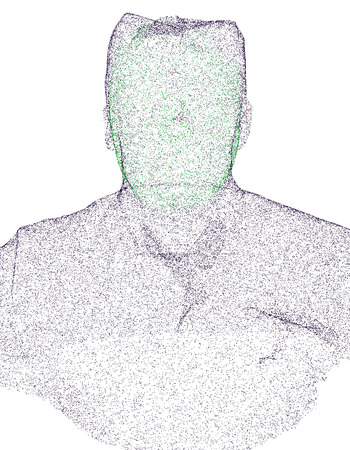} &
        \includegraphics[align=c,width=.09\linewidth]{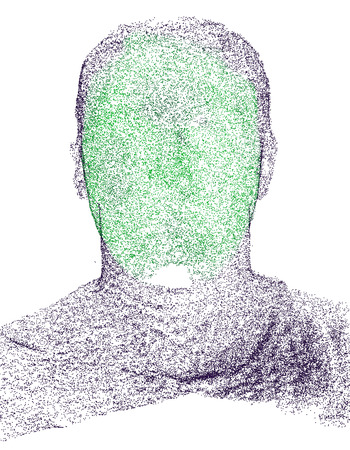} &
        \includegraphics[align=c,width=.09\linewidth]{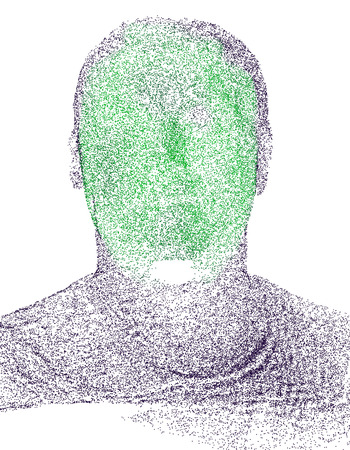} &
        \includegraphics[align=c,width=.09\linewidth]{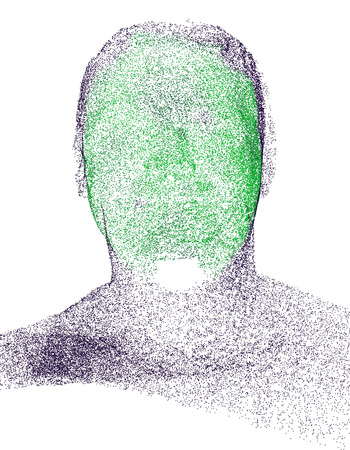} &
        \includegraphics[align=c,width=.09\linewidth]{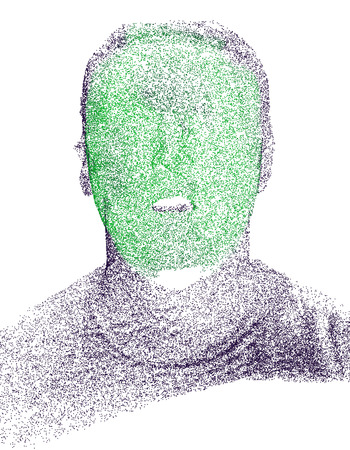} &
        \includegraphics[align=c,width=.09\linewidth]{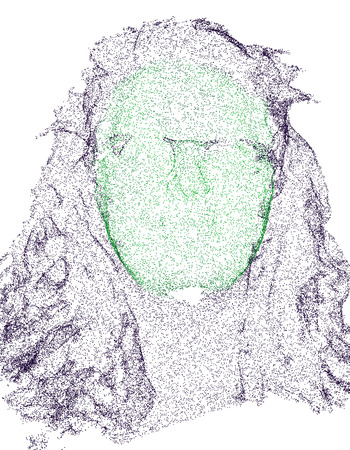} &
        \includegraphics[align=c,width=.09\linewidth]{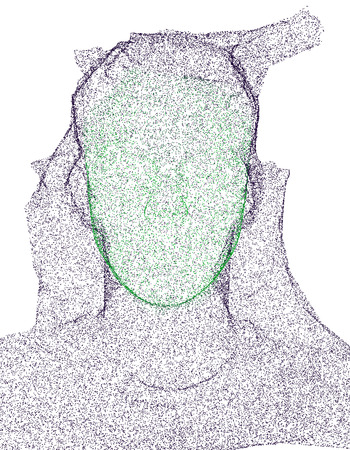}\\
        \hline
        \vspace{.1em}

        \rotatebox[origin=c]{90}{SMF} &
        \includegraphics[align=c,width=.09\linewidth]{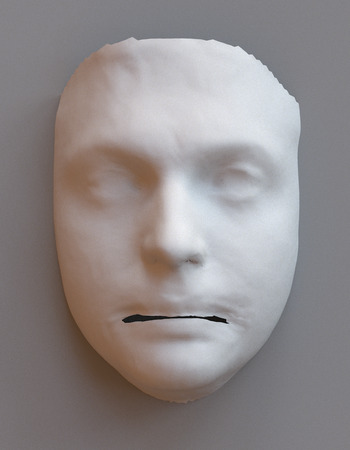} &
        \includegraphics[align=c,width=.09\linewidth]{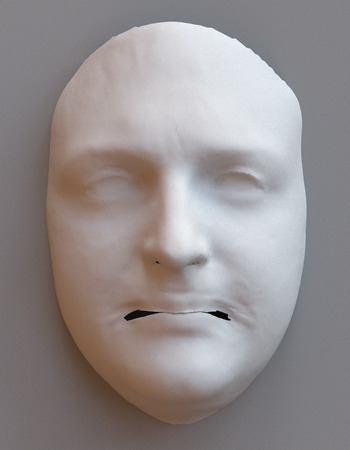} &
        \includegraphics[align=c,width=.09\linewidth]{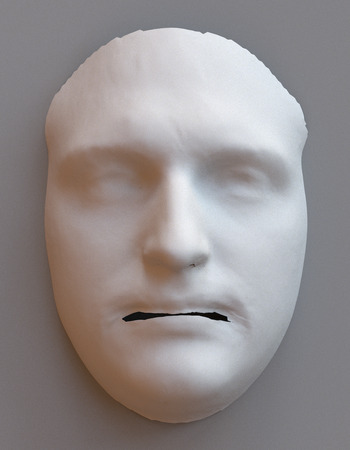} &
        \includegraphics[align=c,width=.09\linewidth]{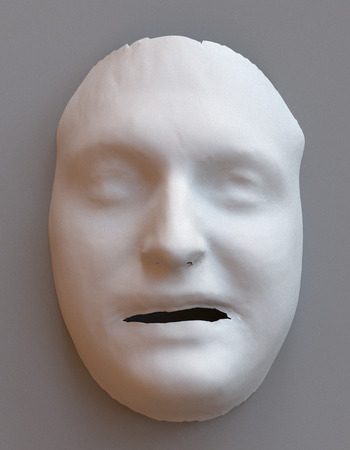} &
        \includegraphics[align=c,width=.09\linewidth]{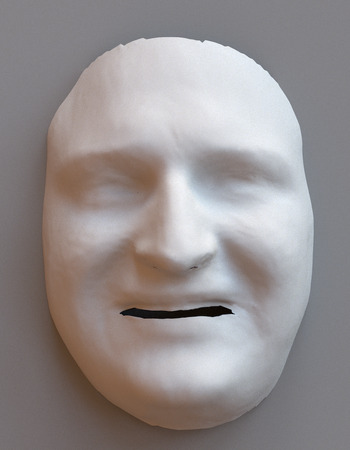} &
        \includegraphics[align=c,width=.09\linewidth]{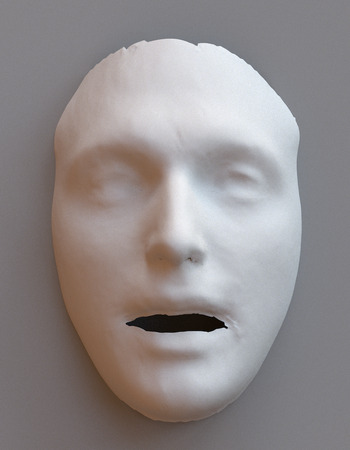} &
        \includegraphics[align=c,width=.09\linewidth]{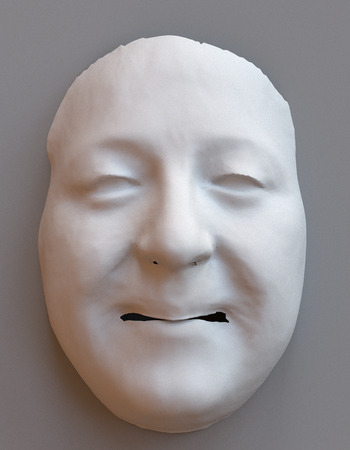} &
        \includegraphics[align=c,width=.09\linewidth]{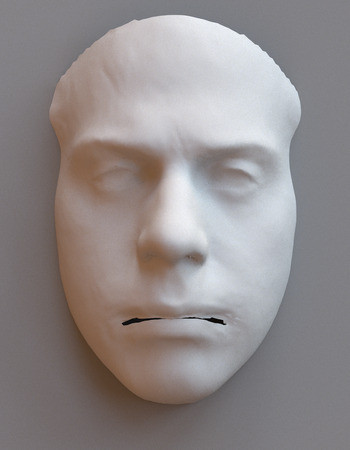}\\
        \hline
        \vspace{.1em}

        \rotatebox[origin=c]{90}{No att.} &
        \includegraphics[align=c,width=.09\linewidth]{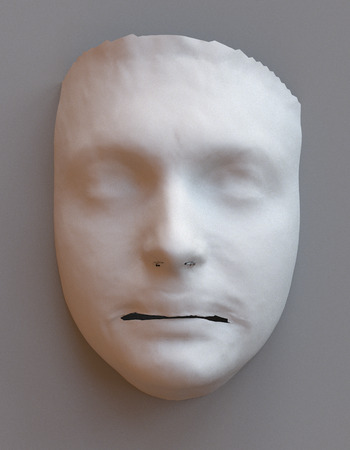} &
        \includegraphics[align=c,width=.09\linewidth]{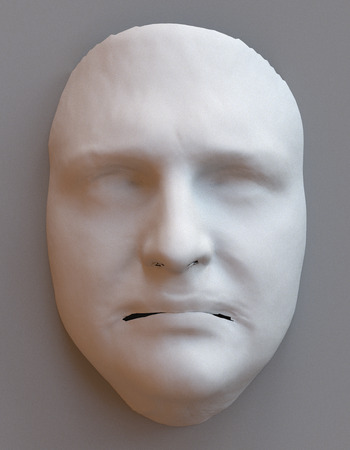} &
        \includegraphics[align=c,width=.09\linewidth]{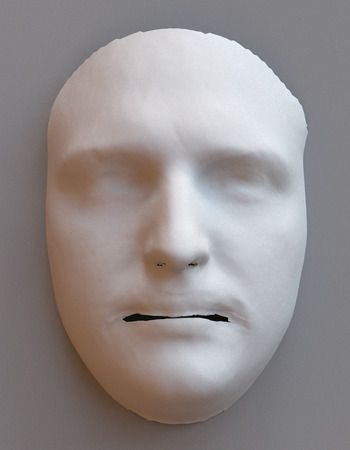} &
        \includegraphics[align=c,width=.09\linewidth]{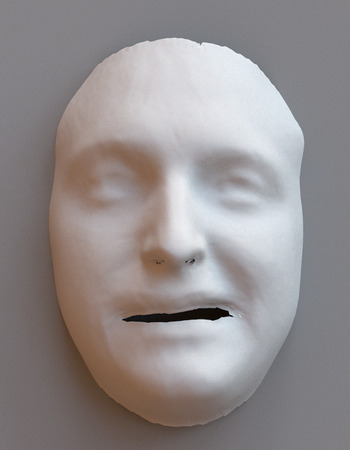} &
        \includegraphics[align=c,width=.09\linewidth]{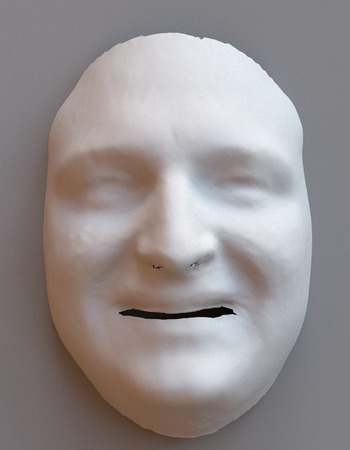} &
        \includegraphics[align=c,width=.09\linewidth]{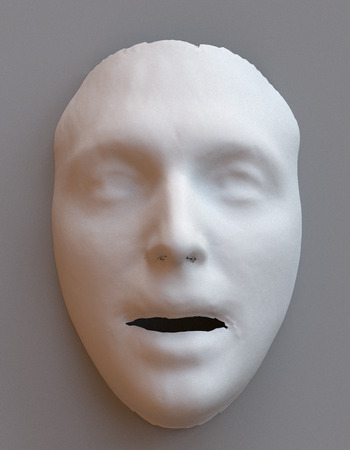} &
        \includegraphics[align=c,width=.09\linewidth]{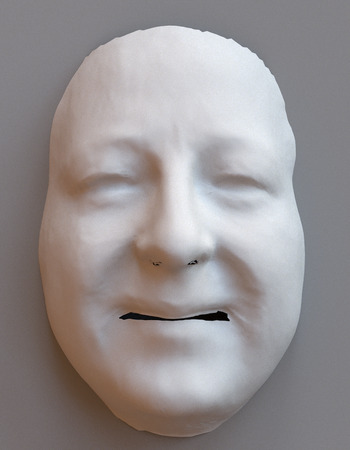} &
        \includegraphics[align=c,width=.09\linewidth]{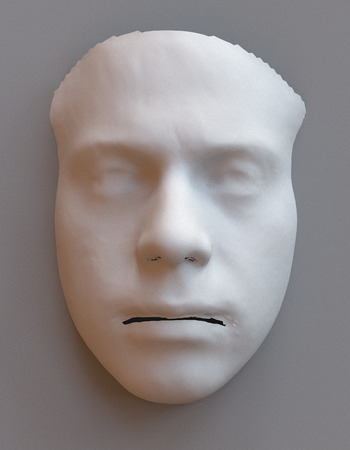}
    \end{tabular}
    }
    \caption{\textbf{In the wild registrations with and without attention:} the scans of subject A were acquired over a period of two years using three different cameras (two different body capture stages and a commodity depth sensor in a smartphone). The scan of subject B was also acquired using a smartphone depth camera, but using a lower resolution setting. The scan of subject C is from a state of the art facial scanning light stage. SMF provides consistent high-quality registrations even from low-resolution scans comprising large areas of the body, hair, or background. In particular, the six scans of subject A show consistent representation of the identity. The attention mechanism can be seen to improve details in the registrations. }
    \label{fig:registration_itw}
\end{figure*}

\begin{figure*}[t]
    \centering
    \setlength\tabcolsep{1.5pt}
    {\small
    \begin{tabular}{ccccccccc}
        A (complex) & $t = 0.25$ & $t = 0.5$ & $t = 0.75$ & C (neutral) & $t = 0.25$ & $t = 0.5$ & $t = 0.75$ & C (surprised)\\
        \includegraphics[align=c,width=.09\linewidth]{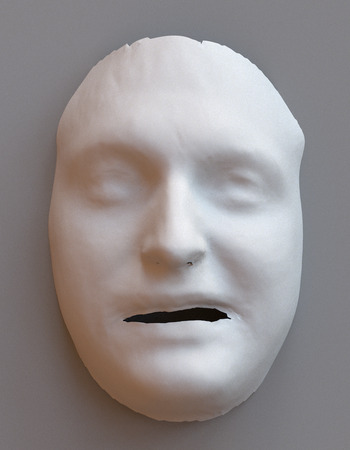} &
        \includegraphics[align=c,width=.09\linewidth]{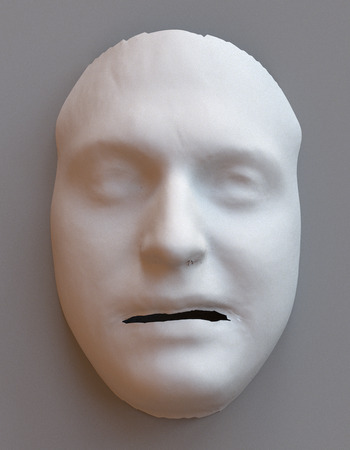} &
        \includegraphics[align=c,width=.09\linewidth]{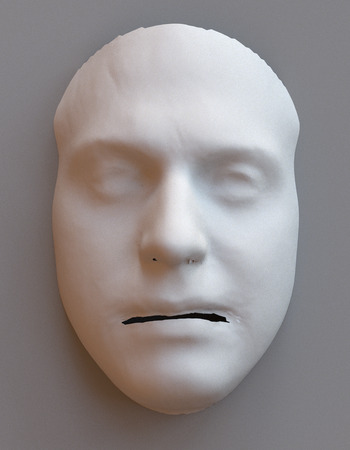} &
        \includegraphics[align=c,width=.09\linewidth]{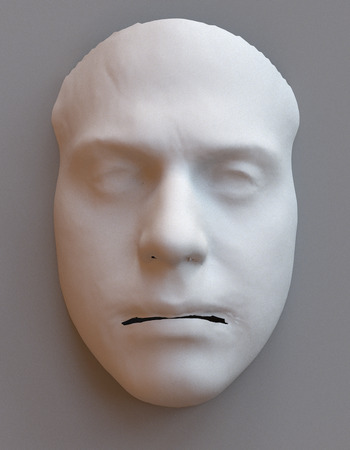} &
        \includegraphics[align=c,width=.09\linewidth]{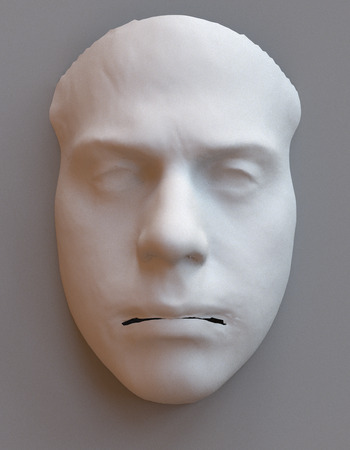} &
        \includegraphics[align=c,width=.09\linewidth]{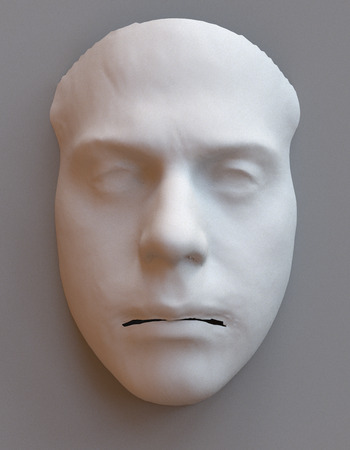} &
        \includegraphics[align=c,width=.09\linewidth]{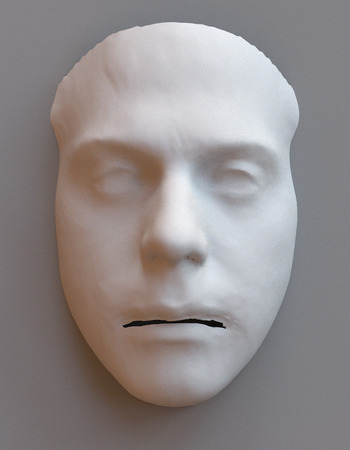} &
        \includegraphics[align=c,width=.09\linewidth]{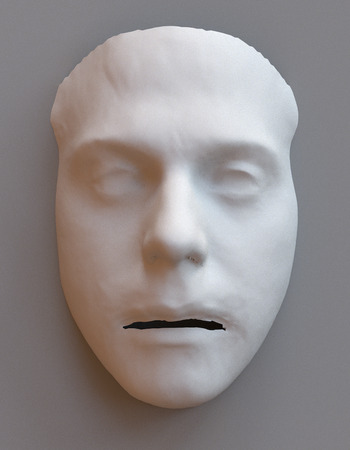} &
        \includegraphics[align=c,width=.09\linewidth]{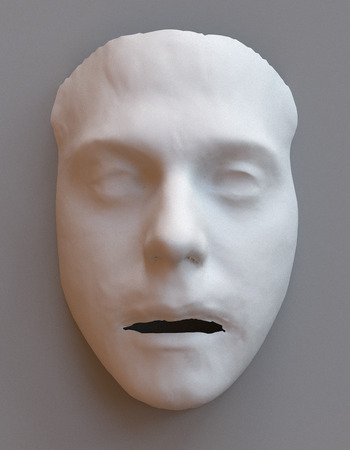}\\
        \textbf{Native} & & & &  \textbf{Native} & & & & \textbf{Transferred}
    \end{tabular}
    }
    \caption{\textbf{Interpolation, transfer, and morphing in the wild:}  From A "complex" to C "neutral" to C "surprised" transferred from A.}
    \label{fig:morphing_itw}
\end{figure*}

\subsection{Face modeling and registration in the wild}
\label{sec:in_the_wild}

We now evaluate SMF on the difficult tasks of registration and manipulation of scans found "in the wild", \ie in uncontrolled environments, with arbitrary sensor types and acquisition noise. We collected the scans of three subjects, referred to A, B, and C, in various conditions. For subject A (male, Caucasian), we obtained crops of two body scans, acquired at over a year and half's interval using two different body capture setups that produce meshes, in two different environments. The first scan shows a crop of the subject squatting while raising his right eyebrow, the second is of the subject jumping with a neutral face. We further acquired four high density point clouds of subject A performing different facial expressions : neutral, smiling (happy), surprise, and a "complex" compound expression consisting of raising the right eyebrow while opening and twisting the mouth to the left. Scanning was done in an uncontrolled environment using a commodity sensor, namely the embedded depth camera of an iPhone 11 Pro. Subject B (female, Caucasian) was captured posing with a light smile in a different uncontrolled environment, also with an iPhone 11 Pro, but using a lower resolution point cloud. Finally, subject C (male, Caucasian) was captured in a neutral pose using a state of the art light stage setup that outputs very high resolution meshes. All in all, the scans represent four different cameras, in five different environments, at five different levels of detail and surface quality, and across two different modalities (mesh and point cloud).

We use the pre-trained SMF model with and without attention to further extend the ablation study of Section \ref{sec:ablation_encoder}. Scans were rigidly aligned with the cropped LSFM mean using landmarks. For meshes (body scans, light stage scan), we sample $2^{16}$ input points at random on the surface of the triangular mesh. For point clouds, we select $2^{16}$ points.

Figure \ref{fig:registration_itw} shows the raw scans, registration from SMF, predicted attention mask for SMF, and registration for SMF trained without visual attention. We can see SMF produced very consistent registrations for subject A across modalities, resolution, and time: it is clear, from the registrations, that the scans came from the same subject, even for the low-resolution face and shoulders region of the first body scan, for which important facial features and the elevated position of the right eyebrow were captured. Comparing the neutral iPhone scan and the neutral body scan further shows identity was robustly captured at the two different resolutions. The highly non-linear complex expression was, also, accurately captured, and so were the more standard happy and surprise expressions. Performance was stable for lower-resolution raw point clouds too as shown with the registration of subject B. SMF produced a sharp detailed registration of the high quality light-stage scan of subject C, correctly capturing the shape of the nose, the sharpness and inflexion of the eyebrows, and the angle of the mouth.

Compared to SMF, SMF trained without our attention mechanism still produced high quality registrations but with fewer details. The two body scans and the light stage scans show clear differences, especially in the eyes. The happy expression of subject B was not captured as accurately, and the shape of the face appears elongated. Looking at the attention masks, we can see our visual attention mechanism discarded points from the body, the inside of the mouth (A surprise), environment noise (C neutral), and hair and partial occlusions (B happy, for which it removed most of the glasses).

\paragraph{Morphing and editing in the wild}

We now show our pre-trained model can be used for shape morphing and editing, such as expression transfer, by linearly interpolating in $\mathcal{S}^{255}$ between the predicted identity and expression vectors of the raw scans. We select the "A complex", "A surprise" and "C neutral" scans and register both of them with our pre-trained SMF model, keeping their predicted identity and expression embeddings. We first interpolate the identity and expression jointly between "A complex" and "C neutral" to produce a smooth morphing. We then keep the identity vector fixed to that of "C neutral" and linearly interpolate between the expression vectors of "C neutral" and "A surprise", this produces a smooth expression transfer. Both experiments are shown as a continuous transformation in Figure \ref{fig:morphing_itw}.

As apparent from Figure \ref{fig:morphing_itw}, our model is able to smoothly interpolate between subjects and expressions of scans captured, in the wild, across different modalities and resolutions. The morphing from A complex to C neutral produces smooth facial motions without discontinuities. Our model is further able to, not only transfer expressions in the wild, but smoothly interpolate between expression vectors of different subjects for a fixed identity. The interpolation transfer again produces a smooth natural-looking transition between the neutral scan of C, with the mouth and eyebrows smoothly moving from a resting position to a surprise expression, while keeping the facial features of subject C.

\section{Conclusion and Future Work}

In this paper, we present Shape My Face (SMF), a novel learning-based algorithm that treats the registration task as a surface-to-surface translation problem. Our model is based on an improved point cloud encoder made highly robust with a novel visual attention mechanism, and on our mesh inception decoders that leverage graph convolutions to learn a compact non-linear morphable model of the human face. We further improve robustness to noise in face scans by blending the output of the mesh convolutions with a specialized statistical model of the mouth in a seamless way. Our model learns to produce high quality registrations both in sample and out of sample, thanks to the improved weight sharing and stochastic training approach that prevent the model from overfitting any particular discretization of the training scans.

We introduce a large scale morphable model, coined as SMF+, by training SMF on 9 comprehensive human 3D facial databases. Our experimental evaluation shows SMF+ can generate thousands of diverse realistic-looking faces from random noise across a wide range of age, ethnicities, genders, and (extreme) facial expressions. We evaluate SMF+ on shape editing and translation tasks and show our model can be used for identity and expression transfer and interpolation. Finally, we show SMF can also accurately register and interpolate between facial scans captured in uncontrolled conditions for unseen subjects and sensors, allowing for shape editing entirely in the wild. In particular, we demonstrated smooth interpolation and transfer of expression and identity between a very high quality mesh acquired in controlled conditions with a sophisticated facial capture environment, and a noisy point cloud produced by consumer-grade electronics.

Future work will investigate improving the reproduction of high frequency details in the scans, and registering texture and geometry simultaneously.

\bibliographystyle{spbasic}      %
\bibliography{paper_clean}   %

\end{document}